\documentclass[10pt,twocolumn,letterpaper]{article}
\usepackage[OT1]{fontenc}

\usepackage[pagenumbers]{cvpr} 

\usepackage{cuted}

\usepackage{currfile}

\usepackage{caption}

\usepackage{amsmath,amssymb,amsthm,mathtools}
\usepackage{booktabs,multirow,makecell,tabularx}
\usepackage{xspace}
\usepackage{tikz}
\usepackage{placeins}
\usetikzlibrary{positioning,calc,decorations.pathreplacing,arrows.meta}
\newcommand{\ours}{ProbConv\xspace}
\newcommand{\Compact}{\operatorname{Compact}}
\newcommand{\TopK}{\operatorname{TopK}}
\newcommand{\sg}{\operatorname{sg}}
\newcommand{\E}{\mathbb{E}}
\newtheorem{proposition}{Proposition}
\newcommand{\cobrafit}[1]{\setbox0\hbox{#1}\ifdim\wd0>\linewidth\resizebox{\linewidth}{!}{\box0}\else\box0\fi}
\usepackage{enumitem}
\newcommand{\sd}[1]{{\scriptsize$\,\pm$#1}}

\definecolor{cvprblue}{rgb}{0.21,0.49,0.74}
\usepackage[pagebackref,breaklinks,colorlinks,allcolors=cvprblue]{hyperref}
\newcommand{\appxtag}{App.}
\newcommand{\appxtags}{Apps.}
\newcommand{\Appxtag}{Appendix}
\newcommand{\Appxtags}{Appendices}
\newcommand{\maintag}{Main}
\newcommand{\Maintag}{Main}
\newcommand{\appxsecstyle}{\arabic}
\newif\ifappxSnumbers
\appxSnumbersfalse

\newif\ifinappendix
\protected\def\toappx{\ifinappendix\else\appxtag~\fi}
\protected\def\Toappx{\ifinappendix\else\Appxtag~\fi}
\protected\def\tomain{\ifinappendix\maintag~\fi}
\protected\def\Tomain{\ifinappendix\Maintag~\fi}
\protected\def\appxsec{\ifinappendix Sec.\else\appxtag\fi}
\protected\def\appxsecs{\ifinappendix Secs.\else\appxtags\fi}
\protected\def\Appxsec{\ifinappendix Section\else\Appxtag\fi}
\protected\def\Appxsecs{\ifinappendix Sections\else\Appxtags\fi}

\makeatletter
\newcommand{\appendixrefs}{%
  \global\inappendixtrue
  \renewcommand{\thesection}{\appxsecstyle{section}}%
  \ifappxSnumbers\gdef\appx@numprefix{S}\else\global\let\appx@numprefix\@empty\fi
  \appx@renumber{figure}\appx@renumber{table}\appx@renumber{equation}\appx@renumber{algocf}%
  \appx@renumber{theorem}\appx@renumber{proposition}\appx@renumber{lemma}\appx@renumber{corollary}%
  \crefalias{section}{appendix}%
  \crefalias{subsection}{subappendix}%
  \crefalias{subsubsection}{subsubappendix}%
  \crefalias{figure}{appfigure}%
  \crefalias{table}{apptable}%
  \crefalias{equation}{appequation}%
  \crefalias{algocf}{appalgorithm}%
  \crefalias{algorithm}{appalgorithm}%
  \crefalias{theorem}{apptheorem}%
  \crefalias{proposition}{appproposition}%
  \crefalias{lemma}{applemma}%
  \crefalias{corollary}{appcorollary}}
\newcommand{\appx@renumber}[1]{%
  \@ifundefined{c@#1}{}{%
    \setcounter{#1}{0}%
    \@namedef{the#1}{\appx@numprefix\arabic{#1}}%
    \@namedef{theH#1}{S\arabic{#1}}}}

\newcommand{\appx@names}[6]{%
  \crefname{#1}{\tomain #3}{\tomain #4}\Crefname{#1}{\Tomain #5}{\Tomain #6}%
  \crefname{#2}{\toappx #3}{\toappx #4}\Crefname{#2}{\Toappx #5}{\Toappx #6}}
\AtEndPreamble{%
  \appx@names{section}{appendix}{Sec.}{Secs.}{Section}{Sections}%
  \appx@names{figure}{appfigure}{Fig.}{Figs.}{Figure}{Figures}%
  \appx@names{table}{apptable}{Tab.}{Tabs.}{Table}{Tables}%
  \appx@names{equation}{appequation}{Eq.}{Eqs.}{Equation}{Equations}%
  \appx@names{algorithm}{appalgorithm}{Alg.}{Algs.}{Algorithm}{Algorithms}%
  \appx@names{theorem}{apptheorem}{Thm.}{Thms.}{Theorem}{Theorems}%
  \appx@names{proposition}{appproposition}{Proposition}{Propositions}{Proposition}{Propositions}%
  \appx@names{lemma}{applemma}{Lem.}{Lems.}{Lemma}{Lemmas}%
  \appx@names{corollary}{appcorollary}{Cor.}{Cors.}{Corollary}{Corollaries}%
  \crefname{appendix}{\appxsec}{\appxsecs}\Crefname{appendix}{\Appxsec}{\Appxsecs}%
  \creflabelformat{appequation}{\textup{(#2#1#3)}}%
}
\makeatother

\def\confName{CVPR}
\def\confYear{2027}

\title{Train What You Run: Probabilistic Convolutions\\for Learning Token Compaction in State Space Models}

\author{
Piotr Kubaty$^{1,2}$\quad
Mikhail Saladkou$^{1}$\quad
Jacek Tabor$^{1,3}$\quad
\L{}ukasz Struski$^{1}$\\[0.5em]
{\normalsize $^{1}$Faculty of Mathematics and Computer Science, Jagiellonian University, Krak\'{o}w, Poland}\\
{\normalsize $^{2}$Doctoral School of Exact and Natural Sciences, Jagiellonian University, Krak\'{o}w, Poland}\\
{\normalsize $^{3}$Centre for Credible Artificial Intelligence, Warsaw University of Technology, Poland}
}

\begin{document}
\maketitle

\begin{abstract}
Visual token compaction reduces inference cost by passing only selected
image patches or video tokens to later layers. In sequence models with
local convolutions, each removal changes which tokens are neighbors.
Learning which tokens to keep therefore requires a training signal that
accounts for the new neighborhoods. We introduce \ours, which computes
the exact expected convolution before activation at a kept position under
independent keep decisions. The operator is differentiable in the keep
probabilities.
With binary decisions, \ours and step gating reproduce the compacted Mamba
block. Across image, vision--language and video backbones, \ours is
strongest under strong compaction: its advantage over matched training
controls grows as more tokens are removed. In Cobra, it also outperforms
adaptations of published vision--language pruning methods at matched
inference budgets. \ours also improves over
matched zero-mask training in video frame selection and speech frame dropping.
Our code is available at: \url{https://github.com/gmum/probconv}.
\end{abstract}

\section{Introduction}
\label{sec:intro}

\begin{figure}[t]
\centering
\input{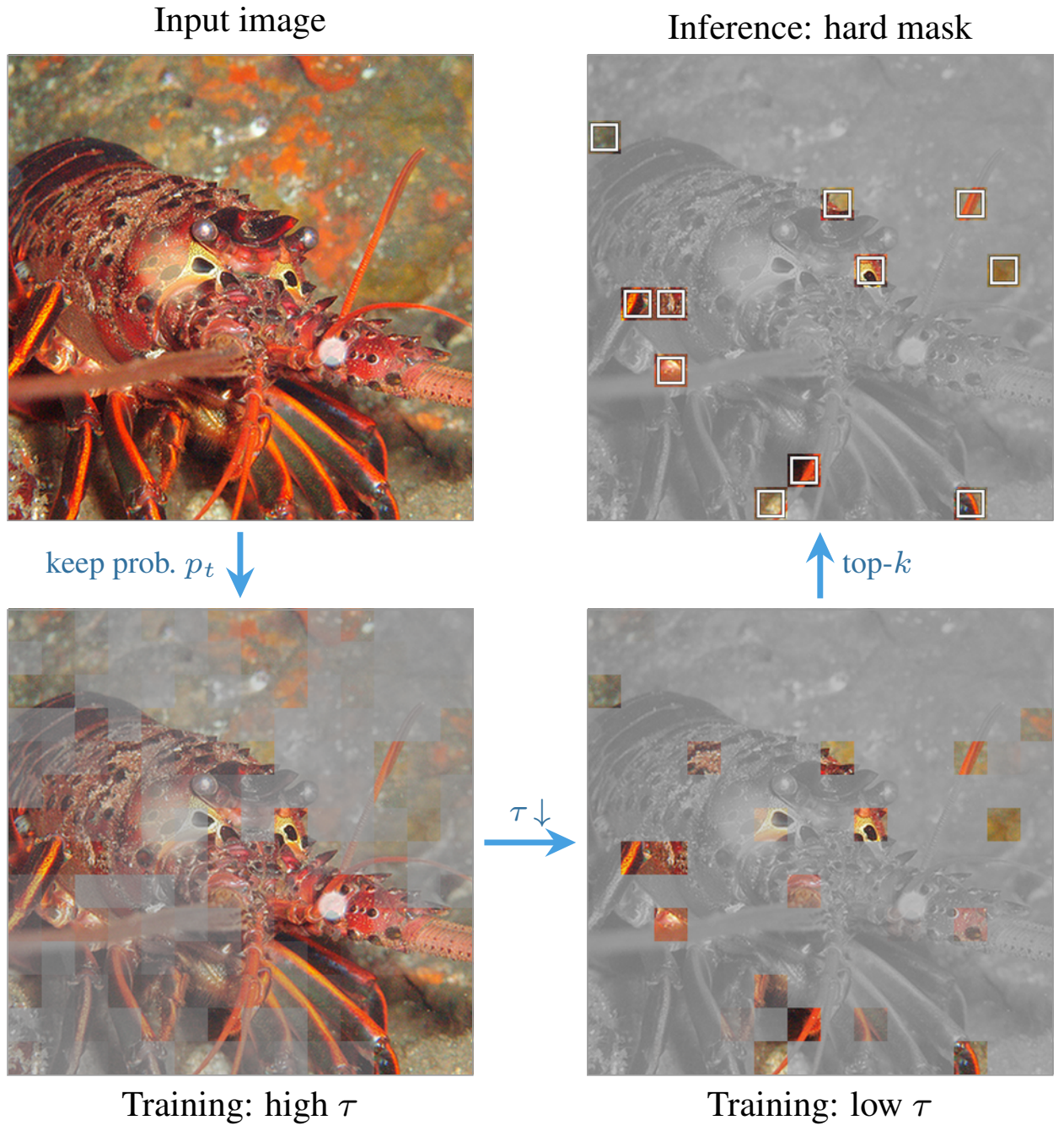}
\caption{\textbf{Keep probabilities in training, a hard mask at inference.}
Illustrative example with schematic probabilities. During training
(bottom), the selector gives each patch a keep probability $p_t$, drawn
from gray ($p_t{=}0$) to full color ($p_t{=}1$), and \ours averages each
convolution window over the keep decisions. At a high LapSum temperature
$\tau$ many patches are partly kept; as $\tau$ decreases, few remain. At
inference (top right), the top-$k$ mask keeps the outlined patches, and
later layers process only these.}
\label{fig:process}
\end{figure}

Many image predictions survive the removal of most intermediate patch tokens:
a per-image search on frozen Vision Mamba finds small patch subsets that
preserve the dense prediction~\cite{zhu2024vim} (\cref{sec:supp_vim}).
Token compaction aims to learn that choice:
a selector keeps a subset of tokens for each input, and later layers process
the shorter sequence (\cref{fig:process}). The challenge is to learn which tokens remain useful
once those layers operate on the selected subset.

\begin{figure*}[t]
\centering
\definecolor{pcGreen}{HTML}{1B9E77}%
\definecolor{pcGreenL}{HTML}{DDF1EA}%
\definecolor{pcPink}{HTML}{E7298A}%
\definecolor{pcPinkL}{HTML}{FBE0EE}%
\definecolor{pcInk}{RGB}{64,64,64}%
\definecolor{pcTok}{RGB}{226,226,226}%
\definecolor{pcLine}{RGB}{150,150,150}%
\definecolor{pcWin}{RGB}{198,222,242}%
\begin{tikzpicture}[x=1mm,y=1mm,font=\footnotesize,line join=round,
  tok/.style={draw=pcLine,fill=pcTok,minimum width=4.6mm,minimum height=4.2mm,inner sep=0pt,font=\scriptsize,line width=0.4pt},
  stok/.style={tok,minimum width=3.3mm,minimum height=3.3mm},
  win/.style={tok,fill=pcWin,draw=pcInk!70},
  zero/.style={tok,fill=pcPinkL,draw=pcPink,text=pcPink},
  gone/.style={tok,fill=white,draw=pcLine,dashed,text=pcLine},
  flow/.style={draw=pcInk,line width=0.6pt,-{Stealth[length=1.6mm,width=1.2mm]}},
  lab/.style={font=\scriptsize,text=pcInk,inner sep=0.5pt,align=center},
  ttl/.style={font=\scriptsize,text=black,inner sep=0.5pt,anchor=west},
  eq/.style={font=\scriptsize,inner sep=0.5pt},
  head/.style={font=\footnotesize\bfseries,inner sep=0pt},
  state/.style={circle,draw=pcInk,fill=pcTok,minimum size=5.6mm,inner sep=0pt,font=\scriptsize},
  newpanel/.style={draw=pcGreen,rounded corners=1.2pt,line width=1.1pt},
]
\node[head,rotate=90] at (1.8,30.2) {zero mask};
\node[head,rotate=90,text=pcGreen] at (1.8,4.6) {\ours};
\begin{scope}[shift={(-11,0)}]
\node[lab,align=left,anchor=west] at (42.6,18) {What\\[-1pt]changes};
\draw[pcLine,dashed,line width=0.4pt] (42,22.8) -- (174.4,22.8);
\draw[pcLine,dashed,line width=0.4pt] (42,13.2) -- (174.4,13.2);

\begin{scope}[shift={(17.2,8)}]
  \node[ttl] at (0,22.2) {same selector};
  \node[ttl] at (0,19.0) {and budget $k$};
  \node[ttl] at (0,14.0) {keep prob.\ $p_t$};
  \foreach \i/\p in {1/0.95,2/0.85,3/0.30,4/0.90,5/0.15,6/0.85}{
    \fill[pcInk!50] ({(\i-1)*3.75+0.6},6.2) rectangle ++(2.5,{\p*6});
    \node[tok,minimum width=3.5mm] at ({(\i-1)*3.75+1.85},3.6) {$x_\i$};
  }
  \node[ttl] at (0,-1.2) {LapSum: $\textstyle\sum p_t{=}k$};
\end{scope}
\draw[draw=pcInk,line width=0.6pt] (40.0,17.8) -- (40.6,17.8);
\draw[flow] (40.6,17.8) |- (43.4,31.4);
\draw[flow] (40.6,17.8) |- (43.4,5.2);

\begin{scope}[shift={(44.5,23.5)}]
  \node[ttl] at (0,12.6) {window of $y_4$ when $x_3$ is dropped};
  \node[tok] at (2.4,7.9) {$x_1$}; \node[win] at (7.4,7.9) {$x_2$};
  \node[zero] at (12.4,7.9) {$0$}; \node[win] at (17.4,7.9) {$x_4$};
  \node[eq,anchor=west] at (21.4,7.9) {$y_4=w_0x_4+w_1\!\cdot\!0+w_2x_2$};
  \node[lab,anchor=west,text=pcPink] at (0,2.4) {the gap stays in the window};
\end{scope}
\begin{scope}[shift={(44.5,-3)}]
  \node[lab,anchor=east] at (4.6,11.4) {$p_3$};
  \node[win] at (7.6,11.4) {$x_2$}; \node[win] at (12.6,11.4) {$x_3$}; \node[win] at (17.6,11.4) {$x_4$};
  \node[lab,anchor=east] at (4.6,6.1) {$\bar p_3$};
  \node[win] at (7.6,6.1) {$x_1$}; \node[win] at (12.6,6.1) {$x_2$}; \node[win] at (17.6,6.1) {$x_4$};
  \node[lab,anchor=west] at (20.6,11.4) {keep $x_3$};
  \node[lab,anchor=west] at (20.6,6.1) {drop $x_3$};
  \draw[decorate,decoration={brace,amplitude=1.1mm},pcInk,line width=0.5pt] (33.4,13.6) -- (33.4,3.9);
  \node[eq,anchor=west,align=left] at (35.2,8.75) {$y_4$: average of\\[-1pt]the compacted\\[-1pt]windows};
  \node[lab,anchor=west,text=pcGreen] at (0,1.1) {exact expectation, one linear scan};
\end{scope}
\draw[newpanel] (43.6,-3.4) rectangle (100.4,12.4);

\draw[flow] (101.4,31.4) -- (105.4,31.4);
\draw[flow] (101.4,5.2) -- (105.4,5.2);
\begin{scope}[shift={(107,23.5)}]
  \node[ttl] at (0,12.6) {state at the dropped $x_3$};
  \node[state] (h2) at (3.2,7.9) {$h_2$};
  \node[state,draw=pcPink,fill=pcPinkL,text=pcPink,minimum size=4.2mm] (h3) at (20,7.9) {$h_3$};
  \draw[flow] (h2) -- node[above=0.5mm,eq] {$\Delta_3$} (h3);
  \node[eq,anchor=west] at (0,2.4) {$h_3=\bar A_3h_2+\bar B_3\!\cdot\!0$};
\end{scope}
\begin{scope}[shift={(107,-3)}]
  \node[state] (g2) at (3.2,10.2) {$h_2$};
  \node[state,draw=pcGreen,fill=pcGreenL] (g3) at (20,10.2) {$h_3$};
  \draw[flow] (g2) -- node[above=0.5mm,eq] {$p_3\Delta_3$} (g3);
  \node[eq,anchor=west] at (0,4.4) {$p_3=0$: $h_3=h_2$};
  \node[lab,anchor=west,text=pcGreen] at (0,1.1) {step gating};
\end{scope}
\draw[newpanel] (106,-3.4) rectangle (134.2,12.4);

\draw[flow] (135.2,31.4) -- (139.2,31.4);
\draw[flow] (135.2,5.2) -- (139.2,5.2);
\begin{scope}[shift={(140.4,23.5)}]
  \node[ttl] at (0,12.6) {training vs.\ inference};
  \node[lab,anchor=west] at (0,7.9) {train};
  \node[lab,anchor=west] at (0,2.4) {deploy};
  \foreach \i/\s in {1/stok,2/stok,3/zero,4/stok,5/zero,6/stok}{\node[\s,minimum width=3.3mm,minimum height=3.3mm] at ({12.2+(\i-1)*3.9},7.9) {};}
  \foreach \j in {1,2,3,4}{\node[stok] at ({12.2+(\j-1)*3.9},2.4) {};}
  \node[eq,text=pcPink,font=\small] at (31.3,2.4) {$\neq$};
\end{scope}
\begin{scope}[shift={(140.4,-3)}]
  \node[lab,anchor=west] at (0,11.4) {train};
  \node[lab,anchor=west] at (0,6.1) {deploy};
  \foreach \i/\s in {1/stok,2/stok,3/gone,4/stok,5/gone,6/stok}{\node[\s,minimum width=3.3mm,minimum height=3.3mm] at ({12.2+(\i-1)*3.9},11.4) {};}
  \foreach \j in {1,2,3,4}{\node[stok] at ({12.2+(\j-1)*3.9},6.1) {};}
  \node[eq,text=pcGreen,font=\small] at (31.3,6.1) {$=$};
  \node[lab,anchor=west] at (0,1.1) {at binary $p$ (Prop.~\ref{prop:block})};
\end{scope}

\node[lab] at (80.4,18) {expected compacted window;\\[-1pt]$\partial\mathcal L/\partial p_3$ compares keep and drop};
\node[lab] at (120.1,18) {a dropped token does\\[-1pt]not update the state};
\node[lab] at (157.4,18) {training runs the\\[-1pt]deployed network};
\end{scope}
\end{tikzpicture}%
\caption{\textbf{Removing a token changes the convolution window.}
Top: zero masking leaves the token's slot in place. Bottom: \ours averages
the keep and drop windows, and its gradient compares them (\cref{eq:pgrad}).
Step gating skips dropped updates, so the block matches physical compaction
for binary decisions (\cref{prop:block}).}
\label{fig:teaser}
\end{figure*}
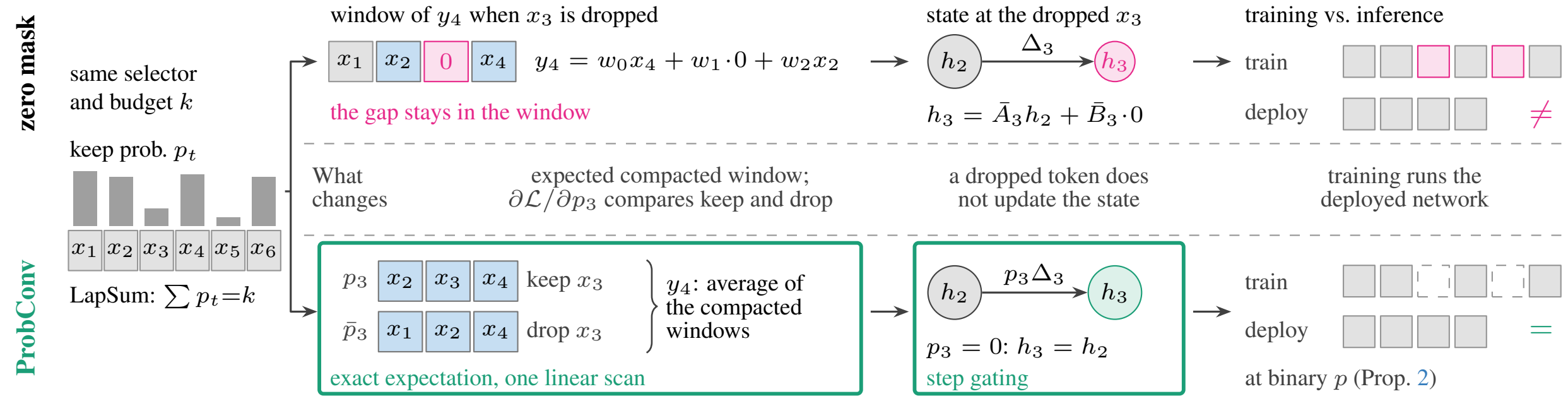

Removing a token in Mamba changes two computations. The convolutions read
local windows along the token sequence, so closing a gap changes which
tokens fall at each offset. The selective scan also skips the update that
would have occurred at the removed position~\cite{gu2023mamba,dao2024mamba2}.
Zero masking illustrates the difference: it leaves a slot in the convolution
window and can still update the recurrent state (\cref{fig:teaser}).

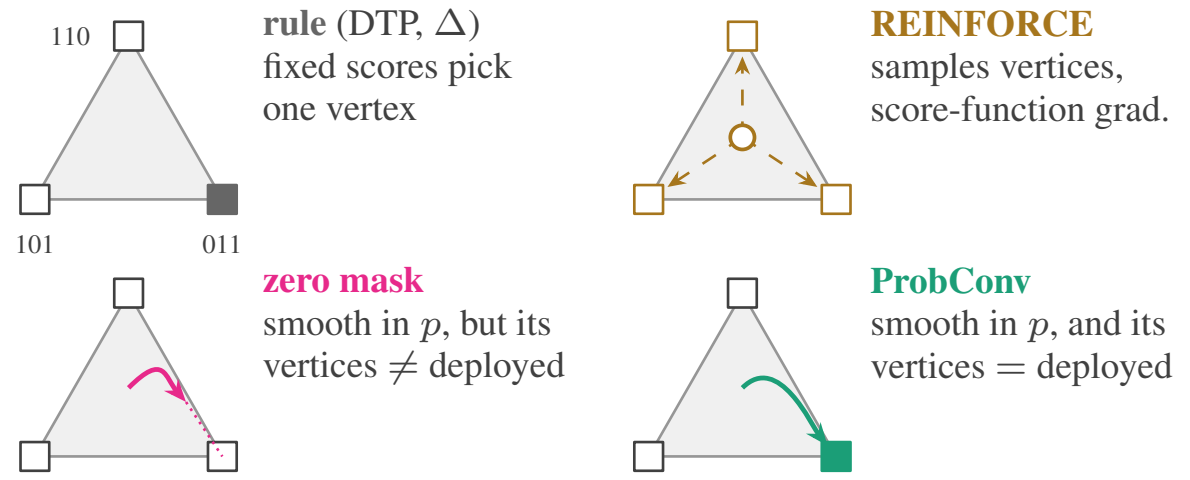
\begin{figure}[t]
\centering
\definecolor{pcGreen}{HTML}{1B9E77}%
\definecolor{pcPink}{HTML}{E7298A}%
\definecolor{pcBrown}{HTML}{A6761D}%
\definecolor{pcGrey}{HTML}{666666}%
\definecolor{pcInk}{RGB}{64,64,64}%
\definecolor{pcLine}{RGB}{150,150,150}%
\begin{tikzpicture}[x=1mm,y=1mm,font=\scriptsize,line join=round,
  vtx/.style={draw=pcInk,fill=white,minimum size=1.9mm,inner sep=0pt,line width=0.5pt},
  dsc/.style={font=\scriptsize,text=pcInk,inner sep=0pt,anchor=north west,align=left},
  traj/.style={line width=0.9pt,-{Stealth[length=1.5mm,width=1.2mm]}},
]
\newcommand{\cell}[3]{%
  \coordinate (A#1) at (#2+7.2,#3+12.4);
  \coordinate (B#1) at (#2+1.0,#3+1.6);
  \coordinate (C#1) at (#2+13.4,#3+1.6);
  \fill[pcLine!14] (A#1) -- (B#1) -- (C#1) -- cycle;
  \draw[pcLine,line width=0.5pt] (A#1) -- (B#1) -- (C#1) -- cycle;
}
\cell{r}{0}{17}
\node[vtx] at (Ar) {}; \node[vtx] at (Br) {};
\node[vtx,fill=pcGrey,draw=pcGrey] at (Cr) {};
\node[font=\tiny,text=pcInk,anchor=east] at ($(Ar)+(-1.3,0)$) {110};
\node[font=\tiny,text=pcInk,anchor=north] at ($(Br)+(0,-1.1)$) {101};
\node[font=\tiny,text=pcInk,anchor=north] at ($(Cr)+(0,-1.1)$) {011};
\node[dsc] at (16.0,31.2) {\textbf{\color{pcGrey}rule} (DTP, $\Delta$)\\fixed scores pick\\one vertex};
\cell{f}{40.6}{17}
\coordinate (Pf) at ($(Bf)!0.5!(Cf)!0.38!(Af)$);
\foreach \v in {Af,Bf,Cf}{\draw[pcBrown,line width=0.5pt,dashed,-{Stealth[length=1.2mm,width=1mm]}] (Pf) -- ($(Pf)!0.8!(\v)$);}
\node[circle,draw=pcBrown,fill=white,minimum size=1.6mm,inner sep=0pt,line width=0.7pt] at (Pf) {};
\node[vtx,draw=pcBrown] at (Af) {}; \node[vtx,draw=pcBrown] at (Bf) {}; \node[vtx,draw=pcBrown] at (Cf) {};
\node[dsc] at (56.2,31.2) {\textbf{\color{pcBrown}REINFORCE}\\samples vertices,\\score-function grad.};
\cell{z}{0}{0}
\coordinate (Sz) at ($(Bz)!0.5!(Cz)!0.42!(Az)$);
\coordinate (Ez) at ($(Cz)+(-2.4,3.6)$);
\draw[traj,pcPink] (Sz) .. controls +(2.4,2.4) and +(-1.4,1.6) .. (Ez);
\node[vtx] at (Az) {}; \node[vtx] at (Bz) {}; \node[vtx] at (Cz) {};
\draw[pcPink,line width=0.5pt,densely dotted] (Ez) -- (Cz);
\node[dsc] at (16.0,14.2) {\textbf{\color{pcPink}zero mask}\\smooth in $p$, but its\\vertices $\neq$ deployed};
\cell{p}{40.6}{0}
\coordinate (Sp) at ($(Bp)!0.5!(Cp)!0.42!(Ap)$);
\draw[traj,pcGreen,shorten >=0.9mm] (Sp) .. controls +(2.4,2.4) and +(-1.6,2.0) .. (Cp);
\node[vtx] at (Ap) {}; \node[vtx] at (Bp) {};
\node[vtx,fill=pcGreen,draw=pcGreen] at (Cp) {};
\node[dsc] at (56.2,14.2) {\textbf{\color{pcGreen}\ours}\\smooth in $p$, and its\\vertices $=$ deployed};
\end{tikzpicture}%
\caption{\textbf{Where each approach trains the selector.} Keep
probabilities with a fixed budget form a hypersimplex ($L{=}3$, $k{=}2$
shown); its vertices are binary masks. Rules choose a vertex and REINFORCE
samples vertices. A zero mask is differentiable in the keep probabilities
but computes a different network even at a vertex. \ours is differentiable
and, with step gating, recovers the compacted Mamba block at each vertex
(\cref{prop:block}).}
\label{fig:hypersimplex}
\end{figure}

Token selection methods differ in how they choose and train on masks
(\cref{fig:hypersimplex}). DTP uses an inference-time score without a learned
selector~\cite{park2026dtp}. DyVM moves sampled survivors into a prefix
during training and uses top-$k$ at inference~\cite{wu2025dyvm}. Its
training and inference masks can therefore differ. OCSampler uses
reinforcement learning (REINFORCE) on compacted clips, with a clip-level reward shared across frame
decisions~\cite{lin2022ocsampler,williams1992reinforce}. Straight-through
training with zero masks gives selector gradients through the zero-filled
sequence. Can we train the selector by differentiation, exactly recover
the compacted forward pass for binary masks, and keep the cost linear in
sequence length?

\ours tracks the expected contents of a convolution window as tokens are
kept or dropped. Its gradient compares the windows produced by the two
choices. Step gating lets a dropped position pass the recurrent state
through unchanged and gives the selector a second gradient path. At binary
keep values, the two operations together reproduce the compacted block
(\cref{fig:process}, top right).
The same construction supports learned token scores and successive
compaction stages (\cref{sec:method}).

\ours is strongest under strong compaction, where removing tokens changes
the neighborhoods of the remaining ones most. In Vim, Cobra and VideoMamba,
its advantage over matched training controls grows as the budget tightens
(\cref{fig:budget}). In Cobra, it also improves on DTP, FastV and DART adapted
to the same backbone at matched inference budgets~\cite{park2026dtp,chen2024fastv,wen2025dart},
and its lead over learned selectors persists at matched training cost. For X3D frame selection, \ours trained on sampled
masks matches or exceeds the mean mAP of retrained OCSampler with REINFORCE. In speech, \ours improves over matched zero-mask training, while
REINFORCE gives similar or lower word error rates. Compaction speeds up
batched inference in the measured vision setups. Controls that keep the
selector and budget fixed isolate the effect of the training computation,
and ablations show how the convolution gradient and step gating affect
accuracy across backbones. The supplement
covers additional video, genomic and long-context experiments.

\paragraph{Contributions.}
\begin{itemize}[leftmargin=*,itemsep=1pt,topsep=2pt,parsep=0pt]
\item \textbf{Expected compacted convolution.} We derive the exact expected
convolution under independent keep decisions. For a fixed kernel width and
channel count, its computational cost grows linearly with sequence length (\cref{sec:probconv}).
\item \textbf{Training through compaction.} With step gating, the Mamba
forward matches physical compaction for binary masks. Gradients through the
convolution and the scan train a budgeted selector (\cref{sec:gating,sec:training}).
\item \textbf{Evaluation.} We compare against pruning methods and matched
controls in image classification, vision--language tasks, video recognition,
frame selection and speech recognition. We measure training and inference
costs and ablate the convolution gradient and step gating (\cref{sec:experiments}).
\end{itemize}

\section{Learning through token compaction}
\label{sec:method}
\label{sec:setup}

Let $x_{1:L}\in\mathbb{R}^{L\times D}$ be the projected token sequence and
$m\in\{0,1\}^L$ a keep mask. Compaction gathers the retained indices
$i_1<\dots<i_n$ in order: $\Compact_m(x)=(x_{i_1},\dots,x_{i_n})$.
A Mamba block~\cite{gu2023mamba} applies a causal depthwise convolution with
$K$ taps $w_r\in\mathbb{R}^D$, bias $b$ and SiLU activation $\sigma$:
\begin{equation}
  v_t = \sigma\Big(b + \sum_{r=0}^{K-1} w_r \odot x_{t-r}\Big).
  \label{eq:conv}
\end{equation}
The selective scan then updates its state and reads out the result:
\begin{equation}
  h_t = \exp(\Delta_t A)\, h_{t-1} + \Delta_t B_t v_t,\qquad y_t = C_t h_t + D v_t ,
  \label{eq:scan}
\end{equation}
The selective parameters depend on the block's input. In Mamba,
$B_t$ and $C_t$ are projected from $v_t$.
Mamba-2 applies the convolution jointly to the projected $(x,B,C)$ and
uses a scalar $A$ per head~\cite{dao2024mamba2}. Compaction changes the
window in \cref{eq:conv} and the positions at which \cref{eq:scan} updates.

\subsection{The expected convolution window}
\label{sec:probconv}

Let $p=(p_1,\dots,p_L)$ be the selector's keep values and model the decisions as
independent draws $m_t\sim\mathrm{Bernoulli}(p_t)$. At a kept position $t$,
the compacted convolution reads $s^{(0)}_t=x_t$ and the $r$-th preceding
kept token $s^{(r)}_t$ at offset $r>0$. A missing predecessor contributes
zero. With $a^{(r)}_t$ the expected $r$-th most recent kept token through
position $t$, the window gives
\begin{equation}
\begin{aligned}
  \text{\ours}(x,p)_t &= \E\Big[\sum_{r=0}^{K-1} w_r\odot s^{(r)}_t \,\Big|\, m_t=1\Big]\\
  &= w_0\odot x_t + \sum_{r=1}^{K-1} w_r\odot a^{(r)}_{t-1},
\end{aligned}
\label{eq:probconv}
\end{equation}
Independence lets us compute the preceding window without conditioning on
the decision at $t$. The same formula extends to $p_t=0$ by continuity;
the compacted convolution requires a readout only at kept positions.

\begin{proposition}
\label{prop:recurrence}
With $a^{(r)}_0=0$, the expectations obey the linear recurrence
\begin{align}
  a^{(1)}_t &= p_t\, x_t + (1-p_t)\, a^{(1)}_{t-1}, \nonumber\\
  a^{(r)}_t &= p_t\, a^{(r-1)}_{t-1} + (1-p_t)\, a^{(r)}_{t-1}, \quad 2\le r\le K-1 .
  \label{eq:recurrence}
\end{align}
\end{proposition}

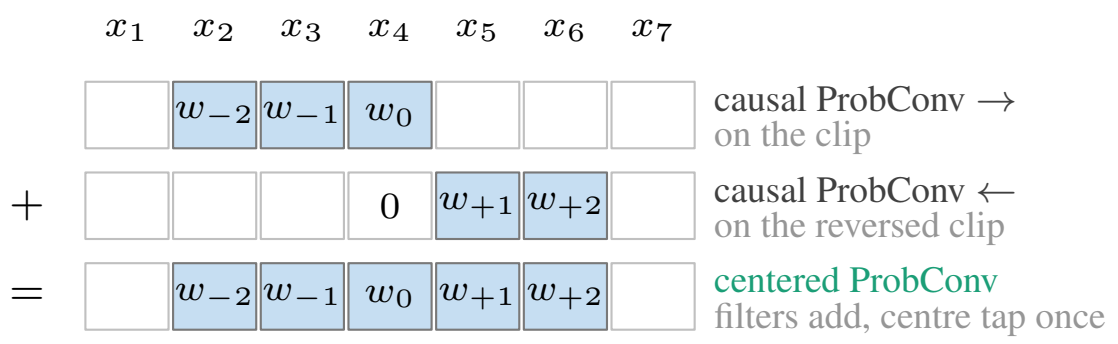
\begin{figure}[t]
\centering
\definecolor{pcGreen}{HTML}{1B9E77}%
\definecolor{pcInk}{RGB}{64,64,64}%
\definecolor{pcTok}{RGB}{226,226,226}%
\definecolor{pcLine}{RGB}{150,150,150}%
\definecolor{pcWin}{RGB}{198,222,242}%
\begin{tikzpicture}[x=1mm,y=1mm,font=\scriptsize,line join=round,
  tok/.style={draw=pcLine,fill=pcTok,minimum width=5.5mm,minimum height=4.4mm,inner sep=0pt,font=\scriptsize,line width=0.4pt},
  win/.style={tok,fill=pcWin,draw=pcInk!70},
  nil/.style={tok,fill=white,draw=pcLine!60},
  lab/.style={font=\scriptsize,text=pcInk,inner sep=0.5pt,anchor=west,align=left},
  op/.style={font=\footnotesize,inner sep=0pt},
]
\foreach \i in {1,...,7}{\node[font=\scriptsize] at ({2+\i*5.8},22.4) {$x_\i$};}
\foreach \i/\w/\s in {1//nil,2/w_{-2}/win,3/w_{-1}/win,4/w_0/win,5//nil,6//nil,7//nil}{\node[\s] at ({2+\i*5.8},16.8) {$\w$};}
\node[lab] at (46.4,17.9) {causal \ours{} $\rightarrow$};
\node[lab,text=pcLine] at (46.4,15.3) {on the clip};
\node[op] at (1.2,10.8) {$+$};
\foreach \i/\w/\s in {1//nil,2//nil,3//nil,4/0/nil,5/w_{+1}/win,6/w_{+2}/win,7//nil}{\node[\s] at ({2+\i*5.8},10.8) {$\w$};}
\node[lab] at (46.4,11.9) {causal \ours{} $\leftarrow$};
\node[lab,text=pcLine] at (46.4,9.3) {on the reversed clip};
\node[op] at (1.2,4.8) {$=$};
\foreach \i/\w/\s in {1//nil,2/w_{-2}/win,3/w_{-1}/win,4/w_0/win,5/w_{+1}/win,6/w_{+2}/win,7//nil}{\node[\s] at ({2+\i*5.8},4.8) {$\w$};}
\node[lab,text=pcGreen] at (46.4,5.9) {centered \ours};
\node[lab,text=pcLine] at (46.4,3.3) {filters add, centre tap once};
\end{tikzpicture}%
\caption{\textbf{A centered convolution is two causal ones.} Taps of a
convolution with two taps on each side of the output frame $x_4$: a causal
pass over the clip and a causal pass over the reversed clip, whose centre tap
is zero, add up to the centered filter. With \ours in both passes the sum is
the expected centered convolution of the compacted clip, because the kept
frames before and after a kept frame are independent
(\cref{eq:supp_centered,fig:supp_centered}).}
\label{fig:centered}
\end{figure}

Keeping a token inserts it at the front of the window and shifts older
entries back one slot. Dropping it leaves the window in place. Averaging
these updates gives \cref{eq:recurrence}. Keeping every token ($p_t=1$ for
all $t$) recovers ordinary convolution. With any binary mask, the window
holds the actual kept predecessors, so the output, with bias and
activation, equals the compacted convolution at every kept position. At fractional $p$, the
expectation in \cref{eq:probconv} is exact before activation. A centered
convolution, as in X3D-S, is a causal pass plus a causal pass over the
reversed sequence, so \ours covers it as well (\cref{fig:centered}). The
derivation is in \cref{sec:supp_derivation}.

Stack the window states as $a_t$, let $S$ shift them down by one slot, and
let $e_1 x_t$ insert $x_t$ into the first slot. For the adjoint
$\bar a_t=\partial\mathcal{L}/\partial a_t$, the probability gradient is
\begin{equation}
  \frac{\partial \mathcal{L}}{\partial p_t}
  = \big\langle \bar a_t,\; (S a_{t-1} + e_1 x_t) - a_{t-1}\big\rangle ,
  \label{eq:pgrad}
\end{equation}
The two terms inside the state difference are the windows obtained by
keeping and dropping $t$. The adjoint weights their difference by its
effect on the loss.

\begin{figure*}[t]
\centering
\includegraphics[width=\textwidth]{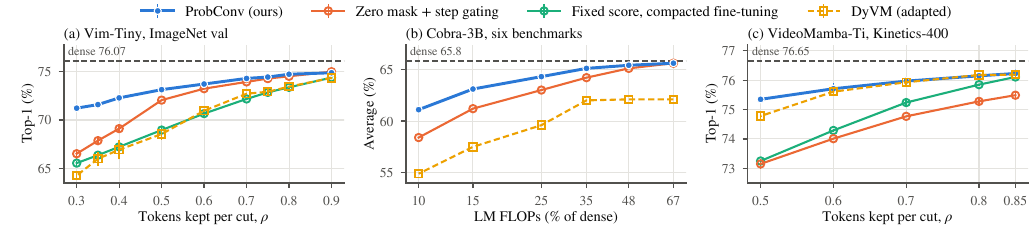}
\caption{\textbf{The lead over zero-mask training grows as budgets tighten.}
Accuracy of (a)~Vim-Tiny, (b)~Cobra-3B and (c)~VideoMamba-Ti at each
budget (compute per budget: \cref{tab:supp_cost}), for the selector trained with \ours or
with zero masking and step gating, and for a fixed score with compacted
fine-tuning (in Vim, the channel mean of the backward scan output, the
unclipped token importance of ToP-ViM~\cite{zhan2024top}; in VideoMamba, the
token norm; none in Cobra, whose backbone stays frozen); every arm runs
physically compacted at inference. DyVM (adapted): in Vim and VideoMamba, our
selector, objective and schedule with DyVM's training forward; in Cobra, DyVM's
predictor and objective (\cref{sec:vim,sec:vision}). Mean over three seeds
(\cref{tab:supp_vim_val,tab:vision,tab:supp_cobra_sweep,tab:videomamba}).}
\label{fig:budget}
\end{figure*}

\subsection{Skipping recurrent updates}
\label{sec:gating}

We gate the step in \cref{eq:scan} by replacing $\Delta_t$ with
$p_t\Delta_t$. If $p_t=0$, the decay is one and the write is zero.
If $p_t=1$, the scan uses the original step. Differentiating this gate
gives the selector a second gradient path. With fractional keep values,
step gating is a relaxation of the state update.

\begin{proposition}
\label{prop:block}
For binary $p$, a Mamba or Mamba-2 block that uses \ours in place of its
convolution and the gated step $p_t\Delta_t$ produces, at every kept position,
the output of the same block applied to $\Compact_m(x)$.
\end{proposition}

Bidirectional blocks apply the same construction in both directions and
reverse the keep values with the backward sequence. Block equivalence
extends through a stack with several cuts if each mask is nested in the
previous one, operations between blocks agree on retained tokens, and
positional information is treated as in deployment
(\cref{sec:supp_block_proof}).

\subsection{Training at a token budget}
\label{sec:training}

A selector assigns scores $s_t$ and uses $m=\TopK_k(s)$ at inference.
During training, LapSum~\cite{struski2025lapsum} yields differentiable
keep values $p=\mathrm{LapSum}_k(s/\tau)$ with $\sum_t p_t=k$. At each cut
we can use the straight-through values $\tilde m=m+p-\sg(p)$, with
$\sg$ the stop-gradient operator. The forward uses the hard mask at the
specified budget. The backward differentiates the relaxed convolution
and scan at that mask and carries both gradients through $p$ to the
selector. Across cuts, each new hard decision acts on the previously
retained tokens, and the corresponding relaxed keep values multiply
across stages. Always-retained tokens have keep value one and are outside
the pruning budget (\cref{sec:supp_protocols}).

A soft forward feeds the operators fractional keep values. In Cobra, an
annealed training schedule encourages these values towards binary
decisions, followed by hard top-$k$ at inference
(\cref{sec:supp_cobra}). The sum constraint sets the expected number of
kept tokens under the independent model of \cref{sec:probconv}.
The hard top-$k$ mask enforces the realized budget. For frame selection,
we can use a binary mask sampled from the policy in the forward pass.
The backward differentiates the relaxed operator at that mask and passes
the gradient through the policy's marginal keep probabilities (\cref{sec:x3d_main}).

\paragraph{Numerical equivalence and cost.}
Under fixed binary masks, Mamba-1 and Mamba-2 block outputs agree with
physical compaction to within $3.2{\cdot}10^{-6}$ relative error. Cobra's
logits agree to $9.7{\cdot}10^{-7}$ (\cref{tab:exactness}). These checks
compare the same keep decisions. Each update in \cref{eq:recurrence}
takes $O(KD)$ work, so the recurrent evaluation takes $O(LKD)$ work.
The affine updates also admit prefix and reverse scans~\cite{blelloch1990prefix}
(\cref{sec:supp_derivation,sec:supp_kernel}). At inference, the model
gathers the selected tokens and runs its ordinary blocks on the shorter
sequence. \Cref{sec:cost} reports measured costs separately for the
convolution and the backbone.

\section{Experiments}
\label{sec:experiments}

We evaluate patch compaction in Vim-Tiny on ImageNet-1k, visual-token
selection in Cobra-3B and spatiotemporal token compaction in VideoMamba-Ti
on Kinetics-400~\cite{zhu2024vim,russakovsky2015imagenet,zhao2025cobra,li2024videomamba,kay2017kinetics}.
We also test frame selection for X3D-S and frame dropping in a Mamba speech
encoder. At inference, each pruned variant processes the physically
compacted sequence. We compare published methods adapted to each setting
and controls that share the selector and budget. In Vim and VideoMamba each
cut keeps a fraction $\rho$ of the remaining patches, as in DynamicViT and
DyVM~\cite{rao2021dynamicvit,wu2025dyvm}; Cobra keeps a fraction of the
language-model (LM) FLOPs, as in DTP~\cite{park2026dtp}; X3D and speech keep
frames (\cref{sec:supp_protocols}).

\subsection{Image classification with Vim}
\label{sec:vim}

\paragraph{Matched selector.}
We train a selector at two cuts in Vim-Tiny, as in ToP-ViM~\cite{zhan2024top},
and fine-tune for three epochs on ImageNet-1k.
On the official validation set, at $\rho=0.3$ (18 of 196 patches remain),
\ours achieves $71.22\%$ top-1 and the same selector trained
with zero-mask convolution and step gating achieves $66.52\%$
($+4.69\pm0.56$ points, seed-paired). At ToP-ViM's $\rho=0.8$ they score
$74.72\%$ and $74.51\%$. Across the nine budgets the lead grows from a
tie at $\rho=0.9$ to $4.7$ points at the tightest (\cref{fig:budget}a).
DyVM~\cite{wu2025dyvm} combines several mechanisms. A ratio loss holds its
Gumbel-sampled predictor near the budget, token distillation from the dense
model recovers accuracy, and block skipping saves further compute. Its
training forward moves the kept tokens ahead of the removed ones, so the scan
reads them as at inference. To isolate how DyVM trains token selection, we keep
our selector, objective and schedule and take only this forward. It scores
$0.5$--$7.0$ points below \ours, and the gap grows as budgets tighten
(\cref{tab:vim,tab:supp_vim_val}).

\begin{figure}[t]
\centering
\includegraphics[width=\linewidth]{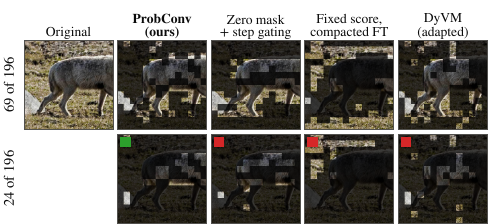}
\caption{\textbf{With 24 patches left, only \ours classifies the coyote
correctly.} Patches kept by the Vim-Tiny networks of \cref{fig:budget}a at
$\rho=0.35$ (69 and 24 of 196 after blocks 10 and 20, seed 0); dimmed patches
are dropped. Green and red squares mark correct and wrong predictions; the
other arms predict \emph{timber wolf}. The fixed score keeps mostly background,
and DyVM (adapted) keeps more background than \ours. The zero mask also keeps
patches on the coyote, but it trains on a forward that differs from the
deployed network (\cref{tab:main_signal}). More images:
\cref{fig:supp_vim_masks}.}
\label{fig:vim_masks}
\end{figure}

\Cref{fig:vim_masks} shows the patches that each arm keeps on one validation
image.

The per-image search cited in \cref{sec:intro} illustrates what a frozen
backbone can preserve on selected inputs. It uses a dense target for each
image, whereas the learned selector makes its choice from the current
input at inference. We report the search protocol and harder examples
in \cref{sec:supp_vim}.

\subsection{Visual-token selection in Cobra}
\label{sec:vision}

We train a visual-token selector before the first language-model block of
frozen Cobra-3B. Baselines include layer-0 $\Delta$, adaptations of DTP,
FastV and DART, and learned selectors on the same
backbone~\cite{park2026dtp,chen2024fastv,wen2025dart}. As with DTP, FastV and
DART, DyVM enters Cobra as a published method: its predictor trains with its
own objective, alone and with its rearranged forward. The forward gives no gain
at either budget, so \cref{fig:budget}b shows the predictor alone. At $10\%$ of LM FLOPs,
\ours scores $61.1$ against $58.6$ for layer-0 $\Delta$, the strongest
training-free rule at that budget (\cref{tab:main_cobra}). The gap shrinks
at the milder budget. We use DTP's best reachable sites at each budget.
DTP leads only at $67\%$ of LM FLOPs in the sweep, where its original
pruning sites are reachable.

Against zero masking with step gating, \ours ties at $67\%$ of LM FLOPs and
leads by 0.3, 0.9, 1.3, 1.9 and 2.7 points at 48, 35, 25, 15 and $10\%$
(\cref{fig:budget}b). The gains on GQA, VQAv2 and POPE
are significant in every seed pair at both main budgets (per-question
tests, Holm correction across benchmarks). VSR and VizWiz stay near their
floors across rules and show no consistent gain.

At $10\%$ of LM FLOPs, \ours leads when learned baselines receive enough
updates to match either its PyTorch GPU-hours or its CUDA training time.
With the same LoRA setup~\cite{hu2022lora}, \ours scores $61.7$ against
$60.1$ for layer-0 $\Delta$, the strongest competing rule in that setting.
\Cref{tab:vision} gives the full comparisons. \Cref{sec:supp_cobra} also
shows the tokens kept by each rule on 19 fixed examples.

\begin{table}[t]
\caption{\textbf{\ours leads at the tight budget, also at equal training time.}
Cobra-3B mean score (\%) on GQA, VQAv2, TextVQA, POPE, VSR and
VizWiz~\cite{hudson2019gqa,goyal2017vqav2,singh2019textvqa,li2023pope,liu2023vsr,gurari2018vizwiz}.
Percentages in the headers denote remaining LM FLOPs (dense: 65.8\% mean
score). The upper block uses training-free rules. Lower rows train the
selector on a frozen backbone (three seeds, mean $\pm$ standard deviation).
DyVM predictor: DyVM's predictor and objective, without or with its
rearranged forward. Equal time gives learned baselines $1.4$--$1.8\times$ as many updates as
our CUDA run. Training-free rules have no training
time to match. LoRA and GPU-hour comparisons: \cref{tab:vision}.}
\label{tab:main_cobra}
\centering\footnotesize
\setlength{\tabcolsep}{3.5pt}
\begin{tabular}{@{}lccc@{}}
\toprule
& \multicolumn{2}{c}{2,000 steps} & Equal time \\
\cmidrule(lr){2-3}\cmidrule(l){4-4}
Selection rule & 10\% & 25\% & 10\% \\
\midrule
Uniform & 57.4 & 62.0 & -- \\
Layer-0 $\Delta$ & 58.6 & 63.0 & -- \\
DTP & 57.5 & 61.4 & -- \\
FastV & 56.2 & 62.3 & -- \\
DART & 54.9 & 58.7 & -- \\
\midrule
DyVM predictor & 54.9\sd{4.6} & 59.6\sd{3.6} & 55.4\sd{1.7} \\
DyVM predictor + forward & 54.3\sd{0.9} & 59.0\sd{3.4} & 54.7\sd{0.9} \\
Zero mask & 56.8\sd{2.8} & 62.8\sd{0.5} & 57.9\sd{0.3} \\
Zero mask + step gating & 58.4\sd{0.2} & 63.0\sd{0.4} & 58.4\sd{0.5} \\
\ours (ours) & \textbf{61.1}\sd{0.2} & \textbf{64.3}\sd{0.1} & \textbf{61.1}\sd{0.1} \\
\bottomrule
\end{tabular}
\end{table}

\subsection{Video recognition with VideoMamba}
\label{sec:videomamba}

\begin{figure*}[t]
\centering
\begin{minipage}[c]{0.655\textwidth}
\includegraphics[width=\linewidth]{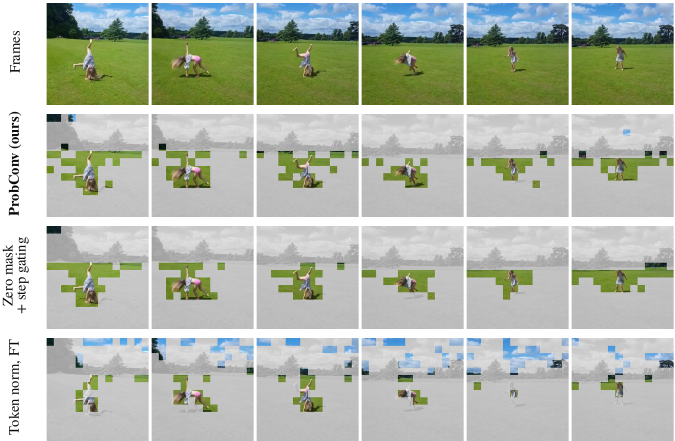}
\end{minipage}\hfill
\begin{minipage}[c]{0.325\textwidth}
\caption{\textbf{\ours keeps the tokens around the moving person.}
VideoMamba-Ti at $\rho=0.5$ on a Kinetics-400 validation clip
(\emph{cartwheeling}, correct for all three rules), six of its eight
frames. Colored tokens reach the last block; gray ones are dropped. The
cuts after blocks 6, 12 and 18 drop 784, 392 and 196 of the 1,568 patch
tokens. The zero mask often keeps neighboring tokens along the scan, and
the token norm keeps many sky tokens. More clips:
\cref{fig:supp_videomamba_tokens}.}
\label{fig:videomamba_tokens}
\end{minipage}
\end{figure*}

We fine-tune the released VideoMamba-Ti checkpoint for three epochs on
Kinetics-400, compacting tokens at three cuts. At the mildest budget,
\ours and compacted fine-tuning with token-norm selection give similar
accuracy. At $\rho=0.5$, \ours retains $75.35\%$ top-1,
$1.30$ points below the dense model and $2.10$ points above fine-tuning
with the fixed norm rule (\cref{tab:main_video}). The advantage over
the matched learned zero-mask control also increases across these budgets
(\cref{fig:budget}c). As in speech, the zero-mask selector keeps runs of
consecutive tokens along the scan (\cref{fig:videomamba_tokens}). On HMDB51, both methods start from a dense
checkpoint fine-tuned on each split. Averaged over the three splits, \ours
leads the matched zero-mask control by 1.06 and 1.08 points at $\rho=0.5$
and $0.7$, in 17 of 18 seed pairs (\cref{sec:supp_videomamba}).

As published context, R-MeeTo reports $76.5\%$ top-1 at $18\%$ fewer
FLOPs, and DyVM reports $74.3\%$ at $38\%$ fewer FLOPs~\cite{shi2024rmeeto,wu2025dyvm}. Both use 30 epochs and their own FLOP
conventions, and DyVM also skips blocks. The full table gives both cost measures
and the HMDB51 results (\cref{tab:videomamba}).

\begin{table}[t]
\caption{\textbf{The lead over the matched control grows as $\rho$ decreases.}
VideoMamba-Ti top-1 (\%) on Kinetics-400 with $4{\times}3$ views and
physical compaction at inference (dense: 76.65\%). Fine-tuned rows use
three epochs (mean $\pm$ standard deviation over three seeds).
Token-norm fine-tuning compacts the sequence during training.
DyVM forward: our selector trained with DyVM's training forward
(\cref{sec:vim}). Further results: \cref{tab:videomamba}.}
\label{tab:main_video}
\centering\footnotesize
\setlength{\tabcolsep}{3pt}
\begin{tabular}{@{}lccc@{}}
\toprule
& \multicolumn{3}{c}{Tokens kept per cut} \\
\cmidrule(l){2-4}
Selection rule & $\rho=0.5$ & $\rho=0.7$ & $\rho=0.85$ \\
\midrule
Token norm, no FT & 65.81 & 73.75 & 75.87 \\
Token norm, FT & 73.25\sd{0.07} & 75.24\sd{0.11} & 76.11\sd{0.08} \\
Zero mask + step gating & 73.15\sd{0.17} & 74.77\sd{0.02} & 75.49\sd{0.11} \\
DyVM forward & 74.78\sd{0.13} & 75.93\sd{0.11} & 76.19\sd{0.12} \\
\ours (ours) & \textbf{75.35}\sd{0.04} & \textbf{75.97}\sd{0.13} & \textbf{76.23}\sd{0.14} \\
\bottomrule
\end{tabular}
\end{table}

\begin{table}[t]
\caption{\textbf{Zero-filled clips fall below uniform sampling.}
ActivityNet-v1.3 validation mAP (\%) with a frozen X3D-S per budget.
Learned policies share the skim network architecture and head and
deploy their top $N$ (mean $\pm$ standard deviation over three seeds).
OCSampler reproduction: its REINFORCE training, reimplemented in our code.
The all-frame row feeds all ten candidates to the classifier for that
column. Bold: the best selection in each column. The released OCSampler policy was trained for six frames. We
use its top $N$ with our classifier. Full results: \cref{tab:video_x3d}.}
\label{tab:main_x3d}
\centering\footnotesize
\setlength{\tabcolsep}{3pt}
\begin{tabular}{@{}lccc@{}}
\toprule
& \multicolumn{3}{c}{Frames kept of 10} \\
\cmidrule(l){2-4}
Frame selection & 2 & 4 & 6 \\
\midrule
All 10 frames & 73.69 & 79.36 & 80.20 \\
Uniform & 61.78 & 72.22 & 75.59 \\
OCSampler, released & 68.99 & 75.98 & \textbf{79.09} \\
\midrule
OCSampler reproduction & 69.45\sd{0.13} & 75.59\sd{0.31} & 78.74\sd{0.27} \\
Zero mask, top-$N$ & 60.53\sd{0.92} & 71.87\sd{0.63} & 77.07\sd{0.32} \\
\ours, top-$N$ (ours) & 67.38\sd{0.09} & 74.60\sd{0.27} & 77.57\sd{0.10} \\
\ours, sampled (ours) & \textbf{69.89}\sd{0.29} & \textbf{76.06}\sd{0.21} & 78.79\sd{0.09} \\
\bottomrule
\end{tabular}
\end{table}

\begin{figure}[t]
\centering
\includegraphics[width=\linewidth]{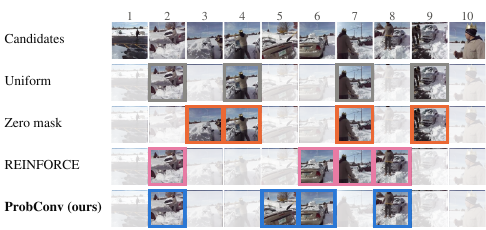}
\caption{\textbf{\ours keeps frames like REINFORCE.} Four of ten candidates kept by each
rule (framed) on one ActivityNet-v1.3 validation video, \emph{Removing ice
from car}, which every rule classifies correctly. \ours and REINFORCE share
three frames, the zero-mask policy at most one with either. The video is fixed by a rule
that ignores the predictions: the most motion (MGSampler's measure) among the
single-label videos on which the four rules keep different frames.}
\label{fig:x3d_frames}
\end{figure}

\subsection{Frame selection for a video CNN}
\label{sec:x3d_main}

We follow OCSampler on ActivityNet-v1.3 and keep $N$ of ten candidate
frames for X3D-S~\cite{lin2022ocsampler,feichtenhofer2020x3d,caba2015activitynet}.
At each budget, learned selectors share the skim architecture, policy
head and frozen classifier. Selection compacts the clip, so the temporal
convolutions of X3D-S read only the kept frames. \ours replaces each of
them with its centered variant (\cref{fig:centered}). At inference, each
learned policy keeps its top-scoring frames.

Zero-mask training scores below uniform selection at two and four frames
(\cref{tab:main_x3d}). \ours trained on policy-sampled masks gains $0.44$
mAP points over the OCSampler reproduction at two frames and leads in all three
seeds. Mean gains at four and six frames are $0.47$ and $0.05$ points,
with mixed results across seeds. Deterministic top-$N$ training trails
the reproduction in every seed.
At six frames, the released OCSampler sampler leads all our trained rules.
\ours keeps nearly the same frames as the reproduction: per validation video, the
two share 3.4 of four frames and the top-1 class in 89\% of videos, as
much as two of its seeds. The zero-mask policy shares 1.8 frames with
either, close to random selection
(\cref{fig:x3d_frames,tab:supp_x3d_selection}). Estimator variants and
joint training are in \cref{sec:supp_video_frames}.

\subsection{Frame dropping in a speech encoder}
\label{sec:asr_main}

We train Mamba-based HuBERT with character CTC on LibriSpeech
train-clean-100~\cite{lin2026mambahubert,graves2006ctc,panayotov2015librispeech}.
After layer 6, the last six Mamba layers process only the kept half of
the frames. Removing frames changes which neighbors the temporal
convolutions read. Before final CTC decoding, we restore dropped frames
using their layer-6 representations~\cite{zhu2024skipformer}. We train causal and
bidirectional encoders from scratch and also adapt a dense causal encoder.
The learned selector refines intermediate CTC scores~\cite{lee2021interctc}.
Baselines include uniform subsampling and Soundwave's CTC-peak
rule~\cite{zhang2025soundwave}.

\begin{figure*}[t]
\centering
\includegraphics[width=\textwidth]{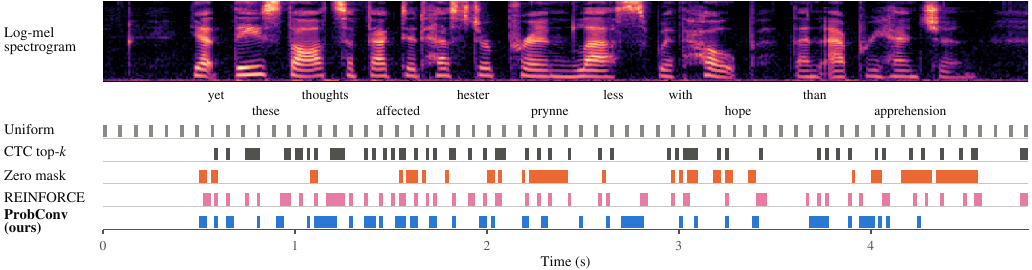}
\caption{\textbf{The zero-mask selector misses character emissions.}
Frames kept by each rule on one LibriSpeech test-clean utterance, below its
log-mel spectrogram and its words, when subword units are decoded from a
quarter of the frames without recovery. The
zero-mask selector keeps long runs of frames. Statistics over 400 utterances:
\cref{fig:asr_patterns}.}
\label{fig:asr_patterns_main}
\end{figure*}

With frame recovery, zero-mask training raises WER by 7.0--19.7 points
over CTC top-$k$. The seed-paired differences between \ours and CTC
top-$k$ range from $-0.3$ to $+0.8$ WER points across the encoders.
\ours beats uniform subsampling in each encoder. REINFORCE also trains on
compacted sequences and gives similar or slightly lower WER
(\cref{tab:main_asr,sec:speech}). Without recovery, uniform subsampling
gives the lowest subword WER. In the subword example, the zero-mask
selector keeps long runs of frames and misses many character emissions.
The selectors trained on compacted sequences keep more frames near
emissions (\cref{fig:asr_patterns_main}).
Frame dropping saves no encoder time here because the convolutional front
end dominates runtime (\cref{sec:speech}).

\begin{table}[t]
\caption{\textbf{\ours reduces WER relative to zero-mask training.}
LibriSpeech test-clean WER (\%, lower is better) for Mamba-based HuBERT
with greedy character CTC and no language model. Half the frames survive
layer 6. The final CTC head sees the restored sequence. Learned rows share
selector architecture, initialization and budget. Mean $\pm$ standard
deviation over five seeds for causal scratch and three for the other
columns. Bold: the lowest WER among the selection rules. Full results: \cref{sec:speech}.}
\label{tab:main_asr}
\centering\footnotesize
\setlength{\tabcolsep}{3pt}
\begin{tabular}{@{}lccc@{}}
\toprule
WER (\%) $\downarrow$ & \multicolumn{2}{c}{Causal encoder} & Bidir. \\
\cmidrule(lr){2-3}\cmidrule(l){4-4}
Frame selection & scratch & from dense & scratch \\
\midrule
All frames & 23.9\sd{0.7} & 22.6\sd{0.9} & 14.3\sd{0.3} \\
\midrule
Uniform & 30.3\sd{0.3} & 34.8\sd{1.0} & 20.2\sd{0.4} \\
CTC peaks (Soundwave) & 27.5\sd{1.2} & 28.6\sd{0.6} & 14.8\sd{0.7} \\
\midrule
Zero mask & 36.9\sd{8.3} & 48.2\sd{3.1} & 21.8\sd{2.6} \\
REINFORCE & \textbf{26.2}\sd{2.8} & \textbf{27.9}\sd{0.4} & \textbf{14.2}\sd{0.2} \\
\ours (ours) & 27.6\sd{0.9} & 28.8\sd{0.5} & 15.6\sd{2.0} \\
\bottomrule
\end{tabular}
\end{table}

\subsection{Training and inference cost}
\label{sec:cost}

Physical compaction speeds up batched inference in Vim and VideoMamba and
batched prefill in Cobra (\cref{tab:main_cost}). At batch size one, kernel
launches offset savings in Vim and VideoMamba. Cobra's prefill savings
depend on the GPU and protocol
(\cref{sec:supp_vim,sec:supp_videomamba,sec:supp_cobra}). Selectors with
the same budget deploy sequences of the same length.

Training evaluates the operators on the full-length sequence. For a Mamba-130M
block shape on RTX 4090, our fused operator takes $0.61$\,ms for the
convolution forward and backward, matching dense \texttt{causal-conv1d}
while also differentiating the keep probabilities. On RTX 4090, Vim
training throughput is $2$--$3\%$ lower than with zero masking. Training
costs vary by implementation and backbone
(\cref{sec:supp_kernel,sec:supp_vim,sec:supp_cobra}).

\begin{table}[t]
\caption{\textbf{Physical compaction speeds up batched inference.}
Tightest measured budgets: Vim, $\rho=0.35$ (fp32); VideoMamba,
$\rho=0.5$ (bf16); Cobra, 10\% of LM FLOPs (bf16).
Cobra measures prefill from the input image through the language model.
VideoMamba and Cobra use synthetic inputs. Speedups are relative to
the dense runs.}
\label{tab:main_cost}
\centering\footnotesize
\setlength{\tabcolsep}{3pt}
\begin{tabular}{@{}lllrrr@{}}
\toprule
Host & GPU & Unit & Dense & Compacted & Speedup \\
\midrule
Vim & RTX 4090 & ms & 45.9 & 28.5 & 1.61$\times$ \\
VideoMamba & RTX 4090 & clips/s & 555.3 & 1133.1 & 2.04$\times$ \\
Cobra prefill & H100 & img/s & 49.5 & 140.9 & 2.84$\times$ \\
\bottomrule
\end{tabular}
\end{table}

\subsection{Understanding the gains}
\label{sec:exact_summary}

In Vim at $\rho=0.35$, removing step gating costs 7.17
top-1 points, while stopping the convolution gradient barely changes
accuracy. In Cobra at 10\% of LM FLOPs, stopping that gradient costs
5.8 points with the same forward pass. Removing step gating has little
effect in Cobra (\cref{tab:main_signal}).

In Vim, both variants with a forward that matches compaction score within
0.1 points of their deployed counterparts. For zero masking with step
gating, soft training and \ours without step gating, validation with soft
keep values scores 1.98--5.60 points above compacted inference. On a small set
of POPE questions, \ours gradients rank measured single-token loss changes
better than zero-mask gradients (\cref{sec:supp_cobra_behaviour}).

Numerical agreement alone does not train a selector. Fixed-score compaction
in Vim runs the deployed computation during fine-tuning but gives no
gradient to a learned selector. REINFORCE on the compacted network scores
65.38 against 71.59 for \ours in Vim at $\rho=0.35$, below the zero mask,
and 60.1 against 61.1 in Cobra at 10\% of LM FLOPs. REINFORCE gives
similar or lower WER in speech. In X3D, \ours reaches REINFORCE's mean
mAP with sampled masks, while deterministic top-$N$ training trails it
(\cref{sec:asr_main,sec:x3d_main}).

\paragraph{Further hosts.} The supplement covers frame selection in
Uni-AdaFocus, dynamic chunking in H-Net and LDARNet, and long-context
token selection in ReMamba. \ours improves on the reproduced LDARNet
fine-tuning and ReMamba selector. H-Net and Uni-AdaFocus give mixed results across training signals
(\cref{sec:supp_video_adafocus,sec:supp_genomics,sec:supp_remamba}).

\begin{table}[t]
\caption{\textbf{The useful gradient path depends on the backbone.}
Vim-Tiny: ImageNet validation top-1 and the gap in points (top-1 with soft keep values
minus top-1 with physical compaction) at $\rho=0.35$. Cobra-3B:
six-benchmark mean at 10\% of LM FLOPs. Scores are percentages, mean
$\pm$ standard deviation over three seeds. Dashes: fixed-score compaction
and REINFORCE have no soft keep values, hence no gap; fixed-score compaction
fine-tunes the backbone, which Cobra keeps frozen. $^\dagger$In Cobra, \ours itself
trains with an annealed soft forward (straight-through: $60.6\pm0.2$).
Full ablations: \cref{tab:supp_vim_ablation,tab:supp_cobra_ablation}.}
\label{tab:main_signal}
\centering\footnotesize
\setlength{\tabcolsep}{3pt}
\begin{tabular}{@{}lccc@{}}
\toprule
& \multicolumn{2}{c}{Vim, $\rho=0.35$} & Cobra, 10\% \\
\cmidrule(lr){2-3}\cmidrule(l){4-4}
Training computation & top-1 & gap & average \\
\midrule
\ours (ours) & 71.59\sd{0.12} & $+0.08$\sd{0.05} & 61.1\sd{0.2} \\
Stop convolution gradient & 71.72\sd{0.16} & $-0.04$\sd{0.06} & 55.3\sd{2.4} \\
Remove step gating & 64.42\sd{0.35} & $+5.60$\sd{0.40} & 61.2\sd{0.1} \\
Soft forward & 70.40\sd{0.07} & $+2.46$\sd{0.07} & $=$\,\ours{}$^\dagger$ \\
Zero mask + step gating & 67.87\sd{0.32} & $+1.98$\sd{0.25} & 58.4\sd{0.2} \\
Fixed-score compaction & 66.37\sd{0.11} & -- & -- \\
REINFORCE, compacted & 65.38\sd{0.44} & -- & 60.1\sd{0.3} \\
\bottomrule
\end{tabular}
\end{table}

\section{Related Work}
\label{sec:related}

\paragraph{Token reduction.}
Vision transformers shorten sequences by pruning or merging
tokens~\cite{rao2021dynamicvit,chen2023diffrate,bolya2023tome}.
In vision state space models, ToP-ViM aligns hidden states for
retained tokens, DyVM trains a selector by rearranging retained
tokens, and R-MeeTo merges tokens~\cite{zhan2024top,wu2025dyvm,shi2024rmeeto}.
Other approaches combine pruning with merging~\cite{zhan2024torssm,huang2026tevmamba} or pool and fuse
tokens~\cite{shen2024fambav,park2025mame,wen2025stm,kapse2025fastvim}.
MTR reduces tokens without training, and STORM preserves the scan's
spatial structure~\cite{ma2025mtr,lv2026storm,liu2024vmamba}.
FastV, DART and VTW prune inside the language model, while DTP scores
visual tokens with Mamba's step size~\cite{chen2024fastv,wen2025dart,lin2025vtw,park2026dtp}.
\ours gives a selector access to the change in the convolution
window and, with step gating, recovers the compacted block for
binary decisions. As a control, we also train the same selector with
DyVM's rearranged forward (\cref{sec:vim}).

\paragraph{Learning discrete selectors.}
Straight-through estimators, continuous relaxations and score-function
estimators provide training signals for discrete choices~\cite{bengio2013ste,jang2017gumbel,maddison2017concrete,williams1992reinforce}.
Differentiable top-$k$ methods supply keep values at a specified
budget~\cite{xie2020difftopk,sander2023sparsetopk,struski2025lapsum}.
Given those keep values, \ours specifies the convolution as an
expectation over the changing neighborhoods. Our scan gate applies
the state-skipping principle of Skip RNN~\cite{campos2018skiprnn}
to Mamba's selective step.

\paragraph{Frame selection and dynamic chunking.}
OCSampler trains a frame selector with REINFORCE on compacted
clips~\cite{lin2022ocsampler}. On masks sampled from the same policy,
we use the compacted forward and differentiate the \ours relaxation.
It also supplies a control variate for REINFORCE
(\cref{sec:supp_video_frames}). The learned routers in H-Net and
LDARNet keep chunk boundaries for processing by the main
network~\cite{hwang2026hnet,ledneva2026ldarnet}. The expectation of
the causal smoothing dechunker follows the same rank-one recurrence
as the convolution window (\cref{prop:dechunk}).

\section{Conclusion}
\label{sec:conclusion}

Removing tokens changes what each convolution reads and which steps the
recurrent state takes. \ours follows this change exactly in expectation at
linear cost. With step gating it reproduces the compacted Mamba block for
binary decisions, so a selector learns on the network it will deploy. This
matters most at tight budgets. There the advantage over zero-mask training
widens in Vim, Cobra and VideoMamba. In Cobra, \ours also outperforms adapted
pruning methods and keeps its lead when learned selectors train for the same
time. X3D frame selection and speech frame dropping show the same advantage
over zero masking, and every gain keeps the inference savings of compaction.

\paragraph{Limitations.} The equivalence with compaction holds only for
binary decisions. \ours treats keep decisions as independent, although a
fixed budget couples them, and a joint model could sharpen the training
signal. Training keeps every token in the sequence, so it runs at the dense
length and the savings appear at deployment. Our image and video experiments
use short fine-tuning schedules, and the speed-up depends on batch size and
hardware (\cref{sec:supp_scope}).

\section*{LLM Usage Statement}
Anthropic Claude, used through Claude Code, and OpenAI Codex assisted with code implementation and debugging,
experiment planning, result analysis, and manuscript drafting and revision. OpenAI ChatGPT assisted with
editing and reviewing the manuscript. The research question and the idea of the probabilistic convolution
came from the authors, without the use of LLMs. The authors are responsible for the final manuscript,
software, experimental protocols and reported results.

{
    \small
    \bibliographystyle{ieeenat_fullname}
    \bibliography{references}
}

\clearpage
\maketitlesupplementary
\appendix
\appendixrefs

\noindent Removing tokens changes which inputs a convolution reads at each offset. We
examine when this change matters for learning a selector. In Vim, VideoMamba
and Cobra, the gap between \ours and zero-mask training is largest when few
tokens survive (\cref{tab:supp_vim_val,tab:videomamba,tab:supp_cobra_sweep}).
Two diagnostics describe the selections: in VideoMamba, zero-mask training
keeps runs of consecutive tokens along the scan, and in Cobra, the three
predecessors of a retained token can span several rows of the image grid
(\cref{fig:supp_cobra_window}). We also examine how the same operator compares
with established selection and routing methods for video, speech, genomic
sequences and long prompts.

For comparisons between \ours and zero-mask training, we use the same selector
architecture and inference budget. Comparisons with existing methods follow
the protocol specified for each host. In tables and figures, ``reported'' denotes the original
authors' result,
``reproduced'' denotes our run of their method, and
``(ours)'' marks \ours and
its variants. Where we use a published component or
adapt a pruning rule to another backbone, we name the component or adaptation
explicitly.

\section{Experimental protocols}
\label{sec:supp_protocols}

\paragraph{Budgets.} In Vim and VideoMamba, each cut retains a fraction $\rho$ of the available
patch tokens, as in DynamicViT and DyVM~\cite{rao2021dynamicvit,wu2025dyvm}.
After successive cuts, approximately $\rho$, $\rho^2$, \dots\ of the original
patches remain. We round each count to a whole token and always retain the
class token. VideoMamba uses DyVM's three cuts, after
blocks 6, 12 and 18, with $\rho\in\{0.5,0.6,0.7,0.8,0.85\}$ ($0.7$ is DyVM's
published budget). The short Vim protocol cuts twice, after blocks 10 and 20 as in
ToP-ViM \cite{zhan2024top}, with $\rho\in\{0.9,0.8,0.75,0.7,0.6,0.5,0.4,0.35,0.3\}$ ($0.8$
is ToP-ViM's setting; 147/111 keeps one patch more at the second cut than
$\rho=0.75$ gives). Following DTP~\cite{park2026dtp}, we express Cobra's budgets as fractions of
the dense language model's FLOPs: 67\%, 48\%, 35\%, 25\%, 15\% and 10\% for a
30-token text prompt. Each budget fixes the number of retained visual tokens.
For example, the 25\% and 10\% budgets retain 160 and 46 of 729 visual tokens,
respectively. The benchmarks use longer prompts, so the LM FLOP shares measured on them are
67.5, 48.7, 36.0, 26.1, 16.2 and 11.3\%.

\paragraph{Compute at each budget.} \Cref{tab:supp_cost} lists, for every
budget, the tokens kept after each cut, multiply-accumulates
counted analytically and FLOPs in the convention of the compared papers.
For Vim and VideoMamba, the analytic MACs include projections, convolutions,
selective scans in both directions, patch embedding and the head. They exclude
the selector. The Cobra count adds the language-model prefill with 30 text
tokens to the vision backbone and projector, whose cost is measured with
PyTorch's FLOP counter. The published FLOP figures for Vim and VideoMamba use
the R-MeeTo counter as used by DyVM, which counts one scan direction per block
and includes our selector. For Cobra, DTP's convention counts the language
model's prefill MACs with 30 text tokens. At a
fixed $\rho$ the cost does not depend on the method that selects the tokens:
the learned rules (\ours, the zero mask, the
straight-through variants) deploy the same network and cost the same; fixed
scores save only the selector, 0.3\% of Vim-Tiny's FLOPs. Rules that prune
inside the network, such as DTP and FastV in Cobra, keep more tokens in early
layers at the same FLOPs, so Cobra's budgets are matched in FLOPs. In Cobra the
vision backbone and the projector cost 617 GMACs at every budget, against
2,011 GMACs for the dense language model, so 10\% of the language-model
FLOPs is 31\% of the cost of the whole model.

\begin{table*}[t]
\caption{\textbf{With common cut locations, the token budget fixes the cost of the
compacted backbone.} The table gives retained tokens (of 196 patches in
Vim-Tiny, 1,568 in VideoMamba-Ti and 729 visual tokens in Cobra-3B) and
compute per image or video view. Analytic MACs cover the backbone, and the
published FLOP columns use the conventions of the compared methods
(\cref{sec:supp_protocols}). Cobra's nominal budgets use a 30-token text
prompt. Parentheses give the LM FLOP shares measured on the six benchmarks.}
\label{tab:supp_cost}
\centering\footnotesize
\begin{tabular}{@{}llrrrrrr@{}}
\toprule
& & \multicolumn{2}{c}{Tokens kept} & \multicolumn{2}{c}{GMACs, analytic} & \multicolumn{2}{c}{GFLOPs, published convention} \\
\cmidrule(lr){3-4}\cmidrule(lr){5-6}\cmidrule(l){7-8}
Host & Budget & per cut & final \% & per input & \% & per input & \% \\
\midrule
Vim-Tiny & dense & 196 & 100 & 1.53 & 100 & 1.45 & 100 \\
 & $\rho=0.3$, 2 cuts & 59 / 18 & 9.2 & 0.87 & 56.8 & 0.83 & 57.1 \\
 & $\rho=0.35$, 2 cuts & 69 / 24 & 12.2 & 0.91 & 59.4 & 0.86 & 59.6 \\
 & $\rho=0.4$, 2 cuts & 78 / 31 & 15.8 & 0.94 & 61.8 & 0.90 & 62.1 \\
 & $\rho=0.5$, 2 cuts & 98 / 49 & 25.0 & 1.03 & 67.5 & 0.98 & 67.8 \\
 & $\rho=0.6$, 2 cuts & 118 / 71 & 36.2 & 1.12 & 73.4 & 1.07 & 73.7 \\
 & $\rho=0.7$, 2 cuts & 137 / 96 & 49.0 & 1.21 & 79.5 & 1.16 & 79.8 \\
 & $\rho=0.75$, 2 cuts & 147 / 111 & 56.6 & 1.26 & 82.8 & 1.20 & 83.1 \\
 & $\rho=0.8$, 2 cuts & 157 / 126 & 64.3 & 1.31 & 86.1 & 1.25 & 86.4 \\
 & $\rho=0.9$, 2 cuts & 176 / 158 & 80.6 & 1.41 & 92.7 & 1.35 & 93.0 \\
\midrule
VideoMamba-Ti & dense & 1568 & 100 & 12.15 & 100 & 11.54 & 100 \\
 & $\rho=0.4$ & 627 / 251 / 100 & 6.4 & 5.07 & 41.8 & 4.86 & 42.1 \\
 & $\rho=0.5$ & 784 / 392 / 196 & 12.5 & 5.82 & 47.9 & 5.57 & 48.3 \\
 & $\rho=0.6$ & 941 / 564 / 339 & 21.6 & 6.72 & 55.3 & 6.42 & 55.6 \\
 & $\rho=0.7$ & 1098 / 768 / 538 & 34.3 & 7.78 & 64.1 & 7.44 & 64.5 \\
 & $\rho=0.8$ & 1254 / 1004 / 803 & 51.2 & 9.03 & 74.3 & 8.63 & 74.8 \\
 & $\rho=0.85$ & 1333 / 1133 / 963 & 61.4 & 9.73 & 80.1 & 9.29 & 80.5 \\
 & $\rho=0.9$ & 1411 / 1270 / 1143 & 72.9 & 10.48 & 86.2 & 10.01 & 86.7 \\
\midrule
Cobra-3B & dense & 729 & 100 & 2628 & 100 & 2011 & 100 \\
 & 10\% LM (11.3) & 46 & 6.3 & 818 & 31.1 & 201 & 10.0 \\
 & 15\% LM (16.2) & 84 & 11.5 & 919 & 35.0 & 302 & 15.0 \\
 & 25\% LM (26.1) & 160 & 21.9 & 1120 & 42.6 & 503 & 25.0 \\
 & 35\% LM (36.0) & 236 & 32.4 & 1321 & 50.3 & 705 & 35.0 \\
 & 48\% LM (48.7) & 334 & 45.8 & 1581 & 60.2 & 964 & 48.0 \\
 & 67\% LM (67.5) & 479 & 65.7 & 1965 & 74.8 & 1349 & 67.1 \\
\bottomrule
\end{tabular}

\end{table*}

\paragraph{Speech: frame dropping in Mamba-based HuBERT.}
We use the Mamba block and convolutional front end of Mamba-based HuBERT
\cite{lin2026mambahubert} (12 layers of width 384, 17.2M parameters, 50 frames
per second), either causal or with ExtBiMamba layers. The authors released no self-supervised checkpoints, so we train character
CTC~\cite{graves2006ctc} from scratch on LibriSpeech
train-clean-100~\cite{panayotov2015librispeech} and decode without a language
model. The absolute error rates reflect this 100-hour training setup without
pre-training. All selection rules share the encoder architecture, training
data, optimizer and frame budget. After layer 6 the encoder keeps a fraction $\rho$ of the frames, and
layers 7--12 run on the compacted sequence at inference. An intermediate CTC
head at layer 6 (weight 0.3, \cite{lee2021interctc}) provides blank posteriors,
and dropped frames re-enter the final CTC with their layer-6 representation
(\cref{sec:training}). Baselines select frames with the same budget: uniform
subsampling, \emph{CTC top-$k$} (lowest intermediate blank posterior), and the
\emph{CTC-peak} rule of Soundwave \cite{zhang2025soundwave} (CTC spikes first,
then the lowest blank posterior). The selector starts from the intermediate CTC non-blank log-odds and adds the
output of a two-layer MLP on detached layer-6 features. We initialize the
MLP's last layer at zero, so all learned configurations start from the CTC
top-$k$ ranking. We train the selector with a zero mask, \ours without step
gating, full \ours, or \ours without the convolution gradient. Models train for 80k steps (AdamW, peak learning rate
$5{\cdot}10^{-4}$, batches of 87.5\,s of audio). In the retrofit scenario of
dynamic Vision Mamba pruning \cite{wu2025dyvm}, we instead fine-tune the trained
dense or CTC top-$k$ model of the same seed for 20k steps with every
selection rule. In a third scenario the dropped frames do not return: the
final CTC reads the compacted sequence and predicts 300 BPE units
\cite{sennrich2016bpe} learned on the training transcripts. Trained from
scratch, most of these models stayed on the all-blank plateau of CTC, so we fine-tune
the dense character model of the same seed for 40k steps (peak learning rate
$10^{-4}$) with a new output head, keeping the character intermediate CTC that
provides the selection prior. We report greedy-decoding WER on test-clean and
test-other.

\paragraph{Genomics: dynamic chunking in H-Net.}
We train a one-stage H-Net \cite{hwang2026hnet} for DNA language modeling on
hg38 with the train/validation split of Basenji \cite{kelley2018basenji}:
an encoder and a decoder of two Mamba-2 layers (width 256) around a main network
of six Mamba-2 layers (width 384); 14.8M parameters, contexts of 4096 bases,
20k steps. The published training (``H-Net routing'') uses straight-through
boundaries, the smoothing dechunker and a ratio loss (weight 0.03) towards a
target of $N$ bases per chunk. In our arms the router is trained through \ours
and step gating in the main network and the expected dechunker
(\cref{prop:dechunk}); all models are evaluated with the published hard
forward pass in bits per base on 512 validation windows. To compare at
matched compression we sweep $N\in\{2,3,4,6,8,10\}$ for the published routing
and $N\in\{2,3,4,6,8,10,12\}$ for our arms, whose routers compress less than
the target, and include an isotropic model of eight Mamba-2 layers without
chunking.

\paragraph{Genomics: LDARNet on the Nucleotide Transformer benchmark.}
LDARNet-110M \cite{ledneva2026ldarnet} applies bidirectional dynamic chunking
in front of a BiMamba-2 main network. We fine-tune the released checkpoint on the 18 downstream tasks of the
Nucleotide Transformer benchmark~\cite{dallatorre2024nt} and reproduce the
authors' published protocol in full. We run ten-fold cross-validation on the
training split, select the epoch with the best validation MCC and report its
MCC on the fixed test set. We use the published learning-rate schedule and task-specific learning rates,
a batch size of 64, and weight decay of 0.1 on all parameters except biases
and norms. We train for up to 100 epochs and stop after five epochs without an
improvement in validation MCC. The baseline fine-tunes the router with the published
mechanism; our arms train it through \ours and step gating in both directions
of the main network and the expected bidirectional dechunker. This amounts to
540 fine-tuning runs (six H100 GPUs for 36 hours). As controls we fine-tune
with the router frozen, add the ratio loss of pre-training ($\alpha{=}0.03$,
$N{=}4$) to the fine-tuning objective, and continue the masked language model
pre-training of the released checkpoint on hg38 for 1000 steps of 512
sequences of 4096 bases, with either routing, before the full fine-tuning
protocol.

\paragraph{Images: patch compaction in Vim-Tiny.}
We start from the released ImageNet-1k Vim-Tiny checkpoint \cite{zhu2024vim}
(BiMamba-v2, 24 blocks of width 192, patch size 16, $224{\times}224$ inputs,
196 patches and one class token; 76.1\% reported top-1). Reductions follow
blocks 10 and 20, as in ToP-ViM, with patch budgets $(k_1,k_2)$ set by $\rho$
(\cref{sec:supp_protocols}) that exclude the class
token, which always survives and keeps its relative order, so that its index
in the compacted sequence varies per image. The selector adds a learned
residual (LayerNorm--Linear(192,64)--GELU--Linear(64,1), last layer initialized
at zero) to a detached, normalized score of the backward-direction mixer mapped
back to the original token order. Nested exact-budget LapSum gives
$p^{(2)}=p^{(1)}\odot q^{(2)}$ with $\sum_i p^{(1)}_iq^{(2)}_i=k_2$; the hard
second-stage selection is restricted to the hard first-stage survivors, and the
straight-through mask is $m=\TopK_k(r)+p-\sg(p)$. The selective scan uses
$\Delta_i=m_i\operatorname{softplus}(d_i+b)$, so a removed position leaves the
state unchanged. In the \ours arm all 28 convolutions after block 10 (both
directions) use the expected compacted convolution on the full grid; the
zero-mask arm runs ordinary convolutions on probability-weighted inputs and
shares the gated scan; the compaction arm has no learned residual and gathers
the survivors of the score in training. Each arm fine-tunes backbone and
selector for three epochs over 1,271,167 ImageNet-1k training images
(micro-batch 64, effective batch 128, fp32 with TF32 disabled). The first
epoch uses the soft forward pass with the LapSum temperature decreasing
linearly from 1 to 0.25, the next two the straight-through forward at 0.25.
The loss is cross-entropy plus 0.5 times the distillation loss from the
frozen dense model (KL divergence at temperature 2, scaled by 4). AdamW
($\beta=(0.9,0.999)$, weight decay $10^{-4}$ on weight matrices only) uses
learning rates $3{\cdot}10^{-5}$ for the backbone and $3{\cdot}10^{-4}$ for the
selector with a linear warm-up over the first epoch and the cosine decay of a
30-epoch schedule, stopped after three epochs, and gradients are clipped at
norm 1. Ten images per class, chosen by a fixed filename hash, are held out
from fine-tuning as the development set (10,000 images); the dense model
scores 83.25\% top-1 on it and 76.07\% on the 50,000 images of the official
validation set. Every cell of \cref{tab:vim} uses three seeds and one code
base in one container (PyTorch 2.6). In the tables, ``pruned, no
fine-tuning'' compacts the released checkpoint with the base score of the
selector before any fine-tuning.

\subsection{Numerical agreement and evaluation scope}
\label{sec:exactness}
\label{sec:supp_scope}
\label{sec:limitations}

\begin{table}[t]
\caption{\textbf{\ours with step gating matches physical compaction at fixed binary
decisions.} Entries give relative errors in the stated precision. Italic rows
show discrepancies for the alternative computations. For DyVM, the fixed-mask
row tests forward equivalence, while the sampled-mask row compares the
training selection with the deployed top-$k$ selection.}
\label{tab:exactness}
\centering
\scriptsize
\begin{tabularx}{\linewidth}{@{}>{\raggedright\arraybackslash}X r@{}}
\toprule
Check (RTX 4090, fp32 unless noted) & relative error \\
\midrule
Mamba-1 (HuBERT block): $p{=}\mathbf{1}$ vs.\ dense & $1.4\cdot10^{-7}$ \\
Mamba-1: binary $p$ vs.\ physical compaction & $2.6\cdot10^{-7}$ \\
Mamba-2: binary $p$ vs.\ physical compaction & $3.2\cdot10^{-6}$ \\
ASR encoder, STE training forward vs.\ compacted inference & $2.9\cdot10^{-8}$ \\
Expected dechunker vs.\ Monte Carlo (causal / bidir.) & $4.1\cdot10^{-3}$ / $5.4\cdot10^{-3}$ \\
H-Net / LDARNet: STE forward vs.\ published forward & $3.4\cdot10^{-3}$ / $6.1\cdot10^{-4}$ \\
Cobra-3B (64 blocks, logits): STE training forward vs.\ compacted inference & $9.7\cdot10^{-7}$ \\
DyVM (Vim-Ti, $\rho{=}0.7$): rearranged forward at fixed decisions vs.\ compaction & $6.4\cdot10^{-7}$ \\
\emph{Zero mask vs.\ compaction (Mamba-1 / Mamba-2)} & 53\% / 57\% \\
\emph{Zero mask + step gating vs.\ compaction, Cobra-3B logits (25\% / 10\%)} & 21\% / 21\% \\
\emph{DyVM (Vim-Ti, $\rho{=}0.7$): sampled training decisions vs.\ deployed top-$k$} & 27\% \\
\bottomrule
\end{tabularx}

\end{table}

\paragraph{Interpreting forward agreement.}
With fixed binary keep decisions, Mamba-1 and Mamba-2 block outputs agree
with compaction to within $3.2{\cdot}10^{-6}$ relative error. At the retained positions, the zero-mask block outputs differ from physical
compaction by 53--57\% in relative error.
With the same binary mask, DyVM's rearranged forward also agrees with physical
compaction. Its 27\% relative error compares outputs under two masks, one
sampled during training and one selected by top-$k$ at deployment
(\cref{tab:exactness}).

\paragraph{Optimization and hardware.}
The short Vim and Kinetics-400 protocols test readaptation of released
checkpoints. ToP-ViM's published
hidden-state alignment result is reported for context. Our experiments with
its released score evaluate that score with compaction
(\cref{tab:supp_vim_val}). For Cobra, the DyVM adaptations learn on a
frozen backbone, and \ours uses an annealed soft forward, 0.5 points above
straight-through training at 10\% of LM FLOPs. We document
these adaptations and cost-matched comparisons in \cref{sec:supp_cobra}.

\paragraph{Keep distributions and runtime.}
A fixed-size subset couples the keep decisions, whereas \ours models them as
independent. The fractional step gate also relaxes the state update. Training
uses the full sequence length. Runtime depends on the operator implementation, host, batch size and hardware.
The inference measurements follow backbone-specific protocols
(\cref{sec:supp_kernel,sec:supp_vim,sec:supp_videomamba,sec:supp_cobra}).
The speech and genomics protocols have additional task-specific constraints
(\cref{sec:speech,sec:supp_genomics}).

\FloatBarrier
\section{Images: Vim-Tiny}
\label{sec:supp_vim}

\begin{table*}[t]
\caption{\textbf{The top-1 gap between \ours and zero masking widens at tight patch
budgets.} Results are for Vim-Tiny on 10,000 held-out ImageNet-1k training
images after three epochs of fine-tuning. All methods use physical compaction
at inference. Values with $\pm$ give the mean and standard deviation over
three seeds. \Cref{tab:supp_vim_val} evaluates the same models on the official
validation set.}
\label{tab:vim}
\centering
\scriptsize
\setlength{\tabcolsep}{3pt}
\resizebox{\textwidth}{!}{\begin{tabular}{@{}l ccccccccc@{}}
\toprule
Patches kept per cut ($\rho$) & 0.3 & 0.35 & 0.4 & 0.5 & 0.6 & 0.7 & 0.75 & 0.8 & 0.9 \\
Patches kept after blocks 10 / 20 & 59 / 18 & 69 / 24 & 78 / 31 & 98 / 49 & 118 / 71 & 137 / 96 & 147 / 111 & 157 / 126 & 176 / 158 \\
Final retention & 9.2\% & 12.2\% & 15.8\% & 25.0\% & 36.2\% & 49.0\% & 56.6\% & 64.3\% & 80.6\% \\
\midrule
Dense (all 196 patches) & 83.25 & 83.25 & 83.25 & 83.25 & 83.25 & 83.25 & 83.25 & 83.25 & 83.25 \\
Pruned, no fine-tuning & 26.93 & 31.41 & 37.01 & 47.94 & 58.41 & 67.50 & 71.52 & 74.59 & 79.72 \\
\midrule
Compaction in training & 70.77{\scriptsize$\pm$0.25} & 71.73{\scriptsize$\pm$0.28} & 72.74{\scriptsize$\pm$0.39} & 74.47{\scriptsize$\pm$0.40} & 76.84{\scriptsize$\pm$0.29} & 78.38{\scriptsize$\pm$0.08} & 79.02{\scriptsize$\pm$0.16} & 79.67{\scriptsize$\pm$0.18} & 80.47{\scriptsize$\pm$0.05} \\
Compaction in training, ToP-ViM score & 72.13{\scriptsize$\pm$0.52} & 73.22{\scriptsize$\pm$0.35} & 74.08{\scriptsize$\pm$0.20} & 75.94{\scriptsize$\pm$0.40} & 77.58{\scriptsize$\pm$0.09} & 79.05{\scriptsize$\pm$0.21} & 79.66{\scriptsize$\pm$0.12} & 80.29{\scriptsize$\pm$0.25} & 81.02{\scriptsize$\pm$0.06} \\
Learned, zero mask & 72.42{\scriptsize$\pm$0.39} & 74.10{\scriptsize$\pm$0.53} & 75.38{\scriptsize$\pm$0.30} & 78.45{\scriptsize$\pm$0.11} & 79.62{\scriptsize$\pm$0.11} & 80.24{\scriptsize$\pm$0.23} & 80.46{\scriptsize$\pm$0.12} & 80.98{\scriptsize$\pm$0.06} & \textbf{81.36}{\scriptsize$\pm$0.14} \\
Learned, DyVM forward & 69.61{\scriptsize$\pm$0.39} & 71.50{\scriptsize$\pm$0.56} & 72.68{\scriptsize$\pm$0.95} & 74.38{\scriptsize$\pm$0.67} & 77.13{\scriptsize$\pm$0.32} & 78.92{\scriptsize$\pm$0.02} & 79.31{\scriptsize$\pm$0.34} & 79.91{\scriptsize$\pm$0.16} & 80.74{\scriptsize$\pm$0.13} \\
Learned, \ours{} (ours) & \textbf{76.95}{\scriptsize$\pm$0.26} & \textbf{77.81}{\scriptsize$\pm$0.26} & \textbf{78.73}{\scriptsize$\pm$0.42} & \textbf{79.29}{\scriptsize$\pm$0.22} & \textbf{80.13}{\scriptsize$\pm$0.04} & \textbf{80.72}{\scriptsize$\pm$0.32} & \textbf{80.75}{\scriptsize$\pm$0.36} & \textbf{81.07}{\scriptsize$\pm$0.09} & 81.15{\scriptsize$\pm$0.56} \\
$\Delta$ \ours{} $-$ zero mask, seed-paired & $+4.53${\scriptsize$\pm$0.64} & $+3.71${\scriptsize$\pm$0.38} & $+3.35${\scriptsize$\pm$0.63} & $+0.84${\scriptsize$\pm$0.30} & $+0.51${\scriptsize$\pm$0.14} & $+0.47${\scriptsize$\pm$0.30} & $+0.29${\scriptsize$\pm$0.25} & $+0.09${\scriptsize$\pm$0.12} & $-0.21${\scriptsize$\pm$0.44} \\
\midrule
\multicolumn{10}{@{}l}{\emph{Training forward minus deployed network (soft $-$ hard)}} \\
Learned, zero mask & $+1.76${\scriptsize$\pm$0.23} & $+1.57${\scriptsize$\pm$0.41} & $+1.35${\scriptsize$\pm$0.33} & $+0.57${\scriptsize$\pm$0.36} & $+0.36${\scriptsize$\pm$0.03} & $-0.19${\scriptsize$\pm$0.15} & $0.00${\scriptsize$\pm$0.04} & $-0.23${\scriptsize$\pm$0.10} & $-0.07${\scriptsize$\pm$0.01} \\
Learned, DyVM forward & $-0.14${\scriptsize$\pm$0.06} & $-0.07${\scriptsize$\pm$0.15} & $-0.10${\scriptsize$\pm$0.10} & $-0.05${\scriptsize$\pm$0.06} & $-0.08${\scriptsize$\pm$0.03} & $-0.08${\scriptsize$\pm$0.01} & $-0.04${\scriptsize$\pm$0.07} & $-0.04${\scriptsize$\pm$0.03} & $-0.07${\scriptsize$\pm$0.08} \\
Learned, \ours{} (ours) & $+0.09${\scriptsize$\pm$0.18} & $+0.10${\scriptsize$\pm$0.21} & $-0.21${\scriptsize$\pm$0.14} & $-0.43${\scriptsize$\pm$0.08} & $-0.26${\scriptsize$\pm$0.12} & $-0.18${\scriptsize$\pm$0.16} & $-0.12${\scriptsize$\pm$0.13} & $-0.05${\scriptsize$\pm$0.15} & $-0.05${\scriptsize$\pm$0.06} \\
\bottomrule
\end{tabular}
}
\end{table*}

\begin{figure*}[p]
\centering
\includegraphics[width=\textwidth]{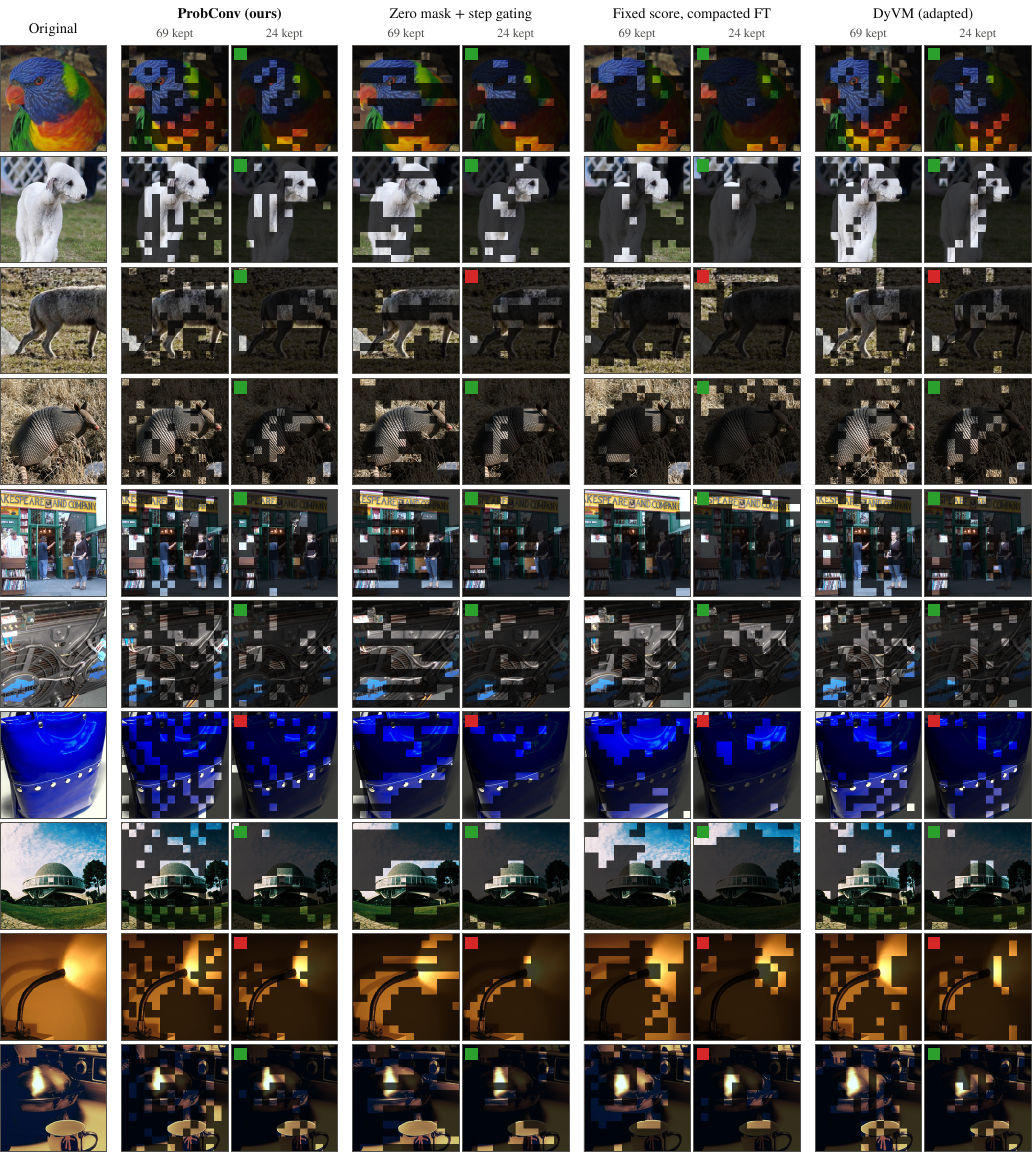}
\caption{\textbf{Different training rules retain different patches at the same budget.}
The ten evenly spaced ImageNet validation images use the Vim-Tiny models from
\cref{fig:budget}a with 69 and 24 patches after the two cuts (seed 0). Dimmed
patches are dropped. Green and red frames mark correct and incorrect
predictions after the last cut. From the top: lorikeet, Bedlington terrier,
coyote (\cref{fig:vim_masks}), armadillo, bookshop, electric fan, mailbag,
planetarium, spotlight and wok.}
\label{fig:supp_vim_masks}
\end{figure*}

\paragraph{Additional comparisons with a matched selector.}
At 12.2\% retention, using DyVM's rearrangement in the short matched-selector
experiment gives 71.50\% on the development split, $6.31\pm0.81$ points below
\ours ($5.59\pm0.84$ on validation).
The ``DyVM forward'' configuration keeps our selector, objective and schedule
and uses DyVM's rearranged training forward. It moves the retained patches
ahead of the dropped patches and masks the dropped patches. On the development split, top-1 with soft keep values minus top-1 after
compaction is $-0.07$ points for the rearranged variant, $+0.10$ for \ours and
$+1.57$ for zero masking (lower block of \cref{tab:vim,tab:supp_vim_val}). This comparison uses a
common selector architecture and short schedule. Across the nine
budgets the rearranged variant stays $0.5$--$7.0$ points below \ours on
validation; it falls below compaction in training at $\rho\le0.5$ and stays
within $0.6$ points of it at the milder budgets, and its soft-to-deployed
difference stays within $0.15$ points. \Cref{fig:supp_vim_masks} shows the
patches that each arm of \cref{fig:budget}a keeps at $\rho=0.35$ on ten
validation images. With DyVM's own selector and recipe, \cref{tab:supp_vim_dyvm_recipe}
trains every arm for 30 epochs with DyVM's recipe and
three cuts after blocks 6, 12 and 18.
As in DyVM, the cuts keep the fractions $\rho$, $\rho^2$ and $\rho^3$ of the
196 patches (78/31/12, 98/49/24 and 137/96/67 at $\rho=0.4$, 0.5 and 0.7). The recipe uses Mixup, CutMix and RandAugment, along with distillation from
the dense model through prediction KL and token MSE. The backbone learning
rate is $3{\cdot}10^{-5}$. The selector learning rate is $3{\cdot}10^{-6}$ at
$\rho=0.4$ and 0.5, and $10^{-5}$ at $\rho=0.7$. The DyVM configuration keeps
its predictor, Gumbel-softmax decisions, ratio loss and rearranged training
forward, and omits block selection. \ours and the zero mask with step gating
use our selector and LapSum with a temperature schedule. For these methods,
the budgets fall from (195, 194, 193) to the target by epoch 12. ``Without
budget schedule'' uses the target from the first step. DyVM ``with our budget
schedule'' uses that schedule as the target of its ratio loss. ``Without token
MSE'' retains only the KL term. At $\rho=0.4$ (12 of 196 patches after the
last cut) \ours reaches $68.40\pm0.23\%$, against $67.01\pm0.23\%$ for DyVM and
$63.76\pm0.42\%$ for the zero mask; at $\rho=0.5$ it leads DyVM by 0.46 points,
and at $\rho=0.7$ DyVM is 0.18 points ahead. The parts of the recipe that differ
between the arms do not explain these differences: without its budget schedule
\ours changes by $0.00$, $+0.34$ and $+0.15$ points at $\rho=0.4$, 0.5 and 0.7,
DyVM does not gain from our schedule ($-0.21$, $-0.41$, $+0.07$), and dropping
the token-MSE distillation helps the zero mask more than \ours at $\rho=0.4$ and
0.5 ($+1.91$ and $+1.37$ against $+0.28$ and $-0.05$) and both at $\rho=0.7$
($+0.42$ and $+0.69$). Without the token MSE, \ours is the best arm at
$\rho=0.7$ (74.28), ahead of DyVM with either schedule (73.77 and 73.84).

\paragraph{Official validation set.} \Cref{tab:supp_vim_val} evaluates the
networks of \cref{tab:vim} on the 50,000 images of the official ImageNet-1k
validation set, which the dense model did not see. Relative to the held-out training images, top-1 accuracy on the official
validation set is 5.2--6.5 points lower for the fine-tuned networks and 7.2
points lower for the dense model. The relative performance of the training
rules follows the same pattern. \ours and zero-mask training are level at the mildest cut
($-0.11\pm0.18$ at $\rho=0.9$), and \ours leads by 4.69 points with 18 of 196
patches left (\cref{tab:supp_vim_val}). The spread over seeds is smaller than
on the development set at eight of the nine budgets, and compaction in
training stays below the zero mask at every budget.
Fine-tuned for the whole 30-epoch schedule at ToP-ViM's setting (157/126),
\ours reaches $76.35\pm0.18\%$, above the dense model ($76.07\%$) and the
$75.1\%$ that ToP-ViM reports for 30 epochs from the same checkpoint (three seeds;
ToP-ViM's number averages several runs with
pruning after blocks 10 and 20, which removes 13.6\% of the FLOPs in ToP-ViM's convention with our
selector counted).
Compaction in training with ToP-ViM's score stays 0.1--4.5 points below \ours
at every budget.
The ``ToP-ViM score'' ranks tokens by the mean of the clipped SSM output, as
in ToP-ViM's released code. During training, physical compaction uses this
score in place of our base score and omits the learned residual. Each budget
uses three seeds. At 69/24 patches, extending fine-tuning to 30 epochs narrows \ours's lead from
3.7--4.0 to 1.9--2.0 points. \ours reaches $74.78\pm0.03\%$, compared with
$72.81\pm0.12\%$ for the zero mask and $72.88\pm0.03\%$ for compaction with
ToP-ViM's score. It leads both methods in every seed.
At 69/24, we use one seed to vary the selector learning rate ($10^{-4}$ or
$10^{-3}$ instead of $3{\cdot}10^{-4}$) and the final LapSum temperature (0.1
or 0.5 instead of 0.25) separately for both training methods. The best final
temperature is 0.5 for both. Over three seeds at that temperature, \ours
reaches $72.31\pm0.04\%$ and the zero mask $69.55\pm0.22\%$ (last block of
\cref{tab:supp_vim_val}). Tuning improves top-1 accuracy by 0.7 points for \ours and 1.7 points for the
zero mask. \ours still leads by $2.77$ points on average, with paired gains of
2.56--3.04 points across seeds. The lead with the default settings is 3.72
points. With this setting at all nine
budgets, \ours leads the zero mask in every seed at every budget, by 2.93 to
0.18 points from the strongest to the mildest budget, including $+0.25$ at
$\rho=0.9$, where the default settings tie ($-0.11$); tuning helps the zero mask
most where it is weakest ($+2.25$ at 59/18, against $+0.49$ for \ours).

\begin{table*}[t]
\caption{\textbf{At tight budgets, \ours's advantage over zero-mask training carries
over to the official validation set.} We evaluate Vim-Tiny with physical
compaction on 50,000 ImageNet-1k validation images and report top-1 accuracy
as mean $\pm$ standard deviation over three seeds. The final block uses a
final LapSum temperature of 0.5 in place of 0.25, the best value at 69/24
patches. \ours leads at every tested budget with this temperature. Bold marks
the best fine-tuned result in each column of the main block.}
\label{tab:supp_vim_val}
\centering
\scriptsize
\setlength{\tabcolsep}{3pt}
\resizebox{\textwidth}{!}{\begin{tabular}{@{}l ccccccccc@{}}
\toprule
Patches kept per cut ($\rho$) & 0.3 & 0.35 & 0.4 & 0.5 & 0.6 & 0.7 & 0.75 & 0.8 & 0.9 \\
Patches kept after blocks 10 / 20 & 59 / 18 & 69 / 24 & 78 / 31 & 98 / 49 & 118 / 71 & 137 / 96 & 147 / 111 & 157 / 126 & 176 / 158 \\
Final retention & 9.2\% & 12.2\% & 15.8\% & 25.0\% & 36.2\% & 49.0\% & 56.6\% & 64.3\% & 80.6\% \\
\midrule
Dense (all 196 patches) & 76.07 & 76.07 & 76.07 & 76.07 & 76.07 & 76.07 & 76.07 & 76.07 & 76.07 \\
Pruned, no fine-tuning & 24.36 & 29.08 & 33.77 & 43.18 & 52.85 & 61.10 & 64.82 & 67.98 & 72.89 \\
\midrule
Compaction in training & 65.55{\scriptsize$\pm$0.27} & 66.37{\scriptsize$\pm$0.11} & 67.23{\scriptsize$\pm$0.16} & 68.96{\scriptsize$\pm$0.35} & 70.66{\scriptsize$\pm$0.21} & 72.17{\scriptsize$\pm$0.07} & 72.85{\scriptsize$\pm$0.08} & 73.37{\scriptsize$\pm$0.13} & 74.37{\scriptsize$\pm$0.11} \\
Compaction in training, ToP-ViM score & 66.70{\scriptsize$\pm$0.18} & 67.63{\scriptsize$\pm$0.30} & 68.31{\scriptsize$\pm$0.12} & 69.83{\scriptsize$\pm$0.14} & 71.46{\scriptsize$\pm$0.11} & 72.77{\scriptsize$\pm$0.12} & 73.27{\scriptsize$\pm$0.13} & 73.95{\scriptsize$\pm$0.09} & 74.73{\scriptsize$\pm$0.14} \\
Learned, zero mask & 66.52{\scriptsize$\pm$0.53} & 67.87{\scriptsize$\pm$0.32} & 69.11{\scriptsize$\pm$0.33} & 72.05{\scriptsize$\pm$0.13} & 73.23{\scriptsize$\pm$0.06} & 73.93{\scriptsize$\pm$0.10} & 74.27{\scriptsize$\pm$0.12} & 74.51{\scriptsize$\pm$0.10} & \textbf{74.98}{\scriptsize$\pm$0.08} \\
Learned, DyVM forward & 64.26{\scriptsize$\pm$0.48} & 66.00{\scriptsize$\pm$0.74} & 66.93{\scriptsize$\pm$1.00} & 68.55{\scriptsize$\pm$0.57} & 70.95{\scriptsize$\pm$0.16} & 72.71{\scriptsize$\pm$0.09} & 72.91{\scriptsize$\pm$0.35} & 73.44{\scriptsize$\pm$0.07} & 74.34{\scriptsize$\pm$0.14} \\
Learned, \ours{} (ours) & \textbf{71.22}{\scriptsize$\pm$0.06} & \textbf{71.59}{\scriptsize$\pm$0.12} & \textbf{72.28}{\scriptsize$\pm$0.45} & \textbf{73.13}{\scriptsize$\pm$0.05} & \textbf{73.71}{\scriptsize$\pm$0.13} & \textbf{74.29}{\scriptsize$\pm$0.19} & \textbf{74.43}{\scriptsize$\pm$0.13} & \textbf{74.72}{\scriptsize$\pm$0.06} & 74.87{\scriptsize$\pm$0.25} \\
$\Delta$ \ours{} $-$ zero mask, seed-paired & $+4.69${\scriptsize$\pm$0.56} & $+3.72${\scriptsize$\pm$0.22} & $+3.17${\scriptsize$\pm$0.62} & $+1.08${\scriptsize$\pm$0.16} & $+0.48${\scriptsize$\pm$0.17} & $+0.36${\scriptsize$\pm$0.28} & $+0.16${\scriptsize$\pm$0.04} & $+0.21${\scriptsize$\pm$0.05} & $-0.11${\scriptsize$\pm$0.18} \\
\midrule
\multicolumn{10}{@{}l}{\emph{Training forward minus deployed network (soft $-$ hard)}} \\
Learned, zero mask & $+2.03${\scriptsize$\pm$0.35} & $+1.98${\scriptsize$\pm$0.25} & $+1.68${\scriptsize$\pm$0.30} & $+0.58${\scriptsize$\pm$0.25} & $+0.38${\scriptsize$\pm$0.09} & $+0.11${\scriptsize$\pm$0.12} & $-0.11${\scriptsize$\pm$0.10} & $-0.01${\scriptsize$\pm$0.11} & $+0.03${\scriptsize$\pm$0.02} \\
Learned, DyVM forward & $-0.05${\scriptsize$\pm$0.03} & $-0.05${\scriptsize$\pm$0.03} & $-0.03${\scriptsize$\pm$0.06} & $-0.02${\scriptsize$\pm$0.06} & $-0.03${\scriptsize$\pm$0.02} & $-0.03${\scriptsize$\pm$0.02} & $+0.02${\scriptsize$\pm$0.03} & $+0.02${\scriptsize$\pm$0.03} & $0.00${\scriptsize$\pm$0.01} \\
Learned, \ours{} (ours) & $+0.05${\scriptsize$\pm$0.17} & $+0.08${\scriptsize$\pm$0.05} & $-0.05${\scriptsize$\pm$0.07} & $-0.34${\scriptsize$\pm$0.16} & $-0.21${\scriptsize$\pm$0.03} & $-0.08${\scriptsize$\pm$0.02} & $-0.06${\scriptsize$\pm$0.08} & $-0.04${\scriptsize$\pm$0.05} & $-0.04${\scriptsize$\pm$0.05} \\
\midrule
\multicolumn{10}{@{}l}{\emph{Short protocol, final temperature 0.5 instead of 0.25 (best of the tuning grid for both arms)}} \\
Learned, zero mask, final $T{=}0.5$ & 68.77{\scriptsize$\pm$0.46} & 69.55{\scriptsize$\pm$0.22} & 70.38{\scriptsize$\pm$0.62} & 72.73{\scriptsize$\pm$0.02} & 73.67{\scriptsize$\pm$0.05} & 74.03{\scriptsize$\pm$0.12} & 74.40{\scriptsize$\pm$0.10} & 74.51{\scriptsize$\pm$0.10} & 74.92{\scriptsize$\pm$0.13} \\
Learned, \ours{} (ours), final $T{=}0.5$ & 71.71{\scriptsize$\pm$0.08} & 72.31{\scriptsize$\pm$0.04} & 72.92{\scriptsize$\pm$0.09} & 73.29{\scriptsize$\pm$0.32} & 74.04{\scriptsize$\pm$0.12} & 74.56{\scriptsize$\pm$0.15} & 74.58{\scriptsize$\pm$0.15} & 74.85{\scriptsize$\pm$0.16} & 75.17{\scriptsize$\pm$0.09} \\
\bottomrule
\end{tabular}
}
\end{table*}

\paragraph{A larger backbone.} \Cref{tab:supp_vim_small} repeats the
comparison on Vim-Small \cite{zhu2024vim} with the same short protocol,
selector and cuts, three seeds, at four budgets: 59/18, 69/24
and 98/49 patches (9.2, 12.2 and 25.0\%, as for Vim-Tiny) and 109/61, which
removes 29.3\% of the FLOPs with our selector counted, against 29.4\% for the
Vim-S result of ToP-ViM \cite{zhan2024top}. The advantage carries over to Vim-Small. At all four tested budgets and in
every seed, \ours improves on zero-mask training, by $1.74\pm0.25$,
$1.70\pm0.27$, $1.46\pm0.58$ and $0.79\pm0.47$ points from the strongest to
the mildest budget. As on Vim-Tiny, the lead grows
with compression, the zero mask scores 0.6--1.5 points higher with soft keep
values than after compaction, and \ours does not. At 12.2\% the
larger network loses less against its dense model than Vim-Tiny (3.45 points
against 4.48). We report the validation set only: the Vim-S checkpoint has
seen the development images in training (93.02\% dense top-1 on them).

\begin{table}[h]
\caption{\textbf{\ours improves Vim-Small top-1 over zero-mask training at all four
tested budgets.} Results on the official ImageNet-1k validation set give the
mean and standard deviation over three seeds after three epochs of
fine-tuning. Inference uses physical compaction. ``Soft $-$ compacted''
subtracts compacted accuracy from accuracy with soft keep values.}
\label{tab:supp_vim_small}
\centering
\small
\setlength{\tabcolsep}{4pt}
\resizebox{\linewidth}{!}{%
\begin{tabular}{@{}lcccc@{}}
\toprule
Patches after blocks 10 / 20 & 59 / 18 (9.2\%) & 69 / 24 (12.2\%) & 98 / 49 (25.0\%) & 109 / 61 \\
\midrule
Dense & 80.48 & 80.48 & 80.48 & 80.48 \\
No fine-tuning & 55.98 & 61.15 & 70.37 & 72.53 \\
Learned, zero mask & 75.14\sd{0.19} & 75.33\sd{0.29} & 76.13\sd{0.51} & 77.03\sd{0.28} \\
Learned, \ours (ours) & \textbf{76.88}\sd{0.09} & \textbf{77.03}\sd{0.02} & \textbf{77.59}\sd{0.07} & \textbf{77.81}\sd{0.25} \\
\midrule
\ours $-$ zero mask, paired & $+1.74$\sd{0.25} & $+1.70$\sd{0.27} & $+1.46$\sd{0.58} & $+0.79$\sd{0.47} \\
Soft $-$ compacted, zero mask & $+0.57$\sd{0.13} & $+1.06$\sd{0.35} & $+1.48$\sd{0.56} & $+1.14$\sd{0.29} \\
Soft $-$ compacted, \ours & $+0.01$\sd{0.05} & $+0.03$\sd{0.04} & $+0.01$\sd{0.03} & $+0.01$\sd{0.05} \\
\bottomrule
\end{tabular}}
\end{table}

\paragraph{Training trajectory.} The paired development-set gap at 69/24 is $2.00\pm0.29$ points after the
first, soft-forward epoch and $3.77\pm0.23$ after the second, when training
switches to binary masks. It is $3.71\pm0.38$ points after the final epoch.
After the soft epoch, both methods score 1.9--2.9 points higher with the soft
forward pass than with the deployed network. Once training switches to a
binary forward pass, this gap falls to 0.06--0.10 points for \ours and remains
at 1.3--1.6 points for the zero mask.

\begin{table}[t]
\caption{\textbf{\ours leads the adapted DyVM selector at the two tighter budgets.}
Vim-Tiny top-1 is measured on the official ImageNet validation set after 30
epochs using the DyVM training recipe. The three cuts follow blocks 6, 12 and
18. Values give the mean and standard deviation over three seeds. Bold marks
the highest score in each column. Each indented row changes one component of
its parent configuration (\cref{sec:supp_vim}).}
\label{tab:supp_vim_dyvm_recipe}
\centering
\small
\setlength{\tabcolsep}{4pt}
\resizebox{\linewidth}{!}{%
\begin{tabular}{@{}lccc@{}}
\toprule
Patches kept per cut ($\rho$) & 0.4 & 0.5 & 0.7 \\
Patches after the last cut & 12 (6.1\%) & 24 (12.2\%) & 67 (34.2\%) \\
\midrule
\ours (ours) & 68.40\sd{0.23} & 70.90\sd{0.20} & 73.59\sd{0.09} \\
\quad without budget schedule & 68.40\sd{0.30} & \textbf{71.24}\sd{0.13} & 73.74\sd{0.11} \\
\quad without token MSE & \textbf{68.68}\sd{0.21} & 70.85\sd{0.46} & \textbf{74.28}\sd{0.09} \\
DyVM \cite{wu2025dyvm}, adapted & 67.01\sd{0.23} & 70.44\sd{0.07} & 73.77\sd{0.07} \\
\quad with our budget schedule & 66.80\sd{0.30} & 70.03\sd{0.08} & 73.84\sd{0.04} \\
Zero-mask conv.\ + step gating & 63.76\sd{0.42} & 67.96\sd{0.13} & 72.32\sd{0.27} \\
\quad without token MSE & 65.67\sd{0.40} & 69.33\sd{0.34} & 72.74\sd{0.21} \\
\bottomrule
\end{tabular}}
\end{table}

\paragraph{Ablations.} \Cref{tab:supp_vim_ablation} changes one part of the
training signal at a time, with the protocol of \cref{tab:vim} at all four
budgets. Its first three columns state whether training runs the deployed
compacted computation (binary keep values in the forward pass), which operator
carries the gradient to the selector through the convolution taps (\ours, the
zero mask or none), and whether the selector also receives a gradient through
the step gate.
The ``soft forward'' configuration uses the soft forward pass in all three
epochs. ``Compaction in training'' uses the fixed score and omits the learned
residual. Both DyVM configurations train our selector with DyVM's rearranged
forward, which moves retained patches ahead of the masked, dropped patches.
Only the step-gated configuration also uses our step gating. Step gating matters most. Without it, \ours loses 0.20, 0.95, 2.12
and 7.18 points on the validation set from the mildest to the strongest
budget (seed-paired), and the zero mask loses a further 0.68--6.25 points.
Both deployed networks then fall up to 5.60 and 7.84 points below their
training forward. Stopping the gradient through the convolution leaves the forward pass
unchanged. On Vim, the resulting top-1 differences range from $-0.01$ to
$+0.13$ points on validation and from $-0.24$ to $+0.24$ on development. Training
straight-through from the first step, without the soft epoch, costs 0.22 to
1.19 points against \ours (seed-paired means at the four budgets). With the
convolution gradient also blocked, the expectation enters neither the forward
pass nor the gradient: this variant is level with \ours at 69/24
($71.55\pm0.16$ against $71.59\pm0.12$) but loses $1.23\pm1.10$, $1.19\pm0.34$
and $0.15\pm0.09$ points at 98/49, 118/71 and 147/111, in every seed at the
last two. On Vim the expectation in the convolution therefore helps through
the soft epoch, while the gradient through the taps adds nothing once that
epoch has run. How the selector
receives its gradient matters, however. DyVM's rearranged forward also trains
the deployed computation but passes the gradient through masked convolution
taps; it stays 5.6 and 4.6 points below \ours at 69/24 and 98/49, and adding
step gating to it lowers it further at every budget (63.73, 67.15, 69.39 and
72.28 at the four budgets of \cref{tab:supp_vim_ablation}; also 62.70 against
64.26 at 59/18 and 73.06 against 73.44 at 157/126). A soft
forward pass in all three epochs, never straight-through, is 0.33 and 0.34
points ahead of \ours at 147/111 and 118/71, level at 98/49 and 1.20 points
behind at 69/24, where its deployed network falls 2.46 points below its
training forward. The cost of training on a computation other than the
deployed one thus grows with compression, as for the zero mask. In summary,
\ours and its variant without the convolution gradient differ by at most 0.13
points at every budget; the variant adds a stop-gradient inside the operator,
so we keep the full operator as the default. \ours and its variant without the convolution gradient lead the comparison at
the two tightest budgets, 69/24 and 98/49 patches, the regime that compaction
targets. At 69/24, the other variants with step gating trail full \ours by
0.04 to 1.19 points, and the remaining training signals by 3.72 to 9.97
points.

\begin{table*}[t]
\caption{\textbf{At the two tightest budgets, no training signal exceeds \ours and its
variant without the convolution gradient.} Results give validation top-1 on
ImageNet-1k after three epochs, with mean and standard deviation over three
seeds. The lower block subtracts physically compacted accuracy from accuracy
with soft keep values. All rows use physical compaction at inference, except
for the soft evaluations used to compute the differences. Dashes mark omitted
training components. The fixed-score row also uses physical compaction during
fine-tuning. The lower block leaves the training-signal columns empty.}
\label{tab:supp_vim_ablation}
\centering
\scriptsize
\begin{tabular}{@{}l ccc cccc@{}}
\toprule
 & \multicolumn{3}{c}{training signal} & \multicolumn{4}{c}{patches kept after blocks 10 / 20 (final retention)} \\
\cmidrule(lr){2-4}\cmidrule(l){5-8}
Selector trained & \makecell{deployed\\forward} & \makecell{grad.\ via\\conv.\ taps} & \makecell{grad.\ via\\step gate} & \makecell{69 / 24\\(12.2\%)} & \makecell{98 / 49\\(25.0\%)} & \makecell{118 / 71\\(36.2\%)} & \makecell{147 / 111\\(56.6\%)} \\
\midrule
\ours{} (ours) & \checkmark & \ours & \checkmark & 71.59{\scriptsize$\pm$0.12} & 73.13{\scriptsize$\pm$0.05} & 73.71{\scriptsize$\pm$0.13} & 74.43{\scriptsize$\pm$0.13} \\
\ours{}, no conv.\ grad.\ (ours) & \checkmark & -- & \checkmark & 71.72{\scriptsize$\pm$0.16} & 73.12{\scriptsize$\pm$0.19} & 73.81{\scriptsize$\pm$0.14} & 74.45{\scriptsize$\pm$0.13} \\
\ours{}, straight-through from step 1 (ours) & \checkmark & \ours & \checkmark & 71.28{\scriptsize$\pm$0.13} & 71.94{\scriptsize$\pm$0.88} & 73.46{\scriptsize$\pm$0.18} & 74.21{\scriptsize$\pm$0.15} \\
\ours{}, straight-through from step 1, no conv.\ grad.\ (ours) & \checkmark & -- & \checkmark & 71.55{\scriptsize$\pm$0.16} & 71.90{\scriptsize$\pm$1.14} & 72.51{\scriptsize$\pm$0.42} & 74.28{\scriptsize$\pm$0.22} \\
\ours{}, soft forward (ours) & -- & \ours & \checkmark & 70.40{\scriptsize$\pm$0.07} & 73.12{\scriptsize$\pm$0.15} & 74.04{\scriptsize$\pm$0.20} & 74.76{\scriptsize$\pm$0.09} \\
\ours{} w/o step gating (ours) & -- & \ours & -- & 64.42{\scriptsize$\pm$0.35} & 71.01{\scriptsize$\pm$0.29} & 72.75{\scriptsize$\pm$0.24} & 74.23{\scriptsize$\pm$0.10} \\
Zero-mask conv.\ + step gating & -- & mask & \checkmark & 67.87{\scriptsize$\pm$0.32} & 72.05{\scriptsize$\pm$0.13} & 73.23{\scriptsize$\pm$0.06} & 74.27{\scriptsize$\pm$0.12} \\
Zero-mask conv., no step gating & -- & mask & -- & 61.62{\scriptsize$\pm$0.90} & 69.32{\scriptsize$\pm$0.34} & 71.76{\scriptsize$\pm$0.13} & 73.59{\scriptsize$\pm$0.10} \\
DyVM forward (rearranged tokens) & \checkmark & mask & -- & 66.00{\scriptsize$\pm$0.74} & 68.55{\scriptsize$\pm$0.57} & 70.95{\scriptsize$\pm$0.16} & 72.91{\scriptsize$\pm$0.35} \\
DyVM forward + step gating & \checkmark & mask & \checkmark & 63.73{\scriptsize$\pm$1.04} & 67.15{\scriptsize$\pm$0.92} & 69.39{\scriptsize$\pm$0.46} & 72.28{\scriptsize$\pm$0.33} \\
Compaction in training (fixed rule) & \checkmark & -- & -- & 66.37{\scriptsize$\pm$0.11} & 68.96{\scriptsize$\pm$0.35} & 70.66{\scriptsize$\pm$0.21} & 72.85{\scriptsize$\pm$0.08} \\
\midrule
\multicolumn{8}{@{}l}{\emph{Training forward minus deployed network (soft $-$ hard)}} \\
\ours{} (ours) &  &  &  & $+0.08${\scriptsize$\pm$0.05} & $-0.34${\scriptsize$\pm$0.16} & $-0.21${\scriptsize$\pm$0.03} & $-0.06${\scriptsize$\pm$0.08} \\
\ours{}, no conv.\ grad.\ (ours) &  &  &  & $-0.04${\scriptsize$\pm$0.06} & $-0.34${\scriptsize$\pm$0.02} & $-0.35${\scriptsize$\pm$0.01} & $-0.15${\scriptsize$\pm$0.07} \\
\ours{}, straight-through from step 1 (ours) &  &  &  & $+0.21${\scriptsize$\pm$0.04} & $-0.21${\scriptsize$\pm$0.08} & $-0.30${\scriptsize$\pm$0.12} & $-0.10${\scriptsize$\pm$0.03} \\
\ours{}, straight-through from step 1, no conv.\ grad.\ (ours) &  &  &  & $-0.03${\scriptsize$\pm$0.03} & $-0.47${\scriptsize$\pm$0.07} & $-0.31${\scriptsize$\pm$0.06} & $-0.20${\scriptsize$\pm$0.10} \\
\ours{}, soft forward (ours) &  &  &  & $+2.46${\scriptsize$\pm$0.07} & $+0.79${\scriptsize$\pm$0.09} & $+0.31${\scriptsize$\pm$0.07} & $+0.09${\scriptsize$\pm$0.02} \\
\ours{} w/o step gating (ours) &  &  &  & $+5.60${\scriptsize$\pm$0.40} & $+2.05${\scriptsize$\pm$0.14} & $+1.19${\scriptsize$\pm$0.11} & $+0.43${\scriptsize$\pm$0.02} \\
Zero-mask conv.\ + step gating &  &  &  & $+1.98${\scriptsize$\pm$0.25} & $+0.58${\scriptsize$\pm$0.25} & $+0.38${\scriptsize$\pm$0.09} & $-0.11${\scriptsize$\pm$0.10} \\
Zero-mask conv., no step gating &  &  &  & $+7.84${\scriptsize$\pm$0.55} & $+3.15${\scriptsize$\pm$0.18} & $+2.02${\scriptsize$\pm$0.06} & $+0.84${\scriptsize$\pm$0.20} \\
DyVM forward (rearranged tokens) &  &  &  & $-0.05${\scriptsize$\pm$0.03} & $-0.02${\scriptsize$\pm$0.06} & $-0.03${\scriptsize$\pm$0.02} & $+0.02${\scriptsize$\pm$0.03} \\
DyVM forward + step gating &  &  &  & $-0.12${\scriptsize$\pm$0.02} & $-0.02${\scriptsize$\pm$0.03} & $+0.00${\scriptsize$\pm$0.03} & $-0.02${\scriptsize$\pm$0.01} \\
\bottomrule
\end{tabular}

\end{table*}

\paragraph{Software stacks.} The dense model and the model pruned to 69/24 without fine-tuning score the
same on every software stack we used (83.25\% and 31.41\%). Training results
vary slightly with the convolution and scan implementations, so the numbers of
\cref{tab:vim,tab:supp_vim_val} come from one code base in one container.

\paragraph{Cost.} On an RTX 4090 at 69/24 (micro-batch 64), training through
\ours processes 447.5 images per second with the soft forward pass and 445.7
with straight-through, against 457.0 and 458.5 for the zero mask, an overhead
of 2.1\% and 2.8\%. The GH200 port of the pilot, with a pure-PyTorch scan and
micro-batch 32, had an overhead of 8.2\% and 11.6\% (230.4 and 208.3 against
249.2 and 232.4 images per second). At inference both arms run the same
compacted network. On an RTX 4090 in fp32, compacted Vim-Tiny processes a batch of 128 images in
28.5\,ms, versus 45.9\,ms for dense Vim-Tiny ($1.61\times$,
\cref{tab:main_cost}). At batch sizes 1 and 16, launch overhead dominates and
the compacted model is 10--14\% slower. The three cuts after blocks 6, 12 and
18 leave 98, 49 and 24 patches, respectively. The two-cut configuration below
also ends with 24 of the original 196 patches. The two-cut network of
\cref{fig:budget}a at $\rho=0.35$ (69 and 24 patches after blocks 10 and 20)
needs 31.5\,ms against 46.7\,ms for the dense model in the same measurement
($1.48\times$; in bf16 23.9 against 32.1\,ms, $1.34\times$), and at batch size 1
it is 8\% slower than the dense model (19.5 against 18.0\,ms).

\begin{figure*}[t]
\centering
\includegraphics[width=0.49\textwidth]{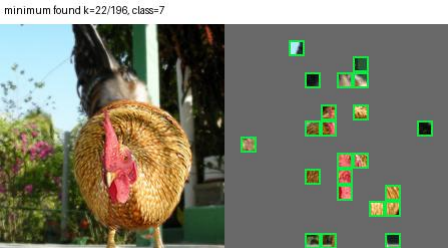}\hfill
\includegraphics[width=0.49\textwidth]{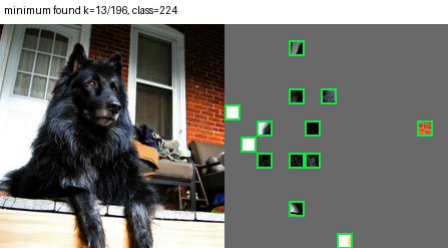}\\[2mm]
\includegraphics[width=0.49\textwidth]{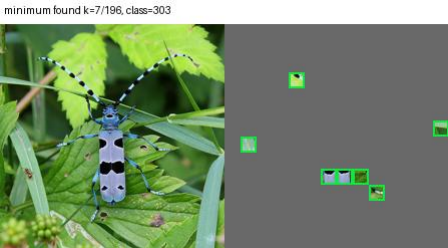}\hfill
\includegraphics[width=0.49\textwidth]{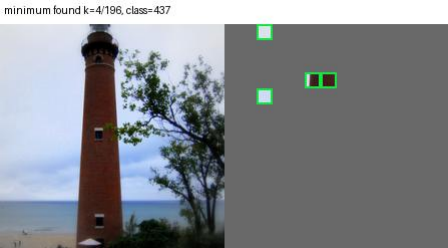}
\caption{\textbf{Four per-image selections preserve the dense prediction with 4--22
patches.} Panels show images 0, 2, 3 and 4 from \cref{tab:vim_oracle}, using
22, 13, 7 and 4 of 196 patches, respectively.}
\label{fig:supp_vim_oracle}
\end{figure*}

\paragraph{Per-image oracle.} The oracle uses the frozen Vim-Tiny with the class
token in the middle of the sequence and a single reduction after block 12 of
24; both directions of every later block use \ours, and the class token is
always kept. For each image, free patch logits and a continuous budget $k$ are
optimized jointly with
\begin{equation}
  \mathcal L_{\mathrm{oracle}}
  =\frac{\|h_{\mathrm{pruned}}-h_{\mathrm{full}}\|_2^2}
  {\|h_{\mathrm{full}}\|_2^2}+\frac{k}{196},
  \label{eq:supp_vim_oracle}
\end{equation}
where $h$ is the final representation, with exact-budget probabilities and a
final straight-through phase. We evaluate hard top-$k$ selections by physically compacting the sequence for
each integer budget $k$, starting at 1. \Cref{tab:vim_oracle} reports the
jointly learned $k$ and the shortest prefix of the learned ranking that
preserves the dense model's top-1 class. ``Agrees''
states whether the network compacted to the learned $k$ keeps that class, and
``retention'' is the smallest $k$ as a share of the 196 patches. The learned
budget averages 25.7 patches and preserves the class in nine of ten images;
the smallest prefixes have a median of 13.5 patches, but one hard image needs
162, so a small fixed count is no universal guarantee. The normalized
representation distance is 0.199 at the learned budgets and 0.676 at the
smallest same-class prefixes: class agreement is a much weaker target than the
representation. \Cref{fig:supp_vim_oracle} shows the masks of four of these images.

\begin{table}[h]
\caption{\textbf{Small per-image selections can preserve the dense Vim-Tiny
prediction.} For ten evenly spaced ImageNet images and the frozen Vim-Tiny,
``smallest $k$'' is the shortest prefix of the learned ranking that preserves
the dense model's top-1 class under physical compaction. ``Agrees'' records
whether the learned budget preserves that class.}
\label{tab:vim_oracle}
\centering
\small
\resizebox{\linewidth}{!}{%
\begin{tabular}{@{}rrrcrr@{}}
\toprule
Image & class & learned $k$ & agrees & smallest $k$ & retention \\
\midrule
0 &   7 & 26 & yes & 22  & 11.22\% \\
1 & 123 & 24 & yes & 11  &  5.61\% \\
2 & 224 & 24 & yes & 13  &  6.63\% \\
3 & 303 & 20 & yes &  7  &  3.57\% \\
4 & 437 & 20 & yes &  4  &  2.04\% \\
5 & 527 & 25 & yes & 14  &  7.14\% \\
6 & 632 & 29 & no  & 162 & 82.65\% \\
7 & 761 & 30 & yes & 15  &  7.65\% \\
8 & 894 & 33 & yes & 15  &  7.65\% \\
9 & 999 & 26 & yes &  9  &  4.59\% \\
\midrule
mean / median & & 25.7 / 25.5 & 9/10 & 27.2 / 13.5 & 13.88 / 6.89\% \\
\bottomrule
\end{tabular}}
\end{table}

\FloatBarrier
\section{Video}
\label{sec:supp_video}

\noindent We apply \ours to video in three hosts: token compaction inside the
bidirectional Mamba blocks of VideoMamba (\cref{sec:supp_videomamba}), frame
selection in front of a convolutional network (\cref{sec:supp_video_frames}),
and frame selection for the local network of Uni-AdaFocus
(\cref{sec:supp_video_adafocus}).

\subsection{Token compaction in VideoMamba}
\label{sec:supp_videomamba}

\begin{table*}[t]
\caption{\textbf{VideoMamba gains most over zero masking at the tightest Kinetics-400
budgets.} VideoMamba-Ti top-1 uses four temporal views and three spatial views
per clip with physical compaction. Fine-tuned results give the mean and
standard deviation over three seeds. HMDB51 first averages over its three
official splits. Two Kinetics columns match published FLOP budgets, and dashes
indicate budgets that the original studies did not report. Bold marks the best
score among our runs. Rules without fine-tuning are deterministic and have no
$\pm$.}
\label{tab:videomamba}
\centering
\scriptsize
\resizebox{\textwidth}{!}{\begin{tabular}{@{}l ccccccc cc cc@{}}
\toprule
 & \multicolumn{7}{c}{Kinetics-400, 3 seeds} & \multicolumn{2}{c}{K400 at reported FLOPs} & \multicolumn{2}{c}{HMDB51, 3 splits} \\
\cmidrule(lr){2-8}\cmidrule(lr){9-10}\cmidrule(l){11-12}
Tokens kept per cut ($\rho$) & 0.4 & 0.5 & 0.6 & 0.7 & 0.8 & 0.85 & 0.9 & 0.681 & 0.861 & 0.5 & 0.7 \\
MACs removed & 58\% & 52\% & 45\% & 36\% & 26\% & 20\% & 14\% & 38\% & 19\% & 52\% & 36\% \\
GMACs per view & 5.07 & 5.82 & 6.72 & 7.78 & 9.03 & 9.73 & 10.48 & 7.56 & 9.89 & 5.82 & 7.78 \\
GFLOPs, R-MeeTo counter & 4.86 & 5.57 & 6.42 & 7.44 & 8.63 & 9.29 & 10.01 & 7.23 & 9.44 & 5.57 & 7.44 \\
\midrule
Dense (12.15 GMACs, 11.54 GFLOPs) & 76.65 & 76.65 & 76.65 & 76.65 & 76.65 & 76.65 & 76.65 & 76.65 & 76.65 & 72.77 & 72.77 \\
\midrule
Token norm, no fine-tuning & 58.16 & 65.81 & 70.74 & 73.75 & 75.49 & 75.87 & \textbf{76.33} & 73.32 & 75.99 & 62.31 & 70.78 \\
Token norm, compacted fine-tuning & 71.33{\scriptsize$\pm$0.15} & 73.25{\scriptsize$\pm$0.07} & 74.29{\scriptsize$\pm$0.04} & 75.24{\scriptsize$\pm$0.11} & 75.85{\scriptsize$\pm$0.03} & 76.11{\scriptsize$\pm$0.08} & 76.12{\scriptsize$\pm$0.10} & 75.08{\scriptsize$\pm$0.13} & 76.13{\scriptsize$\pm$0.09} & 67.89{\scriptsize$\pm$0.40} & 71.42{\scriptsize$\pm$0.33} \\
Learned, zero-mask conv.\ + step gating & 72.07{\scriptsize$\pm$0.10} & 73.15{\scriptsize$\pm$0.17} & 74.01{\scriptsize$\pm$0.02} & 74.77{\scriptsize$\pm$0.02} & 75.28{\scriptsize$\pm$0.11} & 75.49{\scriptsize$\pm$0.11} & 75.79{\scriptsize$\pm$0.05} & 74.66{\scriptsize$\pm$0.09} & 75.57{\scriptsize$\pm$0.07} & 70.41{\scriptsize$\pm$0.38} & 71.64{\scriptsize$\pm$0.25} \\
Learned, DyVM forward (our selector) & 73.86{\scriptsize$\pm$0.23} & 74.78{\scriptsize$\pm$0.13} & 75.61{\scriptsize$\pm$0.11} & 75.93{\scriptsize$\pm$0.11} & \textbf{76.18}{\scriptsize$\pm$0.03} & 76.19{\scriptsize$\pm$0.12} & 76.30{\scriptsize$\pm$0.19} & 75.85{\scriptsize$\pm$0.06} & \textbf{76.26}{\scriptsize$\pm$0.08} & 69.67{\scriptsize$\pm$0.35} & 71.87{\scriptsize$\pm$0.32} \\
Learned, \ours{} (ours) & \textbf{74.66}{\scriptsize$\pm$0.22} & \textbf{75.35}{\scriptsize$\pm$0.04} & \textbf{75.71}{\scriptsize$\pm$0.21} & \textbf{75.97}{\scriptsize$\pm$0.13} & 76.15{\scriptsize$\pm$0.01} & \textbf{76.23}{\scriptsize$\pm$0.14} & 76.24{\scriptsize$\pm$0.04} & \textbf{76.00}{\scriptsize$\pm$0.06} & 76.22{\scriptsize$\pm$0.06} & \textbf{71.47}{\scriptsize$\pm$0.24} & \textbf{72.72}{\scriptsize$\pm$0.23} \\
\midrule
\multicolumn{12}{@{}l}{\emph{Reported, same released checkpoint, 30 epochs, FLOPs in the authors' convention}} \\
R-MeeTo \cite{shi2024rmeeto}, token merging, 18\% removed & -- & -- & -- & -- & -- & -- & -- & -- & 76.5 & -- & -- \\
DyVM \cite{wu2025dyvm}, tokens and blocks, 38\% removed & -- & -- & -- & -- & -- & -- & -- & 74.3 & -- & -- & -- \\
\bottomrule
\end{tabular}
}
\end{table*}

\paragraph{Host.} VideoMamba-Ti \cite{li2024videomamba} (7M parameters)
splits an 8-frame $224{\times}224$ clip into $14{\times}14$ patches per frame,
orders the 1568 patch tokens frame by frame after a class token, and processes
them with 24 bidirectional Mamba blocks, whose forward and backward scans each
have their own causal convolution of width 4. We start from the released
Kinetics-400 checkpoint (8 frames, 76.9 top-1 reported).

\paragraph{Compaction.} As in DynamicViT and DyVM \cite{rao2021dynamicvit,wu2025dyvm}, selection happens in
three stages: after blocks 6, 12 and 18 the network keeps the fractions
$\rho$, $\rho^2$ and $\rho^3$ of the patch tokens, and the class token is
always kept. With $\rho\in\{0.85,0.7,0.5\}$ this removes 20\%, 36\% and 52\%
of the multiply-accumulates of a view. At inference every arm gathers the kept
tokens, so the deployed network is the unmodified VideoMamba on a shorter
sequence; at $\rho=0.7$, blocks 7--12, 13--18 and 19--24 run on 1099, 769 and
539 tokens. The \emph{token norm} rule scores tokens by the norm of the
residual stream; we evaluate it without training and after fine-tuning with
physical compaction, where only the backbone learns. The \emph{learned} selector adds the output of a two-layer MLP, whose last
layer starts at zero, to the standardized token norm. It is trained with a
straight-through estimator and nested, exact-budget LapSum probabilities at
temperature 0.25 (\cref{sec:training}). It is trained
either through a zero mask in the convolutions together with step gating, or
through \ours with step gating in both scan directions, the backward one on
the reversed sequence (\cref{sec:gating}). At binary $p$ the latter computes
exactly the compacted network (\cref{prop:block}).

\paragraph{Training and evaluation.} All arms share the optimizer: AdamW with
learning rates $3{\cdot}10^{-5}$ for the backbone and $3{\cdot}10^{-4}$ for
the selector, weight decay $10^{-4}$, a cosine schedule to $0.1\times$ after
warm-up, batch 32, drop path 0.1, label smoothing 0.1, and bfloat16 with \ours
in float32. A training clip takes one random frame from each of 8 segments of
32 stored, uniformly spaced frames, with a random resized crop (area 25--100\%)
and a horizontal flip. On Kinetics-400 \cite{kay2017kinetics} we fine-tune for
3 epochs (warm-up 0.3 epochs) on the 239{,}789 training clips available from
the CVDF mirror, with three seeds. We evaluate on 19{,}877 of the 19{,}906 validation videos with
VideoMamba's 4 temporal $\times$ 3 spatial views and averaged softmax, where
the released model scores 76.65. On each of the three official splits of
HMDB51 \cite{kuehne2011hmdb} (3570 training and 1530 test videos) we first
fine-tune the dense model with a new head for 10 epochs, then each arm from it
for 5 epochs (warm-up 1 epoch) with three seeds. The seeds share
the test set of a split, whose binomial standard error is about 1.1 points. MACs are counted analytically per view (projections,
convolutions and scans of both directions, patch embedding and head); the dense
model has 12.15 GMACs. The FLOP counter of R-MeeTo \cite{shi2024rmeeto} hooks
the projections, the convolution and the scan (as $9LEN+2LE$) but counts only
one of the two scan directions of a block. On our forward pass, selector
included, it gives 11.54 GFLOPs for the dense model (11.31 without the patch
embedding, against 11.34 reported by DyVM \cite{wu2025dyvm}) and reductions of
19.5, 35.5 and 51.7\% at $\rho=0.85$, 0.7 and 0.5, within 0.5 points of our
MAC count.

\paragraph{Inference time.} We time the deployed networks (selectors, top-$k$, gather and the unchanged
blocks) on one RTX 4090 with bf16 autocast, on random clips already on the
GPU. After warm-up, we time forward passes with CUDA events and report the
median of 200 passes at batch size 1 and of 50 at each of batch sizes 32 and
64. At batch 32 the dense
model processes 555 clips per second and the networks at $\rho=0.85$, 0.7 and
0.5 process 651, 783 and 1,133 ($1.17$, $1.41$ and $2.04\times$, close to the
MAC reduction); batch 64 gives 535, 621, 755 and 1,093. At batch 1 the latency
is 18.7\,ms for the dense model and 19.8, 19.6 and 21.0\,ms for the compacted
ones: the forward pass is bound by kernel launches, and the three selectors
and gathers add kernels.

\paragraph{Results.} \Cref{tab:videomamba} gives the full results; see
\cref{sec:videomamba} for the Kinetics-400 comparison. Averaged over the
three HMDB51 splits, \ours reaches $72.72\%$ top-1 with $36\%$ fewer MACs and
$71.47\%$ with $52\%$ fewer MACs, 0.05 and 1.30 points below the dense
models. The seed-paired gains over zero-mask convolution and step gating are
$1.08\pm0.33$ and $1.06\pm0.26$ points, positive in 17 of the 18 pairs of
seed and split (the exception: $-0.07$ on split 2 at $\rho=0.7$). The gain
varies across splits at $\rho=0.7$ ($+2.00$, $+0.41$ and $+0.83$ on splits 1,
2 and 3) and is stable at $\rho=0.5$ ($+1.15$, $+1.07$, $+0.96$). The DyVM
forward with our selector reaches $69.67\pm0.35$ and $71.87\pm0.32$ on HMDB51,
1.80 and 0.85 points below \ours at $\rho=0.5$ and 0.7. At
$\rho=0.7$, compacted fine-tuning of the token norm is close to the zero mask
(71.42 against 71.64). On Kinetics-400 at $\rho=0.4$ (58\% fewer MACs), \ours reaches $74.66\pm0.22$.
It leads the DyVM forward by 0.80 points, the zero mask by 2.59 and compacted
fine-tuning by 3.33. The untuned token norm scores 58.16. At $\rho=0.9$ (14\%
fewer MACs), the gap over the zero mask narrows to 0.45 points. Fine-tuned
methods score 75.79--76.30, and the untuned token norm reaches 76.33. Changing the selector learning rate of
the zero mask by a factor of three in either direction ($10^{-4}$ or
$10^{-3}$, one seed) does not help it: 72.94 and 72.91 at $\rho=0.5$ (default
73.15) and 75.43 and 75.58 at $\rho=0.85$ (default 75.49).

\paragraph{Matched to the reported FLOPs.} R-MeeTo and DyVM report results at
18\% and 38\% fewer FLOPs by their own counters. With $\rho=0.861$ and
$0.681$, R-MeeTo's counter on our network gives 9.44 and 7.23 GFLOPs ($-18.2$
and $-37.3\%$; in DyVM's convention, without the patch embedding, $-18.5$ and
$-38.1\%$, against 7.02 of 11.34 GFLOPs reported by DyVM).
At $\rho=0.4$, 0.6, 0.8 and 0.9, we use the analytic form of the counter to
obtain GFLOPs. It agrees with the original counter to within 0.01 GFLOPs at
the other six points. The reported rows use the authors' FLOP counts and
30-epoch schedules. DyVM also skips blocks. At the FLOP budget reported by DyVM, all fine-tuned configurations exceed
DyVM's reported top-1 of 74.3 after 30 epochs. After three epochs, zero
masking reaches 74.66, compacted fine-tuning reaches 75.08 and \ours leads
with $76.00\pm0.06$. At R-MeeTo's FLOPs, \ours reaches $76.22\pm0.06$ against 76.5
reported by R-MeeTo after 30 epochs; the token norm without any training
already gives 75.99 and compacted fine-tuning 76.13, so at this mild budget
training the selector adds little and token merging is 0.3 points better. With
DyVM's rearranged forward our selector reaches $75.85\pm0.06$ at DyVM's FLOPs
and $76.26\pm0.08$ at R-MeeTo's, within 0.15 points of \ours at both.

\paragraph{Ablations.} \Cref{tab:supp_videomamba_ablation} changes one part
of the training signal of \ours at a time on HMDB51 split 1, with the same
three seeds.
As in Vim, step gating matters most, and more so under stronger compression:
without it \ours loses 2.20 points at $\rho=0.7$ and 5.01 at $\rho=0.5$, on
every seed (1.83--2.75 and 4.64--5.42), and falls below the zero mask with step
gating. Blocking the gradient through the convolution taps, with the forward
pass unchanged, costs 1.00 points at $\rho=0.7$ and 1.07 at $\rho=0.5$
(0.52--1.24 and 0.52--1.50 on every seed).

A replicate set of runs with the same seeds gives the same picture: \ours
ahead of every other arm at both budgets, by 1.44 and 0.68 over the zero mask,
and step gating worth 1.57 and 4.55 points. Only the effect of the convolution gradient is smaller in the replicate set
(0.46 and $0.24\pm0.42$ points). The same seed of the same arm
differs between the two sets by up to 1.18 points (median 0.33), as much as the
spread over seeds and about the binomial standard error of the test set, so we
read the differences between the sets as run-to-run variation.

\begin{table}[t]
\caption{\textbf{In VideoMamba, both components contribute to accuracy, with the larger
effect coming from step gating.} Results give mean $\pm$ standard deviation
over three seeds for top-1 accuracy on HMDB51 split 1 after five epochs, with
4 temporal $\times$ 3 spatial views. $\Delta$ compares each variant with \ours
at the same selection granularity, pairing seeds. A dash marks the \ours
reference row. Whole frames are n/a at $\rho=0.7$, where 0.7, 0.49 and 0.343
of the 8 frames are not whole frames. The indented rows change one \ours
component. The compaction baseline fine-tunes with the fixed token-norm rule.}
\label{tab:supp_videomamba_ablation}
\centering
\scriptsize
\resizebox{\linewidth}{!}{%
\setlength{\tabcolsep}{3pt}
\begin{tabular}{@{}l cc cc@{}}
\toprule
 & \multicolumn{2}{c}{$\rho=0.5$ ($-52$\% MACs)} & \multicolumn{2}{c}{$\rho=0.7$ ($-36$\% MACs)} \\
\cmidrule(lr){2-3}\cmidrule(l){4-5}
Selector trained & top-1 & $\Delta$ & top-1 & $\Delta$ \\
\midrule
\ours{} (ours) & 72.59{\tiny$\pm$0.19} & -- & 74.81{\tiny$\pm$0.32} & -- \\
\quad no conv.\ grad. & 71.53{\tiny$\pm$0.50} & $-1.07${\tiny$\pm$0.50} & 73.81{\tiny$\pm$0.62} & $-1.00${\tiny$\pm$0.42} \\
\quad w/o step gating & 67.58{\tiny$\pm$0.56} & $-5.01${\tiny$\pm$0.39} & 72.61{\tiny$\pm$0.46} & $-2.20${\tiny$\pm$0.48} \\
Zero mask + step gating & 71.44{\tiny$\pm$0.51} & $-1.15${\tiny$\pm$0.44} & 72.81{\tiny$\pm$0.62} & $-2.00${\tiny$\pm$0.70} \\
Compaction (token norm) & 68.41{\tiny$\pm$0.51} & $-4.18${\tiny$\pm$0.69} & 72.64{\tiny$\pm$0.33} & $-2.18${\tiny$\pm$0.51} \\
\midrule
\multicolumn{5}{@{}l}{\emph{Whole frames (4, 2 and 1 of 8 kept; replicate set)}} \\
\ours{} (ours) & 63.64{\tiny$\pm$1.11} & -- & n/a & n/a \\
Zero mask + step gating & 62.16{\tiny$\pm$1.26} & $-1.48${\tiny$\pm$1.78} & n/a & n/a \\
\bottomrule
\end{tabular}
}
\end{table}

\paragraph{Whole frames.} Keeping whole frames (4, 2 and 1 of 8 at
$\rho=0.5$, the same cost as tokens) gives $63.64\pm1.11$ with \ours and
$62.16\pm1.26$ with the zero mask on HMDB51 (three seeds, replicate
set), 9.0 and 9.3 points below token selection
(\cref{tab:supp_videomamba_ablation}). Because
VideoMamba orders tokens frame by frame, removing whole frames changes the
convolution neighborhood only of the three tokens next to each boundary
between kept frames, so we expected the two arms to agree. The seed-paired difference, $+1.48\pm1.78$ (2.88, 2.09 and $-0.52$), is within
the variation across seeds.

\begin{figure*}[t]
\centering
\includegraphics[width=\textwidth]{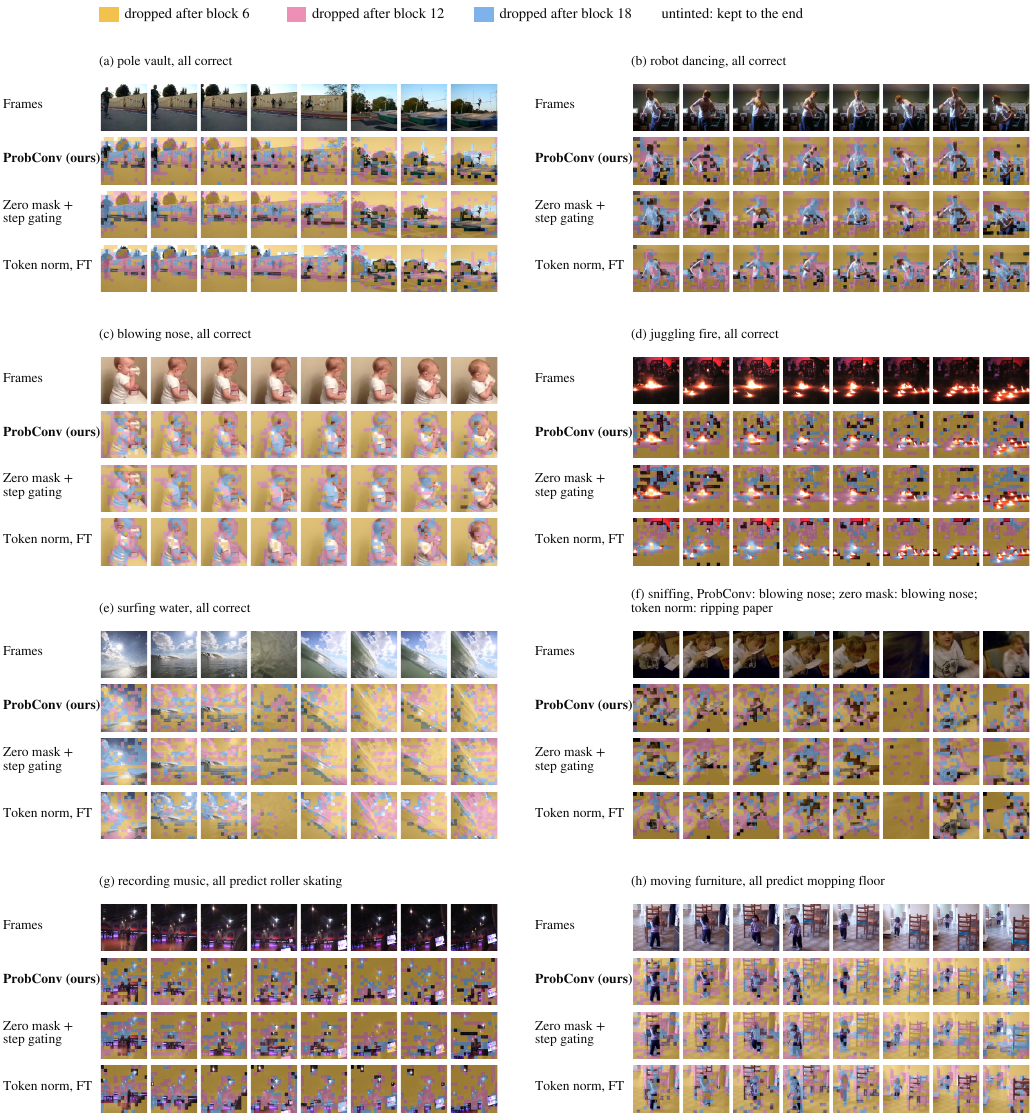}
\caption{\textbf{The three selection rules retain different subsets at each cut.}
VideoMamba-Ti uses $\rho=0.5$ on eight randomly drawn Kinetics-400 validation
clips. Yellow, pink and blue mark tokens dropped after blocks 6, 12 and 18,
respectively. Tokens in their original colors survive all three cuts. Each
clip shows the ground-truth class and each rule's prediction.}
\label{fig:supp_videomamba_tokens}
\end{figure*}

\paragraph{Tokens kept.} \Cref{fig:videomamba_tokens,fig:supp_videomamba_tokens} show when each token
is dropped by \ours, zero masking with step gating and compacted fine-tuning
with token-norm selection. For each clip, we show the center crop of the second temporal view from the
Kinetics-400 runs at $\rho=0.5$ (seed 1). The eight clips of \cref{fig:supp_videomamba_tokens} are
drawn with Python's \texttt{random.Random(0).sample} from the validation list
in annotation order. The clip of \cref{fig:videomamba_tokens} was picked for
legibility from 48 clips of every 50th validation clip (the 16 with the most
motion, the 16 on which \ours and the zero mask keep the most different tokens
and 16 evenly spaced). We analyze every 50th validation clip (398 clips). For tokens that survive all
cuts, the preceding token in scan order, corresponding to the left neighbor,
also survives in 57\% of cases with zero masking, 37\% with \ours and 41\%
with token-norm compaction. For the neighbor above, the corresponding rates
are 40\%, 35\% and 33\%. With the zero mask, the
convolution of a kept token reads its raw neighbors, so its input is
informative only when they are kept too. Seed 2 gives the same shares, and at
$\rho=0.7$ the left neighbor is kept for 74\% with the zero mask and 58\% with
\ours (above: 64 and 57\%). Across the 398 clips, the intersection over union between the top-$k$ training
masks (LapSum at the training temperature) and the masks used for physically
compacted inference is 0.985 for \ours. For the zero mask, it is 0.725 because
the second and third cuts compute scores on a different network during
training.

\subsection{Frame selection in a convolutional video network}
\label{sec:supp_video_frames}

\paragraph{Host.} OCSampler \cite{lin2022ocsampler} selects $N$ of $T$
candidate frames in one step: a skim network, MobileNetV2
\cite{sandler2018mobilenetv2} with temporal shift \cite{lin2019tsm} at
$128^2$, scores every candidate with a linear head, and a classifier runs on
the $N$ selected frames in temporal order. We use X3D-S
\cite{feichtenhofer2020x3d} (Kinetics-400 weights, $160^2$) as the
classifier, with $T{=}10$ and $N{=}4$, on ActivityNet-v1.3
\cite{caba2015activitynet} (10{,}014 of the 10{,}024 training and 4{,}916 of
the 4{,}926 validation videos), with the split files and the candidate
sampling of OCSampler: the video is cut into $T$ equal segments, and a
candidate is a random frame of its segment in training and the middle frame
at test time. X3D-S has no temporal stride, and its temporal convolutions are
depthwise, $5{\times}1{\times}1$ in the stem and $3{\times}3{\times}3$ in every
block. The selected frames form a compacted clip, so each temporal
convolution of the deployed classifier sees the neighbors that survive
selection, not the neighbors in the video.

\paragraph{\ours in X3D-S.} We apply the centered \ours of \cref{eq:supp_centered} to every temporal
convolution, with two taps on each side in the stem and one in every block. In
the squeeze-and-excitation units and the global pool, we weight each frame by
its keep probability. At binary
keep probabilities the outputs at the kept frames equal X3D-S on the
compacted clip, which we check numerically; at $p\equiv1$ the network is the
dense X3D-S. BatchNorm uses its running statistics, so the dropped frames never
enter batch statistics. The zero-mask control multiplies the input of every
temporal convolution by the keep probabilities instead, so a kept frame sees
zeros where its neighbors were dropped.

\paragraph{Learning signals.} Let $S\sim\pi_\theta$ be a selected set of $N$
candidates, $m\in\{0,1\}^T$ its mask, $p=\E_\pi[m]$ the exact marginals of
the policy (differentiable in $\theta$), and $r(S)$ the OCSampler reward of
the classifier on the compacted clip: the probability of the true class,
minus the largest probability when the prediction is wrong. Let $g(p)$ be the
same reward of the \ours network at keep probabilities $p$; at binary keep
probabilities it computes the compacted clip, so $g(m_S)=r(S)$. All arms share
the skim network, the policy head, the data and the optimizer, and differ
only in the gradient:
\begin{itemize}
\item \emph{REINFORCE} (OCSampler, our reimplementation):
$(r(S)-r(S_{\mathrm{TSN}}))\nabla\log\pi(S)+\xi\nabla H$, with the reward of
the uniformly sampled clip as baseline, the entropy $H$ of the scores and
$\xi{=}0.1$; $\pi$ draws $N$ candidates without replacement from the softmax
of the scores, as in OCSampler.
\item \emph{\ours, top-$N$}: the cross-entropy of the \ours network at
$\TopK_N(s)+p-\sg(p)$, with exact-budget LapSum probabilities $p$
\cite{struski2025lapsum} (\cref{sec:training}); \emph{$\lambda{=}0$}: the same
with the gradient to $p$ only through the weighted means; \emph{soft}: the
\ours network at the LapSum probabilities, without straight-through.
\item \emph{\ours, sampled}: the same cross-entropy at $m+p-\sg(p)$ with
$m\sim\pi$ and $p$ the marginals of $\pi$, \ie, the gradient of \ours on the
masks the policy draws.
\item \emph{\ours baseline}: REINFORCE with $g(p)$ in place of the uniform
clip.
\item \emph{MuProp} \cite{gu2016muprop} with the \ours network as the model of
the reward: the control variate $c(m)=g(p)+a^\top(m-p)$,
$a=\partial g/\partial p$, gives
$(r(S)-\eta\,c(m))\nabla\log\pi(S)+\eta\,a^\top\nabla p+\xi\nabla H$. Since
$c$ is linear in $m$ and $\E_\pi[m]=p$ exactly, the estimator is unbiased for
every $\eta$; we use $\eta{=}1$ and the variance-minimizing
$\eta^*=-\E\langle g_{\mathrm{sf}},g_{\mathrm{cv}}\rangle/\E\|g_{\mathrm{cv}}\|^2$,
estimated from the previous minibatches and clipped to $[0,2]$.
\item \emph{REINFORCE $+\,\lambda\,$\ours}: the sum of the REINFORCE loss and
$\lambda$ times the top-$N$ \ours loss.
\end{itemize}
Besides the sampler of OCSampler, we use conditional Poisson sampling
\cite{hajek1964rejective}, $\pi(S)\propto\prod_{t\in S}e^{s_t}$, and
systematic sampling \cite{madow1949systematic} with LapSum marginals. The
marginals of sampling without replacement come from a dynamic program over
the $2^T$ subsets. On a toy problem with all sets enumerated ($T{=}6$,
$N{=}2$, float64), the expectations of REINFORCE, of the \ours baseline and
of MuProp at $\eta\in\{0,\tfrac12,1,2\}$ equal the exact gradient up to a
relative error of $2\cdot10^{-15}$.

\paragraph{Protocol.} We follow the two stages of OCSampler. Stage~I trains
X3D-S on $N$ random candidates in temporal order, \ie, on the compacted clips
it will run, and the skim network on all $T$ candidates (SGD with Nesterov
momentum, batch 64, cosine schedule, 50 epochs, learning rates $0.005$ and
$0.01$). Stage~II freezes the classifier and trains the skim network and the
policy head for 50 epochs with the same optimizer. Every learned rule is
evaluated by the top-$N$ scores of its policy, with the classifier run dense
on the compacted clip and one center crop per video, and we report mAP of the
final checkpoint on the validation set. Hyperparameters are chosen on a fixed
subset of 1000 training videos, with both stages retrained on the remaining
ones: for every arm the learning rate in $\{3,10,30\}\cdot10^{-4}$, and
additionally $\tau\in\{0.3,1,3\}$ for top-$N$ \ours, $\eta\in\{1,\eta^*\}$
for MuProp, and $\lambda\in\{0.1,0.3,1\}$ with the policy or LapSum
marginals for the sum. We then retrain each selected configuration on the full training set with
three seeds, and use the validation set only to report mAP.
Per video, X3D-S costs 0.31, 0.63 and 0.94 GFLOPs on 2, 4 and 6 frames and
1.56 on 10, and the skim network 1.02 GFLOPs on 10 frames at $128^2$ (fvcore).

\begin{table}[t]
\caption{\textbf{\ours trained on sampled masks reaches the mAP of retrained
OCSampler.} Results give ActivityNet-v1.3 validation mAP for X3D-S with $N$ of
ten frames. Learned policies use top-$N$ selection at inference. $\Delta$
gives the mean and range of paired differences from REINFORCE across three
seeds. The released six-frame OCSampler policy also supplies its top $N$
candidates to our classifier. $\Delta$ is not given for the training-free
rules, which have no seeds, and for REINFORCE itself.}
\label{tab:video_x3d}
\centering
\resizebox{\linewidth}{!}{%
\begin{tabular}{@{}l c cc cc@{}}
\toprule
 & $N{=}2$ & \multicolumn{2}{c}{$N{=}4$} & $N{=}4$, 2nd & $N{=}6$ \\
\cmidrule(lr){3-4}
Frame selection & & mAP & $\Delta$ to REINFORCE & classifier & \\
\midrule
\multicolumn{6}{@{}l}{\emph{Training-free rules with the stage-I classifier}} \\
All 10 frames, dense & 73.69 & 79.36 &  & 79.49 & 80.20 \\
Uniform & 61.78 & 72.22 &  & 72.00 & 75.59 \\
Random (3 draws) & 60.13 & 71.03 &  & 71.13 & 75.70 \\
MGSampler \cite{zhi2021mgsampler} & 61.66 & 72.60 &  & 72.35 & 76.23 \\
OCSampler, released sampler & 68.99 & 75.98 &  & \textbf{76.06} & \textbf{79.09} \\
\midrule
\multicolumn{6}{@{}l}{\emph{Learned in stage II, classifier frozen, mean $\pm$ std over 3 seeds}} \\
REINFORCE (OCSampler) & 69.45{\scriptsize$\pm$0.13} & 75.59{\scriptsize$\pm$0.31} &  & 75.60{\scriptsize$\pm$0.12} & 78.74{\scriptsize$\pm$0.27} \\
Zero mask, top-$N$ & 60.53{\scriptsize$\pm$0.92} & 71.87{\scriptsize$\pm$0.63} & $-3.72$ {\scriptsize[$-4.17$, $-3.46$]} & 71.89{\scriptsize$\pm$0.85} & 77.07{\scriptsize$\pm$0.32} \\
\ours, soft forward & 69.30{\scriptsize$\pm$0.26} & 75.42{\scriptsize$\pm$0.16} & $-0.17$ {\scriptsize[$-0.71$, $+0.14$]} & 75.00{\scriptsize$\pm$0.50} & 78.55{\scriptsize$\pm$0.12} \\
\ours, top-$N$, $\lambda{=}0$ & 68.02{\scriptsize$\pm$0.11} & 74.83{\scriptsize$\pm$0.17} & $-0.76$ {\scriptsize[$-1.31$, $-0.45$]} & 74.36{\scriptsize$\pm$0.15} & 77.90{\scriptsize$\pm$0.07} \\
\ours, top-$N$ & 67.38{\scriptsize$\pm$0.09} & 74.60{\scriptsize$\pm$0.27} & $-0.99$ {\scriptsize[$-1.27$, $-0.61$]} & 74.40{\scriptsize$\pm$0.35} & 77.57{\scriptsize$\pm$0.10} \\
REINFORCE, \ours baseline & 69.43{\scriptsize$\pm$0.34} & 75.56{\scriptsize$\pm$0.04} & $-0.02$ {\scriptsize[$-0.40$, $+0.18$]} & 75.68{\scriptsize$\pm$0.25} & 78.63{\scriptsize$\pm$0.16} \\
MuProp with \ours, $\eta^*$ & 69.56{\scriptsize$\pm$0.25} & 75.85{\scriptsize$\pm$0.06} & $+0.26$ {\scriptsize[$-0.14$, $+0.54$]} & 75.90{\scriptsize$\pm$0.19} & 78.55{\scriptsize$\pm$0.05} \\
REINFORCE $+\,0.3\,$\ours & 69.52{\scriptsize$\pm$0.23} & 75.82{\scriptsize$\pm$0.05} & $+0.23$ {\scriptsize[$-0.18$, $+0.48$]} & 75.71{\scriptsize$\pm$0.21} & 78.68{\scriptsize$\pm$0.18} \\
\ours, sampled masks & \textbf{69.89}{\scriptsize$\pm$0.29} & \textbf{76.06}{\scriptsize$\pm$0.21} & $+0.47$ {\scriptsize[$-0.13$, $+0.82$]} & 75.92{\scriptsize$\pm$0.23} & 78.79{\scriptsize$\pm$0.09} \\
\bottomrule
\end{tabular}
}
\end{table}

\paragraph{Results.} Training through \ours on the masks that the policy
samples gives the best selection, 76.06 mAP against 75.59 for the
REINFORCE of OCSampler (\cref{tab:video_x3d}), and matches the released
OCSampler sampler on our classifier (75.98; with its own classifier and 6 frames, the released sampler reaches 75.74). Its lead over REINFORCE appears at every learning rate ($+0.47$, $+0.39$ and
$+1.67$ at $3$, $10$ and $30\cdot10^{-4}$, with the same seeds), in all three
seeds at the two larger rates and in two of three at the smallest. Sampled
\ours gives 76.06--76.18 mAP across the tested learning rates and 75.87--76.09
across the samplers at the learning rate of \cref{tab:video_x3d_estimators}. The same
\ours gradient at the top-$N$ mask is 0.99 points below REINFORCE, below it
in every seed and at every learning rate and temperature: the gradient
through the compacted network helps when it is taken on the masks the policy
draws, and a deterministic top-$N$ mask in training costs about 1.5 points.
With deterministic top-$N$ selection in both training runs, \ours scores 74.60
mAP, 2.73 points above the zero mask at 71.87.
Selecting with a zero-filled clip is 3.72 points below REINFORCE and below the
uniform rule, the largest gap in the table, so in this host the mismatch
between the trained and the deployed network matters more than the estimator.
The other learned rules gain 2.4--3.8 points over the uniform rule,
against the 7.1 points that separate it from all ten frames.

\paragraph{Which frames are kept.} \Cref{tab:supp_x3d_selection} compares
the frames kept at $N{=}4$ by the rules of \cref{tab:main_x3d} on all 4,916
validation videos, for the three seeds (\cref{fig:x3d_frames} shows seed 0).
\ours on sampled masks keeps nearly the same frames as REINFORCE: per video
the two share 3.37 of the four frames and the top-1 class in 88.8\% of the
videos, as much as two REINFORCE runs with different seeds (3.37 and
89.0\%). Both spread the four frames over the clip and keep the first or the
last candidate in 42--44\% of the videos. The zero-mask policy shares 1.75--1.80
frames with either, close to the 1.60 of a random choice, keeps the first or
the last candidate in 10.6\% of the videos and three consecutive candidates
more often (37.7\% against 33.1--33.9\%). Of the frames kept by the zero-mask policy, 73.6\% fall inside annotated
action instances, compared with 73.4\% for both REINFORCE and \ours, 69.9\%
for the uniform rule and 65.6\% for a random choice. Its lower mAP therefore does not
come from missing the action.

\begin{table*}[t]
\caption{\textbf{Sampled \ours shares about as many frames with REINFORCE as two
REINFORCE runs share with each other.} Each policy of \cref{tab:main_x3d}
keeps four of ten candidates on the 4,916 ActivityNet-v1.3 validation videos.
The columns give the share of videos with runs of at least two, three or four
adjacent candidates, the share that retain either endpoint, and the share of
retained frames inside annotated action intervals. The final columns give
shared frames and top-1 agreement with REINFORCE, paired by seed. REINFORCE's
own row compares different seeds. Values with $\pm$ give mean and standard
deviation over three seeds. The uniform rule is deterministic. The random row
gives exact values for a uniformly random subset of four candidates, which no
classifier evaluates (dash).}
\label{tab:supp_x3d_selection}
\centering\small
\begin{tabular}{@{}lccccccc@{}}
\toprule
& \multicolumn{3}{c}{Consecutive kept candidates} & & & \multicolumn{2}{c}{Against REINFORCE} \\
\cmidrule(lr){2-4}\cmidrule(l){7-8}
Frame selection & 2 & 3 & 4 & First or last & In action & Shared frames & Same top-1 \\
\midrule
Random 4 of 10 & 83.3 & 23.3 & 3.3 & 66.7 & 65.6 & 1.60 & -- \\
Uniform & 0.0 & 0.0 & 0.0 & 0.0 & 69.9 & 1.69\sd{0.01} & 72.4\sd{0.2} \\
Zero mask, top-$N$ & 92.6\sd{0.8} & 37.7\sd{1.5} & 9.0\sd{0.9} & 10.6\sd{1.7} & 73.6\sd{0.2} & 1.80\sd{0.07} & 72.1\sd{1.1} \\
REINFORCE (OCSampler) & 90.1\sd{0.2} & 33.1\sd{1.1} & 8.0\sd{0.3} & 41.7\sd{0.5} & 73.4\sd{0.2} & 3.37\sd{0.01} & 89.0\sd{0.1} \\
\ours, sampled (ours) & 90.5\sd{0.1} & 33.9\sd{0.8} & 8.1\sd{0.5} & 43.9\sd{0.2} & 73.4\sd{0.1} & 3.37\sd{0.03} & 88.8\sd{0.6} \\
\bottomrule
\end{tabular}
\end{table*}

\begin{table}[t]
\caption{\textbf{Sampled \ours gives similar mAP across the tested learning rates and
samplers.} The left block gives mean validation mAP over three seeds for four
of ten X3D-S frames (standard deviation at most 0.38, and 0.50 for the
score-function estimators with systematic sampling). The right block measures
gradient variance during training and at fixed policy weights. At fixed
weights, ``cos'' compares the mean estimate with the exact gradient of its own
objective. Parentheses mark cross-entropy training variances. Dashes mark
sampler settings that do not apply to deterministic top-$N$.}
\label{tab:video_x3d_estimators}
\centering
\resizebox{\linewidth}{!}{%
\begin{tabular}{@{}l ccc cc c cc@{}}
\toprule
 & \multicolumn{3}{c}{OCSampler's sampler} & \multicolumn{2}{c}{$10^{-3}$} & train. & \multicolumn{2}{c}{fixed weights} \\
\cmidrule(lr){2-4}\cmidrule(lr){5-6}\cmidrule(l){8-9}
Gradient & $3\!\cdot\!10^{-4}$ & $10^{-3}$ & $3\!\cdot\!10^{-3}$ & cond.\ Poisson & systematic & tr\,Cov & tr\,Cov & cos \\
\midrule
\multicolumn{9}{@{}l}{\emph{Unbiased for $\nabla\E_\pi[r]$}} \\
REINFORCE (OCSampler) & 75.59 & 75.70 & 74.51 & 75.68 & 71.97 & 15.2 & 27.2 & 0.81 \\
REINFORCE, \ours baseline & 75.56 & 75.70 & 75.58 & \textbf{75.93} & 72.44 & 9.2 & 18.8 & 0.82 \\
MuProp with \ours, $\eta{=}1$ & 75.65 & 75.87 & 75.29 & 75.73 & 72.29 & 8.3 & 17.0 & 0.83 \\
MuProp with \ours, $\eta^*$ & 75.76 & 75.85 & 75.65 & 75.79 & 72.63 & 6.8 & 14.5 & 0.85 \\
\midrule
\multicolumn{9}{@{}l}{\emph{Biased}} \\
REINFORCE $+\,0.3\,$\ours & 75.68 & 75.96 & 75.32 & 75.89 & 73.10 & 13.0 & 9.7 & 0.75 \\
\ours, sampled masks & \textbf{76.06} & \textbf{76.09} & \textbf{76.18} & 75.87 & \textbf{76.02} & (13.2) & 3.0 & 0.91 \\
\ours, top-$N$ ($\tau{=}1$) & 74.86 & 74.49 & 73.87 & -- & -- & (12.7) & 0 & 0.45 \\
\bottomrule
\end{tabular}
}
\end{table}

\paragraph{Estimators.} \Cref{tab:video_x3d_estimators} isolates the
gradient.
``train. tr Cov'' is the trace of the minibatch gradient covariance measured
during training with OCSampler's sampler at $10^{-3}$ over epochs 0--9. This
measurement includes data noise. For the fixed-weight columns, we use the
REINFORCE policy and score all 210 sets for each of 128 training videos with
test-time transforms. This gives the exact gradient $g^*$ of the estimator's
objective: expected reward, or expected cross-entropy for \ours. We draw each
estimate 32 times, resampling only the set. At fixed weights, ``tr Cov'' is
relative to $\|g^*\|^2$, and ``cos'' is the cosine similarity between the mean
estimate and $g^*$. The entries for MuProp with $\eta{=}1$ and for REINFORCE
$+\,0.3\,$\ours with the two other samplers come from a separate run, as do
the fixed-weight results for the latter; in that run, the remaining estimators
differ from the tabulated values by at most 1.1 in ``tr Cov'' and 0.06 in
``cos''. The \ours network is a good model of the reward: at fixed policy weights,
MuProp with $\eta^*$ ($0.77$--$0.90$) reduces the gradient variance by
1.3--2.4$\times$ relative to REINFORCE with the baseline of OCSampler, across
the initial policy, six trained policies and both transforms. During training,
the reduction is 1.6--2.2$\times$. The reduction does not raise accuracy at the learning rates
chosen on held-out data (75.56--75.85 against 75.59), but it removes the
sensitivity to a large step: at $3\cdot10^{-3}$ REINFORCE drops to 74.51 and
the unbiased hybrids keep 75.29--75.65 ($+0.78$ to $+1.14$ in paired seeds).
\ours on sampled masks has a fixed-weight variance of 2.5--3.4 relative to
its gradient, an order of magnitude below the unbiased estimators, and its
mean is within 0.87--0.95 in cosine of the gradient of the expected loss,
against 0.26--0.52 for the top-$N$ straight-through. The score-function estimators are the most sensitive to the sampler. Under
systematic sampling, at most $T$ of the 210 sets have positive probability for
a given score vector, so their training variance rises to 300--550 and their
mAP falls to 71.97--72.63, close to the uniform rule's 72.22. Sampled \ours
keeps 76.02 mAP with the same sampler.

\paragraph{Joint training.} OCSampler freezes the classifier in stage~II
because REINFORCE needs a stationary reward. Training the classifier from the
Kinetics-400 weights together with the selection gives, over three seeds
(sharing the stage-I skim network) and without tuning, $73.55\pm0.08$ for
top-$N$ \ours, $69.46\pm0.17$ for a classifier trained on the uniform clip and
$69.09\pm0.30$ for the zero mask, 4.46 points below \ours on every seed
(4.09--4.84). The joint mode of OCSampler (REINFORCE and cross-entropy on the
sampled clip) reaches $73.17\pm0.80$ (72.26--73.76), within noise of top-$N$
\ours (paired difference $0.39\pm0.84$); it runs with the policy
log-probabilities in double precision, without which its scores became
non-finite in the third epoch (no step was skipped in any reported run). All joint arms stay below the two-stage protocol, which remains the reference
for the frame-selection comparisons.

\paragraph{Other budgets and a second classifier.} At two and six frames, we train a separate stage-I classifier for each budget
and reuse the skim network from the four-frame setting. Held-out data are used
to choose the learning rate from $\{3,10\}\cdot10^{-4}$ and, for top-$N$
\ours, $\tau$ from $\{0.3,1,3\}$. For the second stage-I classifier at
$N{=}4$, trained with a different seed, we reuse the settings chosen with the
first classifier (\cref{tab:video_x3d}). REINFORCE $+\,0.3\,$\ours also uses
its $N{=}4$ configuration at $N{=}2$ and 6. The order of the three main signals
holds in every setting. \ours on sampled masks is above or level with
REINFORCE in paired seeds, by $+0.44$ [$+0.22$, $+0.67$] at $N{=}2$, $+0.32$
[$+0.00$, $+0.54$] with the second classifier and $+0.05$ [$-0.27$, $+0.43$]
at $N{=}6$, and top-$N$ \ours is below REINFORCE in every seed ($-2.07$,
$-1.20$, $-1.17$). The lead of sampled \ours shrinks as $N$ grows: at $N{=}6$
the score-function rules, soft \ours and sampled \ours are within 0.24
points of each other. MuProp with $\eta^*$ again stays close to REINFORCE
($+0.11$, $+0.30$, $-0.19$), and so does REINFORCE $+\,0.3\,$\ours (69.52, 75.71
and 78.68 against 69.45, 75.60 and 78.74). The zero mask costs most when few frames
survive: at $N{=}2$ it is 8.92 points below REINFORCE and 1.25 below the
uniform rule, at $N{=}6$ 1.67 below REINFORCE and above uniform. At its own
budget, $N{=}6$, the released OCSampler sampler is better than every rule we
learn (79.09 against 78.79); at $N{=}2$ it is below REINFORCE (68.99 against
69.45).

\subsection{Frame selection in Uni-AdaFocus}
\label{sec:supp_video_adafocus}

\paragraph{Host.} Uni-AdaFocus \cite{wang2024uniadafocus} runs a cheap
\emph{glance} network on a coarse view of the whole video, and a learned
policy uses its features to choose the frames, and within them the patches,
that a large \emph{local} network processes; the prediction is the sum of the
logits of both networks. We keep the temporal part of its X3D variant
\cite{feichtenhofer2020x3d} and remove the spatial policy, the patches and the
early exit. From $T_0{=}48$ candidate frames per video (ActivityNet-v1.3, the
splits and candidate sampling of \cref{sec:supp_video_frames}, $160^2$), the
glance X3D-S sees $T_G{=}16$, one per bin of three candidates. A temporal
convolution over its detached frame features scores the bins, and the
selector of Uni-AdaFocus turns the scores into $K{=}16$ of the 48 candidates:
it reads the inverse of the cumulative distribution of the scores at $K$
stratified thresholds, interpolating linearly within bins (centered thresholds
at test time). The local X3D-L runs on the selected frames in temporal order,
\ie, on a compacted clip. Per video the glance network costs 2.50 and the
local network 4.82 GFLOPs (fvcore, without the policy network).

\paragraph{What the policy learns from.} The policy loss of Uni-AdaFocus,
$L_{\mathrm{temporal}}$, is a Monte Carlo proxy on the frame logits of the
glance network: it weights them by draws from the softmax of the scores and
applies the cross-entropy. The local network, which runs on the selected
frames, gives the policy no signal. We add one: a \emph{policy pass} of X3D-L
with \ours in every temporal convolution (as in
\cref{sec:supp_video_frames}) on all 48 candidates at keep values
$m+p-\sg(p)$, where $m$ is the mask the selector drew and $p$ its marginals.
At binary keep values the pass computes X3D-L on the compacted clip
(\cref{prop:block}), so its forward is the deployed network and only the
gradient flows to $p$. The loss is the cross-entropy of the pass logits plus
the detached glance logits. The pass runs the local network in evaluation mode with frozen parameters, so
the dropped candidates never enter batch statistics and the networks train
identically in every arm. To limit memory use, the policy pass processes 8 of
the 32 videos in each batch. The linear interpolation of the
selector equals a piecewise constant density over the glance bins, which
gives the marginals over the 48 candidates; with collisions between
thresholds they are only approximately $\min(1,Kq)$ (error up to 0.075 for a
mild policy and 0.75 for a peaked one), so we also use exact-budget LapSum
marginals \cite{struski2025lapsum} with systematic sampling, whose marginals
are exact. \Cref{tab:video_adafocus} compares $L_{\mathrm{temporal}}$ alone, REINFORCE
with the reward of OCSampler and the uniform clip as its baseline, and several
ways to train through \ours. The policy pass is used either alone or with
$L_{\mathrm{temporal}}$. With $L_{\mathrm{temporal}}$, we test \ours, \ours at
$\lambda{=}0$ and a soft forward pass at $p$ without straight-through. The
glance-only variant leaves out X3D-L and replaces $L_{\mathrm{temporal}}$ with
the cross-entropy of X3D-S on its 16 frames, at keep values of 5 sampled
frames. $L_{\mathrm{temporal}}$ alone;
REINFORCE with the reward of OCSampler and the uniform clip as baseline;
$L_{\mathrm{temporal}}$ plus the policy pass with \ours, with \ours at
$\lambda{=}0$, or at $p$ without straight-through (soft);
the policy pass alone; and, without X3D-L, \ours in the glance network: the
cross-entropy of X3D-S on its 16 frames at keep values of 5 sampled frames
replaces $L_{\mathrm{temporal}}$. As hybrids (\cf~\cref{sec:supp_video_frames})
we add the policy pass to the glance \ours loss, weight the policy pass by
0.1 or 0.3 next to $L_{\mathrm{temporal}}$, add REINFORCE to
$L_{\mathrm{temporal}}$, and use MuProp \cite{gu2016muprop} with the \ours
X3D-L at $p=\min(1,Kq)$ as the model of the reward. Its control variate is
linear in the numbers $n_t$ of thresholds that hit each candidate before
collisions are resolved; the strata cover $[0,1)$ once, so $\E[n]=Kq$ exactly
and the estimator is unbiased for every $\eta$ although the reward reads the
mask after collisions (enumerating all hits for $K{=}4$ of 6 candidates with
collisions: relative error below $10^{-14}$). We use the variance-minimizing
$\eta^*$ of \cref{sec:supp_video_frames}.

\paragraph{Training and evaluation.} All networks train together from the
Kinetics-400 weights with the losses of Uni-AdaFocus on the drawn compacted
clip: the cross-entropy of the summed logits and of the mean frame logits of
each network. We use the optimizer of the official Kinetics script, SGD with
momentum 0.9, learning rate $0.0025$ at batch 32 (0.005 at 64), scaled by 0.5
for the glance network and 0.4 for the policy, weight decay $5{\cdot}10^{-5}$,
gradient clipping at 20, dropout 0.5, a cosine schedule over 50 epochs with
one warm-up epoch, and bfloat16. We report the mAP of the final checkpoint
with centered thresholds and one center crop, and the same networks with a
flat policy (``uniform rule''), which separates what the policy adds at test
time from what training with it does to the networks. The ratio of the
policy learning rate is chosen per arm on the 1000 held-out training videos of
\cref{sec:supp_video_frames}, from $\{0.12,0.4,1.2\}$; the hybrids keep the
official 0.4, and the weights of the policy pass are chosen on the same videos.
The chosen configuration is trained on the full training set with three
seeds.
The setting costs 7.32 GFLOPs per video.

\begin{table}[t]
\caption{\textbf{Training the glance policy with \ours improves local X3D-L mAP over
the native temporal loss.} Uni-AdaFocus uses 16 of 48 frames for X3D-L and 16
glance frames for X3D-S on ActivityNet-v1.3, with spatial selection disabled.
``Uniform rule'' evaluates the trained networks with a flat policy. ``Local
only'' uses X3D-L logits alone. Validation mAP gives the mean and standard
deviation over three seeds. ``No policy'' has no learning-rate ratio or
separate flat-policy evaluation. The ``dev'' column gives mAP for one seed on
the 1,000 held-out training videos used to select the policy learning-rate
ratio. The quantile selector is that of Uni-AdaFocus; the policy-pass arms
other than the quantile row use LapSum.}
\label{tab:video_adafocus}
\centering
\resizebox{\linewidth}{!}{\begin{tabular}{@{}l c c ccc@{}}
\toprule
 & & dev & \multicolumn{3}{c}{val, mean over 3 seeds} \\
\cmidrule(lr){3-3}\cmidrule(l){4-6}
Policy signal & lr ratio & mAP & mAP & uniform rule & local only \\
\midrule
No policy (uniform) & -- & 84.04 & 82.73{\scriptsize$\pm$0.10} & -- & 78.04 \\
$L_{\mathrm{temporal}}$ (Uni-AdaFocus) & 0.4 & 86.39 & 83.93{\scriptsize$\pm$0.23} & 82.78 & 79.52 \\
REINFORCE (reward of OCSampler) & 0.4 & 85.79 & 83.58{\scriptsize$\pm$0.66} & 83.18 & 77.09 \\
\midrule
\multicolumn{6}{@{}l}{\emph{Policy pass through the local X3D-L, $m+p-\sg(p)$}} \\
$L_{\mathrm{temporal}}$ + \ours, soft forward ($p$) & 1.2 & 85.57 & 83.78{\scriptsize$\pm$0.20} & 82.62 & 79.56 \\
$L_{\mathrm{temporal}}$ + \ours, quantile selector & 0.12 & \textbf{86.89} & 83.90{\scriptsize$\pm$0.20} & 82.83 & 79.52 \\
$L_{\mathrm{temporal}}$ + \ours, LapSum selector & 0.12 & 85.73 & 83.74{\scriptsize$\pm$0.17} & 82.77 & 79.32 \\
$L_{\mathrm{temporal}}$ + $0.3\,$\ours, LapSum selector & 0.4 & 86.34 & 83.93{\scriptsize$\pm$0.08} & 82.63 & 79.53 \\
$L_{\mathrm{temporal}}$ + \ours, $\lambda{=}0$ & 0.12 & 85.85 & 83.78{\scriptsize$\pm$0.34} & 82.74 & 79.54 \\
\ours alone & 1.2 & 86.25 & 83.63{\scriptsize$\pm$0.30} & 82.39 & 79.52 \\
\midrule
\multicolumn{6}{@{}l}{\emph{Policy signal from the glance X3D-S}} \\
\ours on the glance frames & 0.12 & 86.36 & 84.11{\scriptsize$\pm$0.38} & 82.60 & 80.34 \\
\ours on the glance frames + policy pass & 0.4 & 86.61 & \textbf{84.33}{\scriptsize$\pm$0.24} & 82.56 & 80.72 \\
\midrule
\multicolumn{6}{@{}l}{\emph{Hybrids with REINFORCE}} \\
$L_{\mathrm{temporal}}$ + REINFORCE & 0.4 & 85.95 & 83.69{\scriptsize$\pm$0.34} & 82.98 & 79.36 \\
MuProp with the \ours X3D-L, $\eta^*$ & 0.4 & 85.50 & 83.84{\scriptsize$\pm$0.31} & 83.49 & 78.18 \\
\bottomrule
\end{tabular}
}
\end{table}

\begin{table*}[t]
\caption{\textbf{The glance-trained \ours policy improves the local classifier at every
tested budget.} The local Uni-AdaFocus classifier reads $K$ of 48
ActivityNet-v1.3 frames. The glance network reads 16 frames throughout. The
table gives validation mAP over three seeds, with standard deviations for
fused predictions. ``Local only'' evaluates X3D-L alone. The no-policy row has
no policy learning-rate ratio. The $K{=}16$ columns repeat
\cref{tab:video_adafocus}.}
\label{tab:video_adafocus_budget}
\centering
\footnotesize
\begin{tabular}{@{}l c cc cc cc@{}}
\toprule
 & & \multicolumn{2}{c}{$K{=}4$} & \multicolumn{2}{c}{$K{=}8$} & \multicolumn{2}{c}{$K{=}16$} \\
\cmidrule(lr){3-4}\cmidrule(lr){5-6}\cmidrule(l){7-8}
Policy signal & lr ratio & mAP & local only & mAP & local only & mAP & local only \\
\midrule
No policy (uniform) & -- & 82.59{\scriptsize$\pm$0.11} & 61.57 & 82.74{\scriptsize$\pm$0.05} & 71.80 & 82.73{\scriptsize$\pm$0.10} & 78.04 \\
$L_{\mathrm{temporal}}$ (Uni-AdaFocus) & 0.4 & 82.54{\scriptsize$\pm$0.03} & 64.43 & 82.74{\scriptsize$\pm$0.05} & 73.33 & 83.93{\scriptsize$\pm$0.23} & 79.52 \\
REINFORCE (reward of OCSampler) & 0.4 & 82.51{\scriptsize$\pm$0.10} & 62.83 & 82.79{\scriptsize$\pm$0.42} & 72.78 & 83.58{\scriptsize$\pm$0.66} & 77.09 \\
\midrule
\multicolumn{8}{@{}l}{\emph{Policy pass through the local X3D-L, $m+p-\sg(p)$}} \\
$L_{\mathrm{temporal}}$ + \ours, quantile selector & 0.12 & \textbf{82.75}{\scriptsize$\pm$0.10} & 65.52 & 83.00{\scriptsize$\pm$0.20} & 74.44 & 83.90{\scriptsize$\pm$0.20} & 79.52 \\
$L_{\mathrm{temporal}}$ + \ours, LapSum selector & 0.12 & 82.69{\scriptsize$\pm$0.18} & 65.33 & 82.89{\scriptsize$\pm$0.25} & 74.19 & 83.74{\scriptsize$\pm$0.17} & 79.32 \\
\midrule
\ours on the glance frames & 0.12 & 82.57{\scriptsize$\pm$0.10} & 67.43 & \textbf{83.50}{\scriptsize$\pm$0.33} & 76.32 & \textbf{84.11}{\scriptsize$\pm$0.38} & 80.34 \\
\bottomrule
\end{tabular}

\end{table*}

\paragraph{Results.} $L_{\mathrm{temporal}}$ is a strong baseline in this
host: it is the policy loss that Uni-AdaFocus and its convolutional networks
were designed and tuned with, so it is not obvious which \ours signal should
replace or complement it. We therefore train several
(\cref{tab:video_adafocus}) and select on dev. The two best configurations on
dev are both \ours variants: the policy pass with the quantile selector
(86.89) and \ours in the glance network with the policy pass (86.61), against
86.39 for $L_{\mathrm{temporal}}$. Both hold up on the validation set (three
seeds). The first is level with $L_{\mathrm{temporal}}$ at $K{=}16$ (83.90
against 83.93) and ahead of it at the smaller budgets ($+0.26$ at $K{=}8$ and
$+0.21\pm0.07$, on every seed, at $K{=}4$;
\cref{tab:video_adafocus_budget}); the second reaches 84.33, 0.40 points above
$L_{\mathrm{temporal}}$. Both are ahead of REINFORCE (83.58). All learned
signals lie within 83.58--84.33 mAP. Within each run,
the policy adds 0.35--1.77 points at test time over a flat policy with the same
networks. Trained without a policy, the same networks reach 82.73, 0.85--1.60
points below the learned signals and within the range of the learned networks
under a flat policy (82.39--83.49).

The most consistent result, on dev and on validation, is which frames the
policy gives to the local network. We keep X3D-L's weights fixed within each run when comparing its learned and
flat policies. \ours in the glance network improves local-only mAP by 2.19
points, or 2.49 points with the policy pass (2.05--2.70 across seeds). The
corresponding differences are 1.27 for $L_{\mathrm{temporal}}$, $-1.07$ for
REINFORCE, 0.53 for $L_{\mathrm{temporal}}$ plus REINFORCE, and $-0.37$ for
MuProp. REINFORCE is 3.5 points below the flat policy in one seed. \ours in the glance network computes X3D-S on the deployable
5-frame clip instead of averaging the logits of independent frames as
$L_{\mathrm{temporal}}$ does. Combined with the policy pass it gives the best
local network and the best fused accuracy among the \ours arms (84.33), 0.40
points above $L_{\mathrm{temporal}}$.

The policy pass through X3D-L with \ours does not improve on
$L_{\mathrm{temporal}}$ (83.63--83.93, also at weight 0.3 and with a soft
forward pass), and dropping the convolution term of the gradient
($\lambda{=}0$, 83.78) changes nothing against $\lambda{=}1$ (83.74): in this
host the signal reaches $p$ through the weighted pooling, not through the
compacted convolutions. MuProp with the \ours reward model matches REINFORCE
(83.84 against 83.58), and adding REINFORCE to $L_{\mathrm{temporal}}$ does not
raise it (83.69). On dev, at the official learning
rate, all learned rules except REINFORCE almost never select the first or
last candidate (at most 5\% of the videos against 33--39\% in the middle;
REINFORCE 17\% and 11\%).

\paragraph{Frame budget.} We repeat the main arms with $K{=}4$ and $K{=}8$
of the 48 candidates for X3D-L (\cref{tab:video_adafocus_budget}), three seeds
each, at the learning-rate ratios chosen at $K{=}16$. The local network costs roughly $K/16$ of its 4.82
GFLOPs, the glance network stays at 2.50. The local X3D-L branch benefits from the glance-trained \ours policy at every
budget: 67.43, 76.32 and 80.34 mAP at $K{=}4$, 8 and 16, against 64.43, 73.33
and 79.52 with $L_{\mathrm{temporal}}$ and 61.57, 71.80 and 78.04 without a
policy. At $K{=}4$, fused mAP ranges from 82.51 to 82.75 across all configurations,
including the networks trained without a policy, because X3D-L alone reaches
only 61.6--67.4 mAP and the summed logits largely follow the glance network.
At $K{=}8$, \ours in the glance network reaches 83.50 fused mAP. It improves
on $L_{\mathrm{temporal}}$ in every seed, with a paired gain of $0.76\pm0.38$
points, whereas $L_{\mathrm{temporal}}$ and the networks trained without a
policy both score 82.74. \ours in the policy pass is level with or slightly
ahead of $L_{\mathrm{temporal}}$ at the two smaller budgets ($+0.15$ to
$+0.26$; quantile selector at $K{=}4$: $+0.21\pm0.07$, every seed) and
slightly behind it at $K{=}16$ ($-0.19$ to $-0.03$).

\FloatBarrier
\section{Derivations and variants of \ours}
\label{sec:supp_derivation}

We derive the expected convolution window and its backward pass, extend the
operator to centered kernels, and relate step gating to the expected decay of
the scan. At a retained position, the causal convolution reads the most recent
surviving tokens. Under independent keep decisions, their expected values give
the exact expected pre-activation. Throughout, $c_t=\text{\ours}(x,p)_t$
denotes the output of \cref{eq:probconv}, before $\sigma(b+\cdot)$, and
$q_t=1-p_t$.

\paragraph{Equivalence of a block and a stack.}
\label{sec:supp_block_proof}
Fix a binary mask and let $i_j$ be the $j$-th kept position. By
\cref{eq:recurrence}, the convolution at $i_j$ reads the same projected
tokens as the convolution at compacted index $j$. Its bias and activation
therefore also agree. For Mamba-2 this applies to every convolved input.
Between $i_{j-1}$ and $i_j$, every dropped step is the identity by
\cref{eq:scan} with $\Delta_t$ gated to zero. Given equal initial states,
induction over the retained positions gives identical scan updates and
states at $i_j$ and $j$. The pointwise readout and residual operations agree
at these positions. The reverse direction follows by applying the same
argument to the reversed sequence and mask. For a stack, induct over
blocks, restricting the statement to the new subset after each nested cut.
The argument assumes that intervening operations agree at corresponding
retained positions, including any position-dependent inputs.

\paragraph{A worked example.} Take $K{=}3$ taps and the output at position
$t{=}4$, conditioned on $m_4{=}1$. The taps $w_1,w_2$ read the two most recent
survivors among $x_1,x_2,x_3$, which depend on the eight masks of the prefix
(0 marks a missing survivor):
\begin{center}
\footnotesize
\begin{tabular}{@{}cccc@{}}
\toprule
$(m_1,m_2,m_3)$ & compacted & $(w_1,w_2)$ read & probability\\
\midrule
$(1,1,1)$ & $x_1x_2x_3x_4$ & $(x_3,x_2)$ & $p_1p_2p_3$\\
$(0,1,1)$ & $x_2x_3x_4$ & $(x_3,x_2)$ & $q_1p_2p_3$\\
$(1,0,1)$ & $x_1x_3x_4$ & $(x_3,x_1)$ & $p_1q_2p_3$\\
$(1,1,0)$ & $x_1x_2x_4$ & $(x_2,x_1)$ & $p_1p_2q_3$\\
$(0,0,1)$ & $x_3x_4$ & $(x_3,0)$ & $q_1q_2p_3$\\
$(0,1,0)$ & $x_2x_4$ & $(x_2,0)$ & $q_1p_2q_3$\\
$(1,0,0)$ & $x_1x_4$ & $(x_1,0)$ & $p_1q_2q_3$\\
$(0,0,0)$ & $x_4$ & $(0,0)$ & $q_1q_2q_3$\\
\bottomrule
\end{tabular}
\end{center}
Collecting the coefficient of every $x_j$ gives
\begin{equation}
\begin{aligned}
c_4 = {}& w_0\odot x_4 + w_1\odot\big(p_3x_3+q_3p_2x_2+q_3q_2p_1x_1\big)\\
&+ w_2\odot\big(p_3p_2x_2+(p_3q_2+q_3p_2)\,p_1x_1\big).
\end{aligned}
\label{eq:supp_example}
\end{equation}
The first two masks differ but give the same computation: a kernel with $K$
taps reads only which tokens are the $K{-}1$ most recent survivors, and this
collapses the $2^{t-1}$ masks of a prefix into $K{-}1$ expectations. Three
steps of \cref{eq:recurrence}, $a_1=(p_1x_1,\,0)$,
$a_2=(p_2x_2+q_2p_1x_1,\,p_2p_1x_1)$ and
$a_3=(p_3x_3+q_3a^{(1)}_2,\,p_3a^{(1)}_2+q_3a^{(2)}_2)$, give the two
brackets of \cref{eq:supp_example} as $a^{(1)}_3$ and $a^{(2)}_3$.

\paragraph{Proof of \cref{prop:recurrence} and of \cref{eq:probconv}.} Let
$\hat a_t=(s^{(1)}_{t+1},\dots,s^{(K-1)}_{t+1})$ collect the $K{-}1$ most
recent survivors among positions $\le t$ (zero where fewer survive), so that
$a_t=\E[\hat a_t]$. A kept token becomes the most recent survivor and moves
the older ones down a rank, and a dropped token changes nothing:
\begin{equation}
\hat a_t=\hat a_{t-1}+m_t\big((S\hat a_{t-1}+e_1x_t)-\hat a_{t-1}\big).
\label{eq:supp_pathwise}
\end{equation}
The state $\hat a_{t-1}$ is a function of $m_{<t}$ and hence independent of
$m_t$, so the expectation of \cref{eq:supp_pathwise} is $a_t=M_ta_{t-1}+p_te_1x_t$,
which is \cref{eq:recurrence}; $\hat a_0=0$ encodes the zero for missing
survivors. At a kept position $t$, the tap $w_r$ of the compacted convolution
multiplies $s^{(r)}_t=\hat a^{(r)}_{t-1}$, which does not depend on $m_t$, so
$\E[s^{(r)}_t\mid m_t{=}1]=a^{(r)}_{t-1}$, and linearity gives the second
equality of \cref{eq:probconv}. At $p_t=0$ the conditioning event is null, and
the right-hand side of \cref{eq:probconv} defines $c_t$ by continuity.

\paragraph{Closed form.} Unrolling the recurrence gives
$a_t=\sum_{j\le t}M_{(j,t]}\,p_je_1x_j$ with $M_{(j,t]}=M_t\cdots M_{j+1}$.
Each factor is $M_\ell=q_\ell I+p_\ell S=\E[S^{m_\ell}]$, and the factors
commute and belong to independent decisions, so their product is the
probability generating function of the number
$N_{j,t}=\sum_{\ell=j+1}^{t}m_\ell$ of survivors in $(j,t]$, evaluated at the
shift:
\begin{equation}
M_{(j,t]}=\E\big[S^{N_{j,t}}\big]
=\sum_{n=0}^{K-2}\mathbb{P}(N_{j,t}{=}n)\,S^n ,
\label{eq:supp_pgf}
\end{equation}
truncated because $S^{K-1}=0$. With $S^ne_1=e_{n+1}$,
\begin{equation}
a^{(r)}_t=\sum_{j\le t}p_j\,\mathbb{P}(N_{j,t}{=}r{-}1)\,x_j ,
\label{eq:supp_closed}
\end{equation}
where $p_j\,\mathbb{P}(N_{j,t}{=}r{-}1)$ is the probability that $x_j$
survives and is followed by exactly $r{-}1$ survivors up to $t$, \ie, that it
is the $r$-th most recent survivor. In the example, $a^{(2)}_3$ weighs $x_2$
by $p_2\,\mathbb{P}(m_3{=}1)=p_2p_3$ and $x_1$ by
$p_1\,\mathbb{P}(m_2{+}m_3{=}1)=p_1(p_2q_3+q_2p_3)$. The coefficients are
Poisson--binomial probabilities that depend on every $p_\ell$ between $j$ and
$t$; evaluated pair by pair, they cost $O(L^2)$ per rank, whereas the
recurrence costs $O(LKD)$ time and $O(KD)$ memory.

\paragraph{Backward pass.} Let $g_t=\partial\mathcal{L}/\partial c_t$; with
the fused epilogue, it is the gradient with respect to $\sigma(b+c_t)$ times
$\sigma'(b+c_t)$. The state $a_t$ enters $c_{t+1}$ through the taps and
$a_{t+1}$ through $M_{t+1}$, so its adjoint $\bar a_t$ obeys, from
$\bar a_L=0$, the reverse recurrence
\begin{equation}
\bar a_t=M_{t+1}^{\top}\bar a_{t+1}
+\big(w_1\odot g_{t+1},\dots,w_{K-1}\odot g_{t+1}\big),
\label{eq:supp_adjoint}
\end{equation}
an affine scan in reverse order whose maps $M_t^{\top}=q_tI+p_tS^{\top}$ are
polynomials in the up-shift. Besides \cref{eq:pgrad}, whose inner product runs
over ranks and channels, the gradients are
\begin{equation}
\begin{gathered}
\frac{\partial\mathcal{L}}{\partial x_t}=w_0\odot g_t+p_t\,\bar a^{(1)}_t,\qquad
\frac{\partial\mathcal{L}}{\partial b}=\sum_t g_t,\\
\frac{\partial\mathcal{L}}{\partial w_0}=\sum_t g_t\odot x_t,\qquad
\frac{\partial\mathcal{L}}{\partial w_r}=\sum_t g_t\odot a^{(r)}_{t-1},
\end{gathered}
\label{eq:supp_grads}
\end{equation}
with the sums also running over the batch.

\paragraph{Multilinearity and continuity.} Under independent keep decisions, $c_t$ is a polynomial of degree at most one
in each $p_j$, $j<t$. It is the unique multilinear extension of the compacted
convolution from the vertices. \ours is therefore
smooth on the closed cube $[0,1]^L$, and as $p\to m$ it approaches the
compacted convolution at first order in $\lVert p-m\rVert$. Its derivative in
one coordinate is an exact difference,
\begin{equation}
\frac{\partial c_{t'}}{\partial p_t}=c_{t'}\big|_{p_t=1}-c_{t'}\big|_{p_t=0},
\qquad t<t',
\label{eq:supp_difference}
\end{equation}
the effect of keeping token $t$ rather than dropping it, averaged over the
other decisions; \cref{eq:pgrad} propagates this difference through the
adjoint. At a vertex, where the straight-through estimator evaluates the
gradient, the other decisions are fixed, and \cref{eq:supp_difference} is the
exact change of the layer's compacted output when the decision on token $t$
alone is flipped. The exact difference applies to the convolution before the activation. After
the activation and the rest of the network, the gradient gives a first-order
approximation to the change in loss, which we test in
\cref{fig:supp_cobra_gradfid}.

\paragraph{Reference algorithm and scan.} The sequential reference reads the
state before it writes it:
\begin{center}
\small
\begin{tabular}{@{}l@{}}
\toprule
$a\leftarrow0\in\mathbb{R}^{(K-1)\times D}$\\
\textbf{for} $t=1,\dots,L$\textbf{:}\\
\quad $c_t\leftarrow w_0\odot x_t+\sum_{r=1}^{K-1}w_r\odot a^{(r)}$\\
\quad \textbf{for} $r=K{-}1,\dots,2$\textbf{:}\enspace
$a^{(r)}\leftarrow p_t\,a^{(r-1)}+q_t\,a^{(r)}$\\
\quad $a^{(1)}\leftarrow p_t\,x_t+q_t\,a^{(1)}$\\
\textbf{return} $\sigma(b+c_{1:L})$\\
\bottomrule
\end{tabular}
\end{center}
Updating the ranks in place from $r{=}K{-}1$ downwards keeps the new token
from being counted twice. The algorithm runs in $O(LKD)$ time and $O(KD)$
memory and is the float64 reference of our tests. The parallel forms compose
the affine maps $(M,\beta)$ of \cref{eq:recurrence},
$(M_2,\beta_2)\circ(M_1,\beta_1)=(M_2M_1,\,M_2\beta_1+\beta_2)$, which is
associative; by \cref{eq:supp_pgf}, a composed map is a polynomial of degree
$K{-}2$ in $S$, stored with $K{-}1$ coefficients, plus a $(K{-}1)\times D$
offset, and an exclusive scan returns every $a_{t-1}$. The backward pass scans
\cref{eq:supp_adjoint} in the same algebra with $S^{\top}$.

\paragraph{Numerical check.} \Cref{tab:exactness} checks whole blocks
(\cref{prop:block}); here we check the operator against the enumeration of
all $2^{14}{-}1$ keep patterns of six random sequences ($L{=}14$, $K{=}4$,
$D{=}8$, some $p_t\in\{0,1\}$). In float64, the algorithm above matches it to
$2\cdot10^{-15}$ in the output and $2.5\cdot10^{-14}$ in the gradients with
respect to $x$, $p$, $w$ and $b$ (relative $\ell_2$ error
$\le1.2\cdot10^{-15}$). The float32 CUDA kernel (\cref{sec:supp_kernel}), with
and without the fused bias and SiLU, has relative $\ell_2$ errors
$\le7.4\cdot10^{-8}$ in the output and $\le1.8\cdot10^{-7}$ in the gradients
against the enumeration, and $\le7.4\cdot10^{-8}$ and $\le2.0\cdot10^{-7}$
against the float64 algorithm at the size of a Mamba-130M block ($B{=}8$,
$L{=}2048$, $D{=}1536$): float32 rounding ($2^{-24}\approx6\cdot10^{-8}$). Its
unit tests also check binary $p$ against explicit compaction, $p\equiv1$
against \texttt{conv1d}, first-order convergence to the compacted
convolution as $p$ approaches a vertex (log--log slope $0.93$), half
precision and bitwise determinism.

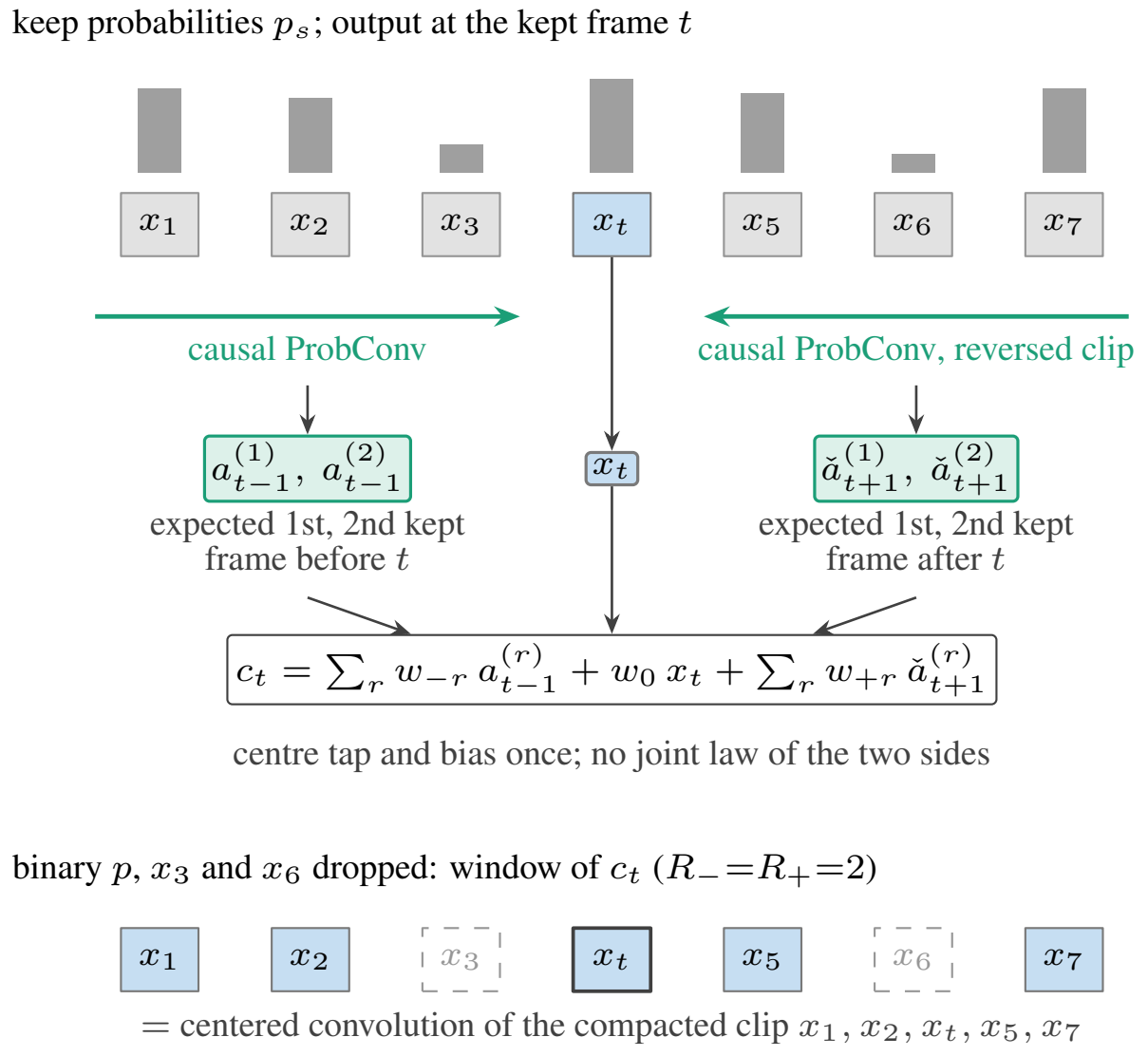
\begin{figure}[t]
\centering
\definecolor{pcGreen}{HTML}{1B9E77}%
\definecolor{pcGreenL}{HTML}{DDF1EA}%
\definecolor{pcInk}{RGB}{64,64,64}%
\definecolor{pcTok}{RGB}{226,226,226}%
\definecolor{pcLine}{RGB}{150,150,150}%
\definecolor{pcWin}{RGB}{198,222,242}%
\begin{tikzpicture}[x=1mm,y=1mm,font=\footnotesize,line join=round,
  tok/.style={draw=pcLine,fill=pcTok,minimum width=5.4mm,minimum height=4.4mm,inner sep=0pt,font=\scriptsize,line width=0.4pt},
  cur/.style={tok,fill=pcWin,draw=pcInk!70},
  win/.style={tok,fill=pcWin,draw=pcInk!70},
  gone/.style={tok,fill=white,draw=pcLine,dashed,text=pcLine},
  outbox/.style={draw=pcGreen,fill=pcGreenL,rounded corners=1pt,inner sep=1.2pt,font=\scriptsize,line width=0.6pt},
  flow/.style={draw=pcInk,line width=0.5pt,-{Stealth[length=1.4mm,width=1.1mm]}},
  scan/.style={draw=pcGreen,line width=0.9pt,-{Stealth[length=1.8mm,width=1.4mm]}},
  lab/.style={font=\scriptsize,text=pcInk,inner sep=0.5pt,align=center},
  ttl/.style={font=\scriptsize,text=black,inner sep=0.5pt,anchor=west},
]
\node[ttl] at (0,53.5) {keep probabilities $p_s$; output at the kept frame $t$};
\foreach \i/\p in {1/0.90,2/0.80,3/0.30,4/1.00,5/0.85,6/0.20,7/0.90}{
  \fill[pcInk!50] ({\i*10.5-1.5},43.2) rectangle ++(3,{\p*6.5});
}
\foreach \i in {1,2,3,5,6,7}{\node[tok] (x\i) at ({\i*10.5},39.6) {$x_\i$};}
\node[cur] (x4) at (42,39.6) {$x_t$};

\draw[scan] (6,33.2) -- (35.6,33.2);
\node[lab,text=pcGreen,anchor=north] at (20.8,32.1) {causal \ours};
\draw[scan] (78,33.2) -- (48.4,33.2);
\node[lab,text=pcGreen,anchor=north] at (63.2,32.1) {causal \ours, reversed clip};

\node[outbox] (past) at (20.8,22.6) {$a^{(1)}_{t-1},\ a^{(2)}_{t-1}$};
\node[outbox,draw=pcInk!70,fill=pcWin] (mid) at (42,22.6) {$x_t$};
\node[outbox] (fut) at (63.2,22.6) {$\check a^{(1)}_{t+1},\ \check a^{(2)}_{t+1}$};
\draw[flow] (20.8,28.4) -- (past.north);
\draw[flow] (63.2,28.4) -- (fut.north);
\draw[flow] (x4.south) -- (mid.north);
\node[lab] at (20.8,17.6) {expected 1st, 2nd kept\\[-1pt]frame before $t$};
\node[lab] at (63.2,17.6) {expected 1st, 2nd kept\\[-1pt]frame after $t$};

\node[draw=pcInk,rounded corners=1pt,inner sep=1.6pt,font=\scriptsize] (c) at (42,8.6)
  {$c_t=\sum_{r} w_{-r}\,a^{(r)}_{t-1}+w_0\,x_t+\sum_{r} w_{+r}\,\check a^{(r)}_{t+1}$};
\draw[flow] (past.south) ++(0,-6.6) -- ($(c.north)+(-14,0)$);
\draw[flow] (mid.south) -- (c.north);
\draw[flow] (fut.south) ++(0,-6.6) -- ($(c.north)+(14,0)$);
\node[lab] at (42,2.6) {centre tap and bias once; no joint law of the two sides};

\node[ttl] at (0,-5.4) {binary $p$, $x_3$ and $x_6$ dropped: window of $c_t$ ($R_-{=}R_+{=}2$)};
\foreach \i/\s in {1/win,2/win,3/gone,5/win,6/gone,7/win}{\node[\s] at ({\i*10.5},-11.6) {$x_\i$};}
\node[win,draw=pcInk,line width=0.7pt] at (42,-11.6) {$x_t$};
\node[lab,anchor=north] at (42,-15) {$=$ centered convolution of the compacted clip $x_1,x_2,x_t,x_5,x_7$};
\end{tikzpicture}%
\caption{\textbf{Two causal passes give the centered \ours window.} One pass tracks the
expected retained predecessors and the reverse pass tracks the expected
successors. The center tap and bias enter once. With a binary mask, these
inputs are exactly those of the centered convolution on the compacted clip.
X3D-S and Uni-AdaFocus use this form in their temporal convolutions.}
\label{fig:supp_centered}
\end{figure}

\paragraph{Bidirectional blocks and centered kernels.} In our bidirectional
hosts, the backward block applies the causal operator to the reversed
sequence and probabilities (\cref{sec:gating}); since reversal commutes with
compaction, \cref{prop:recurrence,prop:block} hold in each direction. A
convolution with $R_-$ taps before and $R_+$ taps after the output token, as
in non-causal models, needs a second set of states:
$\check a_t=M_t\check a_{t+1}+p_te_1x_t$, run from $\check a_{L+1}=0$ down
to $t$, holds the expected nearest survivors among positions $\ge t$. The survivors before $t$ depend on $m_{<t}$ and those
after it on $m_{>t}$, so conditioning on $m_t{=}1$ changes neither, and by
linearity
\begin{equation}
\begin{aligned}
c_t={}&w_0\odot x_t+\sum_{r=1}^{R_-}w_{-r}\odot a^{(r)}_{t-1}\\
&+\sum_{r=1}^{R_+}w_{+r}\odot\check a^{(r)}_{t+1}
\end{aligned}
\label{eq:supp_centered}
\end{equation}
requires no joint law of the two neighborhoods: the centered \ours is the
causal one on the sequence plus the causal one on the reversed sequence, with
the centre tap counted once (\cref{fig:supp_centered}). A centered kernel has taps on both sides of the output position; the weights
on the two sides may differ. The two scans are independent and together take
$O(L(R_-{+}R_+)D)$ work, and the backward pass adds the contributions of the
two adjoint scans to $x_t$ and $p_t$. At binary $p$,
\cref{eq:supp_centered} is the non-causal convolution of the compacted
sequence with zero padding, and at $p\equiv1$ the dense one. The Mamba hosts
need only the causal form, and our CUDA kernel implements only that, for
$K{=}4$. The frame selectors in X3D-S and Uni-AdaFocus use the centered form
(\cref{sec:supp_video_frames,sec:supp_video_adafocus}) through a PyTorch
implementation that builds the operators of \cref{eq:supp_centered} as
matrices over the frames.

\paragraph{Step gating and the expected decay.} Hold the inputs
$\Delta_t,B_t,v_t$ of the scan in \cref{eq:scan} fixed. The compacted scan leaves
the state unchanged at a dropped step,
$h_t=h_{t-1}+m_t\big((e^{\Delta_tA}-1)h_{t-1}+\Delta_tB_tv_t\big)$, and since
$h_{t-1}$ depends only on $m_{<t}$,
\begin{equation}
\E h_t=\big(q_t+p_te^{\Delta_tA}\big)\,\E h_{t-1}+p_t\Delta_tB_tv_t .
\label{eq:supp_decay}
\end{equation}
The expected decay accumulated over $(j,t]$ is
$\prod_{\ell=j+1}^{t}(q_\ell+p_\ell e^{\Delta_\ell A})
=\E\big[e^{A\sum_{\ell}m_\ell\Delta_\ell}\big]$, a product of the same form as
\cref{eq:supp_pgf} with the decay $e^{\Delta_\ell A}$ in place of the shift.
Step gating, $\Delta_t\mapsto p_t\Delta_t$, writes the same expected input
$p_t\Delta_tB_tv_t$ but decays by $e^{p_t\Delta_tA}$, so it accumulates
$e^{A\sum_\ell p_\ell\Delta_\ell}$: the decay of the expected elapsed time
instead of the expected decay. The two agree at the vertices. Between the vertices, convexity of the exponential gives
$e^{p_t\Delta_tA}\le
q_t+p_te^{\Delta_tA}$ elementwise. The step-gated decay
factor is therefore at most the expected decay factor. With a large step and
$p_t=\tfrac12$, step gating nearly erases the previous state, while the
expected decay factor retains about half. \Cref{eq:supp_decay} is the expectation of the scan only if its
inputs do not depend on the mask; inside a block they do, through the
convolution, so neither form is the expectation of the block, and both are
exact at the vertices (\cref{prop:block}). Step gating needs only a rescaled
step, which the selective-scan kernels accept. The expected decay of Mamba-1,
whose state dimensions decay at different rates under one step per channel,
is not a rescaled step; we have not evaluated it.

\FloatBarrier
\section{The operator: implementation, cost and dechunking}
\label{sec:supp_kernel}

\paragraph{Implementation.} In our CUDA kernel, each warp processes one (batch, channel) row in tiles of
256 positions, eight per lane. The forward pass stores the rank-$(K{-}1)$
state at the start of each tile. The backward pass recomputes the state inside
each tile while running the reverse scan of \cref{eq:supp_adjoint}. Tap gradients are reduced
deterministically, inputs may be fp16 or bf16 with fp32 accumulation, and the
bias and SiLU are fused into the epilogue. The rank-one state also implements
the dechunker of \cref{prop:dechunk} in fp32.

\begin{table*}[t]
\caption{\textbf{The fused \ours kernel matches dense \texttt{causal-conv1d} for the
Mamba-130M block shape.} Entries give median forward and forward-plus-backward
times in milliseconds over 100 fp32 runs on RTX 4090, including bias and SiLU.
The dense kernels omit keep probabilities. The three blocks use Vim-Ti,
Mamba-130M and a length-32,768 input.}
\label{tab:kernel}
\centering
\small
\begin{tabular}{@{}lcccccc@{}}
\toprule
 & \multicolumn{2}{c}{Vim-Ti block} & \multicolumn{2}{c}{Mamba-130M block} & \multicolumn{2}{c}{$L{=}32$k} \\
\cmidrule(lr){2-3}\cmidrule(lr){4-5}\cmidrule(lr){6-7}
Implementation (fp32, + bias + SiLU) & fwd & fwd+bwd & fwd & fwd+bwd & fwd & fwd+bwd \\
\midrule
cuDNN \texttt{conv1d} (dense, ignores $p$) & 0.11 & 0.27 & 0.57 & 1.87 & 1.07 & 3.62 \\
\texttt{causal-conv1d} (dense, ignores $p$) & 0.06 & 0.20 & 0.26 & 0.61 & 0.49 & 1.16 \\
\midrule
\ours{} (ours), PyTorch scan & 0.19 & 0.66 & 1.06 & 3.84 & 2.08 & 10.36 \\
\ours{} (ours), earlier CUDA prototype & 0.10 & 0.60 & 0.68 & 4.15 & 1.35 & 36.44 \\
\ours{} (ours), fused CUDA kernel & 0.05 & 0.16 & 0.24 & 0.61 & 0.49 & 1.20 \\
\bottomrule
\end{tabular}

\end{table*}

\Cref{tab:kernel} reports the cost of the forward pass alone (fwd) and of the
forward and backward pass together (fwd+bwd).
We time a Vim-Ti block ($B{=}64$, $D{=}384$, $L{=}197$), a Mamba-130M block
($B{=}8$, $D{=}1536$, $L{=}2048$) and a long sequence ($B{=}1$, $D{=}1536$,
$L{=}32768$), all with $K{=}4$ taps. The dense \texttt{causal-conv1d} baseline
is v1.4.0 of Mamba's fused path, which fuses bias and SiLU. Our fused kernel computes the forward and
backward pass of \ours for a Mamba-130M block in 0.61\,ms, the time of the
dense \texttt{causal-conv1d} kernel of Mamba's fused path (0.61\,ms),
although that kernel ignores the probabilities and computes no gradient with
respect to them. On the Vim-Ti shape our kernel takes 0.79$\times$ its time,
at $L{=}32$k 1.04$\times$. It is 3.1$\times$ faster than cuDNN's dense
depthwise convolution with the same epilogue and 6.8$\times$ faster than our
earlier CUDA prototype, an advantage that grows to 30$\times$ at $L{=}32$k.
Both dense kernels agree with each other to $5\cdot10^{-8}$, and the three
implementations of \ours agree to $9\cdot10^{-7}$ in the output and the
gradients.
In bf16 on the Mamba-130M shape, our kernel takes 0.12\,ms for the forward
pass and 0.34\,ms for forward plus backward. The corresponding times are 0.45
and 1.19\,ms for cuDNN, and 0.15 and 0.36\,ms for \texttt{causal-conv1d}.

\subsection{Expected dechunking}
\label{sec:dechunk}

The recurrence for the most recent retained token also gives the expected
smoothing dechunker used in our routing experiments.
In H-Net \cite{hwang2026hnet}, the main network processes only the boundary
tokens, and a dechunker expands its outputs to the original length. The
router computes $p_t=\tfrac12\big(1-\cos(q_{t-1},k_t)\big)$ and places a
boundary where $b_t=[p_t\ge \tfrac12]$. At boundary $s$, the dechunker
smooths the output $z_s$ using $e_s = w_s z_s + (1-w_s)\, e_{s'}$, where
$s'$ is the previous boundary and $w$ is a clipped copy of $p$. Each
position $t$ copies $e$ from the last boundary $\le t$.

\begin{proposition}
\label{prop:dechunk}
If $b_t\sim\mathrm{Bernoulli}(p_t)$ independently, the expected dechunker output is
$\bar u_t = p_t w_t\, z_t + (1-p_t w_t)\,\bar u_{t-1}$, the rank-one state of
\ours with keep probability $p_t w_t$.
\end{proposition}

\noindent\emph{Proof.} At a boundary, the output is
$u_t = w_t z_t + (1-w_t) u_{t-1}$. Between boundaries, $u_t=u_{t-1}$.
Averaging the two updates over $b_t$ gives the recurrence. \hfill$\square$

The same construction factorizes LDARNet's bidirectional dechunker
\cite{ledneva2026ldarnet} into forward and backward rank-one recurrences
(\cref{sec:supp_dechunk}). Straight-through boundaries reproduce the hard
dechunker in the forward pass. The router receives gradients through both
the main network (\ours and step gating) and the expected dechunker.

\paragraph{Expected bidirectional dechunker.}
\label{sec:supp_dechunk}
LDARNet \cite{ledneva2026ldarnet} averages the output
of the forward smoothing
dechunker (\cref{prop:dechunk}) with a
reverse
smoothing value evaluated at the same chunk, $w_c z_c + (1-w_c)\,
r_{t+1}$,
where $c$ is the last boundary $\le t$ and $r_{t+1}$ the reverse
smoothed value
at the first boundary after $t$. The two factors depend on disjoint sets of
boundaries, so the expectation factorizes into
$\E[w_c z_c] + (1-\E[w_c])\,F_{t+1}$; both expectations are forward rank-one
recurrences with weight $p_t$, and $F$ is a backward recurrence of $z$ with
weight $p_t w_t$. With straight-through boundaries the training forward equals
the published hard dechunker, while the router receives gradients through the
main network (\ours and step gating) and through the exact expected
dechunker.

\FloatBarrier
\section{Speech: learned frame dropping}
\label{sec:speech}\label{sec:supp_speech}

We evaluate frame dropping for CTC recognition both with recovery at the
original frame rate and with decoding on the compacted sequence.
Soundwave and FastLongSpeech use CTC scores for frame selection and
merging, respectively~\cite{zhang2025soundwave,guo2025fastlongspeech}.
We adapt Soundwave's CTC-peak rule and initialize the selector from
intermediate CTC scores~\cite{lee2021interctc}.

With frame recovery, the character CTC head runs at the input frame rate, so
it needs a full-length sequence. We restore dropped positions with their representations at the
selection layer, $o_t = \tilde m_t\, y_t + (1-\tilde m_t)\, x^{\text{cut}}_t$,
as in skip-and-recover encoders \cite{zhu2024skipformer}. In the setting
without recovery, the output layer reads the compacted sequence directly.

\subsection{Setup and main results}
\label{sec:speech_main}
\paragraph{Setup.} We use the convolutional front end and Mamba blocks
of Mamba-based HuBERT \cite{lin2026mambahubert} (12 layers of width 384,
17.2M parameters, 50 frames per second), with either causal or ExtBiMamba
layers. We train the character CTC models \cite{graves2006ctc} from
scratch on LibriSpeech train-clean-100 \cite{panayotov2015librispeech}.
After layer 6, the encoder retains a fraction $\rho$ of the frames.
Layers 7--12 process the compacted sequence at inference.

At the same frame budget, we compare uniform subsampling, \emph{CTC
top-$k$} and Soundwave's \emph{CTC-peak} rule \cite{zhang2025soundwave}.
CTC top-$k$ keeps frames with the lowest blank posteriors from an
intermediate CTC head at layer 6 \cite{lee2021interctc}. The learned
selector adds a two-layer MLP to the CTC top-$k$ score. Its last layer
starts at zero. We train it with a zero mask, with \ours or with
REINFORCE. We also adapt trained dense models to frame dropping, as in
DyVM \cite{wu2025dyvm}. In a separate setting, we use a vocabulary of
300 BPE units \cite{sennrich2016bpe} and decode directly from the
compacted sequence. We report WER with greedy decoding.

\begin{table*}[t]
\caption{\textbf{\ours gives lower WER than zero-mask training in both decoding
setups.} Results give LibriSpeech test-clean WER with greedy CTC decoding and
no language model. The left block restores dropped frames for character CTC.
The right block decodes BPE units from retained frames. Values give the mean
and standard deviation over five seeds, or three for fine-tuning from dense
and bidirectional training.}
\label{tab:asr}
\centering
\scriptsize
\begin{tabular}{@{}l ccc cc@{}}
\toprule
 & \multicolumn{3}{c}{Dropped frames recovered, $\rho{=}0.5$} & \multicolumn{2}{c}{BPE from kept frames} \\
\cmidrule(lr){2-4}\cmidrule(l){5-6}
Frame selection & causal & causal, from dense & bidirectional & $\rho{=}0.25$ & $\rho{=}0.5$ \\
\midrule
Dense (all frames) & 23.9{\scriptsize$\pm$0.7} & 22.6{\scriptsize$\pm$0.9} & 14.3{\scriptsize$\pm$0.3} & 24.2{\scriptsize$\pm$0.7} & 24.2{\scriptsize$\pm$0.7} \\
\midrule
Uniform subsampling & 30.3{\scriptsize$\pm$0.3} & 34.8{\scriptsize$\pm$1.0} & 20.2{\scriptsize$\pm$0.4} & \textbf{26.1}{\scriptsize$\pm$0.2} & \textbf{23.0}{\scriptsize$\pm$0.6} \\
CTC top-$k$ & 27.6{\scriptsize$\pm$1.2} & 28.5{\scriptsize$\pm$0.5} & 14.8{\scriptsize$\pm$0.8} & 32.5{\scriptsize$\pm$0.6} & 24.5{\scriptsize$\pm$0.6} \\
CTC peaks (Soundwave rule) & 27.5{\scriptsize$\pm$1.2} & 28.6{\scriptsize$\pm$0.6} & 14.8{\scriptsize$\pm$0.7} & 32.8{\scriptsize$\pm$0.6} & 24.6{\scriptsize$\pm$0.5} \\
Learned, zero mask & 36.9{\scriptsize$\pm$8.3} & 48.2{\scriptsize$\pm$3.1} & 21.8{\scriptsize$\pm$2.6} & 80.4{\scriptsize$\pm$4.7} & 41.0{\scriptsize$\pm$1.9} \\
Learned, REINFORCE & \textbf{26.2}{\scriptsize$\pm$2.8} & \textbf{27.9}{\scriptsize$\pm$0.4} & \textbf{14.2}{\scriptsize$\pm$0.2} & 26.3{\scriptsize$\pm$0.4} & 24.1{\scriptsize$\pm$0.6} \\
Learned, \ours{} w/o step gating (ours) & 29.2{\scriptsize$\pm$0.8} & 30.1{\scriptsize$\pm$0.1} & 18.0{\scriptsize$\pm$1.5} & 57.2{\scriptsize$\pm$1.6} & 26.6{\scriptsize$\pm$0.7} \\
Learned, \ours{}, no conv.\ grad.\ (ours) & 26.6{\scriptsize$\pm$0.3} & 28.5{\scriptsize$\pm$0.3} & 17.1{\scriptsize$\pm$0.2} & 27.9{\scriptsize$\pm$0.4} & 25.1{\scriptsize$\pm$0.8} \\
Learned, \ours{} (ours) & 27.6{\scriptsize$\pm$0.9} & 28.8{\scriptsize$\pm$0.5} & 15.6{\scriptsize$\pm$2.0} & 34.2{\scriptsize$\pm$1.6} & 25.6{\scriptsize$\pm$0.6} \\
\bottomrule
\end{tabular}

\end{table*}

\paragraph{Decoding with frame recovery.} Compared with CTC top-$k$,
zero-mask training raises WER by 9.3 points in the causal encoder trained
from scratch, 7.0 in the bidirectional encoder and 19.7 when adapting a
trained dense model (\cref{tab:asr}). The mask leaves gaps in the sequence
processed during training, which the deployed encoder closes when it
gathers the kept frames. With \ours, the seed-paired WER differences
from CTC top-$k$ range from $-0.3$ to $+0.8$ points. The Soundwave rule gives similar WER to CTC top-$k$.
Removing step gating increases WER in every setting, by 0.3 to 2.4 points
against \ours (\cref{tab:asr_paired}), indicating that the state update also
matters in this setting. At $\rho{=}0.3$, the
variant without the convolution gradient gives the lowest WER (27.1\%
against 28.9\% for CTC top-$k$). This setting uses a tenfold lower
selector learning rate to prevent divergence in one of five seeds
(\cref{tab:asr_selector}).

\paragraph{Decoding from the compacted sequence.} Without frame recovery,
the final CTC head decodes subwords from the kept frames. At $\rho{=}0.25$, zero-mask training yields 80.4\% WER, and CTC top-$k$ raises
test-clean WER by 8.3 points relative to the dense model; the CTC-peak rule is
similar (\cref{tab:asr}). The selector trained
without the convolution gradient reduces WER by $4.7\pm0.4$ points
relative to CTC top-$k$, with an improvement in every seed. Full \ours
increases it by $1.6\pm1.8$ points. Uniform subsampling gives the lowest test-clean WER at both budgets. At $\rho{=}0.5$, it also improves on the dense
model (23.0 against 24.2\%). Removing step gating raises WER to 57.2\% at $\rho{=}0.25$, compared with
34.2\% for \ours. At $\rho{=}0.5$ the increase is 1.0 point (\cref{tab:asr}).

\begin{figure}[t]
\centering
\includegraphics[width=\linewidth]{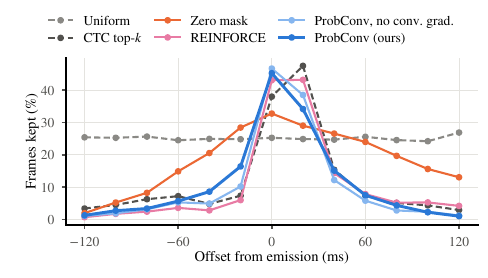}
\caption{\textbf{\ours retains more frames near character emissions than zero-mask
training.} Over 400 randomly sampled LibriSpeech test-clean utterances, the
curves show the retained share of frames at each offset from the nearest
character emission of the dense model's intermediate CTC head. The plotted run
decodes BPE from a quarter of the frames (seed 0).}
\label{fig:asr_patterns}
\end{figure}

\Cref{fig:asr_patterns_main} shows how the selectors cover one utterance, and
\cref{fig:asr_patterns} where their frames fall around character emissions.
CTC top-$k$ and the learned selectors favor frames at or near
character emissions, with 3--5\% of selections in silence compared
with 13\% for uniform subsampling. Their longest gap in speech averages
0.36--0.43\,s, compared with 60\,ms for uniform subsampling. At quarter-rate decoding, the zero-mask selector keeps runs of 4.0 consecutive
frames on average. For this selector, 57\% of the character emissions have no
kept frame within 20\,ms, which is consistent with its high WER. Full \ours keeps longer runs and covers fewer emissions
than the variant without the convolution gradient (70\% against 78\%).

\paragraph{Limitations.} The speech encoders train without
self-supervised pre-training and have substantially higher WERs than the
state of the art. Dropping frames does not reduce wall-clock time in this
small encoder (\cref{sec:speech_more}, \emph{Encoder time}). Uniform
subsampling gives lower WER than the learned and CTC-based rules when
decoding from the compacted sequence.

\subsection{Additional results}
\label{sec:speech_more}

\begin{table*}[t]
\caption{\textbf{With frame recovery, \ours gives lower WER than zero masking on both
test splits.} The Mamba-based HuBERT encoder retains half of the frames after
layer 6 and restores dropped positions before greedy character CTC decoding.
Results give LibriSpeech WER without a language model, as means and standard
deviations over five seeds for causal training from scratch and three
otherwise. ``Fine-tuned from dense'' adapts the trained dense model of the
same seed for 20k steps.}
\label{tab:asr_full}
\centering
\footnotesize
\begin{tabular}{@{}l cc cc cc@{}}
\toprule
 & \multicolumn{2}{c}{Causal, from scratch} & \multicolumn{2}{c}{Causal, fine-tuned from dense} & \multicolumn{2}{c}{Bidirectional, from scratch} \\
\cmidrule(lr){2-3}\cmidrule(lr){4-5}\cmidrule(l){6-7}
Frame selection & clean & other & clean & other & clean & other \\
\midrule
Dense (all frames) & 23.9{\scriptsize$\pm$0.7} & 50.1{\scriptsize$\pm$0.8} & 22.6{\scriptsize$\pm$0.9} & 49.0{\scriptsize$\pm$0.9} & 14.3{\scriptsize$\pm$0.3} & 38.6{\scriptsize$\pm$0.7} \\
\midrule
Uniform subsampling & 30.3{\scriptsize$\pm$0.3} & 56.9{\scriptsize$\pm$0.4} & 34.8{\scriptsize$\pm$1.0} & 59.0{\scriptsize$\pm$1.8} & 20.2{\scriptsize$\pm$0.4} & 46.3{\scriptsize$\pm$0.5} \\
CTC top-$k$ & 27.6{\scriptsize$\pm$1.2} & 54.1{\scriptsize$\pm$1.5} & 28.5{\scriptsize$\pm$0.5} & 53.1{\scriptsize$\pm$0.6} & 14.8{\scriptsize$\pm$0.8} & 38.5{\scriptsize$\pm$0.9} \\
CTC peaks (Soundwave rule) & 27.5{\scriptsize$\pm$1.2} & 54.1{\scriptsize$\pm$1.6} & 28.6{\scriptsize$\pm$0.6} & 53.2{\scriptsize$\pm$0.5} & 14.8{\scriptsize$\pm$0.7} & 38.7{\scriptsize$\pm$0.8} \\
Learned, zero mask & 36.9{\scriptsize$\pm$8.3} & 59.6{\scriptsize$\pm$5.3} & 48.2{\scriptsize$\pm$3.1} & 63.9{\scriptsize$\pm$1.8} & 21.8{\scriptsize$\pm$2.6} & 44.0{\scriptsize$\pm$2.2} \\
Learned, REINFORCE & \textbf{26.2}{\scriptsize$\pm$2.8} & 53.3{\scriptsize$\pm$3.1} & \textbf{27.9}{\scriptsize$\pm$0.4} & \textbf{52.7}{\scriptsize$\pm$0.6} & \textbf{14.2}{\scriptsize$\pm$0.2} & \textbf{38.1}{\scriptsize$\pm$0.2} \\
Learned, \ours{} w/o step gating (ours) & 29.2{\scriptsize$\pm$0.8} & 55.2{\scriptsize$\pm$1.4} & 30.1{\scriptsize$\pm$0.1} & 54.1{\scriptsize$\pm$0.2} & 18.0{\scriptsize$\pm$1.5} & 42.1{\scriptsize$\pm$1.9} \\
Learned, \ours{}, no conv.\ grad.\ (ours) & 26.6{\scriptsize$\pm$0.3} & \textbf{52.1}{\scriptsize$\pm$0.2} & 28.5{\scriptsize$\pm$0.3} & 53.1{\scriptsize$\pm$0.3} & 17.1{\scriptsize$\pm$0.2} & 41.5{\scriptsize$\pm$0.2} \\
Learned, \ours{} (ours) & 27.6{\scriptsize$\pm$0.9} & 54.1{\scriptsize$\pm$1.2} & 28.8{\scriptsize$\pm$0.5} & 53.3{\scriptsize$\pm$0.6} & 15.6{\scriptsize$\pm$2.0} & 39.9{\scriptsize$\pm$2.0} \\
\bottomrule
\end{tabular}

\end{table*}

\begin{table*}[t]
\caption{\textbf{Without frame recovery, training on the compacted network yields
substantially lower WER than zero-mask training.} The final CTC head reads
only the kept sequence. LibriSpeech WER gives the mean and standard deviation
over five seeds. $\Delta$ gives the paired test-clean difference from CTC
top-$k$ for other selection rules. The dense reference retains all frames.}
\label{tab:asr_bpe}
\centering
\small
\begin{tabular}{@{}l ccc ccc@{}}
\toprule
 & \multicolumn{3}{c}{$\rho{=}0.25$} & \multicolumn{3}{c}{$\rho{=}0.5$} \\
\cmidrule(lr){2-4}\cmidrule(l){5-7}
Frame selection & clean & other & $\Delta$ clean & clean & other & $\Delta$ clean \\
\midrule
Dense (all frames) & 24.2{\scriptsize$\pm$0.7} & 49.2{\scriptsize$\pm$0.7} & -- & 24.2{\scriptsize$\pm$0.7} & 49.2{\scriptsize$\pm$0.7} & -- \\
\midrule
Uniform subsampling & \textbf{26.1}{\scriptsize$\pm$0.2} & 51.0{\scriptsize$\pm$0.2} & $-6.4${\scriptsize$\pm$0.4} & \textbf{23.0}{\scriptsize$\pm$0.6} & \textbf{48.1}{\scriptsize$\pm$0.6} & $-1.5${\scriptsize$\pm$0.1} \\
CTC top-$k$ & 32.5{\scriptsize$\pm$0.6} & 54.4{\scriptsize$\pm$0.4} & -- & 24.5{\scriptsize$\pm$0.6} & 49.5{\scriptsize$\pm$0.6} & -- \\
CTC peaks (Soundwave rule) & 32.8{\scriptsize$\pm$0.6} & 54.4{\scriptsize$\pm$0.4} & $+0.2${\scriptsize$\pm$0.2} & 24.6{\scriptsize$\pm$0.5} & 49.5{\scriptsize$\pm$0.5} & $+0.1${\scriptsize$\pm$0.2} \\
Learned, zero mask & 80.4{\scriptsize$\pm$4.7} & 84.7{\scriptsize$\pm$3.5} & $+47.9${\scriptsize$\pm$5.0} & 41.0{\scriptsize$\pm$1.9} & 59.4{\scriptsize$\pm$1.0} & $+16.5${\scriptsize$\pm$2.3} \\
Learned, REINFORCE & 26.3{\scriptsize$\pm$0.4} & \textbf{50.8}{\scriptsize$\pm$0.4} & $-6.2${\scriptsize$\pm$0.5} & 24.1{\scriptsize$\pm$0.6} & 49.1{\scriptsize$\pm$0.6} & $-0.4${\scriptsize$\pm$0.2} \\
Learned, \ours{}, no conv.\ grad.\ (ours) & 27.9{\scriptsize$\pm$0.4} & 52.1{\scriptsize$\pm$0.4} & $-4.7${\scriptsize$\pm$0.4} & 25.1{\scriptsize$\pm$0.8} & 49.8{\scriptsize$\pm$0.8} & $+0.6${\scriptsize$\pm$0.2} \\
Learned, \ours{} (ours) & 34.2{\scriptsize$\pm$1.6} & 56.2{\scriptsize$\pm$1.0} & $+1.6${\scriptsize$\pm$1.8} & 25.6{\scriptsize$\pm$0.6} & 50.3{\scriptsize$\pm$0.6} & $+1.1${\scriptsize$\pm$0.2} \\
\bottomrule
\end{tabular}

\end{table*}

\Cref{tab:asr_full,tab:asr_bpe} give test-clean and test-other for the
settings of \cref{tab:asr}; test-other follows test-clean in every setting.

\begin{table*}[t]
\caption{\textbf{\ours stays close to CTC top-$k$ in the half-rate speech comparisons.}
The first row gives LibriSpeech test-clean WER for CTC top-$k$. Other rows
give WER differences in points to CTC top-$k$ with the same seed (same
initialization and data order), with negative values indicating lower WER.
Values summarize five seeds for causal training from scratch at $\rho{=}0.3$
and $0.5$, and three otherwise. At $\rho{=}0.3$, seed 0 diverged for the zero
mask and both \ours variants, and at $\rho{=}0.5$ the mask degraded on seed 3
(\cref{fig:asr_keep}); a smaller selector learning rate prevents the
divergence (\cref{tab:asr_selector}).}
\label{tab:asr_paired}
\centering
\footnotesize
\begin{tabular}{@{}l ccc cc c@{}}
\toprule
 & \multicolumn{3}{c}{From scratch} & \multicolumn{2}{c}{Fine-tuned, $\rho{=}0.5$} & Bidir. \\
\cmidrule(lr){2-4}\cmidrule(lr){5-6}\cmidrule(l){7-7}
Frame selection & $\rho{=}0.3$ & $\rho{=}0.5$ & $\rho{=}0.7$ & from dense & from CTC top-$k$ & $\rho{=}0.5$ \\
\midrule
CTC top-$k$ (WER) & 28.9{\scriptsize$\pm$2.1} & 27.6{\scriptsize$\pm$1.2} & 26.1{\scriptsize$\pm$1.2} & 28.5{\scriptsize$\pm$0.5} & 26.3{\scriptsize$\pm$1.7} & 14.8{\scriptsize$\pm$0.8} \\
\midrule
Uniform subsampling & $+0.1${\scriptsize$\pm$1.9} & $+2.7${\scriptsize$\pm$1.2} & $+3.3${\scriptsize$\pm$0.9} & $+6.3${\scriptsize$\pm$0.8} & $+3.0${\scriptsize$\pm$0.4} & $+5.4${\scriptsize$\pm$1.1} \\
CTC peaks (Soundwave rule) & $-0.0${\scriptsize$\pm$0.3} & $-0.1${\scriptsize$\pm$0.3} & $+0.1${\scriptsize$\pm$0.5} & $+0.1${\scriptsize$\pm$0.2} & $+0.1${\scriptsize$\pm$0.2} & $+0.0${\scriptsize$\pm$0.1} \\
Learned, zero mask & $+7.2${\scriptsize$\pm$12.7} & $+9.3${\scriptsize$\pm$8.1} & $+2.7${\scriptsize$\pm$0.4} & $+19.7${\scriptsize$\pm$2.9} & $+1.9${\scriptsize$\pm$0.2} & $+7.0${\scriptsize$\pm$1.8} \\
Learned, REINFORCE & $\mathbf{-0.6}${\scriptsize$\pm$1.0} & $\mathbf{-1.3}${\scriptsize$\pm$1.6} & $\mathbf{-0.8}${\scriptsize$\pm$0.3} & $\mathbf{-0.7}${\scriptsize$\pm$0.6} & $\mathbf{-0.1}${\scriptsize$\pm$0.0} & $\mathbf{-0.6}${\scriptsize$\pm$1.0} \\
Learned, \ours{} w/o step gating (ours) & $+6.3${\scriptsize$\pm$14.8} & $+1.6${\scriptsize$\pm$0.8} & $+0.4${\scriptsize$\pm$1.2} & $+1.6${\scriptsize$\pm$0.6} & $+0.3${\scriptsize$\pm$0.5} & $+3.2${\scriptsize$\pm$0.7} \\
Learned, \ours{}, no conv.\ grad.\ (ours) & $+4.8${\scriptsize$\pm$12.6} & $-0.9${\scriptsize$\pm$1.5} & $+0.3${\scriptsize$\pm$0.9} & $-0.1${\scriptsize$\pm$0.6} & $+0.1${\scriptsize$\pm$0.7} & $+2.3${\scriptsize$\pm$0.6} \\
Learned, \ours{} (ours) & $+5.7${\scriptsize$\pm$12.3} & $+0.1${\scriptsize$\pm$1.1} & $-0.3${\scriptsize$\pm$1.4} & $+0.3${\scriptsize$\pm$0.8} & $-0.0${\scriptsize$\pm$0.3} & $+0.8${\scriptsize$\pm$1.2} \\
\bottomrule
\end{tabular}

\end{table*}

\begin{table}[t]
\caption{\textbf{The effect of the convolution gradient depends on the speech encoder
and training setting.} The first row gives LibriSpeech test-clean WER for CTC
top-$k$. Other rows show paired differences in points. $\lambda$ scales the
convolution gradient. At $\rho{=}0.3$, the two columns compare the default
selector learning rate, where one seed diverged for the zero mask and every
\ours variant, with one tenth of that rate. Values summarize five seeds, or
three for intermediate $\lambda$ settings and the bidirectional encoder.}
\label{tab:asr_selector}
\centering
\resizebox{\linewidth}{!}{\begin{tabular}{@{}lcccc@{}}
\toprule
 & \multicolumn{2}{c}{causal, $\rho{=}0.3$} & \multicolumn{2}{c}{$\rho{=}0.5$} \\
\cmidrule(lr){2-3}\cmidrule(l){4-5}
Learned selector & sel.\ LR$\times$1 & sel.\ LR$\times$0.1 & causal & bidir. \\
\midrule
CTC top-$k$ (WER) & 28.9{\scriptsize$\pm$2.1} & 28.9{\scriptsize$\pm$2.1} & 27.6{\scriptsize$\pm$1.2} & 14.8{\scriptsize$\pm$0.8} \\
\midrule
\ours{} (ours), $\lambda{=}1$ & $+5.7${\scriptsize$\pm$12.3} & $+1.3${\scriptsize$\pm$2.5} & $+0.1${\scriptsize$\pm$1.1} & $+0.8${\scriptsize$\pm$1.2} \\
\ours{} (ours), $\lambda{=}0.3$ & $+4.0${\scriptsize$\pm$9.4} & $+2.1${\scriptsize$\pm$7.6} & $+0.9${\scriptsize$\pm$1.1} & $+1.5${\scriptsize$\pm$1.8} \\
\ours{} (ours), $\lambda{=}0.1$ & $+6.7${\scriptsize$\pm$14.6} & $+1.4${\scriptsize$\pm$7.4} & $-0.9${\scriptsize$\pm$2.3} & $+2.1${\scriptsize$\pm$1.4} \\
\ours{}, no conv.\ grad.\ (ours), $\lambda{=}0$ & $+4.8${\scriptsize$\pm$12.6} & $-1.7${\scriptsize$\pm$1.5} & $-0.9${\scriptsize$\pm$1.5} & $+2.3${\scriptsize$\pm$0.6} \\
Zero mask & $+7.2${\scriptsize$\pm$12.7} & $+0.8${\scriptsize$\pm$1.3} & $+9.3${\scriptsize$\pm$8.1} & $+7.0${\scriptsize$\pm$1.8} \\
REINFORCE & $-0.6${\scriptsize$\pm$1.0} & $+0.3${\scriptsize$\pm$1.0} & $-1.3${\scriptsize$\pm$1.6} & $-0.6${\scriptsize$\pm$1.0} \\
\bottomrule
\end{tabular}
}
\end{table}

\begin{figure}[t]
\centering
\includegraphics[width=\linewidth]{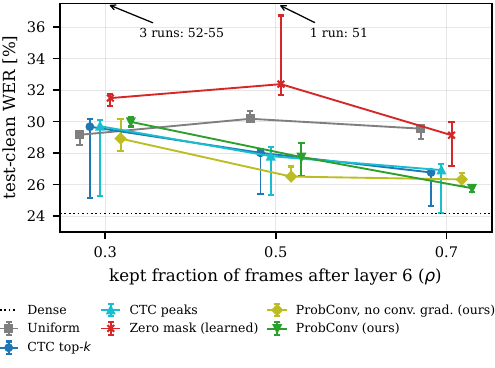}
\caption{\textbf{Among the runs included in the medians, both \ours variants stay at
the level of CTC top-$k$ at every frame budget, and the zero mask stays above
it.} Points give median WER for the causal encoder trained from scratch, over
five seeds except at $\rho{=}0.7$, where we use three. Bars span the range of
the plotted runs. Arrows mark the additional runs above 45\% WER, which are
excluded from the medians. These are seed 0 of the zero mask and both \ours
variants at $\rho{=}0.3$ (52.5--55.0) and seed 3 of the zero mask at
$\rho{=}0.5$ (51.2).}
\label{fig:asr_keep}
\end{figure}

\paragraph{Seed-paired comparisons and the convolution gradient.} Seed-paired
differences between \ours and CTC top-$k$ range from $-0.3$ to
$+0.8$ WER
points; initialized from the CTC top-$k$ model the
learned selector
keeps its accuracy ($-0.0\pm0.3$), as does REINFORCE
($-0.1\pm0.0$). The zero
mask and uniform subsampling increase WER relative to CTC top-$k$ by
$1.9\pm0.2$ and $3.0\pm0.4$ points, respectively. The Soundwave rule is
indistinguishable from CTC top-$k$ ($|\Delta|\le0.1$). Without the convolution
gradient the selector is slightly better in the causal encoder ($-0.9\pm1.5$,
better on four of five seeds)
and worse in the bidirectional one ($+2.3\pm0.6$, against $+0.8\pm1.2$ for full
\ours). Scaling that gradient by $\lambda$ (\cref{tab:asr_selector}) shows a
monotone trend in the bidirectional encoder ($+0.8$, $+1.5$, $+2.1$ and $+2.3$
for $\lambda=1$, 0.3, 0.1 and 0), while in the causal encoder every value lies
within the seed noise ($-0.9$ to $+0.9$); at $\rho{=}0.3$ the values for
$\lambda=0.3$ and 0.1 are dominated by one diverging seed (below). With a bidirectional encoder,
dropping half of the frames costs only 0.5 points with CTC top-$k$, against
3.7 in the causal encoder.

\paragraph{Score-function training.} The selector can also learn on the
compacted network without a pathwise gradient: REINFORCE
\cite{williams1992reinforce} compacts a Bernoulli sample of the exact-budget
probabilities and weights its log-probability by its loss minus the loss of
the top-$k$ network \cite{rennie2017scst}. With one reward per utterance and a few hundred decisions,
training on the exact compacted network is what matters for frame selection,
and a score-function estimator suffices.

\paragraph{Score-function results.} On LibriSpeech test-clean, REINFORCE gives the lowest WER among learned
selectors in eight of the nine settings of
\cref{tab:asr_full,tab:asr_paired,tab:asr_selector,tab:asr_bpe}. At
$\rho{=}0.5$, it reaches 26.2\% WER, a paired difference of $-1.3\pm1.6$
points from CTC top-$k$, with lower WER in every seed. At $\rho{=}0.3$, it
reaches 28.3\% without divergence. It reaches 14.2\% in the bidirectional
encoder and 27.9\% when retrofitted from the dense model. Without frame
recovery at $\rho{=}0.25$, it reaches 26.3\%, close to uniform subsampling at
26.1\%. Its paired differences from CTC top-$k$ are $-0.8\pm0.3$ at
$\rho{=}0.7$ and $-0.1\pm0.0$ when fine-tuning from CTC top-$k$. Only
with the smaller selector learning rate at $\rho{=}0.3$ is \ours without the
convolution gradient better (27.1 against 29.1\%).

\paragraph{Encoder time.} In this 17M-parameter encoder the convolutional
front end takes 88\% (causal) and 76\% (bidirectional) of the encoder time on
an RTX 4090, and the Mamba layers of width 384 are bound by kernel launches
rather than by the sequence length: keeping 25 or 50\% of the frames after
layer 2, 3, 4 or 6 changes the encoder time by $-4$ to $+1$\%, selection
included.

\paragraph{Aggressive budgets.} At $\rho{=}0.7$ REINFORCE is the best selection
rule on average ($-0.8\pm0.3$), ahead of \ours ($-0.3\pm1.4$). At $\rho{=}0.3$ seed 0 diverged for the zero mask and for every \ours variant,
early in training at the peak learning rate, while the other four seeds match
CTC top-$k$; the instability comes from learning the selector at a high
pruning rate and not from the operator. We reduce the selector learning rate by a factor of ten
(\cref{tab:asr_selector}, sel. LR$\times$0.1). At the lower selector learning rate, all seeds converge for $\lambda{=}0$ and
$\lambda=1$. The variant without the convolution gradient gives the lowest
mean WER at this budget (27.1\%), compared with 28.9\% for CTC top-$k$ and
29.0\% for uniform subsampling. Its paired difference from CTC top-$k$ is
$-1.7\pm1.5$ points, with lower WER in four of five seeds. Full \ours and the mask stay within the noise of CTC
top-$k$ ($+1.3\pm2.5$ and $+0.8\pm1.3$). At the lower learning rate, seed 0 still diverges for $\lambda=0.3$ and 0.1,
reaching 40.8\% and 39.7\% WER, respectively. In the other four seeds, WER is
1.2 to 2.1 points below CTC top-$k$, similar to the variant without the
convolution gradient.

\FloatBarrier
\section{Genomics: dynamic chunking in H-Net and LDARNet}
\label{sec:supp_genomics}

We train a one-stage H-Net \cite{hwang2026hnet} DNA language model on hg38
(14.8M parameters, 4096-base contexts, 20k steps) with the published routing,
dechunker and ratio loss towards $N$ bases per chunk, and sweep $N$ to compare
at matched compression. We fine-tune the released LDARNet-110M
\cite{ledneva2026ldarnet} on the 18 Nucleotide Transformer tasks
\cite{dallatorre2024nt} with the authors' full protocol (ten folds, per-task
learning rates, early stopping; 540 runs) and on the nine Genomic Benchmarks
\cite{gresova2023genomic}. In both hosts the baseline trains the router with
the published mechanism; our arms train it through \ours and step gating in
the main network and through the expected dechunker. Protocols are in
\cref{sec:supp_protocols}.

\subsection{Dynamic chunking in H-Net}
\label{sec:hnet}

\begin{figure}[t]
\centering
\includegraphics[width=\linewidth]{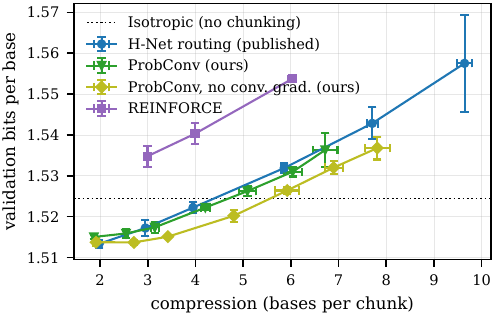}
\caption{\textbf{At comparable compression, both \ours routers reach or improve on the
bits per base of the published routing from about three bases per chunk
onward.} The curves show validation bits per base on hg38 against realized
bases per chunk. Points and error bars give means and standard deviations over
three seeds. \Cref{tab:hnet} gives the target ratios and interpolated
differences. Target ratios $N$ run from 2 to 10 for the published routing, to
12 for \ours and over $\{3,4,6\}$ for REINFORCE.}
\label{fig:hnet_rd}
\end{figure}

Routers trained through expected compaction place more boundaries than the
ratio loss asks for (2.54 and 2.71 bases per chunk at $N{=}3$, against 2.95 for
the published routing), so we compare the arms along the curves of
\cref{fig:hnet_rd} (values in \cref{tab:hnet}). At matched compression, \ours
without the convolution gradient improves on the published routing by 0.0031
bits per base at 2.95 bases per chunk, 0.0052 at 3.95, 0.0058 at 5.86 and
0.0066 at 7.71; from three bases per chunk on, the gain exceeds the seed
standard deviations of both arms. Full \ours is on par up to four bases per
chunk and 0.0018 better at 5.86. With a stronger ratio loss ($\alpha{=}0.3$)
the routers meet the target, and the variant without the convolution gradient
gains 0.0029 and 0.0054 directly; with three times the training budget the
advantage remains (0.0038 at 3.91 bases per chunk with every seed better, and
0.0075 at $N{=}8$; \cref{sec:supp_hnet}). A router trained by REINFORCE, with
the threshold boundaries as the self-critical baseline, meets the target
compression but is 0.018--0.021 bits per base worse than the published routing,
with one reward per window of 4096 bases for about a thousand boundary
decisions.

\subsection{LDARNet on the Nucleotide Transformer benchmark}
\label{sec:ldar}

\begin{figure}[t]
\centering
\includegraphics[width=\linewidth]{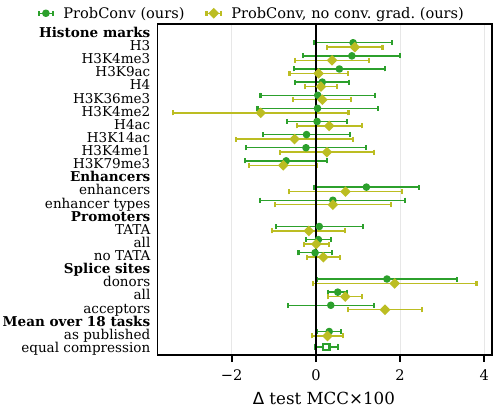}
\caption{\textbf{\ours improves most Nucleotide Transformer tasks, with the clearest
gains on splice sites and enhancers.} The mean gain persists at matched
compression. The plot gives the change in test MCC$\times$100 relative to
reproduced routing on 18 Nucleotide Transformer tasks, with 95\% confidence
intervals. Intervals use ten folds for individual tasks and variation across
tasks for the overall means. For the matched-compression comparison, both
methods use the pretraining ratio loss. \Cref{tab:ldar_full,tab:ldar} give the
task and group results.}
\label{fig:ldar_tasks}
\end{figure}

Our run of the published fine-tuning reproduces the reported results (74.03
against 73.81 mean MCC), so we compare against the authors' method at its
strength. Training the router through expected compaction raises the mean MCC
to 74.34 and improves 14 of the 18 tasks ($p{=}0.009$ over folds;
\cref{fig:ldar_tasks}), mostly splice
sites and enhancers, while histone marks and promoters change by at most 0.15;
the variant without the convolution gradient performs the same. Both \ours routers place more boundaries than the published one, but the gain
does not come from the extra chunks: with the pre-training ratio loss applied
to both routers, compression is similar (4.08 against 4.01 bases per chunk)
and \ours remains 0.25 points better ($p{=}0.02$ over folds), while the ratio
loss itself changes neither routing ($+0.09$ and $+0.02$). The extra chunks leave inference time within 54--59\,ms per batch of 64 on an
H100 for any compression between 3 and 6 bases per chunk. Controls with a
frozen router and with continued pre-training are in \cref{sec:supp_ldar}. On
the nine Genomic Benchmarks, where our published-routing runs reproduce the
reported accuracies (88.30 against 88.48), the gains are small: $+0.21$ for
\ours ($p{=}0.22$ over folds) and $+0.14$ with more chunks without the
convolution gradient ($p{=}0.01$).

\subsection{Ablations: what must be exact}
\label{sec:ablations}

\begin{figure}[t]
\centering
\includegraphics[width=\linewidth]{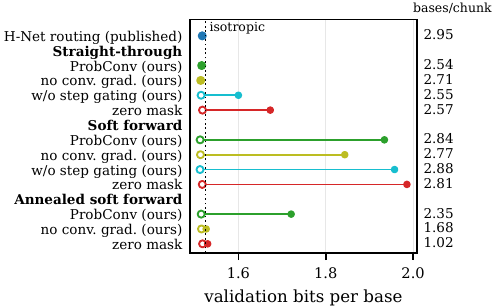}
\caption{\textbf{Straight-through \ours with step gating agrees across H-Net's two
evaluations.} Open markers show validation bits per base using the training
computation, and filled markers show the deployed hard forward. The horizontal
axis gives realized bases per chunk at target $N{=}3$. Points are means over
three seeds. Values in \cref{tab:hnet_ablation}.}
\label{fig:hnet_gap}
\end{figure}

\paragraph{The training forward must be the deployed computation.} In H-Net (\cref{fig:hnet_gap}), the deployed network raises validation loss by
0.16 bits per base for zero masking and by 0.09 for \ours without step gating,
relative to their respective training computations. Using the expectation in
the forward pass gives gaps of 0.33--0.47 bits per base. Straight-through
\ours with step gating gives 1.5159 bits per base in both evaluations at
$N{=}3$.
Straight-through evaluation at the vertices, where \ours is exact, removes the
gap by construction. In \cref{tab:hnet_ablation}, STE denotes this
straight-through training, SOFT the soft forward pass throughout and ANNEAL
the annealed probabilities.

\paragraph{Annealing does not reach the vertices.} A soft forward pass sharpened as $\sigma(\operatorname{logit}(p)/T)$ to
$T{=}0.01$ again gives 1.515--1.518 bits per base under the training
computation, but after hard routing full \ours reaches 1.721 bits per base,
and the variant without the convolution gradient and the zero mask lose most
of their compression.
$T$ decays log-linearly from 1 to 0.01 over the first 80\% of training. The annealed routers of full \ours
and of the mask keep the boundary probabilities near the threshold (mean 0.50,
against 0.32 under straight-through training), where even a low temperature
leaves the relaxation soft. In ReMamba and Cobra the soft forward pass does
reach the vertices (\cref{sec:supp_remamba,sec:supp_cobra}): a learned
temperature or a growing penalty on $p_i(1-p_i)$ drives the selection there.

\paragraph{Which gradients matter.} With the forward pass fixed, the router or
selector receives gradients through the convolution taps, the step gate and
the dechunker. Stopping the convolution gradient changes little on LDARNet and Vim and costs
about one point in VideoMamba (0.2--0.5 in the replicate set;
\cref{tab:supp_vim_ablation,tab:supp_videomamba_ablation}). It improves H-Net
at matched compression. In speech, the effect depends on the encoder and decoding setup. For the
causal encoder, stopping the gradient changes WER by $-0.9\pm1.5$ points,
within the variation across seeds. It raises WER in the bidirectional encoder.
At the tighter budget without frame recovery, stopping the gradient lowers WER
by 6.3 points. Scaling the convolution gradient by $\lambda$ produces monotone
trends on H-Net and in the bidirectional encoder
(\cref{tab:hnet_controls,tab:asr_selector}). In the
genomics and causal speech hosts the benefit thus comes from the exact
compacted forward pass and the gradients through the gated scan and the
expected dechunker. Cobra is the opposite case: without the convolution
gradient its selector falls 5.8 points below \ours, and without step gating it
loses nothing (\cref{tab:supp_cobra_ablation}). The convolution gradient is
therefore host-dependent, from harmful on H-Net to decisive in Cobra.
REINFORCE on the exactly compacted network gives the best learned speech
selector and the weakest H-Net router in these comparisons. It trails \ours by
1.0 point in Cobra. This pattern is consistent with a greater role for the
pathwise gradient when many decisions share a sequence-level reward.

\paragraph{Limitations of the genomics experiments.} Our H-Net is small and
trained for 20k steps, and its \ours routers need a stronger ratio loss to reach
the target compression. The gains on LDARNet are consistent but small, close to
zero on histone marks and promoters and on the nearly saturated Genomic
Benchmarks.

\FloatBarrier
\section{Additional H-Net results}
\label{sec:supp_hnet}

\begin{table*}[t]
\caption{\textbf{At matched compression, both \ours routers match or improve on the
validation loss of the published routing from about three bases per chunk
onward.} Loss is measured in validation bits per base with hard routing and
summarized over three seeds as mean $\pm$ standard deviation. Parentheses give
realized bases per chunk. The last three columns interpolate differences
relative to the published routing. Negative differences favor the compared
method. ``Out of range'' means the matching compression falls outside the
measured curve. REINFORCE ran at $N{=}3$, 4 and 6 only.}
\label{tab:hnet}
\centering
\footnotesize
\begin{tabular}{@{}c ccc ccc@{}}
\toprule
 & \multicolumn{3}{c}{bits/base at target $N$ (bases per chunk)} & \multicolumn{3}{c}{$\Delta$ bits/base at matched compression} \\
\cmidrule(lr){2-4}\cmidrule(l){5-7}
$N$ & \makecell{H-Net routing\\(published)} & \makecell{\ours{}\\(ours)} & \makecell{\ours{}, no conv.\ grad.\\(ours)} & \makecell{\ours{}\\(ours)} & \makecell{no conv.\ grad.\\(ours)} & REINFORCE \\
\midrule
2 & 1.5133{\scriptsize$\pm$0.0010} (1.98) & 1.5150{\scriptsize$\pm$0.0005} (1.87) & 1.5137{\scriptsize$\pm$0.0010} (1.91) & $+0.0018$ & $+0.0004$ & out of range \\
3 & 1.5172{\scriptsize$\pm$0.0020} (2.95) & 1.5159{\scriptsize$\pm$0.0010} (2.54) & 1.5137{\scriptsize$\pm$0.0006} (2.71) & $-0.0003$ & $-0.0031$ & out of range \\
4 & 1.5222{\scriptsize$\pm$0.0013} (3.95) & 1.5174{\scriptsize$\pm$0.0013} (3.15) & 1.5151{\scriptsize$\pm$0.0002} (3.42) & $-0.0012$ & $-0.0052$ & $+0.0179$ \\
6 & 1.5319{\scriptsize$\pm$0.0012} (5.86) & 1.5222{\scriptsize$\pm$0.0006} (4.21) & 1.5202{\scriptsize$\pm$0.0015} (4.80) & $-0.0018$ & $-0.0058$ & $+0.0207$ \\
8 & 1.5428{\scriptsize$\pm$0.0039} (7.71) & 1.5262{\scriptsize$\pm$0.0013} (5.09) & 1.5264{\scriptsize$\pm$0.0008} (5.92) & out of range & $-0.0066$ & out of range \\
10 & 1.5575{\scriptsize$\pm$0.0118} (9.64) & 1.5310{\scriptsize$\pm$0.0012} (6.04) & 1.5320{\scriptsize$\pm$0.0017} (6.90) & out of range & out of range & out of range \\
12 & 1.5623{\scriptsize$\pm$0.0092} (11.44) & 1.5363{\scriptsize$\pm$0.0041} (6.72) & 1.5368{\scriptsize$\pm$0.0028} (7.81) & out of range & out of range & out of range \\
\bottomrule
\end{tabular}

\end{table*}

\paragraph{Compression and the curves.} \Cref{tab:hnet} lists the values
behind \cref{fig:hnet_rd}. The routers achieve different compression at the same target $N$: those
trained through expected compaction place more boundaries than the ratio loss
asks for. At $N{=}3$, they reach 2.54 and 2.71 bases per chunk, compared with
2.95 for the published routing. At $N{=}6$, the corresponding values are 4.21
and 4.80, compared with 5.86. We therefore compare the validation loss at the
compression each router achieves, and \cref{tab:hnet} interpolates each loss
curve at the compression achieved by the published router. Full \ours is on par up to four bases per chunk ($-0.0003$
and $-0.0012$), $0.0018$ better at 5.86, and does not reach 7.71 even at
$N{=}12$. With about two bases per chunk the arms coincide. The published
routing stays below the isotropic baseline (1.5244) up to four bases per chunk
and the variant without the convolution gradient up to 4.8. At $N{=}10$ the
published routing varies strongly across seeds (1.545--1.569). The router
trained by REINFORCE meets the target compression (2.99, 3.99 and 6.03 bases
per chunk at $N{=}3,4,6$).

\paragraph{Annealing.} Annealed to $T{=}0.01$, the variant without the
convolution gradient and the zero mask lose most of the compression (1.68 and
1.02 bases per chunk at $N{=}3$) and end above the isotropic model (1.526 and
1.530). The same pattern appears at $N{=}4$, where full \ours gives 1.714 bits per
base and the two other annealed configurations give 1.526 and 1.532.

\begin{table}[t]
\caption{\textbf{Among the relaxations, only straight-through \ours with step gating
gives the same validation loss under the training computation and after hard
routing.} The table compares validation bits per base at target $N{=}3$ using
the training computation and deployed hard routing. ``STE,'' ``SOFT'' and
``ANNEAL'' indicate a binary, soft or annealed training forward, respectively.
Values give means and standard deviations over three seeds. The isotropic
model has no router and retains every base. ``relax.'' is n/a for the
published routing, which trains its router as in H-Net, and for the isotropic
model.}
\label{tab:hnet_ablation}
\centering
\resizebox{\linewidth}{!}{\begin{tabular}{@{}llcccc@{}}
\toprule
Router training & relax. & training fwd & inference fwd & bases/chunk & seeds \\
\midrule
H-Net routing (published) & n/a & 1.5172{\scriptsize$\pm$0.0020} & 1.5172{\scriptsize$\pm$0.0020} & 2.95{\scriptsize$\pm$0.02} & 3 \\
Isotropic (no chunking) & n/a & 1.5244{\scriptsize$\pm$0.0006} & 1.5244{\scriptsize$\pm$0.0006} & 1 & 3 \\
\midrule
\ours{} (ours) & STE & 1.5159{\scriptsize$\pm$0.0010} & 1.5159{\scriptsize$\pm$0.0010} & 2.54{\scriptsize$\pm$0.03} & 3 \\
\ours{}, no conv.\ grad.\ (ours) & STE & 1.5137{\scriptsize$\pm$0.0006} & 1.5137{\scriptsize$\pm$0.0006} & 2.71{\scriptsize$\pm$0.01} & 3 \\
\ours{} w/o step gating (ours) & STE & 1.5146{\scriptsize$\pm$0.0002} & 1.6000{\scriptsize$\pm$0.0140} & 2.55{\scriptsize$\pm$0.01} & 3 \\
Zero mask & STE & 1.5178{\scriptsize$\pm$0.0007} & 1.6730{\scriptsize$\pm$0.0239} & 2.57{\scriptsize$\pm$0.06} & 3 \\
\midrule
\ours{} (ours) & SOFT & 1.5124{\scriptsize$\pm$0.0005} & 1.9344{\scriptsize$\pm$0.0408} & 2.84{\scriptsize$\pm$0.02} & 3 \\
\ours{}, no conv.\ grad.\ (ours) & SOFT & 1.5131{\scriptsize$\pm$0.0014} & 1.8435{\scriptsize$\pm$0.0862} & 2.77{\scriptsize$\pm$0.05} & 3 \\
\ours{} w/o step gating (ours) & SOFT & 1.5122{\scriptsize$\pm$0.0012} & 1.9576{\scriptsize$\pm$0.0562} & 2.88{\scriptsize$\pm$0.03} & 3 \\
Zero mask & SOFT & 1.5172{\scriptsize$\pm$0.0005} & 1.9859{\scriptsize$\pm$0.0138} & 2.81{\scriptsize$\pm$0.02} & 3 \\
\midrule
\ours{} (ours) & ANNEAL & 1.5148{\scriptsize$\pm$0.0003} & 1.7208{\scriptsize$\pm$0.0360} & 2.35{\scriptsize$\pm$0.17} & 3 \\
\ours{}, no conv.\ grad.\ (ours) & ANNEAL & 1.5155{\scriptsize$\pm$0.0004} & 1.5262{\scriptsize$\pm$0.0039} & 1.68{\scriptsize$\pm$0.10} & 3 \\
Zero mask & ANNEAL & 1.5180{\scriptsize$\pm$0.0003} & 1.5296{\scriptsize$\pm$0.0022} & 1.02{\scriptsize$\pm$0.00} & 3 \\
\bottomrule
\end{tabular}
}
\end{table}

\begin{table}[t]
\caption{\textbf{A larger ratio-loss weight brings \ours closer to H-Net's target
compression.} Values give mean and standard deviation over three seeds for
validation bits per base after hard routing. Parentheses give realized bases
per chunk. $\alpha$ weights the ratio loss, and $\lambda$ scales the
convolution gradient.}
\label{tab:hnet_controls}
\centering
\resizebox{\linewidth}{!}{\begin{tabular}{@{}lcc@{}}
\toprule
Router training & $N{=}3$ & $N{=}4$ \\
\midrule
H-Net routing (published, $\alpha{=}0.03$) & 1.5172{\scriptsize$\pm$0.0020} (2.95) & 1.5222{\scriptsize$\pm$0.0013} (3.95) \\
\midrule
\ours{} (ours), $\alpha{=}0.03$ & 1.5159{\scriptsize$\pm$0.0010} (2.54) & 1.5174{\scriptsize$\pm$0.0013} (3.15) \\
\ours{} (ours), $\alpha{=}0.1$ & 1.5173{\scriptsize$\pm$0.0012} (2.82) & 1.5199{\scriptsize$\pm$0.0004} (3.61) \\
\ours{} (ours), $\alpha{=}0.3$ & 1.5168{\scriptsize$\pm$0.0012} (2.91) & 1.5205{\scriptsize$\pm$0.0009} (3.82) \\
\ours{} (ours), $\lambda{=}0.3$ & 1.5147{\scriptsize$\pm$0.0008} (2.64) & 1.5160{\scriptsize$\pm$0.0007} (3.37) \\
\ours{} (ours), $\lambda{=}0.1$ & 1.5146{\scriptsize$\pm$0.0009} (2.69) & 1.5156{\scriptsize$\pm$0.0009} (3.41) \\
\ours{}, no conv.\ grad.\ (ours), $\alpha{=}0.03$ & 1.5137{\scriptsize$\pm$0.0006} (2.71) & 1.5151{\scriptsize$\pm$0.0002} (3.42) \\
\ours{}, no conv.\ grad.\ (ours), $\alpha{=}0.1$ & 1.5148{\scriptsize$\pm$0.0011} (2.92) & 1.5173{\scriptsize$\pm$0.0003} (3.79) \\
\ours{}, no conv.\ grad.\ (ours), $\alpha{=}0.3$ & 1.5143{\scriptsize$\pm$0.0004} (2.97) & 1.5168{\scriptsize$\pm$0.0003} (3.92) \\
REINFORCE, $\alpha{=}0.03$ & 1.5348{\scriptsize$\pm$0.0025} (2.99) & 1.5403{\scriptsize$\pm$0.0026} (3.99) \\
\bottomrule
\end{tabular}
}
\end{table}

\paragraph{Calibrated compression.} A stronger ratio loss makes the \ours
routers meet the target, so the comparison needs no interpolation
(\cref{tab:hnet_controls}). With $\alpha{=}0.3$ the variant without the
convolution gradient realizes 2.97 and 3.92 bases per chunk and reaches 1.5143
and 1.5168 bits per base, against 1.5172 and 1.5222 for the published routing
at 2.95 and 3.95: direct gains of 0.0029 and 0.0054, which agree with the
interpolated ones. Full \ours with $\alpha{=}0.3$ remains on par (1.5168 at
2.91 bases per chunk). Scaling the gradient through the convolution taps moves
full \ours towards the variant without it: interpolated at 2.95 bases per
chunk, the gain is 0.0003, 0.0020, 0.0022 and 0.0031 for $\lambda=1$, 0.3,
0.1 and 0.

\begin{table}[t]
\caption{\textbf{The stopped-gradient \ours router retains its advantage after longer
training at matched compression.} Results compare H-Net's published routing
and the calibrated \ours variant without the convolution gradient at 20k and
60k steps. Validation bits per base give means and standard deviations over
three seeds. Parentheses give realized bases per chunk. The $N{=}4$ routers
both achieve 3.91 bases per chunk after 60k steps.}
\label{tab:hnet_budget}
\centering
\resizebox{\linewidth}{!}{\begin{tabular}{@{}l cc cc@{}}
\toprule
 & \multicolumn{2}{c}{$N{=}4$} & \multicolumn{2}{c}{$N{=}8$} \\
\cmidrule(lr){2-3}\cmidrule(l){4-5}
Router training & 20k steps & 60k steps & 20k steps & 60k steps \\
\midrule
H-Net routing (published, $\alpha{=}0.03$) & 1.5222{\scriptsize$\pm$0.0013} (3.95) & 1.4839{\scriptsize$\pm$0.0007} (3.91) & 1.5428{\scriptsize$\pm$0.0039} (7.71) & 1.5017{\scriptsize$\pm$0.0066} (7.59) \\
\ours{}, no conv.\ grad.\ (ours), $\alpha{=}0.3$ & 1.5168{\scriptsize$\pm$0.0003} (3.92) & 1.4801{\scriptsize$\pm$0.0010} (3.91) & 1.5325{\scriptsize$\pm$0.0012} (7.51) & 1.4943{\scriptsize$\pm$0.0012} (7.46) \\
\midrule
$\Delta$ bits/base & $-0.0054$ & $-0.0038$ & $-0.0104$ & $-0.0075$ \\
\bottomrule
\end{tabular}
}
\end{table}

\paragraph{Longer training.} With three times the training budget both
routings gain about 0.04 bits per base, and the advantage remains
(\cref{tab:hnet_budget}; its last row is our router minus the published
routing). At $N{=}4$ both reach 3.91 bases per chunk, and
\ours without the convolution gradient is 0.0038 better, with every seed below
every seed of the published routing. At $N{=}8$ it is 0.0075 better at a
slightly lower compression (7.46 against 7.59 bases per chunk), and its
results vary much less across seeds (0.0012 against 0.0066).

\FloatBarrier
\section{Additional LDARNet results}
\label{sec:supp_ldar}

\begin{table*}[t]
\caption{\textbf{\ours improves on reproduced LDARNet routing on most of the 18 tasks.}
Entries give test MCC$\times$100 with 95\% confidence intervals over ten
folds. The $\Delta$ columns give differences relative to the reproduced
router, paired by fold. Bold differences have intervals excluding zero. The
Wilcoxon tests use either task means or paired task--fold results.
``Reported'' refers to the original results, and ``reproduced'' identifies our
run of the published routing. The counts and tests refer to the differences
only (empty cells elsewhere).}
\label{tab:ldar_full}
\centering
\small
\begin{tabular}{@{}l cccc cc@{}}
\toprule
 & \multicolumn{2}{c}{published routing} & \multicolumn{2}{c}{ours} & \multicolumn{2}{c}{$\Delta$ ours $-$ reproduced} \\
\cmidrule(lr){2-3}\cmidrule(lr){4-5}\cmidrule(l){6-7}
Task & reported & reproduced & \ours{} & no conv.\ grad. & \ours{} & no conv.\ grad. \\
\midrule
H3 & 81.7{\scriptsize$\pm$0.8} & 81.4{\scriptsize$\pm$0.5} & \textbf{82.3}{\scriptsize$\pm$0.7} & 82.3{\scriptsize$\pm$0.6} & $+0.88$ & {\boldmath$+0.92$} \\
H3K14ac & 64.0{\scriptsize$\pm$0.7} & \textbf{64.8}{\scriptsize$\pm$0.9} & 64.6{\scriptsize$\pm$0.7} & 64.3{\scriptsize$\pm$0.7} & $-0.22$ & $-0.51$ \\
H3K36me3 & 68.7{\scriptsize$\pm$0.4} & 68.1{\scriptsize$\pm$0.6} & 68.2{\scriptsize$\pm$1.0} & \textbf{68.3}{\scriptsize$\pm$0.3} & $+0.04$ & $+0.15$ \\
H3K4me1 & 59.5{\scriptsize$\pm$0.7} & 59.6{\scriptsize$\pm$0.9} & 59.4{\scriptsize$\pm$0.9} & \textbf{59.9}{\scriptsize$\pm$0.9} & $-0.24$ & $+0.26$ \\
H3K4me2 & 54.2{\scriptsize$\pm$0.9} & \textbf{54.3}{\scriptsize$\pm$0.8} & 54.3{\scriptsize$\pm$1.0} & 53.0{\scriptsize$\pm$1.9} & $+0.04$ & $-1.31$ \\
H3K4me3 & 60.2{\scriptsize$\pm$1.0} & 59.7{\scriptsize$\pm$0.7} & \textbf{60.6}{\scriptsize$\pm$0.6} & 60.1{\scriptsize$\pm$0.6} & $+0.86$ & $+0.39$ \\
H3K79me3 & 69.5{\scriptsize$\pm$0.6} & \textbf{70.3}{\scriptsize$\pm$0.6} & 69.6{\scriptsize$\pm$0.9} & 69.6{\scriptsize$\pm$0.7} & $-0.71$ & $-0.77$ \\
H3K9ac & 64.6{\scriptsize$\pm$0.7} & 64.6{\scriptsize$\pm$0.6} & \textbf{65.2}{\scriptsize$\pm$0.9} & 64.7{\scriptsize$\pm$0.8} & $+0.56$ & $+0.06$ \\
H4 & 81.2{\scriptsize$\pm$0.5} & 81.3{\scriptsize$\pm$0.6} & \textbf{81.5}{\scriptsize$\pm$0.4} & 81.5{\scriptsize$\pm$0.5} & $+0.15$ & $+0.12$ \\
H4ac & 65.7{\scriptsize$\pm$0.7} & 65.4{\scriptsize$\pm$0.6} & 65.4{\scriptsize$\pm$0.8} & \textbf{65.7}{\scriptsize$\pm$0.7} & $+0.03$ & $+0.32$ \\
enhancers & 54.3{\scriptsize$\pm$1.1} & 55.6{\scriptsize$\pm$1.1} & \textbf{56.8}{\scriptsize$\pm$1.3} & 56.3{\scriptsize$\pm$1.0} & $+1.20$ & $+0.70$ \\
enhancers\_types & 44.2{\scriptsize$\pm$2.2} & 44.5{\scriptsize$\pm$1.1} & \textbf{44.9}{\scriptsize$\pm$1.5} & 44.9{\scriptsize$\pm$0.8} & $+0.40$ & $+0.40$ \\
promoter\_all & 94.3{\scriptsize$\pm$0.1} & \textbf{94.3}{\scriptsize$\pm$0.2} & 94.3{\scriptsize$\pm$0.3} & 94.3{\scriptsize$\pm$0.3} & $+0.06$ & $+0.01$ \\
promoter\_no\_tata & 94.4{\scriptsize$\pm$0.3} & 94.4{\scriptsize$\pm$0.2} & 94.4{\scriptsize$\pm$0.3} & \textbf{94.6}{\scriptsize$\pm$0.3} & $-0.02$ & $+0.18$ \\
promoter\_tata & 91.5{\scriptsize$\pm$0.6} & 91.5{\scriptsize$\pm$0.7} & \textbf{91.6}{\scriptsize$\pm$0.7} & 91.4{\scriptsize$\pm$0.4} & $+0.08$ & $-0.17$ \\
splice\_sites\_acceptors & 91.5{\scriptsize$\pm$0.9} & 92.0{\scriptsize$\pm$0.7} & 92.3{\scriptsize$\pm$0.6} & \textbf{93.6}{\scriptsize$\pm$0.7} & $+0.35$ & {\boldmath$+1.64$} \\
splice\_sites\_all & 95.8{\scriptsize$\pm$0.3} & 95.8{\scriptsize$\pm$0.3} & 96.3{\scriptsize$\pm$0.2} & \textbf{96.5}{\scriptsize$\pm$0.3} & {\boldmath$+0.52$} & {\boldmath$+0.70$} \\
splice\_sites\_donors & 93.3{\scriptsize$\pm$1.9} & 94.7{\scriptsize$\pm$1.7} & 96.4{\scriptsize$\pm$0.4} & \textbf{96.6}{\scriptsize$\pm$0.6} & {\boldmath$+1.69$} & $+1.87$ \\
\midrule
Mean over 18 tasks & 73.81 & 74.03 & 74.34 & 74.30 & $+0.32$ & $+0.28$ \\
Tasks improved (Wilcoxon $p$) & & & & & 14/18 (0.018) & 14/18 (0.090) \\
Folds improved (Wilcoxon $p$) & & & & & 109/180 (0.009) & 105/180 (0.011) \\
\bottomrule
\end{tabular}

\end{table*}

\paragraph{Per-task results.} \Cref{tab:ldar_full} gives the per-task results from \cref{fig:ldar_tasks}.
\Cref{tab:ldar} groups the results by task family and gives tests across tasks
and folds. On each task, our reproduction falls within or near the reported
interval. \ours's largest gains are on splice sites (up to $+1.7$, or $+1.9$
without the convolution gradient), H3 and enhancers. Its largest loss is on
H3K79me3 ($-0.7$). The variant without the convolution gradient improves 14 of
18 tasks ($+0.28$; 105 of 180 folds, $p{=}0.011$), and both variants stop early
at the same epoch as the baseline (about 14 epochs). Without the ratio loss,
the gain of \ours is larger on the folds where it compresses at least as much
as the baseline ($+0.65$ against $+0.19$).

\begin{table}[t]
\caption{\textbf{\ours's mean MCC gain on LDARNet persists at comparable compression.}
The upper block groups the 18 Nucleotide Transformer tasks by family and
averages test MCC$\times$100 over tasks after averaging each task over ten
folds. The lower block compares the reproduced router and \ours with the
pre-training ratio loss applied to both. Tests compare \ours with reproduced
routing at the task or task--fold level. Bases per chunk are available for the
reproduced and \ours configurations. Empty cells: the authors report no bases
per chunk, and the second block has only the reproduced routing and \ours.}
\label{tab:ldar}
\centering
\resizebox{\linewidth}{!}{\begin{tabular}{@{}l cccc@{}}
\toprule
 & \multicolumn{2}{c}{published routing} & \multicolumn{2}{c}{ours} \\
\cmidrule(lr){2-3}\cmidrule(l){4-5}
NT tasks (MCC$\times$100) & reported & reproduced & \ours{} & no conv.\ grad. \\
\midrule
Histone marks (10) & 66.93 & 66.96 & 67.10 & 66.92 \\
Enhancers (2) & 49.24 & 50.08 & 50.88 & 50.63 \\
Promoters (3) & 93.40 & 93.42 & 93.46 & 93.42 \\
Splice sites (3) & 93.54 & 94.16 & 95.02 & 95.57 \\
\midrule
Mean over 18 tasks & 73.81 & 74.03 & 74.34 & 74.30 \\
Tasks better (Wilcoxon $p$) & & & 14/18 (0.018) & 14/18 (0.090) \\
Folds better (Wilcoxon $p$) & & & 109/180 (0.009) & 105/180 (0.011) \\
Bases per chunk & & 4.66 & 4.19 & 4.14 \\
\midrule
\multicolumn{5}{@{}l}{\emph{Equal compression: ratio loss of pre-training in both}} \\
Mean over 18 tasks & & 74.12 & 74.37 & \\
Folds better (Wilcoxon $p$) & & & 105/180 (0.023) & \\
Bases per chunk & & 4.01 & 4.08 & \\
\bottomrule
\end{tabular}
}
\end{table}

\begin{table*}[t]
\caption{\textbf{On all 18 tasks, \ours improves significantly with and without the
ratio loss, and the ratio loss alone does not.} Each block reports mean test
MCC$\times$100. $\Delta$ pairs folds against the first row in the block, which
serves as the reference (dashes). The $p$-values use a Wilcoxon signed-rank
test over task--fold pairs. ``Cont.'' identifies the released checkpoint after
1,000 additional masked language modeling steps on hg38 with the specified
router.}
\label{tab:ldar_controls}
\centering
\small
\begin{tabular}{@{}lccccc@{}}
\toprule
Fine-tuning & MCC$\times$100 & $\Delta$ & tasks better & $p$ (folds) & bases/chunk \\
\midrule
\multicolumn{6}{@{}l}{\emph{Router without learning signal} (6 tasks, 60 folds)} \\
published routing & 67.25 & -- & -- & -- & 4.41 \\
frozen router & 66.86 & $-0.39$ & 1/6 & 0.162 & 3.56 \\
\ours{} (ours) & 67.48 & $+0.24$ & 3/6 & 0.448 & 3.72 \\
\ours{}, no conv.\ grad.\ (ours) & 67.23 & $-0.01$ & 3/6 & 0.900 & 3.89 \\
\midrule
\multicolumn{6}{@{}l}{\emph{Ratio loss of pre-training ($\alpha{=}0.03$, $N{=}4$), all tasks} (18 tasks, 180 folds)} \\
published routing & 74.03 & -- & -- & -- & 4.66 \\
+ ratio loss & 74.12 & $+0.09$ & 11/18 & 0.340 & 4.01 \\
\ours{} (ours) & 74.34 & $+0.32$ & 14/18 & 0.009 & 4.19 \\
\ours{} (ours) + ratio loss & 74.37 & $+0.34$ & 12/18 & 0.001 & 4.08 \\
\midrule
\multicolumn{6}{@{}l}{\emph{Ratio loss, variant without the convolution gradient} (5 tasks, 50 folds)} \\
published routing & 61.75 & -- & -- & -- & 4.55 \\
+ ratio loss & 61.74 & $-0.02$ & 2/5 & 0.502 & 4.01 \\
\ours{}, no conv.\ grad.\ (ours) & 61.36 & $-0.39$ & 2/5 & 0.253 & 4.06 \\
\ours{}, no conv.\ grad.\ (ours) + ratio loss & 61.62 & $-0.13$ & 1/5 & 0.421 & 4.00 \\
\midrule
\multicolumn{6}{@{}l}{\emph{Continued MLM pre-training on hg38, then fine-tuning} (18 tasks, 180 folds)} \\
released ckpt, published routing & 74.03 & -- & -- & -- & 4.66 \\
cont.\ (published routing), published & 74.24 & $+0.21$ & 11/18 & 0.064 & 4.80 \\
cont.\ (\ours{}, ours), published & 73.54 & $-0.49$ & 6/18 & 0.006 & 5.79 \\
cont.\ (\ours{}), \ours{} (ours) & 74.14 & $+0.11$ & 11/18 & 0.212 & 4.11 \\
released ckpt, \ours{} (ours) & 74.34 & $+0.32$ & 14/18 & 0.009 & 4.19 \\
\bottomrule
\end{tabular}

\end{table*}

\paragraph{Controls.} \Cref{tab:ldar_controls} tests other explanations of the
gain. ``Tasks better'' counts the tasks on which a configuration scores higher than
the reference row of its block. Each row in the last block identifies the starting checkpoint and the routing
method used for fine-tuning. For a continued checkpoint, the routing used
during continued pre-training appears in parentheses. With the router frozen at its pre-trained parameters, fine-tuning is
worse than with the published routing on five of six tasks ($-0.39$,
$p{=}0.16$). The frozen router keeps 3.56 bases per chunk on these tasks, so
the \ours routers (3.72 and 3.89) stay close to the compression of
pre-training, from which the published fine-tuning drifts to coarser chunks
(4.41). On the five histone tasks on which the \ours routers drifted
furthest from that compression, the ratio loss moves the variant without the
convolution gradient from $-0.39$ to $-0.13$, which is still a tie with the
published routing.

\paragraph{Continued pre-training.} Continuing the pre-training of the
released checkpoint for 1000 steps lowers the validation MLM loss from 0.911
to 0.899 with the published routing and to 0.898 with \ours. With the
published routing it gives a small gain after fine-tuning ($+0.21$, 11 of 18
tasks, $p{=}0.064$). The router pre-trained through \ours, however, drifts to
5.79 bases per chunk under the published fine-tuning (4.66 from the released
checkpoint), and the fine-tuned model loses 0.49 MCC ($p{=}0.006$).
Fine-tuning the continued checkpoint with \ours gives 74.14 MCC$\times$100
($+0.11$, $p{=}0.21$). This is 0.60 points above the published fine-tuning of
the same checkpoint ($p{<}10^{-5}$) and 0.20 points below fine-tuning the
released checkpoint directly with \ours (74.34, $p{=}0.15$). A short continued pre-training thus does not add to the
gain of \ours, which comes from training the router during fine-tuning.

\begin{table*}[t]
\caption{\textbf{On the nine Genomic Benchmarks, \ours matches the reproduced routing
on average, and its paired gain has a 95\% CI excluding zero on two tasks.}
LDARNet-110M test accuracy is reported with 95\% confidence intervals over ten
folds for each of the nine Genomic Benchmarks tasks~\cite{gresova2023genomic}.
$\Delta$ pairs folds against the reproduced router, and bold differences have
intervals excluding zero. Bases per chunk are medians across 90 runs per
routing method. Wilcoxon tests use task means or task--fold pairs.
``Reported'':~\cite{ledneva2026ldarnet}. Empty cells: the authors report no
bases per chunk, the $\Delta$ columns give no bases per chunk, and the counts
and tests refer to the $\Delta$ columns only.}
\label{tab:ldar_gb}
\centering
\small
\begin{tabular}{@{}l cccc cc@{}}
\toprule
 & \multicolumn{2}{c}{published routing} & \multicolumn{2}{c}{ours} & \multicolumn{2}{c}{$\Delta$ ours $-$ reproduced} \\
\cmidrule(lr){2-3}\cmidrule(lr){4-5}\cmidrule(l){6-7}
Task & reported & reproduced & \ours{} & no conv.\ grad. & \ours{} & no conv.\ grad. \\
\midrule
Coding vs.\ intergenomic & 95.6 & 95.5{\scriptsize$\pm$0.1} & \textbf{95.6}{\scriptsize$\pm$0.1} & 95.5{\scriptsize$\pm$0.1} & {\boldmath$+0.09$} & $-0.00$ \\
Drosophila enhancers Stark & 80.3 & 80.5{\scriptsize$\pm$0.4} & 80.3{\scriptsize$\pm$0.7} & \textbf{80.9}{\scriptsize$\pm$0.6} & $-0.21$ & $+0.42$ \\
Human Ensembl regulatory & 93.9 & \textbf{94.3}{\scriptsize$\pm$0.1} & 94.2{\scriptsize$\pm$0.1} & 94.2{\scriptsize$\pm$0.1} & $-0.06$ & $-0.04$ \\
Human OCR Ensembl & 80.1 & 79.8{\scriptsize$\pm$0.4} & \textbf{80.2}{\scriptsize$\pm$0.4} & 80.1{\scriptsize$\pm$0.4} & {\boldmath$+0.36$} & $+0.21$ \\
Human enhancers Cohn & 74.7 & \textbf{74.7}{\scriptsize$\pm$0.3} & 74.3{\scriptsize$\pm$0.5} & 74.7{\scriptsize$\pm$0.5} & $-0.40$ & $+0.00$ \\
Human enhancers Ensembl & 91.8 & 91.8{\scriptsize$\pm$0.1} & 91.8{\scriptsize$\pm$0.2} & \textbf{91.9}{\scriptsize$\pm$0.1} & $+0.01$ & $+0.10$ \\
Human non-TATA promoters & 96.6 & 96.6{\scriptsize$\pm$0.1} & 96.6{\scriptsize$\pm$0.2} & \textbf{96.8}{\scriptsize$\pm$0.1} & $-0.05$ & {\boldmath$+0.12$} \\
Human vs.\ worm & 97.7 & \textbf{97.7}{\scriptsize$\pm$0.1} & 97.7{\scriptsize$\pm$0.1} & 97.7{\scriptsize$\pm$0.1} & $+0.04$ & $+0.00$ \\
Mouse enhancers Ensembl & 85.6 & 83.8{\scriptsize$\pm$2.2} & \textbf{86.0}{\scriptsize$\pm$1.2} & 84.2{\scriptsize$\pm$2.0} & $+2.15$ & $+0.41$ \\
\midrule
Mean over 9 tasks & 88.48 & 88.30 & 88.52 & 88.44 & $+0.21$ & $+0.14$ \\
Bases per chunk (median) & & 3.88 & 3.74 & 3.27 & & \\
Tasks improved (Wilcoxon $p$) & & & & & 5/9 (0.91) & 7/9 (0.04) \\
Folds improved (Wilcoxon $p$) & & & & & 53/90 (0.22) & 57/90 (0.01) \\
\bottomrule
\end{tabular}

\end{table*}

\paragraph{Genomic Benchmarks.} LDARNet is also evaluated on the nine Genomic
Benchmarks tasks \cite{gresova2023genomic}, for which no fine-tuning code is
released. We follow its description: the protocol above with top-1 accuracy
for model selection and early stopping, its learning rates and batch sizes,
and sequences of up to 4096 bases (\cref{tab:ldar_gb}). Our published-routing
runs reproduce the reported accuracies (88.30 against 88.48 on average; within
0.4 points on eight tasks and 1.8 below on the smallest, mouse enhancers with
242 test sequences). The gains here are small. The variant without the
convolution gradient improves 7 of 9 tasks ($+0.14$; Wilcoxon $p{=}0.04$ over
tasks, $p{=}0.01$ over 90 folds), but it also keeps more chunks (3.27 against
3.88 bases per chunk). Full \ours compresses like the published routing (3.74)
and improves 5 of 9 tasks ($+0.21$, $p{=}0.22$ over folds). Its mean rests on
the noisiest task ($+2.2$ on mouse enhancers), and it is significantly better
on open chromatin ($+0.36$, at 3.94 against 3.53 bases per chunk) and on
coding sequences ($+0.09$). Three tasks are close to saturation (above 95\%).
With the learning rate of $10^{-3}$ of the regulatory task, the compression
varies widely across folds (published routing 1.5--23 bases per chunk, \ours
1.0--3.5), while the mean accuracies of the three routings differ by at most
0.06.

\FloatBarrier
\section{Long-context language modeling: token selection in ReMamba}
\label{sec:supp_remamba}

\paragraph{Host.} ReMamba \cite{yuan2025remamba} adapts a Mamba-2.8B language
model \cite{gu2023mamba} to long prompts by selective compression. A first pass
over the prompt yields last-layer hidden states; the leading fraction $\phi$ of
the prompt ($p$ in \cite{yuan2025remamba}) forms a region of $n$ candidates,
each scored by the cosine similarity between a learned key of the candidate and
a learned query of the last prompt token, and the $k=\lceil\rho n\rceil$
highest-scoring candidates are kept. A second pass reads projections of the
kept states in place of the region, followed by the rest of the prompt, and
scales the step $\Delta$ of the kept states by their scores. The second pass
thus runs on a compacted sequence, and the selector learns only through the
scores of the candidates it already keeps. We follow the published fine-tuning:
LoRA of rank 32 for one epoch on 300k OpenOrca instructions
\cite{mukherjee2023orca} and 209k long instructions from OpenOrca and
LongAlpaca-12k \cite{chen2024longlora}, truncated to 6000 tokens, with an
effective batch of 128 and a learning rate of $2{\cdot}10^{-5}$. Prompts longer
than 1600 tokens draw $\phi\sim U(0.1,0.3)$ and $\rho\sim U(0.05,0.2)$ per
example, with the same draw in every arm.

\paragraph{\ours in ReMamba.} We keep the first pass, the region, the scores and
the budget, and run the second pass on the uncompacted sequence with keep
probability $p_i$ for candidate $i$ and one elsewhere. Every Mamba block of the
second pass uses \ours in place of its convolution and gates its step by $p$
(\cref{sec:gating}), so that at binary $p$ the second pass is ReMamba's compacted
one (\cref{prop:block}), and every candidate, kept or not, receives a gradient.
We train the selector with the soft forward pass (\cref{sec:training}): the
second pass uses the exact-budget LapSum probabilities, with a learned
temperature. A second series with the single cosine schedule trains \ours
straight-through: the temperature decays log-linearly from 1 to $10^{-3}$ in
units of the score range of each example, and the straight-through gradient is
scaled by it, which keeps the gradient bounded as the selection sharpens. This
series includes a zero mask, with the same step gate and straight-through
training but ordinary convolutions on the masked inputs in place of \ours.
All arms are evaluated with ReMamba's published compacted inference.

\paragraph{Evaluation.} We use the English LongBench-E \cite{bai2024longbench}
(13 tasks, 3668 examples) with the authors' evaluation code: greedy decoding,
prompts truncated to 5500 tokens, $\phi=0.18$ and $\rho=0.009$. Rescoring the
authors' released predictions reproduces their reported average of 27.86
exactly. The ReMamba paper specifies a cosine learning-rate schedule for its
single epoch \cite{yuan2025remamba}. The released multi-GPU script advances the scheduler once per process at each
optimizer step. The cosine length is set in optimizer steps, so the learning
rate completes a cycle every 1,989 of our 3,977 steps, or every 994 steps on
the authors' eight GPUs. Our run ends at the peak learning rate. We read this as a side effect of multi-GPU training, not as the intended
schedule, and evaluate both versions. The \emph{published schedule} uses the
released script as written, which lets us compare our reproduction with the
released adapter. The other setting uses the \emph{single cosine} over the
epoch described in the paper. Each arm is trained once;
re-evaluating the same adapters in float32 changes the average by at most 0.2
points.
The task families are question answering (Qasper, MultiFieldQA, HotpotQA,
2WikiMQA), summarization (GovReport, MultiNews), few-shot learning (TREC,
TriviaQA, SAMSum), synthetic tasks (PassageCount, PassageRetrieval), and code
completion (LCC, RepoBench-P). As in LongBench-E, length buckets are based on
word counts.

\begin{table*}[t]
\caption{\textbf{Soft-forward \ours improves on reproduced ReMamba under both training
schedules.} The table gives LongBench-E scores for compacted ReMamba
(Mamba-2.8B) at $\phi{=}0.18$ and $\rho{=}0.009$. $\Delta$ uses the
reproduction from the same schedule and series as its reference. The
straight-through series includes the zero-mask run and its reference, which
scores 26.27 (\cref{tab:remamba_tasks}). ``Reproduced'' is our run of the
published selection and ``ours'' the selector trained through the soft forward
pass of \ours. $\Delta$ is empty for the published rows and for the reproduced
selection itself.}
\label{tab:remamba}
\centering
\small
\begin{tabular}{@{}l cccccc r@{}}
\toprule
 & \multicolumn{5}{c}{task family} & & \\
\cmidrule(lr){2-6}
Selection & QA & Summ. & Few-shot & Synth. & Code & Avg. & $\Delta$ \\
\midrule
\multicolumn{8}{@{}l}{\emph{Published models \cite{yuan2025remamba}}} \\
Mamba-2.8B fine-tuned without compression, reported & 21.8 & 19.2 & 37.9 & 2.3 & 38.1 & 24.63 & \\
ReMamba, reported & 21.2 & 19.3 & 46.7 & 5.0 & 44.3 & 27.86 & \\
ReMamba, released adapter, our evaluation & 21.2 & 19.1 & 46.3 & 4.3 & 44.3 & 27.62 & \\
\midrule
\multicolumn{8}{@{}l}{\emph{Published training schedule}} \\
ReMamba, reproduced & 21.2 & 18.5 & 46.9 & 3.3 & 44.1 & 27.48 & \\
soft forward (ours) & 20.7 & 18.9 & 48.3 & 3.4 & 45.6 & \textbf{27.98} & $+0.50$ \\
\midrule
\multicolumn{8}{@{}l}{\emph{Single cosine schedule}} \\
ReMamba, reproduced & 19.3 & 18.5 & 44.0 & 3.2 & 43.6 & 26.13 & \\
soft forward (ours) & 19.3 & 18.8 & 45.7 & 3.9 & 46.0 & \textbf{27.06} & $+0.92$ \\
zero mask & 18.9 & 19.4 & 44.0 & 3.2 & 42.1 & 25.91 & $-0.36$ \\
\bottomrule
\end{tabular}

\end{table*}

\begin{table*}[t]
\caption{\textbf{Few-shot and code tasks show gains under both ReMamba schedules.}
LongBench-E differences use the reproduced selection in the same block as the
reference. Confidence intervals use 2,000 paired bootstrap resamples of
examples stratified by task and length bucket. Bold differences have intervals
excluding zero. The zero mask belongs to the second series and is shown with
the reproduced selection of that series.}
\label{tab:remamba_tasks}
\centering
\footnotesize
\setlength{\tabcolsep}{4pt}
\resizebox{\textwidth}{!}{\begin{tabular}{@{}ll c ccc ccc ccc@{}}
\toprule
 & & reported & \multicolumn{3}{c}{published schedule} & \multicolumn{3}{c}{single cosine schedule} & \multicolumn{3}{c}{single cosine schedule, second series} \\
\cmidrule(lr){4-6}\cmidrule(lr){7-9}\cmidrule(l){10-12}
Family & Task (metric) & ReMamba & reproduced & soft forward & $\Delta$ & reproduced & soft forward & $\Delta$ & reproduced & zero mask & $\Delta$ \\
\midrule
\multirow{4}{*}{QA} & Qasper (F1) & 16.76 & 17.31 & 18.12 & $+0.81$ & 16.48 & 17.50 & $+1.02$ & 16.63 & 17.65 & $+1.02$ \\
 & MultiFieldQA (F1) & 26.39 & 26.83 & 26.08 & $-0.75$ & 25.99 & 26.21 & $+0.22$ & 26.49 & 26.35 & $-0.14$ \\
 & HotpotQA (F1) & 20.56 & 20.83 & 20.61 & $-0.22$ & 18.32 & 17.42 & $-0.90$ & 18.48 & 16.44 & $-2.04$ \\
 & 2WikiMQA (F1) & 21.18 & 19.79 & 18.13 & $-1.66$ & 16.31 & 16.02 & $-0.29$ & 16.91 & 15.25 & $-1.66$ \\
\multirow{2}{*}{Summ.} & GovReport (ROUGE-L) & 19.67 & 18.40 & 18.88 & $+0.48$ & 18.52 & 19.02 & $+0.50$ & 18.27 & 19.29 & $+1.02$ \\
 & MultiNews (ROUGE-L) & 18.86 & 18.62 & 18.97 & $+0.35$ & 18.53 & 18.59 & $+0.06$ & 18.23 & 19.50 & $+1.27$ \\
\multirow{3}{*}{Few-shot} & TREC (accuracy) & 48.67 & 49.00 & 50.67 & $+1.67$ & 49.00 & 49.67 & $+0.67$ & 48.67 & 48.67 & $0.00$ \\
 & TriviaQA (F1) & 57.73 & 58.04 & 59.84 & $+1.80$ & 50.10 & 53.45 & $+3.35$ & 51.51 & 49.69 & $-1.82$ \\
 & SAMSum (ROUGE-L) & 33.65 & 33.54 & 34.43 & $+0.89$ & 32.96 & 34.06 & $+1.10$ & 32.49 & 33.51 & $+1.02$ \\
\multirow{2}{*}{Synth.} & PassageCount & 3.21 & 0.33 & 0.05 & $-0.28$ & 0.36 & 1.14 & $+0.78$ & 0.21 & 0.00 & $-0.21$ \\
 & PassageRetrieval (accuracy) & 6.83 & 6.33 & 6.67 & $+0.34$ & 6.00 & 6.67 & $+0.67$ & 6.33 & 6.33 & $0.00$ \\
\multirow{2}{*}{Code} & LCC (edit sim.) & 48.21 & 47.24 & 51.03 & $+3.79$ & 47.10 & 51.38 & $+4.28$ & 47.40 & 45.59 & $-1.81$ \\
 & RepoBench-P (edit sim.) & 40.40 & 40.92 & 40.23 & $-0.69$ & 40.08 & 40.65 & $+0.57$ & 39.87 & 38.57 & $-1.30$ \\
\midrule
\multicolumn{2}{@{}l}{Average} & 27.86 & 27.48 & 27.98 & {\boldmath$+0.50$} & 26.13 & 27.06 & {\boldmath$+0.92$} & 26.27 & 25.91 & $-0.36$ \\
\multicolumn{2}{@{}l}{\quad 95\% CI over examples} & & \multicolumn{3}{r}{[0.08, 0.94]} & \multicolumn{3}{r}{[0.50, 1.35]} & \multicolumn{3}{r}{[$-0.90$, 0.14]} \\
\multicolumn{2}{@{}l}{\quad $\Delta$ at 0--4k / 4--8k / 8k+ words} & & \multicolumn{3}{r}{$+0.6$ / $+0.5$ / $+0.4$} & \multicolumn{3}{r}{$+0.8$ / $+1.5$ / $+0.6$} & \multicolumn{3}{r}{$-0.9$ / $-0.3$ / $+0.1$} \\
\midrule
\multicolumn{12}{@{}l}{$\Delta$ by family, 95\% CI over examples} \\
\multicolumn{2}{@{}l}{\quad QA} & & \multicolumn{3}{r}{$-0.45$ [$-1.30$, 0.38]} & \multicolumn{3}{r}{$+0.01$ [$-0.69$, 0.73]} & \multicolumn{3}{r}{$-0.71$ [$-1.76$, 0.24]} \\
\multicolumn{2}{@{}l}{\quad Summarization} & & \multicolumn{3}{r}{{\boldmath$+0.41$} [0.02, 0.78]} & \multicolumn{3}{r}{$+0.28$ [$-0.08$, 0.65]} & \multicolumn{3}{r}{{\boldmath$+1.14$} [0.73, 1.53]} \\
\multicolumn{2}{@{}l}{\quad Few-shot} & & \multicolumn{3}{r}{{\boldmath$+1.45$} [0.24, 2.66]} & \multicolumn{3}{r}{{\boldmath$+1.71$} [0.45, 2.96]} & \multicolumn{3}{r}{$-0.27$ [$-1.88$, 1.25]} \\
\multicolumn{2}{@{}l}{\quad Synthetic} & & \multicolumn{3}{r}{$+0.03$ [$-0.26$, 0.46]} & \multicolumn{3}{r}{$+0.72$ [$-0.03$, 1.56]} & \multicolumn{3}{r}{$-0.11$ [$-0.58$, 0.40]} \\
\multicolumn{2}{@{}l}{\quad Code} & & \multicolumn{3}{r}{{\boldmath$+1.55$} [0.39, 2.70]} & \multicolumn{3}{r}{{\boldmath$+2.42$} [1.31, 3.64]} & \multicolumn{3}{r}{{\boldmath$-1.56$} [$-2.91$, $-0.19$]} \\
\bottomrule
\end{tabular}
}
\end{table*}

\paragraph{Results.} With the published schedule, our reproduction scores 27.48. The released
adapter scores 27.62 in our evaluation, and the authors report 27.86
(\cref{tab:remamba}). Training the selector through the soft forward pass of
\ours improves on the reproduced selection with both schedules: by 0.50 with
the published schedule (paired interval over examples $[0.08, 0.94]$; three
further runs with 2.5 and 5 times the initial temperature and with an MLP
selector gain 0.68--0.83, each with an interval that excludes zero) and by
0.92 with the single cosine schedule ($[0.50, 1.35]$;
\cref{tab:remamba_tasks}). In the second series, the zero mask scores 0.36
below its reproduced selection ($[-0.90, 0.14]$), and straight-through \ours,
which differs from it only in the convolution, scores 27.12, 1.21 above the
zero mask ($[0.72, 1.73]$). The single cosine schedule gives lower scores than
the published one (the reproduced selection drops from 27.48 to 26.13).

\paragraph{Where the gain comes from.} The gains lie in the few-shot tasks
($+1.4$ and $+1.7$ with the published and the single cosine schedule; TriviaQA
$+1.8$ and $+3.4$), code completion ($+1.5$ and $+2.4$; LCC $+3.8$ and $+4.3$) and
summarization ($+0.4$ and $+0.3$). The intervals exclude zero for the few-shot
tasks and code completion with both schedules, and for summarization with the
published one. Document question answering does not improve ($-0.5$ and
$0.0$), and the two synthetic tasks stay near chance. ReMamba itself shows the
same profile against Mamba fine-tuned without compression in the reported
results (QA 21.2 against 21.8, few-shot 46.7 against 37.9). The zero mask
gains only in summarization ($+1.1$ over its reproduced selection) and loses
in code completion ($-1.6$).

\paragraph{What matters.} The convolution matters in this host: with the same
step gate and straight-through training, a zero mask in place of \ours loses
1.21 points (27.12 to 25.91). The soft forward pass works in this host because training drives the keep
probabilities to the vertices. The keys and queries spread the cosine scores
about twentyfold, with little change in the learned temperature. By the end of
training, 97\% of the keep mass lies on the top-$k$ candidates, so the soft
forward pass uses a selection close to the hard mask at inference.

\FloatBarrier
\section{Vision--language: visual tokens in Cobra}
\label{sec:supp_cobra}

\begin{table*}[t]
\caption{\textbf{\ours retains the highest mean at 10\% of LM FLOPs after matching
training time.} Entries average the percentage scores on GQA, VQAv2, TextVQA,
POPE, VSR and VizWiz. Values with $\pm$ give the mean and standard deviation
over three seeds. The fixed scoring rules are deterministic and have no $\pm$.
The matched-cost columns give competing learned selectors the GPU-hours of the
reference scan or the runtime of the CUDA kernel. In the LoRA block, the upper
rows keep the scoring rule fixed while training the adapters. Dashes: the
training-free rules have no training cost to match, and ``Dense + LoRA, then
$\Delta$'' is defined only with LoRA. REINFORCE-RLOO draws four samples per
step with a leave-one-out baseline.}
\label{tab:vision}
\centering
\scriptsize
\setlength{\tabcolsep}{4pt}
\cobrafit{\begin{tabular}{@{}cl cc cc cc@{}}
\toprule
 & & \multicolumn{2}{c}{Frozen backbone, 2k steps} & \multicolumn{2}{c}{10\%, matched training cost} & \multicolumn{2}{c}{Same LoRA for every rule} \\
\cmidrule(lr){3-4}\cmidrule(lr){5-6}\cmidrule(l){7-8}
 & Token selection & 10\% & 25\% & scan GPU-h & kernel time & 10\% & 25\% \\
\midrule
 & Dense (all 729 tokens) & 65.8 & 65.8 & -- & -- & 65.2{\scriptsize$\pm$0.8} & 65.2{\scriptsize$\pm$0.8} \\
\midrule
\multirow{7}{*}{\rotatebox[origin=c]{90}{training-free}} & Uniform & 57.4 & 62.0 & -- & -- & 59.9{\scriptsize$\pm$0.2} & 62.5{\scriptsize$\pm$0.1} \\
 & Random & 57.4{\scriptsize$\pm$0.1} & 62.0{\scriptsize$\pm$0.2} & -- & -- & 59.6{\scriptsize$\pm$0.1} & 62.3{\scriptsize$\pm$0.4} \\
 & $\Delta$ of layer 0 & 58.6 & 63.0 & -- & -- & 60.1{\scriptsize$\pm$0.1} & 63.2{\scriptsize$\pm$0.3} \\
 & Dense + LoRA, then $\Delta$ of layer 0 & -- & -- & -- & -- & 58.2{\scriptsize$\pm$0.8} & 62.4{\scriptsize$\pm$0.8} \\
 & DTP \cite{park2026dtp}, best reachable site & 57.5 & 61.4 & -- & -- & 58.7{\scriptsize$\pm$0.1} & 60.9{\scriptsize$\pm$0.4} \\
 & FastV \cite{chen2024fastv} & 56.2 & 62.3 & -- & -- & 57.9{\scriptsize$\pm$0.2} & 62.3{\scriptsize$\pm$0.3} \\
 & DART \cite{wen2025dart} & 54.9 & 58.7 & -- & -- & 57.8{\scriptsize$\pm$0.3} & 59.8{\scriptsize$\pm$0.3} \\
\midrule
\multirow{7}{*}{\rotatebox[origin=c]{90}{learned}} & DyVM predictor \cite{wu2025dyvm} & 54.9{\scriptsize$\pm$4.6} & 59.6{\scriptsize$\pm$3.6} & 56.5{\scriptsize$\pm$0.3} & 55.4{\scriptsize$\pm$1.7} & 58.7{\scriptsize$\pm$1.4} & 59.3{\scriptsize$\pm$1.8} \\
 & DyVM predictor + forward \cite{wu2025dyvm} & 54.3{\scriptsize$\pm$0.9} & 59.0{\scriptsize$\pm$3.4} & 47.6{\scriptsize$\pm$0.2} & 54.7{\scriptsize$\pm$0.9} & 57.5{\scriptsize$\pm$0.4} & 61.5{\scriptsize$\pm$2.7} \\
 & REINFORCE on the compacted network & 60.1{\scriptsize$\pm$0.3} & 63.7{\scriptsize$\pm$0.6} & 60.5{\scriptsize$\pm$0.8} & 60.4{\scriptsize$\pm$0.4} & 59.7{\scriptsize$\pm$1.2} & 62.6{\scriptsize$\pm$0.7} \\
 & REINFORCE-RLOO (4 samples), compacted network & 60.8{\scriptsize$\pm$0.3} & 64.2{\scriptsize$\pm$0.1} & 60.9{\scriptsize$\pm$0.6} & 60.5{\scriptsize$\pm$0.5} & 61.5{\scriptsize$\pm$0.1} & 64.1{\scriptsize$\pm$0.4} \\
 & Learned, zero mask & 56.8{\scriptsize$\pm$2.8} & 62.8{\scriptsize$\pm$0.5} & 56.1{\scriptsize$\pm$2.5} & 57.9{\scriptsize$\pm$0.3} & 56.6{\scriptsize$\pm$2.6} & 60.1{\scriptsize$\pm$2.5} \\
 & Learned, zero-mask conv.\ + step gating & 58.4{\scriptsize$\pm$0.2} & 63.0{\scriptsize$\pm$0.4} & 58.7{\scriptsize$\pm$0.2} & 58.4{\scriptsize$\pm$0.5} & 59.8{\scriptsize$\pm$0.5} & 62.9{\scriptsize$\pm$1.3} \\
 & Learned, \ours{} (ours) & \textbf{61.1}{\scriptsize$\pm$0.2} & \textbf{64.3}{\scriptsize$\pm$0.1} & \textbf{61.1}{\scriptsize$\pm$0.2} & \textbf{61.1}{\scriptsize$\pm$0.1} & \textbf{61.7}{\scriptsize$\pm$0.4} & \textbf{64.2}{\scriptsize$\pm$0.4} \\
\bottomrule
\end{tabular}
}
\end{table*}

\paragraph{Host.} Cobra-3B \cite{zhao2025cobra} encodes a $384{\times}384$
image with DINOv2 and SigLIP \cite{oquab2024dinov2,zhai2023siglip} into a
$27{\times}27$ grid of 14-pixel patches, projects the 729 tokens into the
embedding space of a Mamba-2.8B language model and places them before the text.
We evaluate with Cobra's harness on GQA, VQAv2 (the subsampled split), TextVQA,
POPE, VSR and VizWiz
\cite{hudson2019gqa,goyal2017vqav2,singh2019textvqa,li2023pope,liu2023vsr,gurari2018vizwiz}
and report their average. Our dense model matches the published numbers within
0.16 points; DTP \cite{park2026dtp} at its native site (layers 16 and 45)
reproduces its published 65.8 and 64.9 at $r=0.9$ and $0.5$. Budgets are stated
as the fraction of the language model's FLOPs that remains.

\paragraph{\ours in Cobra.} The selector, a three-layer MLP on each projected
token concatenated with the mean of the tokens, scores the visual tokens at the
projector. In training, every Mamba block runs on the full sequence; its causal
convolution ($K{=}4$) on the 729-token span is replaced by \ours and its step
$\Delta_i$ multiplied by $p_i$ (\cref{sec:gating}), so that at binary $p$ the
network is the compacted one (\cref{prop:block}). Weighted LapSum~\cite{struski2025lapsum} gives the keep probabilities at exact
budget $k$ (measured fp32 budget error $4.6{\cdot}10^{-5}$ to
$1.2{\cdot}10^{-4}$ tokens). Here we use the soft forward pass of
\cref{sec:training}: the LapSum temperature is annealed from 1.5 to 0.2 and a
penalty $\lambda\sum_i p_i(1-p_i)$ grows from $\lambda=0.01$ to 0.5. Unlike the annealed H-Net routers (\cref{sec:ablations}), this training
reaches the vertices: at 25\%, the held-out loss is 0.722 with the soft
forward pass and 0.722 with the hard mask. Training with the
straight-through estimator of \cref{sec:training} instead (hard top-$k$ forward)
gives $60.6\pm0.2$ at 10\% and $64.2\pm0.1$ at 25\%, against $61.1\pm0.2$ and
$64.3\pm0.1$ (\cref{tab:supp_cobra_ablation}). Inference gathers the
hard top-$k$ tokens at the projector and runs Cobra's fused kernels on the
shorter sequence. Training uses batch size 1, learning rate $10^{-4}$ and 2,000
steps on the LLaVA-1.5 mixture \cite{liu2024llava15} without any image of the
evaluation sets. The in-network variant of \cref{tab:supp_cobra_sweep} places
the same selector after layer 2 to 15, the early site of DTP's schedule at the
same budget.

\paragraph{CUDA kernel.} The reference implementation is a PyTorch scan. The
CUDA kernel used here is an earlier, unfused version, not the kernel of
\cref{sec:supp_kernel}. On Cobra's shapes ($B{=}1$, $d_\text{inner}{=}5120$,
$L{=}732$) it agrees with the scan to $1.0{\cdot}10^{-6}$ relative error in the
output and in the gradients with respect to $x$, $p$, the taps and the bias,
for fractional and binary $p$. Forward and backward of one layer take 22.4\,ms
with the scan and 1.09\,ms with the kernel, against 0.28\,ms for a plain
depthwise convolution; a training step takes 1.79--1.89\,s and 0.27--0.28\,s.
Because this unfused kernel is slower than the kernel of
\cref{sec:supp_kernel}, the peers matched to its time receive more steps than
our final kernel would give them, so the kernel-matched comparison is
conservative for \ours.

\paragraph{LoRA.} Cobra's Mamba blocks read \texttt{in\_proj.weight},
\texttt{dt\_proj.weight} and \texttt{out\_proj.weight} directly, bypassing the
forward of a wrapper. Our LoRA module (rank 16) therefore exposes the merged
weight $W+(\alpha/r)BA$ as its \texttt{weight}, so that every call site and the
fused kernels see the adapters. Every rule of the LoRA setting trains the same
adapters for 2,000 steps; ``tune the dense model, then prune'' applies the
training-free rules to the dense model with LoRA. With LoRA, REINFORCE and RLOO
train the adapters with the ordinary gradient of the loss of the sampled
compacted network (RLOO: the mean over its four samples), while the selector
keeps its score-function gradient.

\paragraph{Baselines.} Both zero-mask variants share our selector and the LapSum settings, schedule
and penalty. The ``Zero-mask conv.\ + step gating'' row retains step gating
and replaces \ours with a plain convolution of $x\odot p$. The ``zero mask''
row uses the plain Mamba block with a hard top-$k$ mask in the forward pass
and the LapSum gradient in the backward pass. We floor the mask at $10^{-2}$
as in DyVM because exact zeros give non-finite gradients in the fused kernels. DyVM \cite{wu2025dyvm} trains its
predictor with its own objective on the frozen backbone, with plain masking
(``DyVM predictor'') and with its token rearrangement (``DyVM predictor +
forward''). The training-free rules are uniform and random
selection, the $\Delta$ score of layer 0 at our site, DTP at its best reachable
site and the adaptations of FastV, DART and VTW
\cite{chen2024fastv,wen2025dart,lin2025vtw} to Cobra by the authors of DTP.
\Cref{tab:supp_cobra_audit} lists how each rule deviates from its own regime.

\paragraph{Ablations of \ours.} The three ablations share the selector, LapSum settings, schedule, penalty,
2,000 steps, three seeds and the CUDA kernel. Each changes a single component
of \ours. \emph{No convolution
gradient} detaches $p$ at the input of \ours (\cref{sec:training}): the forward
pass is identical, and the selector learns only through the step gate.
\emph{Without step gating} keeps \ours but not $\Delta_i\mapsto p_i\Delta_i$,
so even at binary $p$ a dropped token decays and writes the scan state and the
training network is not the compacted one. \emph{REINFORCE} runs the stock
Cobra with its fused kernels on the physically compacted tokens of a sample
$m\sim\mathrm{Bernoulli}(p)$ and on the greedy top-$k$ tokens, both without a
graph, and weights $\log P(m)$ by the difference of their losses, the
self-critical estimator of the speech experiments (\cref{sec:supp_speech}); it
takes 0.19\,s per step against 0.29\,s for \ours. \emph{Straight-through
without the convolution gradient} and \emph{straight-through without step
gating} make the first two changes to the straight-through variant of \ours.
\emph{REINFORCE-RLOO} draws four samples per step and uses, for each, the mean
loss of the other three as its baseline. All arms are evaluated with the same hard top-$k$ gather at the projector.
With the soft forward pass the selector learns through the convolution taps,
and step gating can be removed (61.2 against 61.1). With the straight-through
forward, removing either path costs 7 to 10 points (50.3 without the
convolution gradient and 50.4 without step gating against 60.6 at 10\%, 57.3 and
54.0 against 64.2 at 25\%). REINFORCE-RLOO
comes within 0.3 points of \ours at 10\% (60.8) and 0.1 at 25\% (64.2).

\paragraph{Matched cost.} In the GPU-hour comparison, the other learned methods receive the time of
2,000 \ours steps with the PyTorch scan: 4,532\,s at 10\% and 5,080\,s at
25\%, including validation. This gives them 4.7--8$\times$ as many steps
(\cref{tab:supp_cobra_matched}). In the CUDA comparison, they receive the
measured \ours runtime of 1,025\,s, or 1.1--1.9$\times$ as many steps
(\cref{tab:supp_cobra_fk}). We timed RLOO on A100 GPUs only. To estimate its
matched step counts (9,324 for the scan and 2,110 for the kernel), we use the
ratio of REINFORCE runtimes on the two machines. \Cref{tab:supp_cobra_cost} lists the costs.

\begin{figure}[t]
\centering
\includegraphics[width=\linewidth]{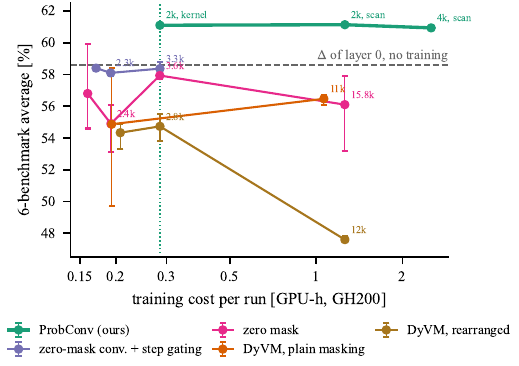}
\caption{\textbf{\ours leads the plotted training-time comparisons at the tight Cobra
budget.} The figure gives the six-benchmark mean at 10\% of LM FLOPs against
the cost of one GH200 run. Labels give optimizer steps, and bars span three
seeds. The dotted line marks the cost of \ours with the CUDA kernel.}
\label{fig:cobra_cost}
\end{figure}

\begin{figure}[t]
\centering
\includegraphics[width=\linewidth]{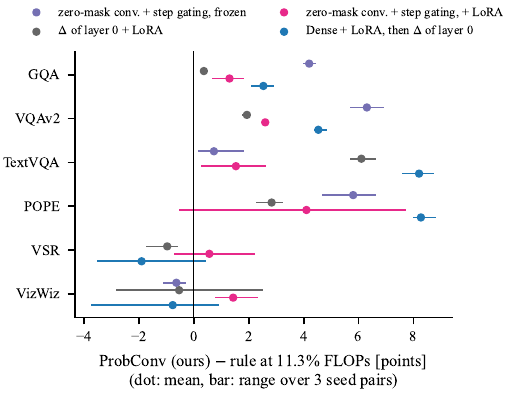}
\caption{\textbf{\ours scores higher on GQA and VQAv2 in every seed pair across the
four comparisons shown.} The plot gives \ours's score minus the comparison
score at a nominal 10\% of LM FLOPs (11.3\% for the benchmark prompts). Dots
give the mean, and bars span the three paired differences.}
\label{fig:supp_cobra_forest}
\end{figure}

\paragraph{More training does not close the gap.} At 10\% of the language model's FLOPs, \ours has the highest six-benchmark
mean under both training-time comparisons.
With the GPU-hours of our PyTorch scan, 4.7--8$\times$ more steps, the
zero-mask arm and DyVM reach only 56.1 and 56.5, zero-mask convolution with
step gating 58.7, REINFORCE 60.5 and RLOO 60.9, all below \ours (61.1), and
the DyVM predictor with rearrangement scores 47.6, with POPE at 51.5--51.9
(\cref{fig:cobra_cost}). Our CUDA kernel matches the scan to $10^{-6}$ in the
output and all gradients and makes a training step 6.6--6.9$\times$ faster;
retrained with it, the selector again scores 61.1, with no benchmark
significantly different. Given its measured time, the zero-mask arms reach 58.4
and 57.9 and rearranged DyVM 54.7; REINFORCE, given the same time (3,777 steps),
reaches 60.4 and RLOO 60.5. Across all six time-matched comparisons, the gains on GQA, VQAv2 and POPE
remain significant in all three seed pairs. The six comparisons include three
at the scan budget and three at the measured kernel time
(\cref{tab:supp_cobra_sig}). The TextVQA gain is significant in 17 of the 18
pairs. \Cref{fig:supp_cobra_forest} shows per-benchmark differences for four
further comparisons. The scores of \ours change little after 500 steps
(\cref{tab:supp_cobra_curve}).

\paragraph{Evaluation.} Learned rules use three seeds. For each pair of runs with the same seed index, we compare the two methods on
the same questions: an exact McNemar test for the binary GQA, POPE and VSR,
and a paired $z$-test for the soft scores of VQAv2, TextVQA and VizWiz. We
apply the Holm correction over the six benchmarks within each seed pair
(\cref{tab:supp_cobra_sig}). The harness
scores POPE by $\arg\max$ over ``yes'' and ``no''; exact bf16 ties, scored as
``yes'', occur in 0.6--1.2\% of the POPE questions for every rule and do not
favor \ours.

\begin{table*}[t]
\caption{\textbf{\ours's lead over matched zero-mask training grows as the Cobra budget
tightens.} The table reports the six-benchmark mean for a frozen Cobra-3B
backbone. Budgets give the share of LM FLOPs retained. Dense: 65.8. Below 50\%
of the FLOPs DTP's native site (16,45) is unreachable, and we report the best
of the sites (2,45), (5,45) and (16,45). The in-network \ours variant uses the
first cut in that schedule. Learned entries average three seeds.}
\label{tab:supp_cobra_sweep}
\centering
\small
\cobrafit{\begin{tabular}{@{}l cccccc@{}}
\toprule
LLM FLOPs & 10\% & 15\% & 25\% & 35\% & 48\% & 67\% \\
\midrule
Uniform & 57.4 & 59.3 & 62.0 & 63.6 & 64.3 & 65.1 \\
Random (3 seeds) & 57.4 & 59.8 & 62.0 & 63.2 & 64.2 & 64.9 \\
$\Delta$ of layer 0 & 58.6 & 60.6 & 63.0 & 64.0 & 64.7 & 65.2 \\
DTP \cite{park2026dtp}, best reachable site & 57.5 & 60.8 & 61.4 & 61.9 & 64.7 & \textbf{65.8} \\
FastV \cite{chen2024fastv} & 56.2 & 59.5 & 62.3 & 63.6 & 64.5 & 65.2 \\
DART \cite{wen2025dart} & 54.9 & 56.9 & 58.7 & 60.1 & 61.4 & 62.7 \\
VTW \cite{lin2025vtw} & 44.6 & 44.8 & 44.9 & 47.5 & 54.1 & 58.9 \\
DyVM predictor \cite{wu2025dyvm} (3 seeds) & 54.9 & 57.5 & 59.6 & 62.0 & 62.1 & 62.1 \\
DyVM predictor + forward \cite{wu2025dyvm} (3 seeds) & 54.3 & 56.6 & 59.0 & 58.3 & 60.6 & 61.9 \\
Learned, zero-mask conv.\ + step gating, at the projector (3 seeds) & 58.4 & 61.2 & 63.0 & 64.2 & 65.1 & 65.6 \\
Learned, \ours{} (ours), in-network site (3 seeds) & 59.4 & 60.2 & 62.3 & 63.9 & 65.2 & 65.7 \\
Learned, \ours{} (ours), at the projector (3 seeds) & \textbf{61.1} & \textbf{63.1} & \textbf{64.3} & \textbf{65.1} & \textbf{65.4} & 65.6 \\
\bottomrule
\end{tabular}
}
\end{table*}

\begin{table*}[t]
\caption{\textbf{The largest gains over the matched zero-mask control occur on GQA,
VQAv2 and POPE.} The table reports six benchmark scores for frozen Cobra-3B at
a nominal 10\% of LM FLOPs. Learned rows average three seeds. The dense row
provides the all-token reference.}
\label{tab:supp_cobra_perbench}
\centering
\small
\cobrafit{\begin{tabular}{@{}l cccccc c@{}}
\toprule
10\% LLM FLOPs, frozen backbone & GQA & VQAv2 & TextVQA & POPE & VSR & VizWiz & Avg. \\
\midrule
Dense (all tokens) & 62.2 & 77.8 & 58.2 & 88.2 & 58.4 & 49.8 & 65.8 \\
\midrule
$\Delta$ of layer 0 & 55.4 & 67.9 & 46.5 & 76.4 & \textbf{55.6} & 49.6 & 58.6 \\
DyVM predictor \cite{wu2025dyvm} & 49.6 & 59.8 & 45.3 & 70.6 & 54.5 & 49.2 & 54.9 \\
Learned, zero mask & 51.8 & 63.7 & 46.4 & 75.1 & 55.0 & 48.9 & 56.8 \\
Learned, zero-mask conv.\ + step gating & 52.1 & 65.5 & 53.5 & 74.2 & 55.3 & \textbf{49.9} & 58.4 \\
Learned, \ours{} (ours) & \textbf{56.3} & \textbf{71.8} & \textbf{54.2} & \textbf{80.0} & 55.2 & 49.3 & \textbf{61.1} \\
\bottomrule
\end{tabular}
}
\end{table*}

\begin{table*}[t]
\caption{\textbf{Cobra's useful gradient paths depend on the training forward.} The
default \ours uses an annealed soft forward, and straight-through training
uses hard top-$k$. ``Deployed forward'' indicates whether a binary mask
reproduces physical compaction. Results use 2,000 steps on frozen Cobra-3B.
Benchmark scores average three seeds, and Avg.\ gives the mean and standard
deviation. Dashes mark absent training components.}
\label{tab:supp_cobra_ablation}
\centering
\small
\cobrafit{\begin{tabular}{@{}l ccc cccccc c c@{}}
\toprule
 & \multicolumn{3}{c}{training signal} & \multicolumn{7}{c}{10\% of the FLOPs} & 25\% \\
\cmidrule(lr){2-4}\cmidrule(lr){5-11}\cmidrule(l){12-12}
Selector trained & \makecell{deployed\\forward} & \makecell{grad.\ via\\conv.\ taps} & \makecell{grad.\ via\\step gate} & GQA & VQAv2 & TextVQA & POPE & VSR & VizWiz & Avg. & Avg. \\
\midrule
\ours{} (ours) & \checkmark & \ours & \checkmark & 56.3 & 71.8 & 54.2 & 80.0 & 55.2 & 49.3 & 61.1{\scriptsize$\pm$0.2} & 64.3{\scriptsize$\pm$0.1} \\
\ours{}, straight-through (ours) & \checkmark & \ours & \checkmark & 55.7 & 70.7 & 52.4 & 79.1 & 55.7 & 49.7 & 60.6{\scriptsize$\pm$0.2} & 64.2{\scriptsize$\pm$0.1} \\
\ours{}, no conv.\ grad.\ (ours) & \checkmark & -- & \checkmark & 48.4 & 58.6 & 52.5 & 69.0 & 53.5 & 50.1 & 55.3{\scriptsize$\pm$2.4} & 61.1{\scriptsize$\pm$2.1} \\
\ours{}, straight-through, no conv.\ grad.\ (ours) & \checkmark & -- & \checkmark & 45.7 & 52.2 & 46.1 & 55.1 & 53.4 & 49.5 & 50.3{\scriptsize$\pm$0.9} & 57.3{\scriptsize$\pm$1.5} \\
\ours{}, straight-through w/o step gating (ours) & \checkmark & \ours & -- & 45.8 & 51.1 & 42.2 & 59.3 & 53.9 & 49.8 & 50.4{\scriptsize$\pm$2.4} & 54.0{\scriptsize$\pm$1.3} \\
\ours{} w/o step gating (ours) & -- & \ours & -- & 55.8 & 71.2 & 54.2 & 81.2 & 55.1 & 49.5 & 61.2{\scriptsize$\pm$0.1} & 64.4{\scriptsize$\pm$0.2} \\
Zero-mask conv.\ + step gating & -- & mask & \checkmark & 52.1 & 65.5 & 53.5 & 74.2 & 55.3 & 49.9 & 58.4{\scriptsize$\pm$0.2} & 63.0{\scriptsize$\pm$0.4} \\
REINFORCE on the compacted network & \checkmark & -- & -- & 55.0 & 69.5 & 53.5 & 77.6 & 56.2 & 49.3 & 60.1{\scriptsize$\pm$0.3} & 63.7{\scriptsize$\pm$0.6} \\
REINFORCE-RLOO (4 samples) on the compacted network & \checkmark & -- & -- & 55.7 & 70.7 & 53.7 & 80.4 & 55.3 & 49.2 & 60.8{\scriptsize$\pm$0.3} & 64.2{\scriptsize$\pm$0.1} \\
\bottomrule
\end{tabular}
}
\end{table*}

\begin{table}[t]
\caption{\textbf{Matching the PyTorch-scan budget leaves \ours ahead of the listed
selectors.} Entries give Cobra-3B's six-benchmark mean over three seeds with a
frozen backbone. Each budget compares 2,000 steps with the reference-scan
GPU-hour setting. Parentheses give step counts.}
\label{tab:supp_cobra_matched}
\centering
\cobrafit{\begin{tabular}{@{}l cc cc@{}}
\toprule
 & \multicolumn{2}{c}{10\% LLM FLOPs} & \multicolumn{2}{c}{25\% LLM FLOPs} \\
\cmidrule(lr){2-3}\cmidrule(l){4-5}
Token selection & 2k steps & scan GPU-h & 2k steps & scan GPU-h \\
\midrule
DyVM predictor \cite{wu2025dyvm} & 54.9 & 56.5 (11.0k) & 59.6 & 59.9 (14.6k) \\
DyVM predictor + forward \cite{wu2025dyvm} & 54.3 & 47.6 (12.2k) & 59.0 & 58.2 (13.7k) \\
Learned, zero mask & 56.8 & 56.1 (15.8k) & 62.8 & 63.5 (16.3k) \\
Learned, \ours{} (ours) & \textbf{61.1} & \textbf{60.9} (4k) & \textbf{64.3} & \textbf{64.5} (4k) \\
\bottomrule
\end{tabular}
}
\end{table}

\begin{table}[t]
\caption{\textbf{Scores on the three tracked Cobra benchmarks change little after 500
steps.} Entries give \ours's mean over three seeds at each training checkpoint
for the two LM FLOP budgets.}
\label{tab:supp_cobra_curve}
\centering
\cobrafit{\begin{tabular}{@{}l ccc ccc@{}}
\toprule
 & \multicolumn{3}{c}{10\% LLM FLOPs} & \multicolumn{3}{c}{25\% LLM FLOPs} \\
\cmidrule(lr){2-4}\cmidrule(l){5-7}
Steps & GQA & TextVQA & POPE & GQA & TextVQA & POPE \\
\midrule
500 & 55.9 & 54.8 & 80.4 & 60.6 & 57.4 & 86.0 \\
1,000 & 56.5 & 53.9 & 81.6 & 60.5 & 57.2 & 85.9 \\
1,500 & 56.3 & 54.0 & 80.8 & 60.3 & 57.2 & 86.0 \\
2,000 & 56.3 & 54.2 & 80.0 & 60.2 & 57.2 & 85.7 \\
4,000 & 55.7 & 54.2 & 79.9 & 60.5 & 57.5 & 85.7 \\
\bottomrule
\end{tabular}
}
\end{table}

\begin{table*}[t]
\caption{\textbf{With the CUDA kernel, \ours reaches the same six-benchmark mean as
with the PyTorch scan (seed 0), and peers given its measured time stay below
it.} At 10\% of LM FLOPs, the two \ours rows show seed 0. Other rows average
three seeds. The measured CUDA time budget is 1,025\,s including validation.
Rows marked ``first pass'' use a preliminary budget of 691\,s.}
\label{tab:supp_cobra_fk}
\centering
\small
\cobrafit{\begin{tabular}{@{}l c cccccc c@{}}
\toprule
10\% LLM FLOPs & Steps & GQA & VQAv2 & TextVQA & POPE & VSR & VizWiz & Avg. \\
\midrule
\ours{} (ours), PyTorch scan (seed 0) & 2,000 & 56.3 & 71.6 & 55.1 & 79.2 & 54.9 & 49.5 & 61.1 \\
\ours{} (ours), CUDA kernel (seed 0) & 2,000 & 56.5 & 71.8 & 54.6 & 79.5 & 54.4 & 49.5 & 61.1 \\
\midrule
Learned, zero-mask conv.\ + step gating & 3,340 & 51.9 & 65.1 & 52.7 & 74.8 & 55.9 & 50.0 & 58.4 \\
Learned, zero mask & 3,570 & 52.7 & 66.0 & 49.2 & 75.6 & 55.1 & 49.0 & 57.9 \\
DyVM predictor + forward \cite{wu2025dyvm} & 2,750 & 48.6 & 58.5 & 49.2 & 67.6 & 54.8 & 49.5 & 54.7 \\
\midrule
Learned, zero-mask conv.\ + step gating, first pass & 2,250 & 52.0 & 65.0 & 52.0 & 74.5 & 55.5 & 49.9 & 58.1 \\
Learned, zero mask, first pass & 2,410 & 50.7 & 60.0 & 42.4 & 74.7 & 53.4 & 48.3 & 54.9 \\
\bottomrule
\end{tabular}
}
\end{table*}

\begin{table}[t]
\caption{\textbf{With the same LoRA for every rule, \ours keeps the highest mean at
both budgets.} The table reports Cobra-3B's six-benchmark mean over three
seeds at each LM FLOP budget. Parentheses give the individual dense scores or
the observed range where shown. Bold marks the highest score within a block.
At 25\% one seed of the zero-mask arm has VizWiz 38.8.}
\label{tab:supp_cobra_lora}
\centering
\small
\cobrafit{\begin{tabular}{@{}l cc@{}}
\toprule
Same LoRA, 2k steps & 10\% & 25\% \\
\midrule
Dense + LoRA, no pruning & \multicolumn{2}{c}{65.2 (65.8 / 65.6 / 64.3)} \\
\midrule
\multicolumn{3}{@{}l}{\emph{Tune the dense model, then prune}} \\
Uniform & 57.0 & 61.5 \\
DTP \cite{park2026dtp} & 57.2 & 59.9 \\
$\Delta$ of layer 0 & \textbf{58.2} & \textbf{62.4} \\
\midrule
\multicolumn{3}{@{}l}{\emph{Pruning-aware LoRA}} \\
DTP \cite{park2026dtp} & 58.7 & 60.9 \\
FastV \cite{chen2024fastv} & 57.9 & 62.3 \\
DART \cite{wen2025dart} & 57.8 & 59.8 \\
Random & 59.6 & 62.3 \\
DyVM predictor \cite{wu2025dyvm} & 58.7 (57.8--60.4) & 59.3 (58.2--61.3) \\
DyVM predictor + forward \cite{wu2025dyvm} & 57.5 & 61.5 (58.7--64.0) \\
Learned, zero mask & 56.6 (55.1--59.6) & 60.1 (57.4--62.3) \\
REINFORCE on the compacted network & 59.7 (58.3--60.6) & 62.6 \\
REINFORCE-RLOO (4 samples) & 61.5 & 64.1 \\
Learned, zero-mask conv.\ + step gating & 59.8 & 62.9 (61.4--63.8) \\
Uniform & 59.9 & 62.5 \\
$\Delta$ of layer 0 & 60.1 & 63.2 \\
Learned, \ours{} (ours) & \textbf{61.7} (61.4--62.2) & \textbf{64.2} (63.8--64.6) \\
\bottomrule
\end{tabular}
}
\end{table}

\begin{table}[t]
\caption{\textbf{CUDA shortens a 2,000-step \ours run relative to the PyTorch scan.}
Training costs are per run on one GH200 at a nominal 10\% of LM FLOPs and
include validation. The table gives optimizer steps, elapsed seconds and
GPU-hours.}
\label{tab:supp_cobra_cost}
\centering
\cobrafit{\begin{tabular}{@{}l ccc@{}}
\toprule
10\% LLM FLOPs, frozen backbone & Steps & s per run & GPU-h per run \\
\midrule
\ours{} (ours), PyTorch scan & 2,000 & 4,532 & 1.26 \\
\ours{} (ours), CUDA kernel & 2,000 & 1,025 & 0.28 \\
Learned, zero-mask conv.\ + step gating & 2,000 & $\approx$614 & 0.17 \\
Learned, zero mask & 2,000 & $\approx$574 & 0.16 \\
DyVM predictor / + forward & 2,000 & $\approx$696 / 746 & 0.19 / 0.21 \\
Peers, scan-GPU-h matched & 11.0k--15.8k & $\approx$4,500 & $\approx$1.3 \\
\bottomrule
\end{tabular}
}
\end{table}

\begin{table*}[t]
\caption{\textbf{Against the matched zero-mask convolution with step gating, \ours's
gains on GQA, VQAv2 and POPE are significant in every seed pair at both
budgets.} Entries give \ours minus the comparison method in points for seed
pairs 0, 1 and 2. Unless indicated, the backbone is frozen and the nominal
budget is 10\% of LM FLOPs. Asterisks mark $p<0.05$ after Holm correction
across the six benchmarks within each seed pair. ``n.s.'' indicates zero
significant pairs. ``Mixed'' indicates significant differences in both
directions.}
\label{tab:supp_cobra_sig}
\centering
\scriptsize
\setlength{\tabcolsep}{2.5pt}
\cobrafit{\begin{tabular}{@{}l cccccc@{}}
\toprule
\ours{} (ours) minus & GQA & VQAv2 & TextVQA & POPE & VSR & VizWiz \\
\midrule
\multicolumn{7}{@{}l}{\emph{Equal steps (2k), frozen backbone}} \\
Zero-mask conv.\ + step gating, 10\% & +4.0$^\ast$/+4.2$^\ast$/+4.4$^\ast$ & +6.9$^\ast$/+6.3$^\ast$/+5.7$^\ast$ & +1.8$^\ast$/+0.2/+0.2 & +4.7$^\ast$/+6.1$^\ast$/+6.6$^\ast$ & n.s. & $-$0.3/$-$0.5/$-$1.1$^\ast$ \\
Zero-mask conv.\ + step gating, 25\% & +2.3$^\ast$/+1.1$^\ast$/+3.3$^\ast$ & +2.6$^\ast$/+1.7$^\ast$/+3.6$^\ast$ & +0.5/0.0/+0.5 & +3.1$^\ast$/+1.4$^\ast$/+3.2$^\ast$ & n.s. & n.s. \\
Zero mask, 10\% & +5.3$^\ast$/+6.6$^\ast$/+1.6$^\ast$ & +8.6$^\ast$/+12.9$^\ast$/+2.6$^\ast$ & +10.1$^\ast$/+11.2$^\ast$/+2.3$^\ast$ & +6.3$^\ast$/+7.3$^\ast$/+1.2$^\ast$ & n.s. & mixed \\
Zero mask, 25\% & +2.9$^\ast$/+1.6$^\ast$/+2.2$^\ast$ & +1.4$^\ast$/+3.0$^\ast$/+2.2$^\ast$ & +2.3$^\ast$/+3.5$^\ast$/+0.1 & +4.1$^\ast$/+1.1$^\ast$/+0.4 & n.s. & mixed \\
\midrule
\multicolumn{7}{@{}l}{\emph{Peers given the GPU-hours of our reference scan (steps in parentheses), 10\% unless noted}} \\
Zero mask (15.8k) & +8.0$^\ast$/+3.2$^\ast$/+3.3$^\ast$ & +14.5$^\ast$/+5.5$^\ast$/+7.9$^\ast$ & +12.7$^\ast$/+6.7$^\ast$/+9.2$^\ast$ & +10.7$^\ast$/+3.3$^\ast$/+5.3$^\ast$ & +0.6/+0.4/$-$2.5$^\ast$ & +1.0/+1.0$^\ast$/+0.0 \\
DyVM predictor (11.0k) & +4.5$^\ast$/+4.9$^\ast$/+2.9$^\ast$ & +6.8$^\ast$/+10.2$^\ast$/+8.3$^\ast$ & +7.1$^\ast$/+8.1$^\ast$/+9.8$^\ast$ & +5.0$^\ast$/+9.7$^\ast$/+6.4$^\ast$ & +1.3/$-$1.7/$-$0.5 & +1.4$^\ast$/$-$0.1/+0.0 \\
DyVM predictor + forward (12.2k) & +14.1$^\ast$/+13.6$^\ast$/+14.1$^\ast$ & +25.2$^\ast$/+24.7$^\ast$/+25.1$^\ast$ & +12.7$^\ast$/+12.1$^\ast$/+11.5$^\ast$ & +27.6$^\ast$/+28.2$^\ast$/+29.1$^\ast$ & +2.3/+2.7$^\ast$/+2.5$^\ast$ & $-$0.6/$-$0.5/$-$1.0 \\
DyVM predictor + forward (2k) & +7.2$^\ast$/+9.2$^\ast$/+7.8$^\ast$ & +13.2$^\ast$/+16.7$^\ast$/+12.6$^\ast$ & +8.6$^\ast$/+5.7$^\ast$/+3.7$^\ast$ & +7.5$^\ast$/+14.9$^\ast$/+12.8$^\ast$ & +0.8/+1.1/+0.8 & +0.6/+0.2/$-$0.7 \\
Zero mask, 25\% (16.3k) & +2.0$^\ast$/+0.1/+1.3$^\ast$ & +1.9$^\ast$/$-$0.2/+2.4$^\ast$ & +0.7/+1.1$^\ast$/+1.7$^\ast$ & +0.4/+0.6$^\ast$/+1.0$^\ast$ & +1.0/$-$0.8/+0.6 & +0.3/+0.3/+0.3 \\
DyVM predictor, 25\% (14.6k) & +1.9$^\ast$/+0.6/+10.3$^\ast$ & +2.2$^\ast$/+1.7$^\ast$/+19.3$^\ast$ & +3.6$^\ast$/+2.1$^\ast$/+14.5$^\ast$ & +0.5/$-$0.1/+20.4$^\ast$ & +0.2/$-$0.2/+0.4 & +0.7/+0.2/+1.5$^\ast$ \\
\midrule
\multicolumn{7}{@{}l}{\emph{Peers matched to the measured time of the CUDA kernel, 10\%}} \\
Zero-mask conv.\ + step gating (3.3k) & +4.7$^\ast$/+4.6$^\ast$/+3.8$^\ast$ & +6.5$^\ast$/+8.1$^\ast$/+5.4$^\ast$ & +1.6$^\ast$/+4.0$^\ast$/$-$0.8 & +3.2$^\ast$/+6.3$^\ast$/+6.3$^\ast$ & $-$0.3/$-$1.2/$-$0.7 & $-$0.3/$-$0.9/$-$0.8 \\
Zero mask (3.6k) & +4.8$^\ast$/+3.3$^\ast$/+2.7$^\ast$ & +6.8$^\ast$/+4.6$^\ast$/+5.8$^\ast$ & +2.0$^\ast$/+4.6$^\ast$/+8.8$^\ast$ & +4.7$^\ast$/+2.9$^\ast$/+5.5$^\ast$ & $-$0.1/+2.5$^\ast$/$-$2.2$^\ast$ & +0.9$^\ast$/+0.5/$-$0.6 \\
DyVM predictor + forward (2.75k) & +7.6$^\ast$/+8.8$^\ast$/+6.8$^\ast$ & +12.4$^\ast$/+15.5$^\ast$/+11.9$^\ast$ & +5.9$^\ast$/+5.3$^\ast$/+3.9$^\ast$ & +10.2$^\ast$/+14.7$^\ast$/+12.4$^\ast$ & +1.0/+1.1/$-$1.0 & +0.2/$-$0.3/$-$0.7 \\
\ours{} (ours), CUDA kernel minus scan (seed 0) & +0.2 & +0.3 & $-$0.5 & +0.3 & $-$0.5 & +0.1 \\
\midrule
\multicolumn{7}{@{}l}{\emph{Ablations of \ours, 2,000 steps (\cref{tab:supp_cobra_ablation})}} \\
No conv.\ gradient, 10\% & +9.7$^\ast$/+9.4$^\ast$/+4.5$^\ast$ & +16.0$^\ast$/+16.7$^\ast$/+7.0$^\ast$ & +4.4$^\ast$/+1.9$^\ast$/$-$1.0 & +10.0$^\ast$/+14.2$^\ast$/+8.9$^\ast$ & +1.9/+3.8$^\ast$/$-$0.7 & $-$0.6/$-$0.5/$-$1.3$^\ast$ \\
No conv.\ gradient, 25\% & +1.4$^\ast$/+3.0$^\ast$/+8.2$^\ast$ & +2.1$^\ast$/+5.7$^\ast$/+12.6$^\ast$ & n.s. & +3.8$^\ast$/+6.5$^\ast$/+10.1$^\ast$ & n.s. & n.s. \\
W/o step gating, 25\% & n.s. & n.s. & n.s. & +0.1/$-$0.0/$-$0.6$^\ast$ & n.s. & n.s. \\
REINFORCE, 10\% & +2.3$^\ast$/+0.7/+0.9$^\ast$ & +2.9$^\ast$/+1.6$^\ast$/+2.0$^\ast$ & +1.2$^\ast$/+0.7/+0.5 & +2.8$^\ast$/+2.3$^\ast$/+2.2$^\ast$ & $-$2.0$^\ast$/+0.2/$-$1.1 & n.s. \\
REINFORCE, 25\% & +1.4$^\ast$/+0.7/$-$0.2 & +2.1$^\ast$/+0.5/+0.2 & +1.3$^\ast$/+0.3/+0.2 & +2.1$^\ast$/+2.0$^\ast$/+0.2 & n.s. & n.s. \\
\midrule
\multicolumn{7}{@{}l}{\emph{Same LoRA for every rule, 10\% (\ours{} with LoRA minus)}} \\
$\Delta$ of layer 0 + LoRA & +0.5/+0.3/+0.3 & +2.0$^\ast$/+1.8$^\ast$/+2.0$^\ast$ & +6.0$^\ast$/+5.7$^\ast$/+6.6$^\ast$ & +3.2$^\ast$/+2.3$^\ast$/+3.0$^\ast$ & $-$1.7$^\ast$/$-$0.6/$-$0.6 & $-$1.3$^\ast$/+2.5$^\ast$/$-$2.8$^\ast$ \\
Uniform + LoRA & +0.9$^\ast$/+0.6/$-$0.2 & +3.7$^\ast$/+3.0$^\ast$/+3.6$^\ast$ & +7.6$^\ast$/+6.7$^\ast$/+7.7$^\ast$ & +3.0$^\ast$/+2.1$^\ast$/0.0 & $-$2.9$^\ast$/+2.1$^\ast$/+1.2 & $-$3.0$^\ast$/$-$0.3/$-$1.8$^\ast$ \\
Zero-mask conv.\ + step gating + LoRA & +1.8$^\ast$/+1.4$^\ast$/+0.7 & +2.5$^\ast$/+2.7$^\ast$/+2.6$^\ast$ & +1.7$^\ast$/+0.3/+2.6$^\ast$ & +5.1$^\ast$/+7.7$^\ast$/$-$0.5 & $-$0.7/+2.2$^\ast$/+0.2 & +0.8/+2.3$^\ast$/+1.2 \\
DTP + LoRA & +2.4$^\ast$/+2.2$^\ast$/+1.7$^\ast$ & +4.0$^\ast$/+4.4$^\ast$/+4.2$^\ast$ & +8.0$^\ast$/+7.9$^\ast$/+8.6$^\ast$ & +4.4$^\ast$/+4.0$^\ast$/+2.9$^\ast$ & $-$0.6/+2.0$^\ast$/+1.7$^\ast$ & $-$1.1$^\ast$/+0.1/$-$2.4$^\ast$ \\
DyVM predictor + forward, LoRA & +2.9$^\ast$/+2.6$^\ast$/+2.6$^\ast$ & +8.8$^\ast$/+9.6$^\ast$/+7.5$^\ast$ & +10.5$^\ast$/+11.0$^\ast$/+11.4$^\ast$ & +3.4$^\ast$/+2.7$^\ast$/+3.5$^\ast$ & $-$1.9$^\ast$/+1.1/$-$0.2 & $-$1.8$^\ast$/+3.2$^\ast$/$-$1.6$^\ast$ \\
Dense + LoRA, then $\Delta$ of layer 0 & +2.6$^\ast$/+2.9$^\ast$/+2.1$^\ast$ & +4.4$^\ast$/+4.4$^\ast$/+4.8$^\ast$ & +8.3$^\ast$/+7.6$^\ast$/+8.7$^\ast$ & +8.0$^\ast$/+8.8$^\ast$/+8.0$^\ast$ & $-$3.5$^\ast$/$-$2.6$^\ast$/+0.4 & $-$3.7$^\ast$/+0.9$^\ast$/+0.5 \\
\bottomrule
\end{tabular}
}
\end{table*}

\begin{table*}[t]
\caption{\textbf{The Cobra comparisons use each baseline under the conditions listed
here.} The table gives the original setting, the implementation evaluated on
Cobra and the additional comparisons for each method.}
\label{tab:supp_cobra_audit}
\centering
\footnotesize
\begin{tabular}{@{}p{0.15\linewidth}p{0.29\linewidth}p{0.31\linewidth}p{0.17\linewidth}@{}}
\toprule
Rule & Native regime & Run here & Remedy \\
\midrule
Uniform, random, $\Delta$ of layer 0 & training-free & frozen backbone, hard gather at the projector & + pruning-aware LoRA \\
DTP \cite{park2026dtp} & training-free, in-network at layers (16,45) & faithful; reproduces its Table~1 & + LoRA; on dense + LoRA \\
FastV, DART, VTW \cite{chen2024fastv,wen2025dart,lin2025vtw} & training-free & DTP's adaptations to Cobra & reference rows only \\
DyVM \cite{wu2025dyvm} & predictor + 30-epoch full fine-tuning, token rearrangement, two distillation losses, three stages & one stage, frozen backbone, its five-term objective (block selection has no causal analogue); plain masking and rearranged & GPU-hour matching; LoRA \\
\bottomrule
\end{tabular}

\end{table*}

\begin{figure}[t]
\centering
\includegraphics[width=\linewidth]{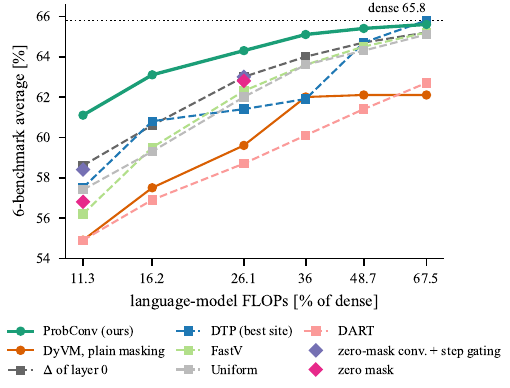}
\caption{\textbf{The gap over zero masking increases as the language-model budget
falls.} The curves show the six-benchmark mean for frozen Cobra-3B. Solid
curves show learned selectors, dashed curves show training-free rules, and
diamonds identify zero-mask training of the matched selector. Learned scores
average three seeds. Values in \cref{tab:supp_cobra_sweep}.}
\label{fig:supp_cobra_budget}
\end{figure}

\paragraph{Results.} \Cref{tab:supp_cobra_sweep} and
\cref{fig:supp_cobra_budget} give the whole budget range: \ours at the
projector leads every rule below 67\% of the FLOPs, and the in-network
variant is 1.7--2.9 points behind it up to 25\%. At 10\% the gain over the
zero-mask arm with step gating comes from GQA ($+4.2$), VQAv2 ($+6.3$) and POPE
($+5.8$) (\cref{tab:supp_cobra_perbench}). At 10\%, under both cost
matchings, the zero-mask and DyVM peers stay below the untrained $\Delta$ score
of layer 0 (58.6; \cref{tab:supp_cobra_matched,tab:supp_cobra_fk}), while
REINFORCE and RLOO reach 60.4--60.9, still below \ours, and the CUDA kernel
reproduces the scan-trained selector on every benchmark (largest difference
0.5, none significant; \cref{tab:supp_cobra_sig}). With LoRA, each
training-free rule run both ways (uniform, DTP and the $\Delta$ score) is better
tuned for pruning than tuned densely and pruned afterwards, by 1.5--2.8 points
at 10\% (\cref{tab:supp_cobra_lora}; its dense row lists the three seeds in
parentheses). Tuned for pruning, random selection, FastV
and DART reach 59.6, 57.9 and 57.8 at 10\%, the learned DyVM predictor and
zero mask 58.7 and 56.6, and REINFORCE and RLOO 59.7 and 61.5, all below \ours
(61.7); at 25\% RLOO comes within 0.1 points (64.1 against 64.2).

\paragraph{Which part of \ours matters.} \Cref{tab:supp_cobra_ablation}
separates the parts of the training signal; its two gradient columns state, as
in \cref{tab:supp_vim_ablation}, which operator carries the gradient through
the convolution taps and whether the step gate carries one. Without the gradient through the
convolution taps the forward pass is unchanged and reaches the vertices (soft
and hard held-out losses 0.773--0.806 and 0.772--0.806 at 10\%, against
0.731--0.734 for \ours), but the selector is worse than the zero mask and
unstable: 54.2, 53.7 and 58.1 at 10\%, 62.9, 61.6 and 58.7 at 25\%, with
the losses on VQAv2 ($-13.2$), POPE ($-11.0$) and GQA ($-7.9$), significant
in all three seed pairs at both budgets (\cref{tab:supp_cobra_sig}). The step
gate alone thus carries little information about which tokens to keep. Without
step gating the selector matches \ours (61.1, 61.1, 61.3 and 64.2, 64.5, 64.6;
within 1.2 points on every benchmark), although its training forward differs
from the deployed network (held-out loss 0.756--0.758 soft against
0.735--0.746 hard at 10\%). With the compacted network, REINFORCE is below \ours on every seed: it scores
59.8--60.3 at 10\%, compared with 61.0--61.3 for \ours, and 63.1--64.2 at
25\%, compared with 64.2--64.4. The largest deficits are on POPE ($-2.4$) and
VQAv2 ($-2.2$), both significant in all three seed pairs at 10\%. It is above
every other learned peer of \cref{tab:vision} except its four-sample variant
RLOO. In Cobra the selector therefore learns from the gradient of
the expected compacted convolution; training on the deployed network is not
sufficient, and the exact scan is not necessary.

\begin{figure}[t]
\centering
\includegraphics[width=\linewidth]{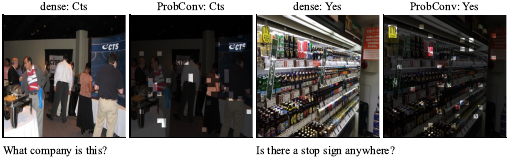}
\caption{\textbf{In the left example, \ours is the only one of the seven visualized
selection rules to read the logo correctly.} The left example asks about the
logo, and the right asks about a stop sign. \ours retains 46 of 729 tokens,
and dimmed patches are dropped. Each panel shows the answers from dense Cobra
and \ours at a nominal 10\% of LM FLOPs; on the left, the zero-mask arms
answer ``Ctas'' and ``Gcts'' and DTP ``Gcts''. Both examples belong to the 19
fixed examples shown with every rule in
\cref{fig:supp_cobra_tokens_a,fig:supp_cobra_tokens_b,fig:supp_cobra_tokens_c,fig:supp_cobra_tokens_ours}.}
\label{fig:cobra_examples}
\end{figure}

\begin{figure}[t]
\centering
\includegraphics[width=\linewidth]{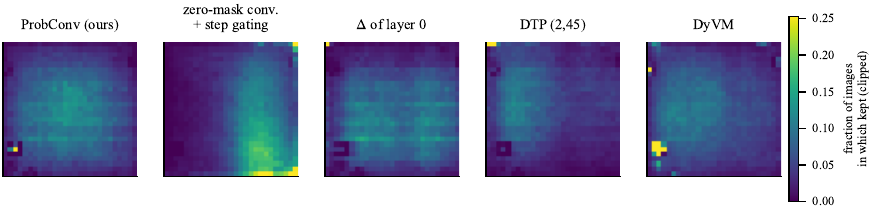}
\caption{\textbf{Cobra selectors differ in how strongly they favor fixed positions.}
The heat maps give the fraction of images in which each $27{\times}27$ grid
position is retained at 10\% of LM FLOPs. Frequencies use 500 POPE, 1,000
TextVQA and 398 GQA images. Brighter positions are selected more often.}
\label{fig:supp_cobra_freq}
\end{figure}

\begin{figure*}[p]
\centering
\includegraphics[width=\textwidth,height=0.86\textheight,keepaspectratio]{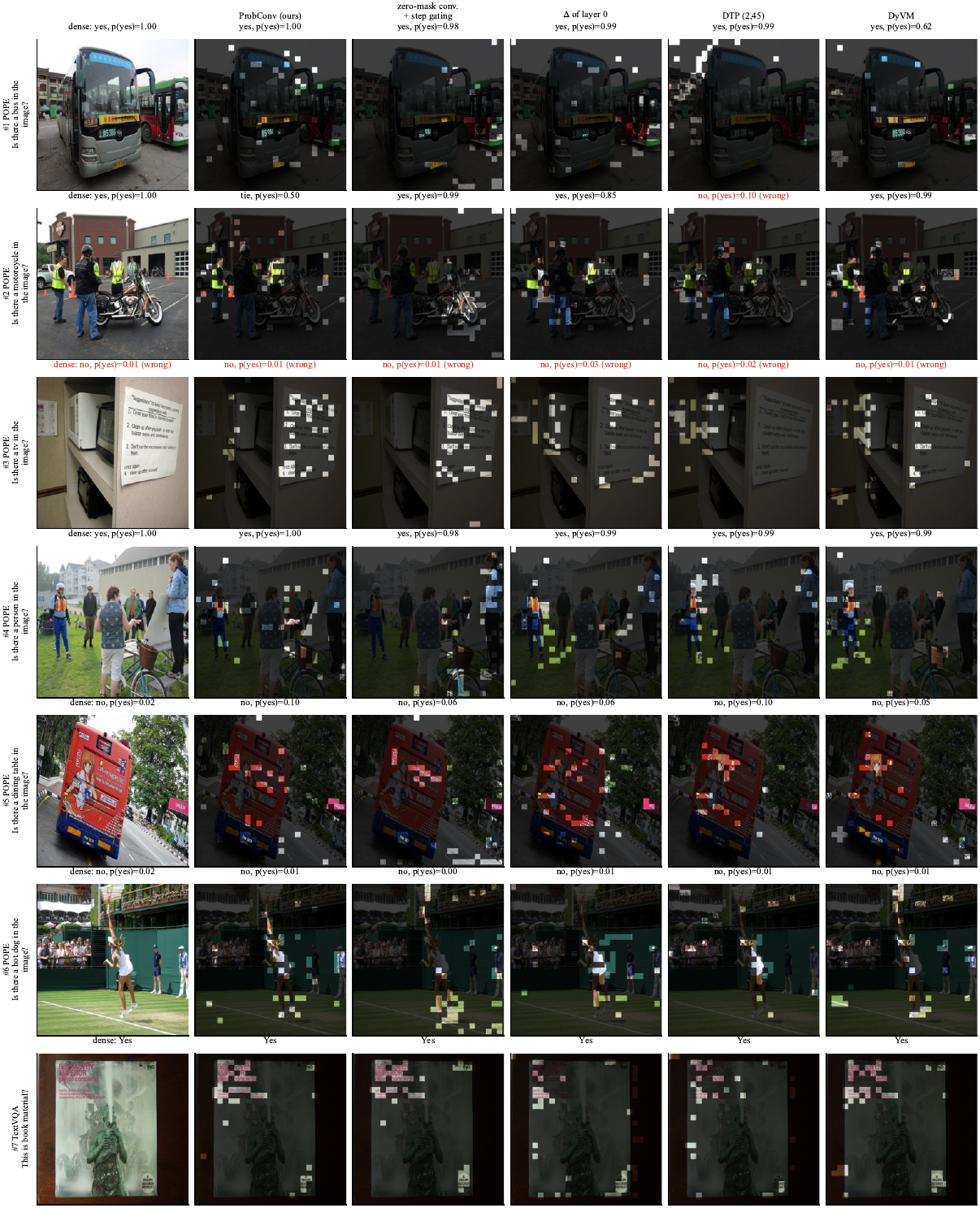}
\caption{\textbf{The same seven POPE questions reveal different selections and answers
across rules.} These are examples 1--7 of the 19 fixed Cobra examples. The
nominal 10\% LM FLOP budget retains 46 of 729 tokens at the projector, or 34
after layer 2 for DTP. Dimmed patches are dropped. Panel titles give each
rule's answer and $p(\text{yes})$. Incorrect answers appear in red. The
evaluation harness scores an exact bf16 tie as ``yes,'' marked ``tie'' here.}
\label{fig:supp_cobra_tokens_a}
\end{figure*}

\begin{figure*}[p]
\centering
\includegraphics[width=\textwidth,height=0.86\textheight,keepaspectratio]{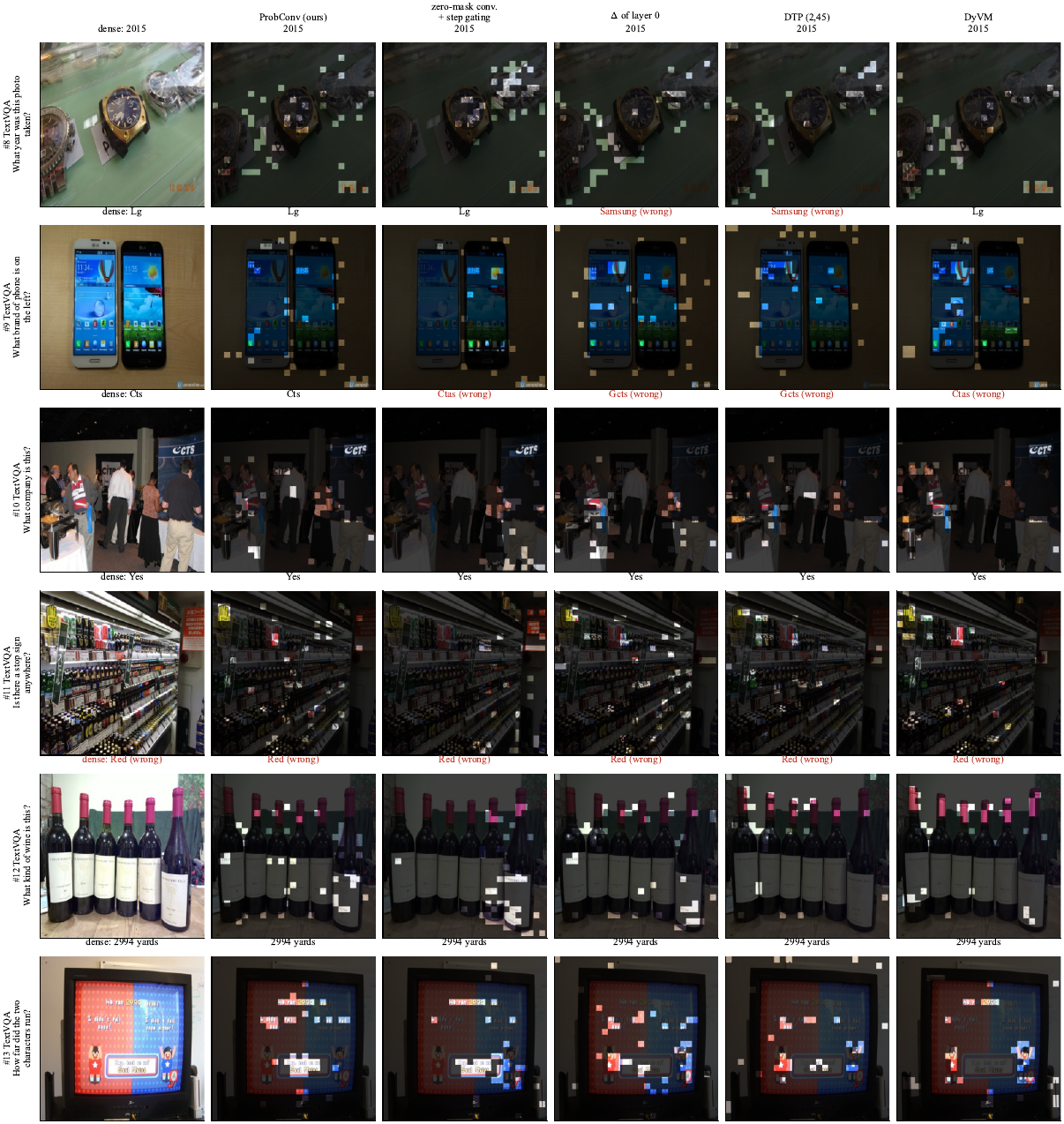}
\caption{\textbf{The six fixed TextVQA examples compare retained patches and answers
across Cobra selectors at a nominal 10\% of LM FLOPs.} Dimmed patches are
dropped, and incorrect answers appear in red. These are examples 8--13 of the
gallery introduced in \cref{fig:supp_cobra_tokens_a}.}
\label{fig:supp_cobra_tokens_b}
\end{figure*}

\begin{figure*}[p]
\centering
\includegraphics[width=\textwidth,height=0.86\textheight,keepaspectratio]{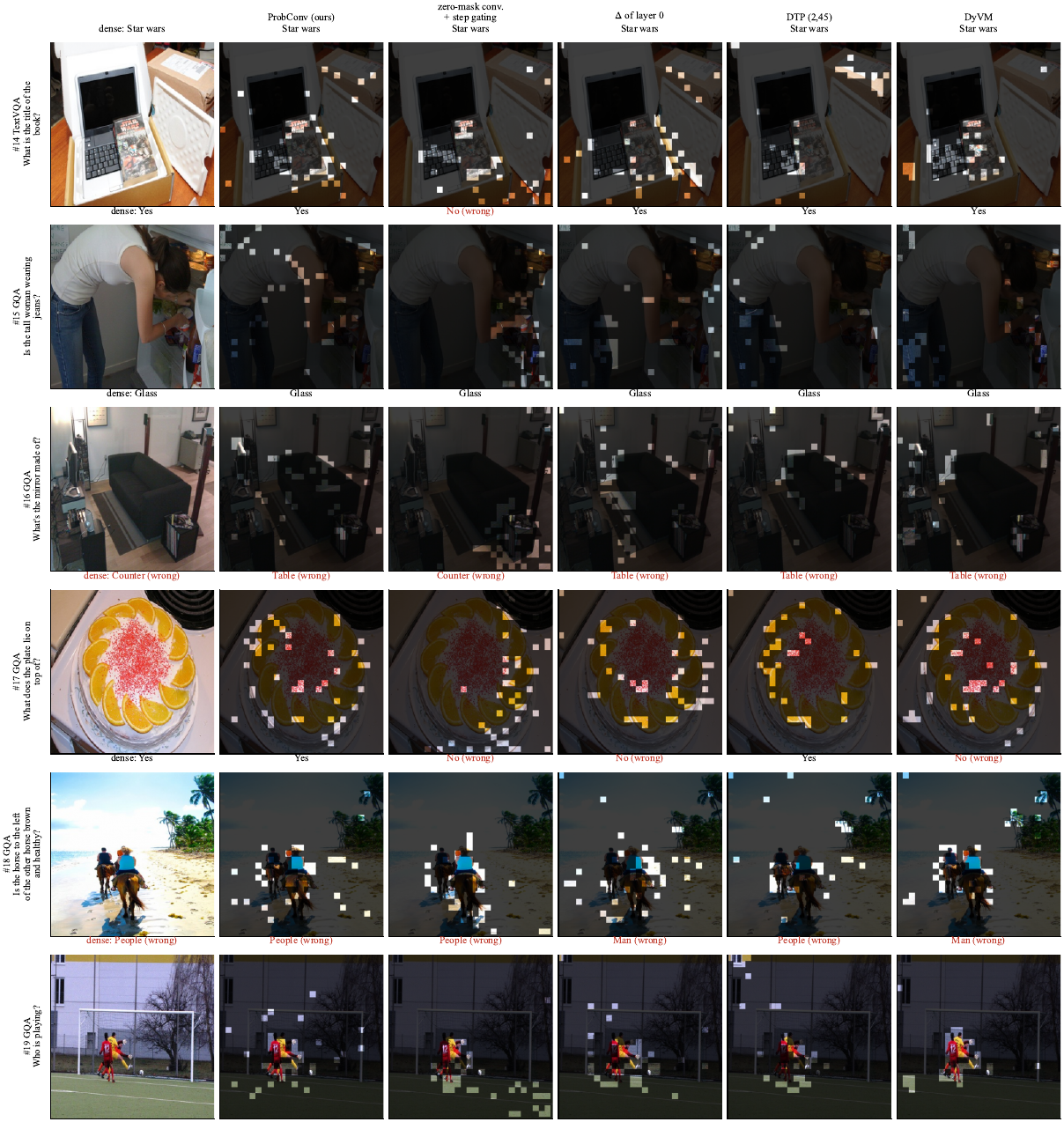}
\caption{\textbf{The remaining TextVQA and GQA examples compare Cobra selectors at a
nominal 10\% of LM FLOPs.} Dimmed patches are dropped, and incorrect answers
appear in red. These are examples 14--19 of the gallery introduced in
\cref{fig:supp_cobra_tokens_a}.}
\label{fig:supp_cobra_tokens_c}
\end{figure*}

\begin{figure*}[p]
\centering
\includegraphics[width=\textwidth,height=0.86\textheight,keepaspectratio]{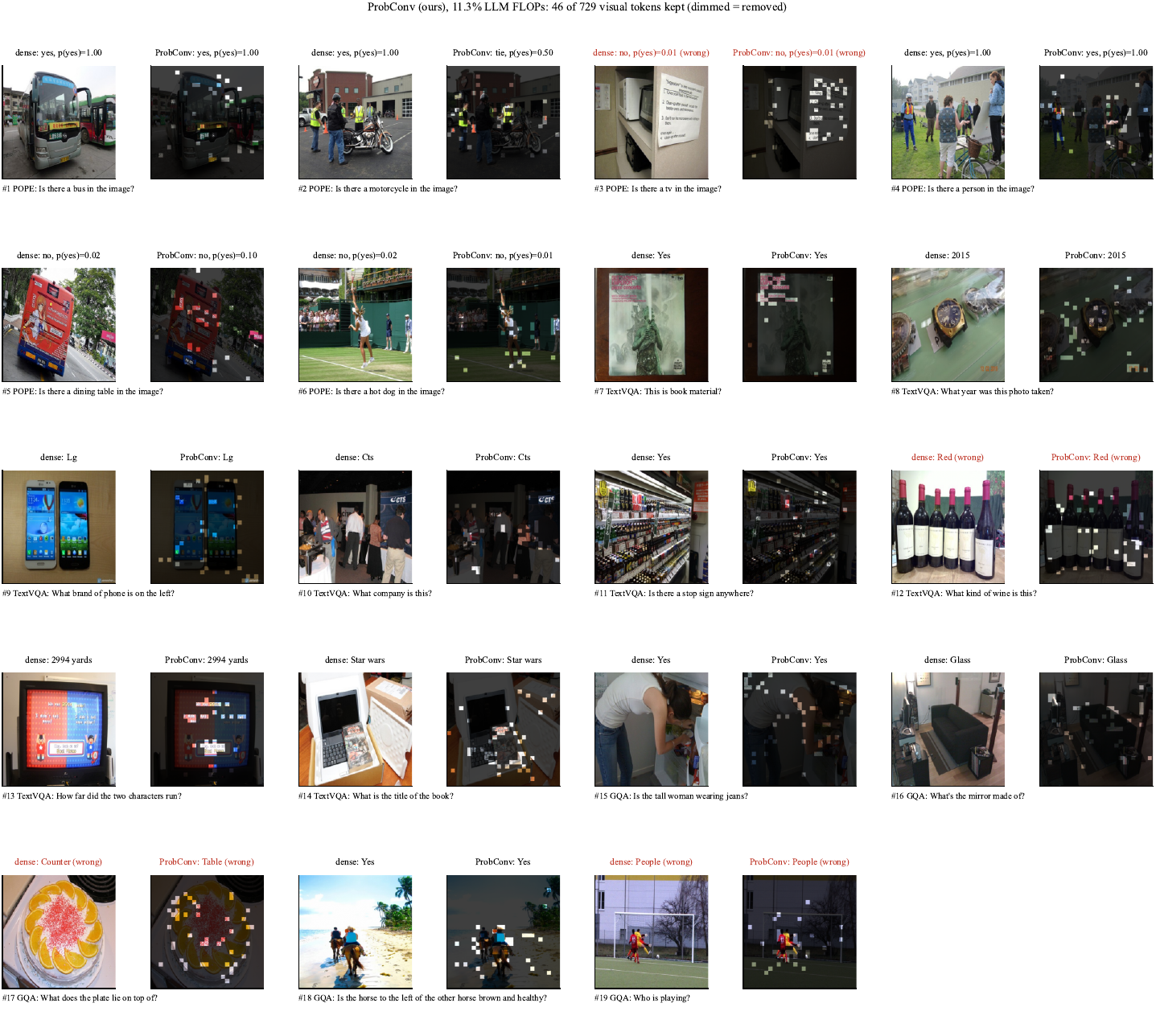}
\caption{\textbf{The full gallery pairs dense Cobra's answers with \ours's compacted
answers.} All 19 fixed examples show the original image beside the selection
at a nominal 10\% of LM FLOPs. The two labels give the model answers. Dimmed
patches are dropped.}
\label{fig:supp_cobra_tokens_ours}
\end{figure*}

\paragraph{What is kept.} \Cref{fig:cobra_examples} shows two examples in
detail.
In the left example, none of the 46 retained patches overlaps the logo. The
vision encoder mixes information across the image before token selection. The 19 examples of
\cref{fig:supp_cobra_tokens_a,fig:supp_cobra_tokens_b,fig:supp_cobra_tokens_c,fig:supp_cobra_tokens_ours}
are fixed in advance: we take every 97th distinct image of POPE, TextVQA and
GQA in index order, at most eight per benchmark, with one question per image. For the 36 projector-level examples, the retained tokens extracted from the
language-model inputs match the displayed sets exactly, and the harness
reproduces the shown POPE probabilities within 0.016. For DTP, five of six
examples match. An occlusion test confirms the patch geometry (the
bottom 6-pixel strip of the image changes no token). \Cref{fig:supp_cobra_freq}
counts how often each position is kept: \ours spreads its budget over the
image, the zero-mask arm concentrates on the last raster positions, the
$\Delta$ score and DTP always keep the first token, and DyVM keeps a fixed
cluster at the bottom left, a SigLIP artifact (\cref{sec:supp_cobra_behaviour}).
\Cref{sec:supp_cobra_behaviour} also tests why the selector of \ours
works: its gradient, its training dynamics and what its tokens cover.

\paragraph{What matters.} As in ReMamba (\cref{sec:supp_remamba}), the soft
forward pass works here because training drives it to the vertices, where it
is the deployed compacted network. With the annealed soft forward at 10\% of the language model's FLOPs, stopping
the convolution gradient lowers the mean from 61.1 to 55.3, while removing
step gating gives 61.2. With a straight-through forward, both paths matter:
the full model scores 60.6, compared with 50.3 and 50.4 when the convolution
gradient or the step gate is removed (\cref{tab:supp_cobra_ablation}). Relative to matched zero-mask training at 10\%, the largest gains are on VQAv2
($+6.3$), POPE ($+5.8$) and GQA ($+4.2$). TextVQA gains 0.7 points against
this control and 7.7 against the $\Delta$ score of layer 0. VSR and VizWiz
show little consistent benefit. With the same LoRA setup for every rule, \ours gives the highest mean at both
budgets (61.7 and 64.2). Four-sample RLOO is within 0.2 points (61.5 and
64.1). The layer-0 $\Delta$ score gives 60.1 and 63.2. At batch size 1 on a GH200 the end-to-end time is
dominated by the vision encoder and kernel launches (prefill 82--90\,ms for
every rule), so pruning the visual span saves little wall-clock time there. On
an H100 (bf16 prefill of the vision encoder, projector, selector and language
model, synthetic images), the prefill takes 66.0\,ms for the dense model and
35.5 and 34.8\,ms at 25\% and 10\% of the language model's FLOPs at batch
size 1. At batch size 32 the compacted networks process 111.9 and 140.9 images per
second against 49.5 for the dense model ($2.26\times$ and $2.84\times$), and the
selector costs no more than the $\Delta$ rule (110.0 and 137.7 images per second).

\FloatBarrier
\section{Vision--language: what the Cobra selectors learn}
\label{sec:supp_cobra_behaviour}

\begin{figure*}[t]
\centering
\includegraphics[width=\textwidth]{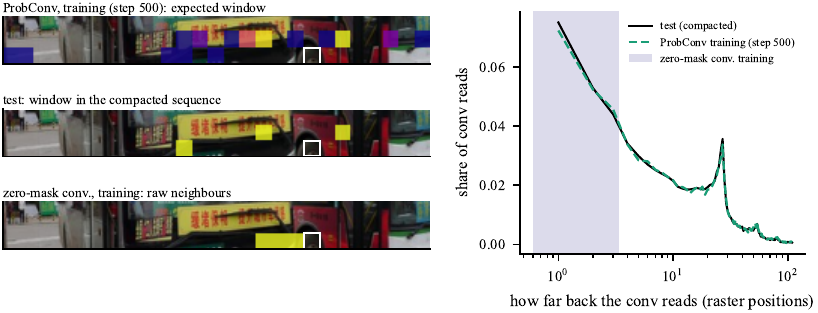}
\caption{\textbf{After compaction, the short convolution reads beyond the three
original predecessor offsets.} \ours follows the observed distribution of
these distances during training. At left, the white square marks a retained
token. The rows show the \ours training window after 500 steps, the physically
compacted window and the zero-mask training window. At right, curves give
predecessor distances in raster positions over 500 POPE images. The solid
curve shows deployment, the dashed curve shows training with \ours, and
shading marks the zero-mask offsets.}
\label{fig:supp_cobra_window}
\end{figure*}

The short convolution in Cobra mixes a retained token with the three preceding
survivors. At 46 of 729 visual tokens, these neighbors can be far apart in
raster order. Across 500 POPE images, the deployed distance distribution has a
peak at 27 positions, one image row. \ours already approximates this
distribution after 500 training steps, while the zero-mask convolution reads
only the three original predecessor offsets (\cref{fig:supp_cobra_window}). We
also examine the gradients of the training computations, the evolution of the
keep probabilities and what the kept tokens cover. Unless noted otherwise, the
diagnostics use seed 0 of the frozen-backbone rules at 10\% of the FLOPs
($k=46$ of 729 tokens) and the first 500 distinct POPE images; each diagnostic
states the number of questions or images it uses. The learned rules are read at their checkpoints
after 500, 1,000, 1,500 and 2,000 steps. Because the visual tokens precede the
text, every selector score depends on the image only.

\begin{figure*}[t]
\centering
\includegraphics[width=\textwidth]{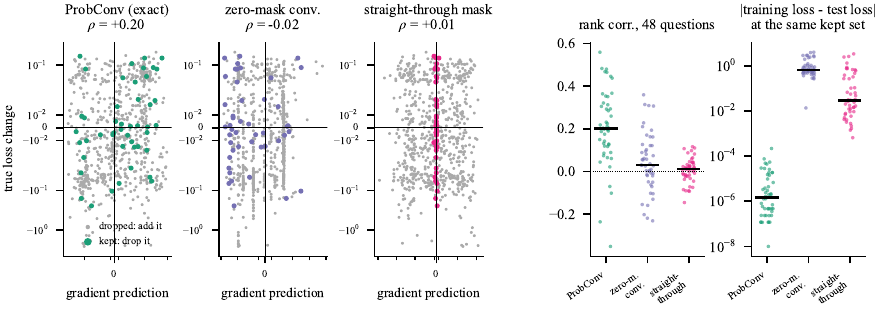}
\caption{\textbf{At 10\% of LM FLOPs, \ours's gradient ranks the measured loss changes
of single-token flips better than the gradients of the two zero-mask
computations.} For one POPE question, the first three panels compare predicted
and measured loss changes for 729 token additions or removals on symmetric log
axes. Colored points mark removals, and gray points mark additions. The fourth
panel gives rank correlations over 48 questions, with bars at the medians. The
right panel gives the absolute loss gap to physical compaction at the same
mask. The example question is the one with the median \ours correlation. Only
the training forward pass of \ours is the deployed network.}
\label{fig:supp_cobra_gradfid}
\end{figure*}

\paragraph{\ours's gradient tracks loss changes when tokens are added or removed.} We take
the 46 tokens kept by the trained \ours selector as the reference set $S_0$.
For each of the 729 tokens we flip its membership (drop it if kept, add it if
dropped), physically compact the sequence, run the deployed network and record
the change of the POPE loss $-\log\mathrm{softmax}_{\{\text{yes},\text{no}\}}$
at the answer position, the readout of the harness. We compare this true
change with the first-order change predicted by the gradient $\partial
L/\partial p_i$ of each relaxation at $p=\mathbb{1}[S_0]$ (held $10^{-6}$ inside
$[0,1]$): \ours, the zero-mask convolution with step gating, and the
straight-through zero mask, i.e.\ exactly the gradients the three rules of
\cref{tab:supp_cobra_perbench} train with. Everything runs in fp32 without
autocast; the spread of the base loss over batch compositions is at most
$6.1{\cdot}10^{-5}$, below the one-token effects. Over 48 POPE questions
(\cref{fig:supp_cobra_gradfid}), the rank correlation between prediction and
truth has median $+0.20$ for \ours and is positive for 92\% of the questions;
for the zero-mask convolution it is $+0.03$ (positive for 65\%) and for the
straight-through mask $+0.01$ (58\%). On the ten tokens with the largest true
effect the predicted sign is right in 62\% of the cases for \ours and at chance
for the two masks (50\% and 52\%). The right panel compares each training computation with the deployed network
at the same kept set. For \ours, the training forward pass computes the
deployed network (median loss difference $1.5{\cdot}10^{-6}$, the fp32 floor;
\cref{prop:block}). The zero-mask convolution and the straight-through mask
compute different networks, with median loss differences of 0.66 and 0.03,
respectively, so their gradients describe those networks and not the deployed
one. A correlation of
$+0.2$ is modest in absolute terms, since a one-token flip is a discrete change
that a first-order term captures only in part, but it is the only one of the
three that is clearly above zero.

\begin{figure*}[t]
\centering
\includegraphics[width=\textwidth]{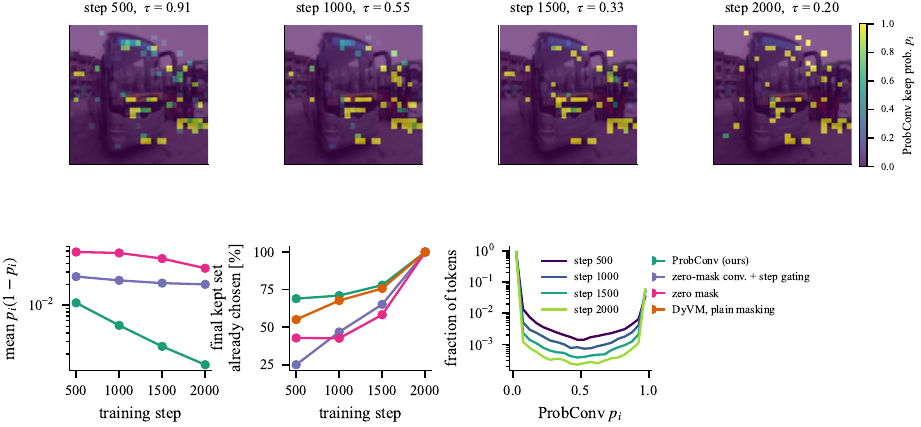}
\caption{\textbf{\ours's keep probabilities sharpen over training.} At 10\% of LM
FLOPs, the upper panels show one POPE image and its LapSum temperatures at
four checkpoints. Below, the left panel gives mean $p_i(1-p_i)$ over 500
images, the middle panel tracks overlap with the final 46 tokens, and the
right panel shows the probability histograms. The left panel includes only the
three LapSum-trained rules; DyVM's sigmoid probabilities carry no budget and
are omitted.}
\label{fig:supp_cobra_crystal}
\end{figure*}

\paragraph{The keep probabilities of \ours reach the vertices early.} Over
training the LapSum probabilities of \ours become binary
(\cref{fig:supp_cobra_crystal}): the mean of $p_i(1-p_i)$ falls from
$1.1{\cdot}10^{-2}$ after 500 steps to $1.4{\cdot}10^{-3}$ after 2,000, and the
share of tokens with $0.1<p_i<0.9$ from 4.6\% to 0.6\%. The two zero-mask arms,
with the same LapSum, temperature schedule and penalty, stay further from the
vertices ($2.0{\cdot}10^{-2}$ and $3.4{\cdot}10^{-2}$ at the end, with 8.6\% and
13.1\% of the tokens undecided). The selection of \ours is also settled early:
69\% of its final 46 tokens are already among its top 46 after 500 steps,
against 25\% for the zero-mask convolution, 43\% for the zero mask and 55\% for
DyVM. This matches the learning curve of \cref{tab:supp_cobra_curve}, which is
flat from 500 steps on, and it is what makes the soft forward pass a good
proxy here: near the vertices it is the compacted network.

\begin{figure*}[t]
\centering
\includegraphics[width=\textwidth]{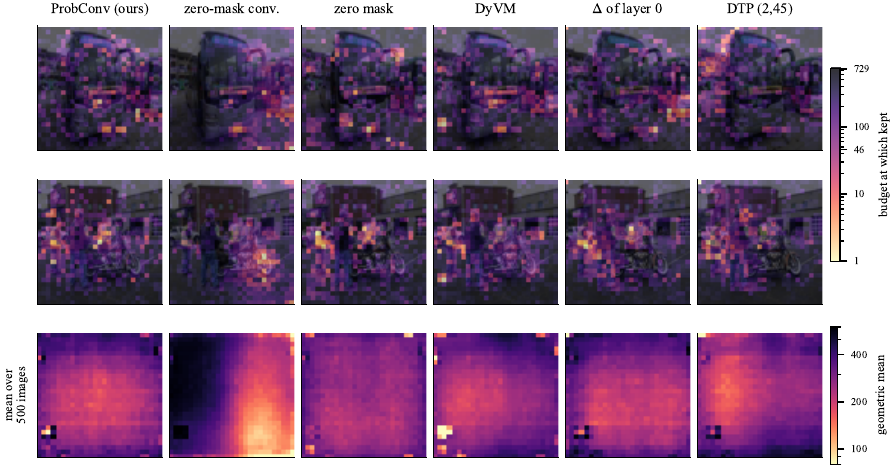}
\caption{\textbf{\ours's geometric-mean rank map is nearly flat, while the zero-mask
convolution ranks the last raster positions first.} The top rows show two POPE
images. Bright positions enter the top-$k$ set at smaller budgets. The bottom
row gives the geometric mean of these ranks over 500 images.}
\label{fig:supp_cobra_survival}
\end{figure*}

\paragraph{One score vector, every budget.} Each rule selects its top-$k$ tokens by score. For a given token, its rank at
the 10\% checkpoint equals the smallest $k$ at which that ranking retains it
(\cref{fig:supp_cobra_survival}). On individual images the learned rules follow the
content. Averaged over 500 images, the rank map of \ours is close
to flat with a mild preference for the centre, while the zero-mask convolution
ranks the last raster positions (right and bottom of the image) first at every
budget, the positional bias already visible in \cref{fig:supp_cobra_freq}.
DyVM ranks a fixed cluster at the bottom left first, and the $\Delta$ score
and DTP the first token.

\begin{figure*}[t]
\centering
\includegraphics[width=\textwidth]{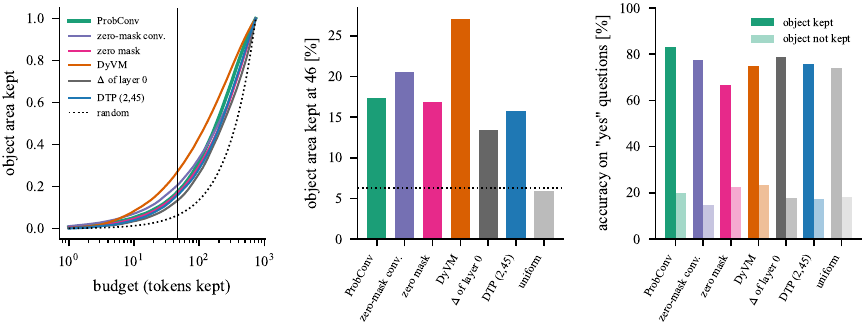}
\caption{\textbf{Object coverage and conditional accuracy rank the selectors
differently.} The left panel shows retained object area across budgets on
4,053 POPE ``yes'' questions. The vertical line marks 46 tokens (10\% of LM
FLOPs), also shown in the middle panel with random selection as a dotted
reference. The right panel shows accuracy with at least one retained token
whose area is at least half covered by the queried object (solid), and without
such a token (light). Objects come from COCO annotations rasterized onto the
token grid.}
\label{fig:supp_cobra_objects}
\end{figure*}

\paragraph{Keeping more of the object does not mean answering better.} For each
of the 4,053 POPE questions whose answer is ``yes'' we rasterize all COCO
instances of the asked category onto the $27{\times}27$ token grid, as Cobra
sees the image, and measure which share of the object area each rule keeps
(\cref{fig:supp_cobra_objects}). At 46 tokens DyVM keeps the most (27.1\%),
followed by the zero-mask convolution (20.6\%), \ours (17.3\%), the zero mask
(16.9\%), DTP (15.8\%), the $\Delta$ score (13.4\%) and uniform selection
(5.9\%; random 6.3\%). Coverage alone thus does not explain the ranking of
\cref{tab:supp_cobra_perbench}. What differs is how useful the kept part is to
the language model: when at least one kept token lies mostly ($\ge50\%$) on the
object, \ours answers 83.3\% of these questions correctly, against 74.8\% for
DyVM, 77.4\% for the zero-mask convolution, 66.6\% for the zero mask, 78.8\% for
the $\Delta$ score, 75.7\% for DTP and 74.1\% for uniform selection; when no
kept token does, every rule is at 15--23\%. This is consistent with a selector
that is trained through the network it is deployed in and therefore picks the
tokens that network can use, rather than the largest share of the object. We measure this association on the ``yes'' half of POPE for one seed. The
questions that satisfy the overlap condition vary across rules.

\paragraph{Where the convolution reads from.} Each Mamba block of Cobra has a
causal convolution with $K=4$, so a kept token mixes with the three tokens
before it in the \emph{compacted} sequence, which can lie far back in raster
order. During training, \ours replaces the convolution by its expectation over
the compactions: the $m$-th tap reads token $j$ with the probability that $j$
is the $m$-th kept token before $t$. The zero-mask convolution instead reads
the raw neighbours $t-1,\dots,t-3$, most of them zeroed
(\cref{fig:supp_cobra_window}). Over all 500 images and the kept tokens of the
final selector, the distance at which the deployed network reads extends
far beyond the three raw neighbours, with a peak at 27, the token directly
above in the image. The expected window of \ours already matches this
distribution after 500 steps, while the zero-mask convolution trains only on
distances 1--3.

\begin{figure*}[t]
\centering
\includegraphics[width=\textwidth]{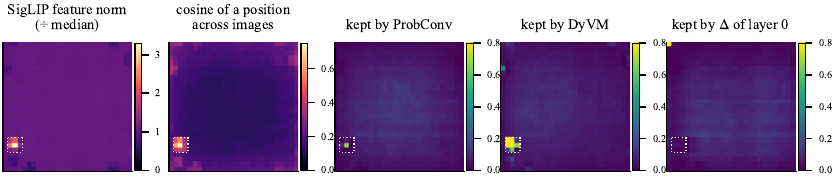}
\caption{\textbf{DyVM's recurring selection cluster overlaps the SigLIP outlier token
at grid position (21,2), marked by the dotted square.} The first two maps
summarize feature norm relative to the median and cross-image cosine
similarity over 300 POPE images. The remaining maps show selection frequencies
at 10\% of LM FLOPs for \ours, adapted DyVM and the layer-0 $\Delta$ score on
the images used in \cref{fig:supp_cobra_freq}.}
\label{fig:supp_cobra_artifact}
\end{figure*}

\paragraph{The fixed cluster of DyVM sits on a high-norm SigLIP token.} The bright cells of
\cref{fig:supp_cobra_freq} at the bottom left sit at grid position (21,2),
where SigLIP produces a high-norm token in every image: on 300 POPE images its
feature norm is the largest of the grid in all 300 and 3.3 times the median,
and its features are nearly the same across images (cosine 0.74, median over
positions 0.15; \cref{fig:supp_cobra_artifact}). DyVM keeps this token in every
image and five of its neighbours in 60--100\% of them, spending about six of its
46 tokens on positions that carry little image-specific content. \ours keeps the token itself in 68\% of the images and each of its neighbors
in at most 10\%, and the $\Delta$ score of layer 0 keeps it in 4\%.
Across these images, the clearest difference between the training rules is
which predecessors the convolution reads after selection
(\cref{fig:supp_cobra_window}).

\end{document}